\documentclass[pdflatex,sn-nature]{sn-jnl}
\usepackage{graphicx}
\usepackage{amsmath,amssymb}
\usepackage{xcolor}
\usepackage{colortbl}
\usepackage{booktabs}
\usepackage{longtable}
\usepackage{array}
\usepackage{needspace}
\usepackage[breakable]{tcolorbox}
\definecolor{PromptBackground}{HTML}{F5F7FA}
\definecolor{PromptBorder}{HTML}{D6DEE8}
\newtcolorbox{agentprompt}{%
  breakable,
  colback=PromptBackground,
  colframe=PromptBorder,
  coltext=black,
  boxrule=0.4pt,
  arc=2mm,
  boxsep=0pt,
  left=3.5mm,right=3.5mm,top=3mm,bottom=3mm,
  before skip=8pt,after skip=10pt,
  fontupper=\fontsize{9.5}{12}\selectfont\ttfamily,
  before upper={\raggedright\setlength{\parskip}{0pt}\setlength{\parindent}{0pt}}
}

\usepackage[nameinlink,sort&compress]{cleveref}
\crefname{figure}{Fig.}{Figs.}
\Crefname{figure}{Figure}{Figures}
\crefname{table}{Table}{Tables}
\Crefname{table}{Table}{Tables}
\crefname{equation}{Eq.}{Eqs.}
\Crefname{equation}{Equation}{Equations}
\crefname{section}{Section}{Sections}
\Crefname{section}{Section}{Sections}
\crefname{subsection}{Section}{Sections}
\Crefname{subsection}{Section}{Sections}
\crefname{suppfigure}{Supplementary Fig.}{Supplementary Figs.}
\Crefname{suppfigure}{Supplementary Figure}{Supplementary Figures}
\crefname{figurepanel}{Fig.}{Figs.}
\Crefname{figurepanel}{Figure}{Figures}
\crefname{suppfigurepanel}{Supplementary Fig.}{Supplementary Figs.}
\Crefname{suppfigurepanel}{Supplementary Figure}{Supplementary Figures}
\crefname{supptable}{Supplementary Table}{Supplementary Tables}
\Crefname{supptable}{Supplementary Table}{Supplementary Tables}
\crefname{suppinfo}{Supplementary Methods}{Supplementary Methods}
\Crefname{suppinfo}{Supplementary Methods}{Supplementary Methods}
\crefname{suppmethodsection}{Supplementary Methods section}{Supplementary Methods sections}
\Crefname{suppmethodsection}{Supplementary Methods section}{Supplementary Methods sections}
\crefname{suppresultsection}{Supplementary Results section}{Supplementary Results sections}
\Crefname{suppresultsection}{Supplementary Results section}{Supplementary Results sections}
\newcounter{suppinfo}
\newcounter{suppmethodsection}
\newcounter{suppresultsection}
\newcounter{suppmethodsubsection}[suppmethodsection]
\newcounter{suppresultsubsection}[suppresultsection]
\renewcommand{\thesuppresultsubsection}{\thesuppresultsection.\arabic{suppresultsubsection}}

\crefname{suppresultsubsection}{Supplementary Results section}{Supplementary Results sections}
\Crefname{suppresultsubsection}{Supplementary Results section}{Supplementary Results sections}
\newcommand{\supplementaryresultsubsection}[2]{%
  \Needspace{4\baselineskip}%
  \refstepcounter{suppresultsubsection}%
  \subsubsection*{\thesuppresultsubsection. #1}\label{#2}%
}
\renewcommand{\thesuppmethodsubsection}{\thesuppmethodsection.\arabic{suppmethodsubsection}}

\crefname{suppmethodsubsection}{Supplementary Methods section}{Supplementary Methods sections}
\Crefname{suppmethodsubsection}{Supplementary Methods section}{Supplementary Methods sections}
\newcommand{\supplementarymethodsubsection}[2]{%
  \Needspace{4\baselineskip}%
  \refstepcounter{suppmethodsubsection}%
  \subsubsection*{\thesuppmethodsubsection. #1}\label{#2}%
}

\newcommand{\supplementaryinfosection}[1]{%
  \refstepcounter{suppinfo}%
  \label{#1}%
}
\newcommand{\supplementarymethodsection}[3][\FloatBarrier]{%
  #1%
  \Needspace{5\baselineskip}%
  \refstepcounter{suppmethodsection}%
  \subsection*{\thesuppmethodsection. #2}\label{#3}%
}
\newcommand{\supplementaryresultsection}[2]{%
  \Needspace{5\baselineskip}%
  \refstepcounter{suppresultsection}%
  \subsection*{\thesuppresultsection. #1}\label{#2}%
}
\makeatletter
\let\manuscript@native@tablecaption\@tablecaption
\long\def\@tablecaption#1#2{%
  \hbox to\hsize{\hss%
    \parbox[t]{\textwidth}{%
      \manuscript@native@tablecaption{#1}{#2}}%
    \hss}%
}
\def\LT@makecaption#1#2#3{%
  \LT@mcol\LT@cols{@{}c@{}}{\hbox to\z@{\hss\parbox[t]\LTcapwidth{%
    \tablecaptionfont\mathversion{normal}%
    \@tablecaption{#1{#2}}{#3}%
    \endgraf\vskip\baselineskip}%
  \hss}}}
\makeatother
\newcommand{\manuscripttablecontinued}{%
  {\tablecaptionfont\mathversion{normal}\bfseries
   \tablename~\thetable\ (continued)}%
}
\newcommand{\manuscripttablefont}{\tablebodyfont\mathversion{normal}}
\newenvironment{supplementarytables}{%
  \begingroup\manuscripttablefont
  \setlength{\tabcolsep}{2pt}\renewcommand{\arraystretch}{1.08}%
  \setlength{\LTcapwidth}{\linewidth}%
  \setlength{\LTpre}{7pt}\setlength{\LTpost}{10pt}%
}{\endgroup}
\newenvironment{supplementarynativetables}{%
  \begingroup\manuscripttablefont
  \setlength{\tabcolsep}{2pt}\renewcommand{\arraystretch}{1.02}%
  \setlength{\LTcapwidth}{\linewidth}%
  \setlength{\LTpre}{7pt}\setlength{\LTpost}{10pt}%
}{\endgroup}

\usepackage{environ}
\usepackage{placeins}
\usepackage{overpic}
\usepackage{helvet}
\input{manuscript_panel_labels}
\renewcommand{\floatpagefraction}{0.85}
\makeatletter
\newcommand{\manuscriptsetfigurename}[1]{\def\@currentlabelname{#1}}
\newcommand{\manuscriptformatfigurecaption}[3]{%
  \mathversion{normal}%
  \@figurecaption{#1}{{\let\textbf\@firstofone #2}\space #3}%
}
\makeatother

\newsavebox{\manuscriptartbox}
\newsavebox{\manuscriptnativeartbox}
\newsavebox{\manuscripttitlebox}
\newsavebox{\manuscriptlegendbox}
\newsavebox{\manuscriptlegendpart}
\newsavebox{\manuscriptstackbox}
\newdimen\manuscriptlegendroom
\newdimen\manuscriptfullheight
\newdimen\manuscriptartroom
\newdimen\manuscriptnaturalheight
\newdimen\manuscriptpagelimit
\newdimen\manuscriptnonartheight
\newif\ifmanuscriptcontinued
\newcommand{\manuscriptminartscale}{1.0}
\newcommand{\manuscriptlegendfont}{\figurecaptionfont}
\newcommand{\manuscriptpanelblock}[5]{%
\begin{minipage}[t]{#2}\vspace{0pt}%
\parbox[t][#1][t]{\linewidth}{{\fontencoding{T1}\fontfamily{phv}\selectfont\fontsize{8}{9}\selectfont\textbf{#3}\hspace{1.5mm}\textbf{#4}}}\par
\nointerlineskip#5%
\end{minipage}}
\newcommand{\manuscriptmeasurestack}{%
  \setbox\manuscriptstackbox=\vbox{%
    \hsize=\linewidth\offinterlineskip
    \hbox to \linewidth{\hfil\copy\manuscriptartbox\hfil}%
    \vskip 4pt\copy\manuscripttitlebox
    \copy\manuscriptlegendbox}%
  \manuscriptfullheight=\dimexpr\ht\manuscriptstackbox+\dp\manuscriptstackbox\relax}

\newcommand{\manuscriptfigurecore}[4]{%
  \begin{figure}[!htbp]
  \refstepcounter{figure}\manuscriptsetfigurename{#1}\label{#2}%
  \edef\manuscriptfigureid{\thefigure}%
  \sbox{\manuscriptnativeartbox}{#3}%
  \ifdim\wd\manuscriptnativeartbox>\linewidth
    \sbox{\manuscriptartbox}{\resizebox{\linewidth}{!}{\usebox{\manuscriptnativeartbox}}}%
  \else
    \sbox{\manuscriptartbox}{\usebox{\manuscriptnativeartbox}}%
  \fi
  \manuscriptnaturalheight=\dimexpr\ht\manuscriptartbox+\dp\manuscriptartbox\relax
  \manuscriptpagelimit=\dimexpr\textheight-6pt\relax
  \setbox\manuscripttitlebox=\vbox{}%
  \setbox\manuscriptlegendbox=\vbox{%
    \hsize=\linewidth\parindent=0pt\parskip=0pt
    \manuscriptformatfigurecaption{\figurename~\thefigure}{#1}{#4}}%
  \manuscriptmeasurestack
  \typeout{PATHXAI-CAPTION-CHECK #2 full=\the\manuscriptfullheight; limit=\the\manuscriptpagelimit}%
  \manuscriptcontinuedfalse
  \ifdim\manuscriptfullheight>\manuscriptpagelimit
    \PackageError{PathXAI}{Figure #2 and its legend exceed one page}{%
      Shorten repeated legend text or rearrange the panel assembly.
      Preserve native artwork size and readable type.}%
  \fi
  \ifdim\manuscriptfullheight>\manuscriptpagelimit
    \manuscriptnonartheight=\dimexpr\manuscriptfullheight-\manuscriptnaturalheight\relax
    \manuscriptartroom=\dimexpr\manuscriptpagelimit-\manuscriptnonartheight-.5pt\relax
    \ifdim\manuscriptartroom<\manuscriptminartscale\manuscriptnaturalheight
      \PackageInfo{PathXAI}{Figure #2 uses an explicit legend continuation to preserve native artwork}%
    \else
      \sbox{\manuscriptartbox}{\resizebox*{!}{\manuscriptartroom}{\usebox{\manuscriptnativeartbox}}}%
      \manuscriptmeasurestack
    \fi
  \fi
  \ifdim\manuscriptfullheight>\manuscriptpagelimit
    \manuscriptcontinuedtrue
    \manuscriptlegendroom=\dimexpr\manuscriptpagelimit-\manuscriptfullheight
      +\ht\manuscriptlegendbox+\dp\manuscriptlegendbox-19pt\relax
    \ifdim\manuscriptlegendroom<0pt
      \PackageError{PathXAI}{Figure #2 artwork and title exceed the page}{Split the panel assembly; do not shrink its text.}%
    \fi
    \ifdim\manuscriptlegendroom<19pt
      \setbox\manuscriptlegendpart=\vbox{}%
    \else
      \setbox\manuscriptlegendpart=\vsplit\manuscriptlegendbox to \manuscriptlegendroom
    \fi
  \fi
  \typeout{PATHXAI-FIGURE #2 natural=\the\manuscriptnaturalheight; art=\the\dimexpr\ht\manuscriptartbox+\dp\manuscriptartbox\relax; full=\the\manuscriptfullheight; limit=\the\manuscriptpagelimit; continued=\ifmanuscriptcontinued yes\else no\fi}%
  \ifmanuscriptcontinued
    \noindent\makebox[\linewidth][c]{\usebox{\manuscriptartbox}}\par\vskip 4pt
    \unvbox\manuscripttitlebox
    \unvbox\manuscriptlegendpart
    {\manuscriptlegendfont\slshape Legend continued on \cpageref{#2-legend}.\par}%
  \else
    \box\manuscriptstackbox
  \fi
  \ifmanuscriptcontinued
    \global\setbox\manuscriptlegendbox=\box\manuscriptlegendbox
    \xdef\manuscriptcontinuedid{\manuscriptfigureid}%
    \global\manuscriptcontinuedtrue
  \else
    \global\manuscriptcontinuedfalse
  \fi
  \end{figure}
  \ifmanuscriptcontinued
    \begin{figure}[!htbp]
    \phantomsection
    \label{#2-legend}%
    \manuscriptlegendfont\parindent=0pt
    {\bfseries\figurename~\manuscriptcontinuedid\quad Legend (continued)\par}\vskip 4pt
    \unvbox\manuscriptlegendbox
    \end{figure}
  \fi
}
\newcommand{\manuscriptnativefigure}[4]{%
  \manuscriptfigurecore{#1}{#2}{#3}{#4}%
}
\NewEnviron{manuscriptfigure}[3]{%
  \manuscriptfigurecore{#2}{#3}{\includegraphics[width=180mm]{#1}}{\BODY}%
}

\newcommand{\manuscriptwidefigureplacement}{p}
\newcommand{\manuscriptwidefigurepart}[5]{%
  \edef\manuscriptopenwidefigure{\noexpand\begin{figure}[\manuscriptwidefigureplacement]}%
  \manuscriptopenwidefigure
  \def\manuscriptpartkind{#1}\def\manuscriptcontinuationkind{continued}%
  \ifx\manuscriptpartkind\manuscriptcontinuationkind
    \addtocounter{figure}{-1}%
  \fi
  \renewcommand{\theHfigure}{\arabic{figure}.#3}%
  \refstepcounter{figure}\manuscriptsetfigurename{#2}\label{#3}%
  \setbox\manuscriptstackbox=\vbox{%
    \hsize=180mm\linewidth=180mm\parindent=0pt\parskip=0pt
    \noindent\ignorespaces#4\unskip\par\vskip 4pt
    \manuscriptformatfigurecaption{\figurename~\thefigure}{#2}{#5}}%
  \manuscriptfullheight=\dimexpr\ht\manuscriptstackbox+\dp\manuscriptstackbox\relax
  \typeout{PATHXAI-WIDE-FIGURE #3 width=180mm; full=\the\manuscriptfullheight; limit=\the\textheight}%
  \ifdim\manuscriptfullheight>\textheight
    \PackageError{PathXAI}{Figure part #3 exceeds the page}{Shorten the legend or split the artwork; native text must not shrink.}%
  \fi
  \noindent\makebox[\linewidth][c]{\usebox{\manuscriptstackbox}}%
  \end{figure}%
}

\unnumbered
\hypersetup{colorlinks=true,linkcolor=blue,citecolor=blue,urlcolor=blue,
  pdfborder={0 0 0}}

\begin{document}
\newcommand{\CuratedCancerFull}{\textcolor{black}{76.75}}
\newcommand{\CuratedCancerFullCI}{\textcolor{black}{72.60--80.56}}
\newcommand{\CuratedCancerFullN}{456}
\newcommand{\CuratedDiagnosticFull}{\textcolor{black}{85.91}}
\newcommand{\CuratedDiagnosticFullCI}{\textcolor{black}{83.76--88.08}}
\newcommand{\CuratedDiagnosticFullN}{456}
\newcommand{\CuratedEntityFull}{66.19}
\newcommand{\CuratedEntityFullCI}{\textcolor{black}{60.99--71.12}}
\newcommand{\CuratedEntityFullN}{352}
\newcommand{\CuratedFindingFull}{45.51}
\newcommand{\CuratedFindingFullCI}{41.98--48.96}
\newcommand{\CuratedFindingFullN}{408}
\newcommand{\CuratedAssertionFull}{45.04}
\newcommand{\CuratedAssertionFullCI}{41.74--48.35}
\newcommand{\CuratedAssertionFullN}{408}
\newcommand{\CuratedCorrectFindingFull}{37.81}
\newcommand{\CuratedCorrectFindingFullCI}{34.55--41.09}
\newcommand{\CuratedCorrectFindingFullN}{408}
\newcommand{\CuratedContradictionFull}{15.51}
\newcommand{\CuratedContradictionFullCI}{11.92--19.21}
\newcommand{\CuratedContradictionFullN}{311}
\newcommand{\CuratedCoverageFull}{\textcolor{black}{91.40}}
\newcommand{\CuratedCoverageFullCI}{\textcolor{black}{90.08--92.71}}
\newcommand{\CuratedCoverageFullN}{456}
\newcommand{\CuratedGradeFull}{\textcolor{black}{43.23}}
\newcommand{\CuratedGradeFullCI}{\textcolor{black}{34.42--65.65}}
\newcommand{\CuratedGradeFullN}{\textcolor{black}{168}}
\newcommand{\CuratedCancerCore}{\textcolor{black}{73.68}}
\newcommand{\CuratedCancerCoreCI}{\textcolor{black}{69.39--77.67}}
\newcommand{\CuratedCancerCoreN}{456}
\newcommand{\CuratedDiagnosticCore}{\textcolor{black}{83.53}}
\newcommand{\CuratedDiagnosticCoreCI}{\textcolor{black}{81.08--86.03}}
\newcommand{\CuratedDiagnosticCoreN}{456}
\newcommand{\CuratedEntityCore}{64.77}
\newcommand{\CuratedEntityCoreCI}{\textcolor{black}{59.53--69.76}}
\newcommand{\CuratedEntityCoreN}{352}
\newcommand{\CuratedFindingCore}{41.60}
\newcommand{\CuratedFindingCoreCI}{38.00--45.07}
\newcommand{\CuratedFindingCoreN}{408}
\newcommand{\CuratedAssertionCore}{40.41}
\newcommand{\CuratedAssertionCoreCI}{37.08--43.81}
\newcommand{\CuratedAssertionCoreN}{408}
\newcommand{\CuratedCorrectFindingCore}{34.25}
\newcommand{\CuratedCorrectFindingCoreCI}{31.13--37.55}
\newcommand{\CuratedCorrectFindingCoreN}{408}
\newcommand{\CuratedContradictionCore}{14.40}
\newcommand{\CuratedContradictionCoreCI}{11.01--17.96}
\newcommand{\CuratedContradictionCoreN}{283}
\newcommand{\CuratedCoverageCore}{\textcolor{black}{87.17}}
\newcommand{\CuratedCoverageCoreCI}{\textcolor{black}{85.63--88.70}}
\newcommand{\CuratedCoverageCoreN}{456}
\newcommand{\CuratedGradeCore}{\textcolor{black}{32.64}}
\newcommand{\CuratedGradeCoreCI}{\textcolor{black}{19.59--57.96}}
\newcommand{\CuratedGradeCoreN}{\textcolor{black}{168}}
\newcommand{\CuratedCancerQwen}{71.49}
\newcommand{\CuratedCancerQwenCI}{\textcolor{black}{67.11--75.59}}
\newcommand{\CuratedCancerQwenN}{456}
\newcommand{\CuratedDiagnosticQwen}{82.34}
\newcommand{\CuratedDiagnosticQwenCI}{79.75--84.86}
\newcommand{\CuratedDiagnosticQwenN}{456}
\newcommand{\CuratedEntityQwen}{61.65}
\newcommand{\CuratedEntityQwenCI}{\textcolor{black}{56.35--66.75}}
\newcommand{\CuratedEntityQwenN}{352}
\newcommand{\CuratedFindingQwen}{40.64}
\newcommand{\CuratedFindingQwenCI}{37.17--44.10}
\newcommand{\CuratedFindingQwenN}{408}
\newcommand{\CuratedAssertionQwen}{41.41}
\newcommand{\CuratedAssertionQwenCI}{38.10--44.73}
\newcommand{\CuratedAssertionQwenN}{408}
\newcommand{\CuratedCorrectFindingQwen}{33.53}
\newcommand{\CuratedCorrectFindingQwenCI}{30.33--36.83}
\newcommand{\CuratedCorrectFindingQwenN}{408}
\newcommand{\CuratedContradictionQwen}{19.30}
\newcommand{\CuratedContradictionQwenCI}{15.35--23.54}
\newcommand{\CuratedContradictionQwenN}{297}
\newcommand{\CuratedCoverageQwen}{\textcolor{black}{84.75}}
\newcommand{\CuratedCoverageQwenCI}{\textcolor{black}{83.02--86.52}}
\newcommand{\CuratedCoverageQwenN}{456}
\newcommand{\CuratedGradeQwen}{\textcolor{black}{24.65}}
\newcommand{\CuratedGradeQwenCI}{\textcolor{black}{12.63--49.86}}
\newcommand{\CuratedGradeQwenN}{\textcolor{black}{168}}
\newcommand{\CuratedCancerDeepSeek}{\textcolor{black}{70.61}}
\newcommand{\CuratedCancerDeepSeekCI}{\textcolor{black}{66.20--74.76}}
\newcommand{\CuratedCancerDeepSeekN}{456}
\newcommand{\CuratedDiagnosticDeepSeek}{\textcolor{black}{82.43}}
\newcommand{\CuratedDiagnosticDeepSeekCI}{\textcolor{black}{79.98--84.78}}
\newcommand{\CuratedDiagnosticDeepSeekN}{456}
\newcommand{\CuratedEntityDeepSeek}{\textcolor{black}{62.50}}
\newcommand{\CuratedEntityDeepSeekCI}{\textcolor{black}{57.21--67.58}}
\newcommand{\CuratedEntityDeepSeekN}{352}
\newcommand{\CuratedFindingDeepSeek}{33.93}
\newcommand{\CuratedFindingDeepSeekCI}{30.49--37.42}
\newcommand{\CuratedFindingDeepSeekN}{408}
\newcommand{\CuratedAssertionDeepSeek}{38.27}
\newcommand{\CuratedAssertionDeepSeekCI}{34.92--41.70}
\newcommand{\CuratedAssertionDeepSeekN}{408}
\newcommand{\CuratedCorrectFindingDeepSeek}{28.23}
\newcommand{\CuratedCorrectFindingDeepSeekCI}{25.15--31.51}
\newcommand{\CuratedCorrectFindingDeepSeekN}{408}
\newcommand{\CuratedContradictionDeepSeek}{26.69}
\newcommand{\CuratedContradictionDeepSeekCI}{21.86--31.64}
\newcommand{\CuratedContradictionDeepSeekN}{268}
\newcommand{\CuratedCoverageDeepSeek}{\textcolor{black}{85.15}}
\newcommand{\CuratedCoverageDeepSeekCI}{\textcolor{black}{83.13--87.01}}
\newcommand{\CuratedCoverageDeepSeekN}{456}
\newcommand{\CuratedGradeDeepSeek}{\textcolor{black}{49.65}}
\newcommand{\CuratedGradeDeepSeekCI}{\textcolor{black}{32.63--73.73}}
\newcommand{\CuratedGradeDeepSeekN}{\textcolor{black}{168}}
\newcommand{\CuratedCancerGLM}{\textcolor{black}{55.92}}
\newcommand{\CuratedCancerGLMCI}{\textcolor{black}{51.23--60.54}}
\newcommand{\CuratedCancerGLMN}{456}
\newcommand{\CuratedDiagnosticGLM}{\textcolor{black}{69.61}}
\newcommand{\CuratedDiagnosticGLMCI}{\textcolor{black}{66.18--72.97}}
\newcommand{\CuratedDiagnosticGLMN}{456}
\newcommand{\CuratedEntityGLM}{\textcolor{black}{46.59}}
\newcommand{\CuratedEntityGLMCI}{\textcolor{black}{41.29--51.95}}
\newcommand{\CuratedEntityGLMN}{352}
\newcommand{\CuratedFindingGLM}{40.84}
\newcommand{\CuratedFindingGLMCI}{37.28--44.22}
\newcommand{\CuratedFindingGLMN}{408}
\newcommand{\CuratedAssertionGLM}{46.69}
\newcommand{\CuratedAssertionGLMCI}{43.23--50.36}
\newcommand{\CuratedAssertionGLMN}{408}
\newcommand{\CuratedCorrectFindingGLM}{34.59}
\newcommand{\CuratedCorrectFindingGLMCI}{31.30--37.88}
\newcommand{\CuratedCorrectFindingGLMN}{408}
\newcommand{\CuratedContradictionGLM}{24.55}
\newcommand{\CuratedContradictionGLMCI}{20.45--29.10}
\newcommand{\CuratedContradictionGLMN}{302}
\newcommand{\CuratedCoverageGLM}{\textcolor{black}{79.96}}
\newcommand{\CuratedCoverageGLMCI}{\textcolor{black}{77.71--82.19}}
\newcommand{\CuratedCoverageGLMN}{456}
\newcommand{\CuratedGradeGLM}{\textcolor{black}{39.76}}
\newcommand{\CuratedGradeGLMCI}{\textcolor{black}{26.54--64.64}}
\newcommand{\CuratedGradeGLMN}{\textcolor{black}{168}}
\newcommand{\CuratedCancerMedGemma}{7.68}
\newcommand{\CuratedCancerMedGemmaCI}{\textcolor{black}{5.40--10.51}}
\newcommand{\CuratedCancerMedGemmaN}{456}
\newcommand{\CuratedDiagnosticMedGemma}{39.14}
\newcommand{\CuratedDiagnosticMedGemmaCI}{36.42--42.05}
\newcommand{\CuratedDiagnosticMedGemmaN}{456}
\newcommand{\CuratedEntityMedGemma}{3.69}
\newcommand{\CuratedEntityMedGemmaCI}{\textcolor{black}{1.98--6.23}}
\newcommand{\CuratedEntityMedGemmaN}{352}
\newcommand{\CuratedFindingMedGemma}{31.00}
\newcommand{\CuratedFindingMedGemmaCI}{27.40--34.31}
\newcommand{\CuratedFindingMedGemmaN}{408}
\newcommand{\CuratedAssertionMedGemma}{31.09}
\newcommand{\CuratedAssertionMedGemmaCI}{27.86--34.40}
\newcommand{\CuratedAssertionMedGemmaN}{408}
\newcommand{\CuratedCorrectFindingMedGemma}{25.31}
\newcommand{\CuratedCorrectFindingMedGemmaCI}{22.21--28.47}
\newcommand{\CuratedCorrectFindingMedGemmaN}{408}
\newcommand{\CuratedContradictionMedGemma}{17.72}
\newcommand{\CuratedContradictionMedGemmaCI}{13.39--22.56}
\newcommand{\CuratedContradictionMedGemmaN}{229}
\newcommand{\CuratedCoverageMedGemma}{\textcolor{black}{68.33}}
\newcommand{\CuratedCoverageMedGemmaCI}{\textcolor{black}{65.14--71.45}}
\newcommand{\CuratedCoverageMedGemmaN}{456}
\newcommand{\CuratedGradeMedGemma}{\textcolor{black}{18.92}}
\newcommand{\CuratedGradeMedGemmaCI}{\textcolor{black}{14.18--40.89}}
\newcommand{\CuratedGradeMedGemmaN}{\textcolor{black}{168}}
\newcommand{\CuratedCancerPathAgent}{\textcolor{black}{29.82}}
\newcommand{\CuratedCancerPathAgentCI}{\textcolor{black}{25.66--34.25}}
\newcommand{\CuratedCancerPathAgentN}{456}
\newcommand{\CuratedDiagnosticPathAgent}{\textcolor{black}{55.25}}
\newcommand{\CuratedDiagnosticPathAgentCI}{\textcolor{black}{52.14--58.49}}
\newcommand{\CuratedDiagnosticPathAgentN}{456}
\newcommand{\CuratedEntityPathAgent}{\textcolor{black}{23.86}}
\newcommand{\CuratedEntityPathAgentCI}{\textcolor{black}{19.50--28.67}}
\newcommand{\CuratedEntityPathAgentN}{352}
\newcommand{\CuratedFindingPathAgent}{15.82}
\newcommand{\CuratedFindingPathAgentCI}{13.13--18.52}
\newcommand{\CuratedFindingPathAgentN}{408}
\newcommand{\CuratedAssertionPathAgent}{13.21}
\newcommand{\CuratedAssertionPathAgentCI}{10.99--15.60}
\newcommand{\CuratedAssertionPathAgentN}{408}
\newcommand{\CuratedCorrectFindingPathAgent}{12.09}
\newcommand{\CuratedCorrectFindingPathAgentCI}{9.97--14.30}
\newcommand{\CuratedCorrectFindingPathAgentN}{408}
\newcommand{\CuratedContradictionPathAgent}{9.92}
\newcommand{\CuratedContradictionPathAgentCI}{4.96--15.29}
\newcommand{\CuratedContradictionPathAgentN}{121}
\newcommand{\CuratedCoveragePathAgent}{\textcolor{black}{72.29}}
\newcommand{\CuratedCoveragePathAgentCI}{\textcolor{black}{69.74--74.88}}
\newcommand{\CuratedCoveragePathAgentN}{456}
\newcommand{\CuratedGradePathAgent}{\textcolor{black}{28.47}}
\newcommand{\CuratedGradePathAgentCI}{\textcolor{black}{24.92--48.36}}
\newcommand{\CuratedGradePathAgentN}{\textcolor{black}{168}}
\newcommand{\CuratedCancerPRISM}{\textcolor{black}{55.92}}
\newcommand{\CuratedCancerPRISMCI}{\textcolor{black}{51.23--60.54}}
\newcommand{\CuratedCancerPRISMN}{456}
\newcommand{\CuratedDiagnosticPRISM}{\textcolor{black}{72.64}}
\newcommand{\CuratedDiagnosticPRISMCI}{\textcolor{black}{69.54--75.63}}
\newcommand{\CuratedDiagnosticPRISMN}{456}
\newcommand{\CuratedEntityPRISM}{\textcolor{black}{49.15}}
\newcommand{\CuratedEntityPRISMCI}{\textcolor{black}{43.81--54.50}}
\newcommand{\CuratedEntityPRISMN}{352}
\newcommand{\OCRCancerFull}{\textcolor{black}{72.50}}
\newcommand{\OCRCancerFullCI}{\textcolor{black}{69.88--75.01}}
\newcommand{\OCRCancerFullN}{1200}
\newcommand{\OCRDiagnosticFull}{\textcolor{black}{86.91}}
\newcommand{\OCRDiagnosticFullCI}{\textcolor{black}{85.81--88.02}}
\newcommand{\OCRDiagnosticFullN}{1200}
\newcommand{\OCREntityFull}{\textcolor{black}{62.25}}
\newcommand{\OCREntityFullCI}{\textcolor{black}{58.25--66.13}}
\newcommand{\OCREntityFullN}{\textcolor{black}{604}}
\newcommand{\OCRFindingFull}{48.76}
\newcommand{\OCRFindingFullCI}{46.45--51.10}
\newcommand{\OCRFindingFullN}{997}
\newcommand{\OCRAssertionFull}{47.18}
\newcommand{\OCRAssertionFullCI}{44.88--49.48}
\newcommand{\OCRAssertionFullN}{997}
\newcommand{\OCRCorrectFindingFull}{41.64}
\newcommand{\OCRCorrectFindingFullCI}{39.31--43.93}
\newcommand{\OCRCorrectFindingFullN}{997}
\newcommand{\OCRContradictionFull}{11.53}
\newcommand{\OCRContradictionFullCI}{9.64--13.52}
\newcommand{\OCRContradictionFullN}{759}
\newcommand{\OCRCoverageFull}{\textcolor{black}{90.26}}
\newcommand{\OCRCoverageFullCI}{\textcolor{black}{89.34--91.15}}
\newcommand{\OCRCoverageFullN}{1200}
\newcommand{\OCRGradeFull}{\textcolor{black}{36.97}}
\newcommand{\OCRGradeFullCI}{\textcolor{black}{31.14--44.76}}
\newcommand{\OCRGradeFullN}{\textcolor{black}{472}}
\newcommand{\OCRCancerCore}{\textcolor{black}{68.17}}
\newcommand{\OCRCancerCoreCI}{\textcolor{black}{65.45--70.80}}
\newcommand{\OCRCancerCoreN}{1200}
\newcommand{\OCRDiagnosticCore}{\textcolor{black}{84.67}}
\newcommand{\OCRDiagnosticCoreCI}{\textcolor{black}{83.44--85.89}}
\newcommand{\OCRDiagnosticCoreN}{1200}
\newcommand{\OCREntityCore}{\textcolor{black}{58.11}}
\newcommand{\OCREntityCoreCI}{\textcolor{black}{54.06--62.08}}
\newcommand{\OCREntityCoreN}{\textcolor{black}{604}}
\newcommand{\OCRFindingCore}{46.70}
\newcommand{\OCRFindingCoreCI}{44.35--49.01}
\newcommand{\OCRFindingCoreN}{997}
\newcommand{\OCRAssertionCore}{43.97}
\newcommand{\OCRAssertionCoreCI}{41.70--46.24}
\newcommand{\OCRAssertionCoreN}{997}
\newcommand{\OCRCorrectFindingCore}{39.40}
\newcommand{\OCRCorrectFindingCoreCI}{37.24--41.62}
\newcommand{\OCRCorrectFindingCoreN}{997}
\newcommand{\OCRContradictionCore}{9.82}
\newcommand{\OCRContradictionCoreCI}{7.96--11.77}
\newcommand{\OCRContradictionCoreN}{723}
\newcommand{\OCRCoverageCore}{\textcolor{black}{86.64}}
\newcommand{\OCRCoverageCoreCI}{\textcolor{black}{85.62--87.63}}
\newcommand{\OCRCoverageCoreN}{1200}
\newcommand{\OCRGradeCore}{\textcolor{black}{30.25}}
\newcommand{\OCRGradeCoreCI}{\textcolor{black}{24.68--38.10}}
\newcommand{\OCRGradeCoreN}{\textcolor{black}{472}}
\newcommand{\OCRCancerQwen}{\textcolor{black}{68.08}}
\newcommand{\OCRCancerQwenCI}{\textcolor{black}{65.36--70.72}}
\newcommand{\OCRCancerQwenN}{1200}
\newcommand{\OCRDiagnosticQwen}{\textcolor{black}{84.78}}
\newcommand{\OCRDiagnosticQwenCI}{\textcolor{black}{83.53--86.01}}
\newcommand{\OCRDiagnosticQwenN}{1200}
\newcommand{\OCREntityQwen}{\textcolor{black}{58.94}}
\newcommand{\OCREntityQwenCI}{\textcolor{black}{54.90--62.89}}
\newcommand{\OCREntityQwenN}{\textcolor{black}{604}}
\newcommand{\OCRFindingQwen}{43.02}
\newcommand{\OCRFindingQwenCI}{40.63--45.42}
\newcommand{\OCRFindingQwenN}{997}
\newcommand{\OCRAssertionQwen}{42.90}
\newcommand{\OCRAssertionQwenCI}{40.61--45.13}
\newcommand{\OCRAssertionQwenN}{997}
\newcommand{\OCRCorrectFindingQwen}{36.57}
\newcommand{\OCRCorrectFindingQwenCI}{34.34--38.88}
\newcommand{\OCRCorrectFindingQwenN}{997}
\newcommand{\OCRContradictionQwen}{16.01}
\newcommand{\OCRContradictionQwenCI}{13.72--18.42}
\newcommand{\OCRContradictionQwenN}{707}
\newcommand{\OCRCoverageQwen}{\textcolor{black}{84.46}}
\newcommand{\OCRCoverageQwenCI}{\textcolor{black}{83.37--85.56}}
\newcommand{\OCRCoverageQwenN}{1200}
\newcommand{\OCRGradeQwen}{\textcolor{black}{24.13}}
\newcommand{\OCRGradeQwenCI}{\textcolor{black}{18.45--32.35}}
\newcommand{\OCRGradeQwenN}{\textcolor{black}{472}}
\newcommand{\OCRCancerDeepSeek}{\textcolor{black}{71.92}}
\newcommand{\OCRCancerDeepSeekCI}{\textcolor{black}{69.28--74.44}}
\newcommand{\OCRCancerDeepSeekN}{1200}
\newcommand{\OCRDiagnosticDeepSeek}{\textcolor{black}{84.50}}
\newcommand{\OCRDiagnosticDeepSeekCI}{\textcolor{black}{83.12--85.87}}
\newcommand{\OCRDiagnosticDeepSeekN}{1200}
\newcommand{\OCREntityDeepSeek}{\textcolor{black}{60.26}}
\newcommand{\OCREntityDeepSeekCI}{\textcolor{black}{56.24--64.19}}
\newcommand{\OCREntityDeepSeekN}{\textcolor{black}{604}}
\newcommand{\OCRFindingDeepSeek}{42.09}
\newcommand{\OCRFindingDeepSeekCI}{39.82--44.43}
\newcommand{\OCRFindingDeepSeekN}{997}
\newcommand{\OCRAssertionDeepSeek}{47.46}
\newcommand{\OCRAssertionDeepSeekCI}{45.08--49.76}
\newcommand{\OCRAssertionDeepSeekN}{997}
\newcommand{\OCRCorrectFindingDeepSeek}{36.14}
\newcommand{\OCRCorrectFindingDeepSeekCI}{33.98--38.40}
\newcommand{\OCRCorrectFindingDeepSeekN}{997}
\newcommand{\OCRContradictionDeepSeek}{23.18}
\newcommand{\OCRContradictionDeepSeekCI}{20.57--25.84}
\newcommand{\OCRContradictionDeepSeekN}{738}
\newcommand{\OCRCoverageDeepSeek}{\textcolor{black}{85.26}}
\newcommand{\OCRCoverageDeepSeekCI}{\textcolor{black}{83.97--86.51}}
\newcommand{\OCRCoverageDeepSeekN}{1200}
\newcommand{\OCRGradeDeepSeek}{\textcolor{black}{38.09}}
\newcommand{\OCRGradeDeepSeekCI}{\textcolor{black}{31.86--46.25}}
\newcommand{\OCRGradeDeepSeekN}{\textcolor{black}{472}}
\newcommand{\OCRCancerGLM}{\textcolor{black}{61.83}}
\newcommand{\OCRCancerGLMCI}{\textcolor{black}{59.02--64.59}}
\newcommand{\OCRCancerGLMN}{1200}
\newcommand{\OCRDiagnosticGLM}{\textcolor{black}{74.50}}
\newcommand{\OCRDiagnosticGLMCI}{\textcolor{black}{72.63--76.42}}
\newcommand{\OCRDiagnosticGLMN}{1200}
\newcommand{\OCREntityGLM}{\textcolor{black}{55.96}}
\newcommand{\OCREntityGLMCI}{\textcolor{black}{51.90--59.96}}
\newcommand{\OCREntityGLMN}{\textcolor{black}{604}}
\newcommand{\OCRFindingGLM}{44.16}
\newcommand{\OCRFindingGLMCI}{41.85--46.47}
\newcommand{\OCRFindingGLMN}{997}
\newcommand{\OCRAssertionGLM}{48.59}
\newcommand{\OCRAssertionGLMCI}{46.28--50.86}
\newcommand{\OCRAssertionGLMN}{997}
\newcommand{\OCRCorrectFindingGLM}{38.13}
\newcommand{\OCRCorrectFindingGLMCI}{35.86--40.42}
\newcommand{\OCRCorrectFindingGLMN}{997}
\newcommand{\OCRContradictionGLM}{22.55}
\newcommand{\OCRContradictionGLMCI}{19.96--25.22}
\newcommand{\OCRContradictionGLMN}{751}
\newcommand{\OCRCoverageGLM}{\textcolor{black}{78.95}}
\newcommand{\OCRCoverageGLMCI}{\textcolor{black}{77.42--80.48}}
\newcommand{\OCRCoverageGLMN}{1200}
\newcommand{\OCRGradeGLM}{\textcolor{black}{40.81}}
\newcommand{\OCRGradeGLMCI}{\textcolor{black}{32.04--50.84}}
\newcommand{\OCRGradeGLMN}{\textcolor{black}{472}}
\newcommand{\OCRCancerMedGemma}{\textcolor{black}{6.67}}
\newcommand{\OCRCancerMedGemmaCI}{\textcolor{black}{5.32--8.23}}
\newcommand{\OCRCancerMedGemmaN}{1200}
\newcommand{\OCRDiagnosticMedGemma}{\textcolor{black}{43.01}}
\newcommand{\OCRDiagnosticMedGemmaCI}{\textcolor{black}{41.35--44.65}}
\newcommand{\OCRDiagnosticMedGemmaN}{1200}
\newcommand{\OCREntityMedGemma}{\textcolor{black}{6.13}}
\newcommand{\OCREntityMedGemmaCI}{\textcolor{black}{4.35--8.35}}
\newcommand{\OCREntityMedGemmaN}{\textcolor{black}{604}}
\newcommand{\OCRFindingMedGemma}{37.44}
\newcommand{\OCRFindingMedGemmaCI}{35.12--39.70}
\newcommand{\OCRFindingMedGemmaN}{997}
\newcommand{\OCRAssertionMedGemma}{37.01}
\newcommand{\OCRAssertionMedGemmaCI}{34.71--39.34}
\newcommand{\OCRAssertionMedGemmaN}{997}
\newcommand{\OCRCorrectFindingMedGemma}{31.78}
\newcommand{\OCRCorrectFindingMedGemmaCI}{29.54--33.92}
\newcommand{\OCRCorrectFindingMedGemmaN}{997}
\newcommand{\OCRContradictionMedGemma}{14.41}
\newcommand{\OCRContradictionMedGemmaCI}{11.90--17.02}
\newcommand{\OCRContradictionMedGemmaN}{610}
\newcommand{\OCRCoverageMedGemma}{\textcolor{black}{67.62}}
\newcommand{\OCRCoverageMedGemmaCI}{\textcolor{black}{65.63--69.62}}
\newcommand{\OCRCoverageMedGemmaN}{1200}
\newcommand{\OCRGradeMedGemma}{\textcolor{black}{20.15}}
\newcommand{\OCRGradeMedGemmaCI}{\textcolor{black}{16.94--26.22}}
\newcommand{\OCRGradeMedGemmaN}{\textcolor{black}{472}}
\newcommand{\OCRCancerPathAgent}{\textcolor{black}{45.67}}
\newcommand{\OCRCancerPathAgentCI}{\textcolor{black}{42.82--48.53}}
\newcommand{\OCRCancerPathAgentN}{1200}
\newcommand{\OCRDiagnosticPathAgent}{\textcolor{black}{76.56}}
\newcommand{\OCRDiagnosticPathAgentCI}{\textcolor{black}{75.29--77.77}}
\newcommand{\OCRDiagnosticPathAgentN}{1200}
\newcommand{\OCREntityPathAgent}{\textcolor{black}{39.24}}
\newcommand{\OCREntityPathAgentCI}{\textcolor{black}{35.32--43.26}}
\newcommand{\OCREntityPathAgentN}{\textcolor{black}{604}}
\newcommand{\OCRFindingPathAgent}{21.89}
\newcommand{\OCRFindingPathAgentCI}{19.84--24.04}
\newcommand{\OCRFindingPathAgentN}{997}
\newcommand{\OCRAssertionPathAgent}{18.72}
\newcommand{\OCRAssertionPathAgentCI}{16.90--20.66}
\newcommand{\OCRAssertionPathAgentN}{997}
\newcommand{\OCRCorrectFindingPathAgent}{17.77}
\newcommand{\OCRCorrectFindingPathAgentCI}{15.97--19.65}
\newcommand{\OCRCorrectFindingPathAgentN}{997}
\newcommand{\OCRContradictionPathAgent}{6.27}
\newcommand{\OCRContradictionPathAgentCI}{3.99--8.83}
\newcommand{\OCRContradictionPathAgentN}{351}
\newcommand{\OCRCoveragePathAgent}{\textcolor{black}{85.34}}
\newcommand{\OCRCoveragePathAgentCI}{\textcolor{black}{84.43--86.26}}
\newcommand{\OCRCoveragePathAgentN}{1200}
\newcommand{\OCRGradePathAgent}{33.33}
\newcommand{\OCRGradePathAgentCI}{\textcolor{black}{32.62--36.33}}
\newcommand{\OCRGradePathAgentN}{\textcolor{black}{472}}
\newcommand{\OCRCancerPRISM}{53.25}
\newcommand{\OCRCancerPRISMCI}{\textcolor{black}{50.38--56.10}}
\newcommand{\OCRCancerPRISMN}{1200}
\newcommand{\OCRDiagnosticPRISM}{75.80}
\newcommand{\OCRDiagnosticPRISMCI}{74.16--77.38}
\newcommand{\OCRDiagnosticPRISMN}{1200}
\newcommand{\OCREntityPRISM}{\textcolor{black}{52.65}}
\newcommand{\OCREntityPRISMCI}{\textcolor{black}{48.58--56.69}}
\newcommand{\OCREntityPRISMN}{\textcolor{black}{604}}
\newcommand{\EvaluationCancerFull}{\textcolor{black}{72.50}}
\newcommand{\EvaluationCancerFullProse}{72.5}
\newcommand{\EvaluationCancerFullCI}{\textcolor{black}{69.88--75.01}}
\newcommand{\EvaluationCancerFullN}{1200}
\newcommand{\EvaluationDiagnosticFull}{\textcolor{black}{86.91}}
\newcommand{\EvaluationDiagnosticFullCI}{\textcolor{black}{85.81--88.02}}
\newcommand{\EvaluationDiagnosticFullN}{1200}
\newcommand{\EvaluationEntityFull}{\textcolor{black}{62.25}}
\newcommand{\EvaluationEntityFullProse}{62.3}
\newcommand{\EvaluationEntityFullCI}{\textcolor{black}{58.25--66.13}}
\newcommand{\EvaluationEntityFullN}{\textcolor{black}{604}}
\newcommand{\EvaluationFindingFull}{48.76}
\newcommand{\EvaluationFindingFullProse}{48.8}
\newcommand{\EvaluationFindingFullCI}{46.45--51.10}
\newcommand{\EvaluationFindingFullN}{997}
\newcommand{\EvaluationAssertionFull}{47.18}
\newcommand{\EvaluationAssertionFullCI}{44.88--49.48}
\newcommand{\EvaluationAssertionFullN}{997}
\newcommand{\EvaluationCorrectFindingFull}{41.64}
\newcommand{\EvaluationCorrectFindingFullProse}{41.6}
\newcommand{\EvaluationCorrectFindingFullCI}{39.31--43.93}
\newcommand{\EvaluationCorrectFindingFullN}{997}
\newcommand{\EvaluationContradictionFull}{11.53}
\newcommand{\EvaluationContradictionFullCI}{9.64--13.52}
\newcommand{\EvaluationContradictionFullN}{759}
\newcommand{\EvaluationCoverageFull}{\textcolor{black}{90.26}}
\newcommand{\EvaluationCoverageFullCI}{\textcolor{black}{89.34--91.15}}
\newcommand{\EvaluationCoverageFullN}{1200}
\newcommand{\EvaluationGradeFull}{\textcolor{black}{36.97}}
\newcommand{\EvaluationGradeFullCI}{\textcolor{black}{31.14--44.76}}
\newcommand{\EvaluationGradeFullN}{\textcolor{black}{472}}
\newcommand{\EvaluationCancerCore}{\textcolor{black}{68.17}}
\newcommand{\EvaluationCancerCoreCI}{\textcolor{black}{65.45--70.80}}
\newcommand{\EvaluationCancerCoreN}{1200}
\newcommand{\EvaluationDiagnosticCore}{\textcolor{black}{84.67}}
\newcommand{\EvaluationDiagnosticCoreCI}{\textcolor{black}{83.44--85.89}}
\newcommand{\EvaluationDiagnosticCoreN}{1200}
\newcommand{\EvaluationEntityCore}{\textcolor{black}{58.11}}
\newcommand{\EvaluationEntityCoreCI}{\textcolor{black}{54.06--62.08}}
\newcommand{\EvaluationEntityCoreN}{\textcolor{black}{604}}
\newcommand{\EvaluationFindingCore}{46.70}
\newcommand{\EvaluationFindingCoreCI}{44.35--49.01}
\newcommand{\EvaluationFindingCoreN}{997}
\newcommand{\EvaluationAssertionCore}{43.97}
\newcommand{\EvaluationAssertionCoreCI}{41.70--46.24}
\newcommand{\EvaluationAssertionCoreN}{997}
\newcommand{\EvaluationCorrectFindingCore}{39.40}
\newcommand{\EvaluationCorrectFindingCoreCI}{37.24--41.62}
\newcommand{\EvaluationCorrectFindingCoreN}{997}
\newcommand{\EvaluationContradictionCore}{9.82}
\newcommand{\EvaluationContradictionCoreCI}{7.96--11.77}
\newcommand{\EvaluationContradictionCoreN}{723}
\newcommand{\EvaluationCoverageCore}{\textcolor{black}{86.64}}
\newcommand{\EvaluationCoverageCoreCI}{\textcolor{black}{85.62--87.63}}
\newcommand{\EvaluationCoverageCoreN}{1200}
\newcommand{\EvaluationGradeCore}{\textcolor{black}{30.25}}
\newcommand{\EvaluationGradeCoreCI}{\textcolor{black}{24.68--38.10}}
\newcommand{\EvaluationGradeCoreN}{\textcolor{black}{472}}
\newcommand{\EvaluationCancerQwen}{\textcolor{black}{68.08}}
\newcommand{\EvaluationCancerQwenCI}{\textcolor{black}{65.36--70.72}}
\newcommand{\EvaluationCancerQwenN}{1200}
\newcommand{\EvaluationDiagnosticQwen}{\textcolor{black}{84.78}}
\newcommand{\EvaluationDiagnosticQwenCI}{\textcolor{black}{83.53--86.01}}
\newcommand{\EvaluationDiagnosticQwenN}{1200}
\newcommand{\EvaluationEntityQwen}{\textcolor{black}{58.94}}
\newcommand{\EvaluationEntityQwenCI}{\textcolor{black}{54.90--62.89}}
\newcommand{\EvaluationEntityQwenN}{\textcolor{black}{604}}
\newcommand{\EvaluationFindingQwen}{43.02}
\newcommand{\EvaluationFindingQwenCI}{40.63--45.42}
\newcommand{\EvaluationFindingQwenN}{997}
\newcommand{\EvaluationAssertionQwen}{42.90}
\newcommand{\EvaluationAssertionQwenCI}{40.61--45.13}
\newcommand{\EvaluationAssertionQwenN}{997}
\newcommand{\EvaluationCorrectFindingQwen}{36.57}
\newcommand{\EvaluationCorrectFindingQwenProse}{36.6}
\newcommand{\EvaluationCorrectFindingQwenCI}{34.34--38.88}
\newcommand{\EvaluationCorrectFindingQwenN}{997}
\newcommand{\EvaluationContradictionQwen}{16.01}
\newcommand{\EvaluationContradictionQwenCI}{13.72--18.42}
\newcommand{\EvaluationContradictionQwenN}{707}
\newcommand{\EvaluationCoverageQwen}{\textcolor{black}{84.46}}
\newcommand{\EvaluationCoverageQwenCI}{\textcolor{black}{83.37--85.56}}
\newcommand{\EvaluationCoverageQwenN}{1200}
\newcommand{\EvaluationGradeQwen}{\textcolor{black}{24.13}}
\newcommand{\EvaluationGradeQwenCI}{\textcolor{black}{18.45--32.35}}
\newcommand{\EvaluationGradeQwenN}{\textcolor{black}{472}}
\newcommand{\EvaluationCancerDeepSeek}{\textcolor{black}{71.92}}
\newcommand{\EvaluationCancerDeepSeekCI}{\textcolor{black}{69.28--74.44}}
\newcommand{\EvaluationCancerDeepSeekN}{1200}
\newcommand{\EvaluationDiagnosticDeepSeek}{\textcolor{black}{84.50}}
\newcommand{\EvaluationDiagnosticDeepSeekCI}{\textcolor{black}{83.12--85.87}}
\newcommand{\EvaluationDiagnosticDeepSeekN}{1200}
\newcommand{\EvaluationEntityDeepSeek}{\textcolor{black}{60.26}}
\newcommand{\EvaluationEntityDeepSeekCI}{\textcolor{black}{56.24--64.19}}
\newcommand{\EvaluationEntityDeepSeekN}{\textcolor{black}{604}}
\newcommand{\EvaluationFindingDeepSeek}{42.09}
\newcommand{\EvaluationFindingDeepSeekCI}{39.82--44.43}
\newcommand{\EvaluationFindingDeepSeekN}{997}
\newcommand{\EvaluationAssertionDeepSeek}{47.46}
\newcommand{\EvaluationAssertionDeepSeekCI}{45.08--49.76}
\newcommand{\EvaluationAssertionDeepSeekN}{997}
\newcommand{\EvaluationCorrectFindingDeepSeek}{36.14}
\newcommand{\EvaluationCorrectFindingDeepSeekCI}{33.98--38.40}
\newcommand{\EvaluationCorrectFindingDeepSeekN}{997}
\newcommand{\EvaluationContradictionDeepSeek}{23.18}
\newcommand{\EvaluationContradictionDeepSeekCI}{20.57--25.84}
\newcommand{\EvaluationContradictionDeepSeekN}{738}
\newcommand{\EvaluationCoverageDeepSeek}{\textcolor{black}{85.26}}
\newcommand{\EvaluationCoverageDeepSeekCI}{\textcolor{black}{83.97--86.51}}
\newcommand{\EvaluationCoverageDeepSeekN}{1200}
\newcommand{\EvaluationGradeDeepSeek}{\textcolor{black}{38.09}}
\newcommand{\EvaluationGradeDeepSeekCI}{\textcolor{black}{31.86--46.25}}
\newcommand{\EvaluationGradeDeepSeekN}{\textcolor{black}{472}}
\newcommand{\EvaluationCancerGLM}{\textcolor{black}{61.83}}
\newcommand{\EvaluationCancerGLMCI}{\textcolor{black}{59.02--64.59}}
\newcommand{\EvaluationCancerGLMN}{1200}
\newcommand{\EvaluationDiagnosticGLM}{\textcolor{black}{74.50}}
\newcommand{\EvaluationDiagnosticGLMCI}{\textcolor{black}{72.63--76.42}}
\newcommand{\EvaluationDiagnosticGLMN}{1200}
\newcommand{\EvaluationEntityGLM}{\textcolor{black}{55.96}}
\newcommand{\EvaluationEntityGLMCI}{\textcolor{black}{51.90--59.96}}
\newcommand{\EvaluationEntityGLMN}{\textcolor{black}{604}}
\newcommand{\EvaluationFindingGLM}{44.16}
\newcommand{\EvaluationFindingGLMCI}{41.85--46.47}
\newcommand{\EvaluationFindingGLMN}{997}
\newcommand{\EvaluationAssertionGLM}{48.59}
\newcommand{\EvaluationAssertionGLMCI}{46.28--50.86}
\newcommand{\EvaluationAssertionGLMN}{997}
\newcommand{\EvaluationCorrectFindingGLM}{38.13}
\newcommand{\EvaluationCorrectFindingGLMCI}{35.86--40.42}
\newcommand{\EvaluationCorrectFindingGLMN}{997}
\newcommand{\EvaluationContradictionGLM}{22.55}
\newcommand{\EvaluationContradictionGLMCI}{19.96--25.22}
\newcommand{\EvaluationContradictionGLMN}{751}
\newcommand{\EvaluationCoverageGLM}{\textcolor{black}{78.95}}
\newcommand{\EvaluationCoverageGLMCI}{\textcolor{black}{77.42--80.48}}
\newcommand{\EvaluationCoverageGLMN}{1200}
\newcommand{\EvaluationGradeGLM}{\textcolor{black}{40.81}}
\newcommand{\EvaluationGradeGLMCI}{\textcolor{black}{32.04--50.84}}
\newcommand{\EvaluationGradeGLMN}{\textcolor{black}{472}}
\newcommand{\EvaluationCancerMedGemma}{\textcolor{black}{6.67}}
\newcommand{\EvaluationCancerMedGemmaCI}{\textcolor{black}{5.32--8.23}}
\newcommand{\EvaluationCancerMedGemmaN}{1200}
\newcommand{\EvaluationDiagnosticMedGemma}{\textcolor{black}{43.01}}
\newcommand{\EvaluationDiagnosticMedGemmaCI}{\textcolor{black}{41.35--44.65}}
\newcommand{\EvaluationDiagnosticMedGemmaN}{1200}
\newcommand{\EvaluationEntityMedGemma}{\textcolor{black}{6.13}}
\newcommand{\EvaluationEntityMedGemmaCI}{\textcolor{black}{4.35--8.35}}
\newcommand{\EvaluationEntityMedGemmaN}{\textcolor{black}{604}}
\newcommand{\EvaluationFindingMedGemma}{37.44}
\newcommand{\EvaluationFindingMedGemmaCI}{35.12--39.70}
\newcommand{\EvaluationFindingMedGemmaN}{997}
\newcommand{\EvaluationAssertionMedGemma}{37.01}
\newcommand{\EvaluationAssertionMedGemmaCI}{34.71--39.34}
\newcommand{\EvaluationAssertionMedGemmaN}{997}
\newcommand{\EvaluationCorrectFindingMedGemma}{31.78}
\newcommand{\EvaluationCorrectFindingMedGemmaCI}{29.54--33.92}
\newcommand{\EvaluationCorrectFindingMedGemmaN}{997}
\newcommand{\EvaluationContradictionMedGemma}{14.41}
\newcommand{\EvaluationContradictionMedGemmaCI}{11.90--17.02}
\newcommand{\EvaluationContradictionMedGemmaN}{610}
\newcommand{\EvaluationCoverageMedGemma}{\textcolor{black}{67.62}}
\newcommand{\EvaluationCoverageMedGemmaCI}{\textcolor{black}{65.63--69.62}}
\newcommand{\EvaluationCoverageMedGemmaN}{1200}
\newcommand{\EvaluationGradeMedGemma}{\textcolor{black}{20.15}}
\newcommand{\EvaluationGradeMedGemmaCI}{\textcolor{black}{16.94--26.22}}
\newcommand{\EvaluationGradeMedGemmaN}{\textcolor{black}{472}}
\newcommand{\EvaluationCancerPathAgent}{\textcolor{black}{45.67}}
\newcommand{\EvaluationCancerPathAgentCI}{\textcolor{black}{42.82--48.53}}
\newcommand{\EvaluationCancerPathAgentN}{1200}
\newcommand{\EvaluationDiagnosticPathAgent}{\textcolor{black}{76.56}}
\newcommand{\EvaluationDiagnosticPathAgentCI}{\textcolor{black}{75.29--77.77}}
\newcommand{\EvaluationDiagnosticPathAgentN}{1200}
\newcommand{\EvaluationEntityPathAgent}{\textcolor{black}{39.24}}
\newcommand{\EvaluationEntityPathAgentCI}{\textcolor{black}{35.32--43.26}}
\newcommand{\EvaluationEntityPathAgentN}{\textcolor{black}{604}}
\newcommand{\EvaluationFindingPathAgent}{21.89}
\newcommand{\EvaluationFindingPathAgentCI}{19.84--24.04}
\newcommand{\EvaluationFindingPathAgentN}{997}
\newcommand{\EvaluationAssertionPathAgent}{18.72}
\newcommand{\EvaluationAssertionPathAgentCI}{16.90--20.66}
\newcommand{\EvaluationAssertionPathAgentN}{997}
\newcommand{\EvaluationCorrectFindingPathAgent}{17.77}
\newcommand{\EvaluationCorrectFindingPathAgentCI}{15.97--19.65}
\newcommand{\EvaluationCorrectFindingPathAgentN}{997}
\newcommand{\EvaluationContradictionPathAgent}{6.27}
\newcommand{\EvaluationContradictionPathAgentCI}{3.99--8.83}
\newcommand{\EvaluationContradictionPathAgentN}{351}
\newcommand{\EvaluationCoveragePathAgent}{\textcolor{black}{85.34}}
\newcommand{\EvaluationCoveragePathAgentCI}{\textcolor{black}{84.43--86.26}}
\newcommand{\EvaluationCoveragePathAgentN}{1200}
\newcommand{\EvaluationGradePathAgent}{33.33}
\newcommand{\EvaluationGradePathAgentCI}{\textcolor{black}{32.62--36.33}}
\newcommand{\EvaluationGradePathAgentN}{\textcolor{black}{472}}
\newcommand{\EvaluationCancerPRISM}{53.25}
\newcommand{\EvaluationCancerPRISMCI}{\textcolor{black}{50.38--56.10}}
\newcommand{\EvaluationCancerPRISMN}{1200}
\newcommand{\EvaluationDiagnosticPRISM}{75.80}
\newcommand{\EvaluationDiagnosticPRISMCI}{74.16--77.38}
\newcommand{\EvaluationDiagnosticPRISMN}{1200}
\newcommand{\EvaluationEntityPRISM}{\textcolor{black}{52.65}}
\newcommand{\EvaluationEntityPRISMCI}{\textcolor{black}{48.58--56.69}}
\newcommand{\EvaluationEntityPRISMN}{\textcolor{black}{604}}
\newcommand{\PublicCancerFull}{\textcolor{black}{73.67}}
\newcommand{\PublicCancerFullCI}{\textcolor{black}{71.48--75.78}}
\newcommand{\PublicCancerFullN}{1656}
\newcommand{\PublicDiagnosticFull}{\textcolor{black}{86.63}}
\newcommand{\PublicDiagnosticFullCI}{\textcolor{black}{85.66--87.65}}
\newcommand{\PublicDiagnosticFullN}{1656}
\newcommand{\PublicFindingFull}{47.82}
\newcommand{\PublicFindingFullCI}{45.94--49.77}
\newcommand{\PublicFindingFullN}{1405}
\newcommand{\PublicAssertionFull}{46.56}
\newcommand{\PublicAssertionFullCI}{44.63--48.44}
\newcommand{\PublicAssertionFullN}{1405}
\newcommand{\PublicCorrectFindingFull}{40.52}
\newcommand{\PublicCorrectFindingFullCI}{38.62--42.42}
\newcommand{\PublicCorrectFindingFullN}{1405}
\newcommand{\PublicContradictionFull}{12.69}
\newcommand{\PublicContradictionFullCI}{10.98--14.53}
\newcommand{\PublicContradictionFullN}{1070}
\newcommand{\PublicCoverageFull}{\textcolor{black}{90.57}}
\newcommand{\PublicCoverageFullCI}{\textcolor{black}{89.81--91.32}}
\newcommand{\PublicCoverageFullN}{1656}
\newcommand{\PublicGradeFull}{\textcolor{black}{38.48}}
\newcommand{\PublicGradeFullCI}{\textcolor{black}{33.57--45.13}}
\newcommand{\PublicGradeFullN}{\textcolor{black}{640}}
\newcommand{\PublicCancerCore}{\textcolor{black}{69.69}}
\newcommand{\PublicCancerCoreCI}{\textcolor{black}{67.41--71.89}}
\newcommand{\PublicCancerCoreN}{1656}
\newcommand{\PublicDiagnosticCore}{\textcolor{black}{84.36}}
\newcommand{\PublicDiagnosticCoreCI}{\textcolor{black}{83.26--85.52}}
\newcommand{\PublicDiagnosticCoreN}{1656}
\newcommand{\PublicFindingCore}{45.22}
\newcommand{\PublicFindingCoreCI}{43.25--47.20}
\newcommand{\PublicFindingCoreN}{1405}
\newcommand{\PublicAssertionCore}{42.93}
\newcommand{\PublicAssertionCoreCI}{41.09--44.79}
\newcommand{\PublicAssertionCoreN}{1405}
\newcommand{\PublicCorrectFindingCore}{37.91}
\newcommand{\PublicCorrectFindingCoreCI}{36.04--39.71}
\newcommand{\PublicCorrectFindingCoreN}{1405}
\newcommand{\PublicContradictionCore}{11.11}
\newcommand{\PublicContradictionCoreCI}{9.41--12.85}
\newcommand{\PublicContradictionCoreN}{1006}
\newcommand{\PublicCoverageCore}{\textcolor{black}{86.79}}
\newcommand{\PublicCoverageCoreCI}{\textcolor{black}{85.91--87.62}}
\newcommand{\PublicCoverageCoreN}{1656}
\newcommand{\PublicGradeCore}{\textcolor{black}{30.01}}
\newcommand{\PublicGradeCoreCI}{\textcolor{black}{24.86--37.21}}
\newcommand{\PublicGradeCoreN}{\textcolor{black}{640}}
\newcommand{\PublicCancerQwen}{\textcolor{black}{69.02}}
\newcommand{\PublicCancerQwenCI}{\textcolor{black}{66.73--71.24}}
\newcommand{\PublicCancerQwenN}{1656}
\newcommand{\PublicDiagnosticQwen}{\textcolor{black}{84.11}}
\newcommand{\PublicDiagnosticQwenCI}{\textcolor{black}{82.95--85.24}}
\newcommand{\PublicDiagnosticQwenN}{1656}
\newcommand{\PublicFindingQwen}{42.33}
\newcommand{\PublicFindingQwenCI}{40.41--44.32}
\newcommand{\PublicFindingQwenN}{1405}
\newcommand{\PublicAssertionQwen}{42.46}
\newcommand{\PublicAssertionQwenCI}{40.58--44.32}
\newcommand{\PublicAssertionQwenN}{1405}
\newcommand{\PublicCorrectFindingQwen}{35.69}
\newcommand{\PublicCorrectFindingQwenCI}{33.92--37.56}
\newcommand{\PublicCorrectFindingQwenN}{1405}
\newcommand{\PublicContradictionQwen}{16.99}
\newcommand{\PublicContradictionQwenCI}{15.03--18.98}
\newcommand{\PublicContradictionQwenN}{1004}
\newcommand{\PublicCoverageQwen}{\textcolor{black}{84.54}}
\newcommand{\PublicCoverageQwenCI}{\textcolor{black}{83.59--85.45}}
\newcommand{\PublicCoverageQwenN}{1656}
\newcommand{\PublicGradeQwen}{\textcolor{black}{23.47}}
\newcommand{\PublicGradeQwenCI}{\textcolor{black}{18.28--30.83}}
\newcommand{\PublicGradeQwenN}{\textcolor{black}{640}}
\newcommand{\PublicCancerDeepSeek}{\textcolor{black}{71.56}}
\newcommand{\PublicCancerDeepSeekCI}{\textcolor{black}{69.32--73.72}}
\newcommand{\PublicCancerDeepSeekN}{1656}
\newcommand{\PublicDiagnosticDeepSeek}{\textcolor{black}{83.93}}
\newcommand{\PublicDiagnosticDeepSeekCI}{\textcolor{black}{82.75--85.18}}
\newcommand{\PublicDiagnosticDeepSeekN}{1656}
\newcommand{\PublicFindingDeepSeek}{39.72}
\newcommand{\PublicFindingDeepSeekCI}{37.79--41.77}
\newcommand{\PublicFindingDeepSeekN}{1405}
\newcommand{\PublicAssertionDeepSeek}{44.79}
\newcommand{\PublicAssertionDeepSeekCI}{42.80--46.70}
\newcommand{\PublicAssertionDeepSeekN}{1405}
\newcommand{\PublicCorrectFindingDeepSeek}{33.85}
\newcommand{\PublicCorrectFindingDeepSeekCI}{32.07--35.64}
\newcommand{\PublicCorrectFindingDeepSeekN}{1405}
\newcommand{\PublicContradictionDeepSeek}{24.12}
\newcommand{\PublicContradictionDeepSeekCI}{21.78--26.49}
\newcommand{\PublicContradictionDeepSeekN}{1006}
\newcommand{\PublicCoverageDeepSeek}{\textcolor{black}{85.23}}
\newcommand{\PublicCoverageDeepSeekCI}{\textcolor{black}{84.16--86.29}}
\newcommand{\PublicCoverageDeepSeekN}{1656}
\newcommand{\PublicGradeDeepSeek}{\textcolor{black}{39.94}}
\newcommand{\PublicGradeDeepSeekCI}{\textcolor{black}{34.02--47.47}}
\newcommand{\PublicGradeDeepSeekN}{\textcolor{black}{640}}
\newcommand{\PublicCancerGLM}{\textcolor{black}{60.21}}
\newcommand{\PublicCancerGLMCI}{\textcolor{black}{57.80--62.57}}
\newcommand{\PublicCancerGLMN}{1656}
\newcommand{\PublicDiagnosticGLM}{\textcolor{black}{73.15}}
\newcommand{\PublicDiagnosticGLMCI}{\textcolor{black}{71.50--74.83}}
\newcommand{\PublicDiagnosticGLMN}{1656}
\newcommand{\PublicFindingGLM}{43.20}
\newcommand{\PublicFindingGLMCI}{41.35--45.15}
\newcommand{\PublicFindingGLMN}{1405}
\newcommand{\PublicAssertionGLM}{48.04}
\newcommand{\PublicAssertionGLMCI}{46.10--49.92}
\newcommand{\PublicAssertionGLMN}{1405}
\newcommand{\PublicCorrectFindingGLM}{37.10}
\newcommand{\PublicCorrectFindingGLMCI}{35.26--38.96}
\newcommand{\PublicCorrectFindingGLMN}{1405}
\newcommand{\PublicContradictionGLM}{23.12}
\newcommand{\PublicContradictionGLMCI}{20.87--25.30}
\newcommand{\PublicContradictionGLMN}{1053}
\newcommand{\PublicCoverageGLM}{\textcolor{black}{79.23}}
\newcommand{\PublicCoverageGLMCI}{\textcolor{black}{77.95--80.46}}
\newcommand{\PublicCoverageGLMN}{1656}
\newcommand{\PublicGradeGLM}{\textcolor{black}{40.85}}
\newcommand{\PublicGradeGLMCI}{\textcolor{black}{33.14--49.71}}
\newcommand{\PublicGradeGLMN}{\textcolor{black}{640}}
\newcommand{\PublicCancerMedGemma}{\textcolor{black}{6.94}}
\newcommand{\PublicCancerMedGemmaCI}{\textcolor{black}{5.77--8.28}}
\newcommand{\PublicCancerMedGemmaN}{1656}
\newcommand{\PublicDiagnosticMedGemma}{\textcolor{black}{41.95}}
\newcommand{\PublicDiagnosticMedGemmaCI}{\textcolor{black}{40.51--43.47}}
\newcommand{\PublicDiagnosticMedGemmaN}{1656}
\newcommand{\PublicFindingMedGemma}{35.57}
\newcommand{\PublicFindingMedGemmaCI}{33.63--37.53}
\newcommand{\PublicFindingMedGemmaN}{1405}
\newcommand{\PublicAssertionMedGemma}{35.29}
\newcommand{\PublicAssertionMedGemmaCI}{33.41--37.20}
\newcommand{\PublicAssertionMedGemmaN}{1405}
\newcommand{\PublicCorrectFindingMedGemma}{29.90}
\newcommand{\PublicCorrectFindingMedGemmaCI}{28.05--31.78}
\newcommand{\PublicCorrectFindingMedGemmaN}{1405}
\newcommand{\PublicContradictionMedGemma}{15.32}
\newcommand{\PublicContradictionMedGemmaCI}{13.05--17.55}
\newcommand{\PublicContradictionMedGemmaN}{839}
\newcommand{\PublicCoverageMedGemma}{\textcolor{black}{67.82}}
\newcommand{\PublicCoverageMedGemmaCI}{\textcolor{black}{66.12--69.51}}
\newcommand{\PublicCoverageMedGemmaN}{1656}
\newcommand{\PublicGradeMedGemma}{\textcolor{black}{19.85}}
\newcommand{\PublicGradeMedGemmaCI}{\textcolor{black}{17.02--25.17}}
\newcommand{\PublicGradeMedGemmaN}{\textcolor{black}{640}}
\newcommand{\PublicCancerPathAgent}{\textcolor{black}{41.30}}
\newcommand{\PublicCancerPathAgentCI}{\textcolor{black}{38.92--43.72}}
\newcommand{\PublicCancerPathAgentN}{1656}
\newcommand{\PublicDiagnosticPathAgent}{\textcolor{black}{70.69}}
\newcommand{\PublicDiagnosticPathAgentCI}{\textcolor{black}{69.34--72.07}}
\newcommand{\PublicDiagnosticPathAgentN}{1656}
\newcommand{\PublicFindingPathAgent}{20.12}
\newcommand{\PublicFindingPathAgentCI}{18.55--21.76}
\newcommand{\PublicFindingPathAgentN}{1405}
\newcommand{\PublicAssertionPathAgent}{17.12}
\newcommand{\PublicAssertionPathAgentCI}{15.65--18.72}
\newcommand{\PublicAssertionPathAgentN}{1405}
\newcommand{\PublicCorrectFindingPathAgent}{16.12}
\newcommand{\PublicCorrectFindingPathAgentCI}{14.74--17.63}
\newcommand{\PublicCorrectFindingPathAgentN}{1405}
\newcommand{\PublicContradictionPathAgent}{7.20}
\newcommand{\PublicContradictionPathAgentCI}{5.08--9.43}
\newcommand{\PublicContradictionPathAgentN}{472}
\newcommand{\PublicCoveragePathAgent}{\textcolor{black}{81.74}}
\newcommand{\PublicCoveragePathAgentCI}{\textcolor{black}{80.71--82.74}}
\newcommand{\PublicCoveragePathAgentN}{1656}
\newcommand{\PublicGradePathAgent}{\textcolor{black}{31.86}}
\newcommand{\PublicGradePathAgentCI}{\textcolor{black}{30.66--35.19}}
\newcommand{\PublicGradePathAgentN}{\textcolor{black}{640}}
\newcommand{\PublicCancerPRISM}{\textcolor{black}{53.99}}
\newcommand{\PublicCancerPRISMCI}{\textcolor{black}{51.55--56.41}}
\newcommand{\PublicCancerPRISMN}{1656}
\newcommand{\PublicDiagnosticPRISM}{\textcolor{black}{74.93}}
\newcommand{\PublicDiagnosticPRISMCI}{\textcolor{black}{73.51--76.42}}
\newcommand{\PublicDiagnosticPRISMN}{1656}
\newcommand{\PrivateOneEntityFull}{\textcolor{black}{67.41}}
\newcommand{\PrivateOneEntityFullProse}{67.4}
\newcommand{\PrivateOneEntityFullCI}{\textcolor{black}{58.81--75.22}}
\newcommand{\PrivateOneEntityFullN}{135}
\newcommand{\PrivateOneFindingFull}{76.56}
\newcommand{\PrivateOneFindingFullProse}{76.6}
\newcommand{\PrivateOneFindingFullCI}{71.22--81.56}
\newcommand{\PrivateOneFindingFullN}{150}
\newcommand{\PrivateOneAssertionFull}{92.67}
\newcommand{\PrivateOneAssertionFullCI}{89.33--95.67}
\newcommand{\PrivateOneAssertionFullN}{150}
\newcommand{\PrivateOneCorrectFindingFull}{75.00}
\newcommand{\PrivateOneCorrectFindingFullCI}{69.67--80.33}
\newcommand{\PrivateOneCorrectFindingFullN}{150}
\newcommand{\PrivateOneContradictionFull}{18.71}
\newcommand{\PrivateOneContradictionFullCI}{13.95--24.15}
\newcommand{\PrivateOneContradictionFullN}{147}
\newcommand{\PrivateOneCoverageFull}{\textcolor{black}{94.35}}
\newcommand{\PrivateOneCoverageFullCI}{\textcolor{black}{92.48--96.06}}
\newcommand{\PrivateOneCoverageFullN}{151}
\newcommand{\PrivateOneGradeFull}{\textcolor{black}{27.43}}
\newcommand{\PrivateOneGradeFullCI}{\textcolor{black}{15.82--72.40}}
\newcommand{\PrivateOneGradeFullN}{76}
\newcommand{\PrivateOneEntityCore}{\textcolor{black}{57.78}}
\newcommand{\PrivateOneEntityCoreCI}{\textcolor{black}{48.98--66.22}}
\newcommand{\PrivateOneEntityCoreN}{135}
\newcommand{\PrivateOneFindingCore}{67.33}
\newcommand{\PrivateOneFindingCoreCI}{62.00--72.67}
\newcommand{\PrivateOneFindingCoreN}{150}
\newcommand{\PrivateOneAssertionCore}{84.33}
\newcommand{\PrivateOneAssertionCoreCI}{80.00--88.33}
\newcommand{\PrivateOneAssertionCoreN}{150}
\newcommand{\PrivateOneCorrectFindingCore}{64.00}
\newcommand{\PrivateOneCorrectFindingCoreCI}{58.33--69.33}
\newcommand{\PrivateOneCorrectFindingCoreN}{150}
\newcommand{\PrivateOneContradictionCore}{24.49}
\newcommand{\PrivateOneContradictionCoreCI}{18.71--30.27}
\newcommand{\PrivateOneContradictionCoreN}{147}
\newcommand{\PrivateOneCoverageCore}{\textcolor{black}{87.45}}
\newcommand{\PrivateOneCoverageCoreCI}{\textcolor{black}{85.39--89.50}}
\newcommand{\PrivateOneCoverageCoreN}{151}
\newcommand{\PrivateOneGradeCore}{17.07}
\newcommand{\PrivateOneGradeCoreCI}{\textcolor{black}{6.49--62.30}}
\newcommand{\PrivateOneGradeCoreN}{76}
\newcommand{\PrivateOneEntityQwen}{54.07}
\newcommand{\PrivateOneEntityQwenCI}{\textcolor{black}{45.29--62.68}}
\newcommand{\PrivateOneEntityQwenN}{135}
\newcommand{\PrivateOneFindingQwen}{71.56}
\newcommand{\PrivateOneFindingQwenCI}{65.89--77.00}
\newcommand{\PrivateOneFindingQwenN}{150}
\newcommand{\PrivateOneAssertionQwen}{79.00}
\newcommand{\PrivateOneAssertionQwenCI}{73.67--84.00}
\newcommand{\PrivateOneAssertionQwenN}{150}
\newcommand{\PrivateOneCorrectFindingQwen}{68.00}
\newcommand{\PrivateOneCorrectFindingQwenCI}{62.33--73.67}
\newcommand{\PrivateOneCorrectFindingQwenN}{150}
\newcommand{\PrivateOneContradictionQwen}{14.49}
\newcommand{\PrivateOneContradictionQwenCI}{9.78--19.57}
\newcommand{\PrivateOneContradictionQwenN}{138}
\newcommand{\PrivateOneCoverageQwen}{83.73}
\newcommand{\PrivateOneCoverageQwenCI}{81.16--86.32}
\newcommand{\PrivateOneCoverageQwenN}{151}
\newcommand{\PrivateOneGradeQwen}{12.97}
\newcommand{\PrivateOneGradeQwenCI}{\textcolor{black}{3.93--57.75}}
\newcommand{\PrivateOneGradeQwenN}{76}
\newcommand{\PrivateOneEntityDeepSeek}{60.00}
\newcommand{\PrivateOneEntityDeepSeekCI}{\textcolor{black}{51.22--68.33}}
\newcommand{\PrivateOneEntityDeepSeekN}{135}
\newcommand{\PrivateOneFindingDeepSeek}{66.44}
\newcommand{\PrivateOneFindingDeepSeekCI}{61.22--71.34}
\newcommand{\PrivateOneFindingDeepSeekN}{150}
\newcommand{\PrivateOneAssertionDeepSeek}{78.67}
\newcommand{\PrivateOneAssertionDeepSeekCI}{74.00--83.33}
\newcommand{\PrivateOneAssertionDeepSeekN}{150}
\newcommand{\PrivateOneCorrectFindingDeepSeek}{61.67}
\newcommand{\PrivateOneCorrectFindingDeepSeekCI}{56.33--66.67}
\newcommand{\PrivateOneCorrectFindingDeepSeekN}{150}
\newcommand{\PrivateOneContradictionDeepSeek}{19.72}
\newcommand{\PrivateOneContradictionDeepSeekCI}{14.79--24.65}
\newcommand{\PrivateOneContradictionDeepSeekN}{142}
\newcommand{\PrivateOneCoverageDeepSeek}{\textcolor{black}{84.88}}
\newcommand{\PrivateOneCoverageDeepSeekCI}{\textcolor{black}{81.89--87.81}}
\newcommand{\PrivateOneCoverageDeepSeekN}{151}
\newcommand{\PrivateOneGradeDeepSeek}{\textcolor{black}{19.32}}
\newcommand{\PrivateOneGradeDeepSeekCI}{\textcolor{black}{7.98--63.82}}
\newcommand{\PrivateOneGradeDeepSeekN}{76}
\newcommand{\PrivateOneEntityGLM}{\textcolor{black}{42.96}}
\newcommand{\PrivateOneEntityGLMCI}{\textcolor{black}{34.48--51.76}}
\newcommand{\PrivateOneEntityGLMN}{135}
\newcommand{\PrivateOneFindingGLM}{63.11}
\newcommand{\PrivateOneFindingGLMCI}{57.11--68.78}
\newcommand{\PrivateOneFindingGLMN}{150}
\newcommand{\PrivateOneAssertionGLM}{63.33}
\newcommand{\PrivateOneAssertionGLMCI}{57.67--69.00}
\newcommand{\PrivateOneAssertionGLMN}{150}
\newcommand{\PrivateOneCorrectFindingGLM}{56.67}
\newcommand{\PrivateOneCorrectFindingGLMCI}{51.00--62.67}
\newcommand{\PrivateOneCorrectFindingGLMN}{150}
\newcommand{\PrivateOneContradictionGLM}{10.94}
\newcommand{\PrivateOneContradictionGLMCI}{6.25--16.02}
\newcommand{\PrivateOneContradictionGLMN}{128}
\newcommand{\PrivateOneCoverageGLM}{\textcolor{black}{75.28}}
\newcommand{\PrivateOneCoverageGLMCI}{\textcolor{black}{71.25--79.25}}
\newcommand{\PrivateOneCoverageGLMN}{151}
\newcommand{\PrivateOneGradeGLM}{\textcolor{black}{23.06}}
\newcommand{\PrivateOneGradeGLMCI}{\textcolor{black}{10.88--68.52}}
\newcommand{\PrivateOneGradeGLMN}{76}
\newcommand{\PrivateOneEntityMedGemma}{2.96}
\newcommand{\PrivateOneEntityMedGemmaCI}{\textcolor{black}{0.81--7.41}}
\newcommand{\PrivateOneEntityMedGemmaN}{135}
\newcommand{\PrivateOneFindingMedGemma}{38.44}
\newcommand{\PrivateOneFindingMedGemmaCI}{33.00--43.89}
\newcommand{\PrivateOneFindingMedGemmaN}{150}
\newcommand{\PrivateOneAssertionMedGemma}{63.00}
\newcommand{\PrivateOneAssertionMedGemmaCI}{56.00--69.33}
\newcommand{\PrivateOneAssertionMedGemmaN}{150}
\newcommand{\PrivateOneCorrectFindingMedGemma}{35.67}
\newcommand{\PrivateOneCorrectFindingMedGemmaCI}{30.33--41.00}
\newcommand{\PrivateOneCorrectFindingMedGemmaN}{150}
\newcommand{\PrivateOneContradictionMedGemma}{41.59}
\newcommand{\PrivateOneContradictionMedGemmaCI}{34.96--48.23}
\newcommand{\PrivateOneContradictionMedGemmaN}{113}
\newcommand{\PrivateOneCoverageMedGemma}{\textcolor{black}{68.37}}
\newcommand{\PrivateOneCoverageMedGemmaCI}{\textcolor{black}{63.53--72.88}}
\newcommand{\PrivateOneCoverageMedGemmaN}{151}
\newcommand{\PrivateOneGradeMedGemma}{18.09}
\newcommand{\PrivateOneGradeMedGemmaCI}{\textcolor{black}{8.61--60.36}}
\newcommand{\PrivateOneGradeMedGemmaN}{76}
\newcommand{\PrivateOneEntityPathAgent}{32.59}
\newcommand{\PrivateOneEntityPathAgentCI}{\textcolor{black}{24.78--41.19}}
\newcommand{\PrivateOneEntityPathAgentN}{135}
\newcommand{\PrivateOneFindingPathAgent}{31.11}
\newcommand{\PrivateOneFindingPathAgentCI}{25.78--36.44}
\newcommand{\PrivateOneFindingPathAgentN}{150}
\newcommand{\PrivateOneAssertionPathAgent}{30.67}
\newcommand{\PrivateOneAssertionPathAgentCI}{26.67--34.67}
\newcommand{\PrivateOneAssertionPathAgentN}{150}
\newcommand{\PrivateOneCorrectFindingPathAgent}{23.33}
\newcommand{\PrivateOneCorrectFindingPathAgentCI}{19.33--27.33}
\newcommand{\PrivateOneCorrectFindingPathAgentN}{150}
\newcommand{\PrivateOneContradictionPathAgent}{23.08}
\newcommand{\PrivateOneContradictionPathAgentCI}{14.29--31.87}
\newcommand{\PrivateOneContradictionPathAgentN}{91}
\newcommand{\PrivateOneCoveragePathAgent}{\textcolor{black}{66.15}}
\newcommand{\PrivateOneCoveragePathAgentCI}{\textcolor{black}{62.02--69.99}}
\newcommand{\PrivateOneCoveragePathAgentN}{151}
\newcommand{\PrivateOneGradePathAgent}{31.75}
\newcommand{\PrivateOneGradePathAgentCI}{\textcolor{black}{23.86--66.49}}
\newcommand{\PrivateOneGradePathAgentN}{76}
\newcommand{\PrivateOneEntityPRISM}{\textcolor{black}{54.81}}
\newcommand{\PrivateOneEntityPRISMCI}{\textcolor{black}{46.02--63.39}}
\newcommand{\PrivateOneEntityPRISMN}{135}
\newcommand{\PrivateTwoEntityFull}{\textcolor{black}{67.65}}
\newcommand{\PrivateTwoEntityFullProse}{67.6}
\newcommand{\PrivateTwoEntityFullCI}{\textcolor{black}{59.10--75.41}}
\newcommand{\PrivateTwoEntityFullN}{136}
\newcommand{\PrivateTwoFindingFull}{78.90}
\newcommand{\PrivateTwoFindingFullProse}{78.9}
\newcommand{\PrivateTwoFindingFullCI}{73.69--83.64}
\newcommand{\PrivateTwoFindingFullN}{151}
\newcommand{\PrivateTwoAssertionFull}{91.50}
\newcommand{\PrivateTwoAssertionFullCI}{88.08--94.59}
\newcommand{\PrivateTwoAssertionFullN}{151}
\newcommand{\PrivateTwoCorrectFindingFull}{77.26}
\newcommand{\PrivateTwoCorrectFindingFullCI}{72.07--82.34}
\newcommand{\PrivateTwoCorrectFindingFullN}{151}
\newcommand{\PrivateTwoContradictionFull}{16.22}
\newcommand{\PrivateTwoContradictionFullCI}{11.82--21.28}
\newcommand{\PrivateTwoContradictionFullN}{148}
\newcommand{\PrivateTwoCoverageFull}{\textcolor{black}{94.36}}
\newcommand{\PrivateTwoCoverageFullCI}{\textcolor{black}{92.65--95.96}}
\newcommand{\PrivateTwoCoverageFullN}{151}
\newcommand{\PrivateTwoGradeFull}{\textcolor{black}{42.18}}
\newcommand{\PrivateTwoGradeFullCI}{\textcolor{black}{25.11--82.72}}
\newcommand{\PrivateTwoGradeFullN}{76}
\newcommand{\PrivateTwoEntityCore}{\textcolor{black}{55.88}}
\newcommand{\PrivateTwoEntityCoreCI}{\textcolor{black}{47.12--64.38}}
\newcommand{\PrivateTwoEntityCoreN}{136}
\newcommand{\PrivateTwoFindingCore}{70.95}
\newcommand{\PrivateTwoFindingCoreCI}{65.74--76.05}
\newcommand{\PrivateTwoFindingCoreN}{151}
\newcommand{\PrivateTwoAssertionCore}{84.11}
\newcommand{\PrivateTwoAssertionCoreCI}{79.69--88.30}
\newcommand{\PrivateTwoAssertionCoreN}{151}
\newcommand{\PrivateTwoCorrectFindingCore}{67.55}
\newcommand{\PrivateTwoCorrectFindingCoreCI}{62.47--72.74}
\newcommand{\PrivateTwoCorrectFindingCoreN}{151}
\newcommand{\PrivateTwoContradictionCore}{18.97}
\newcommand{\PrivateTwoContradictionCoreCI}{14.25--24.02}
\newcommand{\PrivateTwoContradictionCoreN}{145}
\newcommand{\PrivateTwoCoverageCore}{\textcolor{black}{87.80}}
\newcommand{\PrivateTwoCoverageCoreCI}{\textcolor{black}{85.72--89.71}}
\newcommand{\PrivateTwoCoverageCoreN}{151}
\newcommand{\PrivateTwoGradeCore}{\textcolor{black}{24.06}}
\newcommand{\PrivateTwoGradeCoreCI}{\textcolor{black}{9.58--65.86}}
\newcommand{\PrivateTwoGradeCoreN}{76}
\newcommand{\PrivateTwoEntityQwen}{54.41}
\newcommand{\PrivateTwoEntityQwenCI}{\textcolor{black}{45.66--62.97}}
\newcommand{\PrivateTwoEntityQwenN}{136}
\newcommand{\PrivateTwoFindingQwen}{73.29}
\newcommand{\PrivateTwoFindingQwenCI}{67.55--78.76}
\newcommand{\PrivateTwoFindingQwenN}{151}
\newcommand{\PrivateTwoAssertionQwen}{78.04}
\newcommand{\PrivateTwoAssertionQwenCI}{72.41--83.11}
\newcommand{\PrivateTwoAssertionQwenN}{151}
\newcommand{\PrivateTwoCorrectFindingQwen}{69.98}
\newcommand{\PrivateTwoCorrectFindingQwenCI}{64.13--75.72}
\newcommand{\PrivateTwoCorrectFindingQwenN}{151}
\newcommand{\PrivateTwoContradictionQwen}{10.62}
\newcommand{\PrivateTwoContradictionQwenCI}{6.67--14.94}
\newcommand{\PrivateTwoContradictionQwenN}{135}
\newcommand{\PrivateTwoCoverageQwen}{83.86}
\newcommand{\PrivateTwoCoverageQwenCI}{81.22--86.36}
\newcommand{\PrivateTwoCoverageQwenN}{151}
\newcommand{\PrivateTwoGradeQwen}{20.38}
\newcommand{\PrivateTwoGradeQwenCI}{\textcolor{black}{7.07--62.78}}
\newcommand{\PrivateTwoGradeQwenN}{76}
\newcommand{\PrivateTwoEntityDeepSeek}{55.88}
\newcommand{\PrivateTwoEntityDeepSeekCI}{\textcolor{black}{47.12--64.38}}
\newcommand{\PrivateTwoEntityDeepSeekN}{136}
\newcommand{\PrivateTwoFindingDeepSeek}{69.09}
\newcommand{\PrivateTwoFindingDeepSeekCI}{63.80--74.17}
\newcommand{\PrivateTwoFindingDeepSeekN}{151}
\newcommand{\PrivateTwoAssertionDeepSeek}{78.92}
\newcommand{\PrivateTwoAssertionDeepSeekCI}{74.06--83.66}
\newcommand{\PrivateTwoAssertionDeepSeekN}{151}
\newcommand{\PrivateTwoCorrectFindingDeepSeek}{64.68}
\newcommand{\PrivateTwoCorrectFindingDeepSeekCI}{59.49--69.98}
\newcommand{\PrivateTwoCorrectFindingDeepSeekN}{151}
\newcommand{\PrivateTwoContradictionDeepSeek}{17.37}
\newcommand{\PrivateTwoContradictionDeepSeekCI}{12.70--22.26}
\newcommand{\PrivateTwoContradictionDeepSeekN}{143}
\newcommand{\PrivateTwoCoverageDeepSeek}{\textcolor{black}{85.06}}
\newcommand{\PrivateTwoCoverageDeepSeekCI}{\textcolor{black}{81.99--87.90}}
\newcommand{\PrivateTwoCoverageDeepSeekN}{151}
\newcommand{\PrivateTwoGradeDeepSeek}{\textcolor{black}{35.21}}
\newcommand{\PrivateTwoGradeDeepSeekCI}{\textcolor{black}{20.87--68.95}}
\newcommand{\PrivateTwoGradeDeepSeekN}{76}
\newcommand{\PrivateTwoEntityGLM}{\textcolor{black}{43.38}}
\newcommand{\PrivateTwoEntityGLMCI}{\textcolor{black}{34.91--52.15}}
\newcommand{\PrivateTwoEntityGLMN}{136}
\newcommand{\PrivateTwoFindingGLM}{64.70}
\newcommand{\PrivateTwoFindingGLMCI}{58.52--70.49}
\newcommand{\PrivateTwoFindingGLMN}{151}
\newcommand{\PrivateTwoAssertionGLM}{64.13}
\newcommand{\PrivateTwoAssertionGLMCI}{58.39--69.76}
\newcommand{\PrivateTwoAssertionGLMN}{151}
\newcommand{\PrivateTwoCorrectFindingGLM}{58.83}
\newcommand{\PrivateTwoCorrectFindingGLMCI}{52.54--64.79}
\newcommand{\PrivateTwoCorrectFindingGLMN}{151}
\newcommand{\PrivateTwoContradictionGLM}{9.06}
\newcommand{\PrivateTwoContradictionGLMCI}{4.72--13.78}
\newcommand{\PrivateTwoContradictionGLMN}{127}
\newcommand{\PrivateTwoCoverageGLM}{\textcolor{black}{76.72}}
\newcommand{\PrivateTwoCoverageGLMCI}{\textcolor{black}{72.65--80.58}}
\newcommand{\PrivateTwoCoverageGLMN}{151}
\newcommand{\PrivateTwoGradeGLM}{\textcolor{black}{43.85}}
\newcommand{\PrivateTwoGradeGLMCI}{\textcolor{black}{16.38--78.97}}
\newcommand{\PrivateTwoGradeGLMN}{76}
\newcommand{\PrivateTwoEntityMedGemma}{2.94}
\newcommand{\PrivateTwoEntityMedGemmaCI}{\textcolor{black}{0.81--7.36}}
\newcommand{\PrivateTwoEntityMedGemmaN}{136}
\newcommand{\PrivateTwoFindingMedGemma}{42.83}
\newcommand{\PrivateTwoFindingMedGemmaCI}{36.87--48.79}
\newcommand{\PrivateTwoFindingMedGemmaN}{151}
\newcommand{\PrivateTwoAssertionMedGemma}{62.69}
\newcommand{\PrivateTwoAssertionMedGemmaCI}{55.74--69.32}
\newcommand{\PrivateTwoAssertionMedGemmaN}{151}
\newcommand{\PrivateTwoCorrectFindingMedGemma}{40.40}
\newcommand{\PrivateTwoCorrectFindingMedGemmaCI}{34.66--46.25}
\newcommand{\PrivateTwoCorrectFindingMedGemmaN}{151}
\newcommand{\PrivateTwoContradictionMedGemma}{35.14}
\newcommand{\PrivateTwoContradictionMedGemmaCI}{28.38--41.89}
\newcommand{\PrivateTwoContradictionMedGemmaN}{111}
\newcommand{\PrivateTwoCoverageMedGemma}{\textcolor{black}{67.19}}
\newcommand{\PrivateTwoCoverageMedGemmaCI}{\textcolor{black}{62.46--71.81}}
\newcommand{\PrivateTwoCoverageMedGemmaN}{151}
\newcommand{\PrivateTwoGradeMedGemma}{23.63}
\newcommand{\PrivateTwoGradeMedGemmaCI}{\textcolor{black}{11.03--61.34}}
\newcommand{\PrivateTwoGradeMedGemmaN}{76}
\newcommand{\PrivateTwoEntityPathAgent}{31.62}
\newcommand{\PrivateTwoEntityPathAgentCI}{\textcolor{black}{23.92--40.14}}
\newcommand{\PrivateTwoEntityPathAgentN}{136}
\newcommand{\PrivateTwoFindingPathAgent}{33.11}
\newcommand{\PrivateTwoFindingPathAgentCI}{27.37--38.64}
\newcommand{\PrivateTwoFindingPathAgentN}{151}
\newcommand{\PrivateTwoAssertionPathAgent}{32.34}
\newcommand{\PrivateTwoAssertionPathAgentCI}{27.70--37.09}
\newcommand{\PrivateTwoAssertionPathAgentN}{151}
\newcommand{\PrivateTwoCorrectFindingPathAgent}{26.16}
\newcommand{\PrivateTwoCorrectFindingPathAgentCI}{21.52--31.13}
\newcommand{\PrivateTwoCorrectFindingPathAgentN}{151}
\newcommand{\PrivateTwoContradictionPathAgent}{20.22}
\newcommand{\PrivateTwoContradictionPathAgentCI}{12.36--28.09}
\newcommand{\PrivateTwoContradictionPathAgentN}{89}
\newcommand{\PrivateTwoCoveragePathAgent}{\textcolor{black}{67.90}}
\newcommand{\PrivateTwoCoveragePathAgentCI}{\textcolor{black}{63.64--71.80}}
\newcommand{\PrivateTwoCoveragePathAgentN}{151}
\newcommand{\PrivateTwoGradePathAgent}{33.33}
\newcommand{\PrivateTwoGradePathAgentCI}{\textcolor{black}{23.06--62.47}}
\newcommand{\PrivateTwoGradePathAgentN}{76}
\newcommand{\PrivateTwoEntityPRISM}{\textcolor{black}{52.21}}
\newcommand{\PrivateTwoEntityPRISMCI}{\textcolor{black}{43.48--60.84}}
\newcommand{\PrivateTwoEntityPRISMN}{136}
\newcommand{\PrivateJointEntityFull}{\textcolor{black}{69.84}}
\newcommand{\PrivateJointEntityFullCI}{\textcolor{black}{59.79--78.69}}
\newcommand{\PrivateJointEntityFullN}{126}
\newcommand{\PrivateJointFindingFull}{77.82}
\newcommand{\PrivateJointFindingFullCI}{72.75--82.56}
\newcommand{\PrivateJointFindingFullN}{150}
\newcommand{\PrivateJointAssertionFull}{92.06}
\newcommand{\PrivateJointAssertionFullCI}{88.78--95.11}
\newcommand{\PrivateJointAssertionFullN}{150}
\newcommand{\PrivateJointCorrectFindingFull}{76.22}
\newcommand{\PrivateJointCorrectFindingFullCI}{71.17--81.06}
\newcommand{\PrivateJointCorrectFindingFullN}{150}
\newcommand{\PrivateJointContradictionFull}{16.78}
\newcommand{\PrivateJointContradictionFullCI}{12.50--21.58}
\newcommand{\PrivateJointContradictionFullN}{146}
\newcommand{\PrivateJointCoverageFull}{\textcolor{black}{94.35}}
\newcommand{\PrivateJointCoverageFullCI}{\textcolor{black}{92.64--95.95}}
\newcommand{\PrivateJointCoverageFullN}{151}
\newcommand{\PrivateJointEntityCore}{\textcolor{black}{57.94}}
\newcommand{\PrivateJointEntityCoreCI}{\textcolor{black}{47.57--67.82}}
\newcommand{\PrivateJointEntityCoreN}{126}
\newcommand{\PrivateJointFindingCore}{69.16}
\newcommand{\PrivateJointFindingCoreCI}{64.10--74.16}
\newcommand{\PrivateJointFindingCoreN}{150}
\newcommand{\PrivateJointAssertionCore}{84.33}
\newcommand{\PrivateJointAssertionCoreCI}{80.17--88.28}
\newcommand{\PrivateJointAssertionCoreN}{150}
\newcommand{\PrivateJointCorrectFindingCore}{65.83}
\newcommand{\PrivateJointCorrectFindingCoreCI}{60.56--71.11}
\newcommand{\PrivateJointCorrectFindingCoreN}{150}
\newcommand{\PrivateJointContradictionCore}{21.15}
\newcommand{\PrivateJointContradictionCoreCI}{16.08--26.17}
\newcommand{\PrivateJointContradictionCoreN}{143}
\newcommand{\PrivateJointCoverageCore}{\textcolor{black}{87.63}}
\newcommand{\PrivateJointCoverageCoreCI}{\textcolor{black}{85.62--89.59}}
\newcommand{\PrivateJointCoverageCoreN}{151}
\newcommand{\PrivateJointEntityQwen}{56.35}
\newcommand{\PrivateJointEntityQwenCI}{\textcolor{black}{45.98--66.33}}
\newcommand{\PrivateJointEntityQwenN}{126}
\newcommand{\PrivateJointFindingQwen}{72.67}
\newcommand{\PrivateJointFindingQwenCI}{67.03--77.76}
\newcommand{\PrivateJointFindingQwenN}{150}
\newcommand{\PrivateJointAssertionQwen}{78.78}
\newcommand{\PrivateJointAssertionQwenCI}{73.50--83.67}
\newcommand{\PrivateJointAssertionQwenN}{150}
\newcommand{\PrivateJointCorrectFindingQwen}{69.22}
\newcommand{\PrivateJointCorrectFindingQwenCI}{63.94--74.50}
\newcommand{\PrivateJointCorrectFindingQwenN}{150}
\newcommand{\PrivateJointContradictionQwen}{12.44}
\newcommand{\PrivateJointContradictionQwenCI}{8.27--16.85}
\newcommand{\PrivateJointContradictionQwenN}{134}
\newcommand{\PrivateJointCoverageQwen}{83.80}
\newcommand{\PrivateJointCoverageQwenCI}{81.35--86.24}
\newcommand{\PrivateJointCoverageQwenN}{151}
\newcommand{\PrivateJointEntityDeepSeek}{60.32}
\newcommand{\PrivateJointEntityDeepSeekCI}{\textcolor{black}{49.96--70.05}}
\newcommand{\PrivateJointEntityDeepSeekN}{126}
\newcommand{\PrivateJointFindingDeepSeek}{68.00}
\newcommand{\PrivateJointFindingDeepSeekCI}{63.17--72.72}
\newcommand{\PrivateJointFindingDeepSeekN}{150}
\newcommand{\PrivateJointAssertionDeepSeek}{78.89}
\newcommand{\PrivateJointAssertionDeepSeekCI}{74.17--83.44}
\newcommand{\PrivateJointAssertionDeepSeekN}{150}
\newcommand{\PrivateJointCorrectFindingDeepSeek}{63.39}
\newcommand{\PrivateJointCorrectFindingDeepSeekCI}{58.39--68.50}
\newcommand{\PrivateJointCorrectFindingDeepSeekN}{150}
\newcommand{\PrivateJointContradictionDeepSeek}{17.80}
\newcommand{\PrivateJointContradictionDeepSeekCI}{13.39--22.50}
\newcommand{\PrivateJointContradictionDeepSeekN}{140}
\newcommand{\PrivateJointCoverageDeepSeek}{\textcolor{black}{84.97}}
\newcommand{\PrivateJointCoverageDeepSeekCI}{\textcolor{black}{81.98--87.85}}
\newcommand{\PrivateJointCoverageDeepSeekN}{151}
\newcommand{\PrivateJointEntityGLM}{\textcolor{black}{45.24}}
\newcommand{\PrivateJointEntityGLMCI}{\textcolor{black}{35.18--55.59}}
\newcommand{\PrivateJointEntityGLMN}{126}
\newcommand{\PrivateJointFindingGLM}{64.12}
\newcommand{\PrivateJointFindingGLMCI}{58.27--69.60}
\newcommand{\PrivateJointFindingGLMN}{150}
\newcommand{\PrivateJointAssertionGLM}{63.94}
\newcommand{\PrivateJointAssertionGLMCI}{58.44--69.45}
\newcommand{\PrivateJointAssertionGLMN}{150}
\newcommand{\PrivateJointCorrectFindingGLM}{57.94}
\newcommand{\PrivateJointCorrectFindingGLMCI}{52.28--63.67}
\newcommand{\PrivateJointCorrectFindingGLMN}{150}
\newcommand{\PrivateJointContradictionGLM}{9.60}
\newcommand{\PrivateJointContradictionGLMCI}{5.60--14.20}
\newcommand{\PrivateJointContradictionGLMN}{125}
\newcommand{\PrivateJointCoverageGLM}{\textcolor{black}{76.00}}
\newcommand{\PrivateJointCoverageGLMCI}{\textcolor{black}{71.90--80.03}}
\newcommand{\PrivateJointCoverageGLMN}{151}
\newcommand{\PrivateJointEntityMedGemma}{3.17}
\newcommand{\PrivateJointEntityMedGemmaCI}{\textcolor{black}{0.70--8.70}}
\newcommand{\PrivateJointEntityMedGemmaN}{126}
\newcommand{\PrivateJointFindingMedGemma}{40.78}
\newcommand{\PrivateJointFindingMedGemmaCI}{35.11--46.44}
\newcommand{\PrivateJointFindingMedGemmaN}{150}
\newcommand{\PrivateJointAssertionMedGemma}{63.06}
\newcommand{\PrivateJointAssertionMedGemmaCI}{56.67--69.61}
\newcommand{\PrivateJointAssertionMedGemmaN}{150}
\newcommand{\PrivateJointCorrectFindingMedGemma}{38.17}
\newcommand{\PrivateJointCorrectFindingMedGemmaCI}{32.83--43.50}
\newcommand{\PrivateJointCorrectFindingMedGemmaN}{150}
\newcommand{\PrivateJointContradictionMedGemma}{37.84}
\newcommand{\PrivateJointContradictionMedGemmaCI}{31.46--44.52}
\newcommand{\PrivateJointContradictionMedGemmaN}{111}
\newcommand{\PrivateJointCoverageMedGemma}{\textcolor{black}{67.78}}
\newcommand{\PrivateJointCoverageMedGemmaCI}{\textcolor{black}{63.10--72.29}}
\newcommand{\PrivateJointCoverageMedGemmaN}{151}
\newcommand{\PrivateJointEntityPathAgent}{34.13}
\newcommand{\PrivateJointEntityPathAgentCI}{\textcolor{black}{24.86--44.36}}
\newcommand{\PrivateJointEntityPathAgentN}{126}
\newcommand{\PrivateJointFindingPathAgent}{32.22}
\newcommand{\PrivateJointFindingPathAgentCI}{26.67--37.56}
\newcommand{\PrivateJointFindingPathAgentN}{150}
\newcommand{\PrivateJointAssertionPathAgent}{31.61}
\newcommand{\PrivateJointAssertionPathAgentCI}{27.33--35.83}
\newcommand{\PrivateJointAssertionPathAgentN}{150}
\newcommand{\PrivateJointCorrectFindingPathAgent}{24.83}
\newcommand{\PrivateJointCorrectFindingPathAgentCI}{20.67--29.17}
\newcommand{\PrivateJointCorrectFindingPathAgentN}{150}
\newcommand{\PrivateJointContradictionPathAgent}{20.79}
\newcommand{\PrivateJointContradictionPathAgentCI}{12.92--29.21}
\newcommand{\PrivateJointContradictionPathAgentN}{89}
\newcommand{\PrivateJointCoveragePathAgent}{\textcolor{black}{67.03}}
\newcommand{\PrivateJointCoveragePathAgentCI}{\textcolor{black}{62.66--70.82}}
\newcommand{\PrivateJointCoveragePathAgentN}{151}
\newcommand{\PrivateJointEntityPRISM}{\textcolor{black}{55.56}}
\newcommand{\PrivateJointEntityPRISMCI}{\textcolor{black}{45.20--65.58}}
\newcommand{\PrivateJointEntityPRISMN}{126}
\newcommand{\PolicyCancerFull}{\textcolor{black}{72.50}}
\newcommand{\PolicyCancerFullCI}{\textcolor{black}{69.88--75.01}}
\newcommand{\PolicyCancerFullN}{\textcolor{black}{1200}}
\newcommand{\PolicyDiagnosticFull}{\textcolor{black}{86.91}}
\newcommand{\PolicyDiagnosticFullCI}{\textcolor{black}{85.79--88.03}}
\newcommand{\PolicyDiagnosticFullN}{\textcolor{black}{1200}}
\newcommand{\PolicyEntityFull}{\textcolor{black}{62.25}}
\newcommand{\PolicyEntityFullCI}{\textcolor{black}{58.25--66.13}}
\newcommand{\PolicyEntityFullN}{\textcolor{black}{604}}
\newcommand{\PolicyFindingFull}{\textcolor{black}{48.76}}
\newcommand{\PolicyFindingFullCI}{\textcolor{black}{46.50--51.03}}
\newcommand{\PolicyFindingFullN}{\textcolor{black}{997}}
\newcommand{\PolicyAssertionFull}{\textcolor{black}{47.18}}
\newcommand{\PolicyAssertionFullCI}{\textcolor{black}{44.87--49.36}}
\newcommand{\PolicyAssertionFullN}{\textcolor{black}{997}}
\newcommand{\PolicyCorrectFindingFull}{\textcolor{black}{41.64}}
\newcommand{\PolicyCorrectFindingFullCI}{\textcolor{black}{39.44--43.88}}
\newcommand{\PolicyCorrectFindingFullN}{\textcolor{black}{997}}
\newcommand{\PolicyContradictionFull}{\textcolor{black}{11.53}}
\newcommand{\PolicyContradictionFullCI}{\textcolor{black}{9.60--13.55}}
\newcommand{\PolicyContradictionFullN}{\textcolor{black}{759}}
\newcommand{\PolicyCoverageFull}{\textcolor{black}{90.26}}
\newcommand{\PolicyCoverageFullCI}{\textcolor{black}{89.36--91.17}}
\newcommand{\PolicyCoverageFullN}{\textcolor{black}{1200}}
\newcommand{\PolicyGradeFull}{\textcolor{black}{36.97}}
\newcommand{\PolicyGradeFullCI}{\textcolor{black}{31.14--44.76}}
\newcommand{\PolicyGradeFullN}{\textcolor{black}{472}}
\newcommand{\PolicyCancerQwen}{\textcolor{black}{68.08}}
\newcommand{\PolicyCancerQwenCI}{\textcolor{black}{65.36--70.72}}
\newcommand{\PolicyCancerQwenN}{\textcolor{black}{1200}}
\newcommand{\PolicyDiagnosticQwen}{\textcolor{black}{84.78}}
\newcommand{\PolicyDiagnosticQwenCI}{\textcolor{black}{83.53--86.03}}
\newcommand{\PolicyDiagnosticQwenN}{\textcolor{black}{1200}}
\newcommand{\PolicyEntityQwen}{\textcolor{black}{58.94}}
\newcommand{\PolicyEntityQwenCI}{\textcolor{black}{54.90--62.89}}
\newcommand{\PolicyEntityQwenN}{\textcolor{black}{604}}
\newcommand{\PolicyFindingQwen}{\textcolor{black}{43.02}}
\newcommand{\PolicyFindingQwenCI}{\textcolor{black}{40.67--45.30}}
\newcommand{\PolicyFindingQwenN}{\textcolor{black}{997}}
\newcommand{\PolicyAssertionQwen}{\textcolor{black}{42.90}}
\newcommand{\PolicyAssertionQwenCI}{\textcolor{black}{40.66--45.14}}
\newcommand{\PolicyAssertionQwenN}{\textcolor{black}{997}}
\newcommand{\PolicyCorrectFindingQwen}{\textcolor{black}{36.57}}
\newcommand{\PolicyCorrectFindingQwenCI}{\textcolor{black}{34.36--38.81}}
\newcommand{\PolicyCorrectFindingQwenN}{\textcolor{black}{997}}
\newcommand{\PolicyContradictionQwen}{\textcolor{black}{16.01}}
\newcommand{\PolicyContradictionQwenCI}{\textcolor{black}{13.73--18.40}}
\newcommand{\PolicyContradictionQwenN}{\textcolor{black}{707}}
\newcommand{\PolicyCoverageQwen}{\textcolor{black}{84.46}}
\newcommand{\PolicyCoverageQwenCI}{\textcolor{black}{83.41--85.56}}
\newcommand{\PolicyCoverageQwenN}{\textcolor{black}{1200}}
\newcommand{\PolicyGradeQwen}{\textcolor{black}{24.13}}
\newcommand{\PolicyGradeQwenCI}{\textcolor{black}{18.45--32.35}}
\newcommand{\PolicyGradeQwenN}{\textcolor{black}{472}}
\newcommand{\PolicyCancerStatistics}{\textcolor{black}{69.17}}
\newcommand{\PolicyCancerStatisticsCI}{\textcolor{black}{66.47--71.77}}
\newcommand{\PolicyCancerStatisticsN}{\textcolor{black}{1200}}
\newcommand{\PolicyDiagnosticStatistics}{\textcolor{black}{84.87}}
\newcommand{\PolicyDiagnosticStatisticsCI}{\textcolor{black}{83.62--86.13}}
\newcommand{\PolicyDiagnosticStatisticsN}{\textcolor{black}{1200}}
\newcommand{\PolicyEntityStatistics}{\textcolor{black}{61.09}}
\newcommand{\PolicyEntityStatisticsCI}{\textcolor{black}{57.08--65.00}}
\newcommand{\PolicyEntityStatisticsN}{\textcolor{black}{604}}
\newcommand{\PolicyFindingStatistics}{\textcolor{black}{43.03}}
\newcommand{\PolicyFindingStatisticsCI}{\textcolor{black}{40.74--45.35}}
\newcommand{\PolicyFindingStatisticsN}{\textcolor{black}{997}}
\newcommand{\PolicyAssertionStatistics}{\textcolor{black}{40.30}}
\newcommand{\PolicyAssertionStatisticsCI}{\textcolor{black}{38.07--42.47}}
\newcommand{\PolicyAssertionStatisticsN}{\textcolor{black}{997}}
\newcommand{\PolicyCorrectFindingStatistics}{\textcolor{black}{36.01}}
\newcommand{\PolicyCorrectFindingStatisticsCI}{\textcolor{black}{33.94--38.17}}
\newcommand{\PolicyCorrectFindingStatisticsN}{\textcolor{black}{997}}
\newcommand{\PolicyContradictionStatistics}{\textcolor{black}{11.14}}
\newcommand{\PolicyContradictionStatisticsCI}{\textcolor{black}{9.15--13.38}}
\newcommand{\PolicyContradictionStatisticsN}{\textcolor{black}{690}}
\newcommand{\PolicyCoverageStatistics}{\textcolor{black}{84.13}}
\newcommand{\PolicyCoverageStatisticsCI}{\textcolor{black}{83.01--85.29}}
\newcommand{\PolicyCoverageStatisticsN}{\textcolor{black}{1200}}
\newcommand{\PolicyGradeStatistics}{\textcolor{black}{24.93}}
\newcommand{\PolicyGradeStatisticsCI}{\textcolor{black}{19.27--33.11}}
\newcommand{\PolicyGradeStatisticsN}{\textcolor{black}{472}}
\newcommand{\PolicyCancerOverlays}{\textcolor{black}{64.58}}
\newcommand{\PolicyCancerOverlaysCI}{\textcolor{black}{61.80--67.29}}
\newcommand{\PolicyCancerOverlaysN}{\textcolor{black}{1200}}
\newcommand{\PolicyDiagnosticOverlays}{\textcolor{black}{82.73}}
\newcommand{\PolicyDiagnosticOverlaysCI}{\textcolor{black}{81.40--84.05}}
\newcommand{\PolicyDiagnosticOverlaysN}{\textcolor{black}{1200}}
\newcommand{\PolicyEntityOverlays}{\textcolor{black}{56.13}}
\newcommand{\PolicyEntityOverlaysCI}{\textcolor{black}{52.06--60.13}}
\newcommand{\PolicyEntityOverlaysN}{\textcolor{black}{604}}
\newcommand{\PolicyFindingOverlays}{\textcolor{black}{41.72}}
\newcommand{\PolicyFindingOverlaysCI}{\textcolor{black}{39.45--44.01}}
\newcommand{\PolicyFindingOverlaysN}{\textcolor{black}{997}}
\newcommand{\PolicyAssertionOverlays}{\textcolor{black}{41.54}}
\newcommand{\PolicyAssertionOverlaysCI}{\textcolor{black}{39.25--43.83}}
\newcommand{\PolicyAssertionOverlaysN}{\textcolor{black}{997}}
\newcommand{\PolicyCorrectFindingOverlays}{\textcolor{black}{35.12}}
\newcommand{\PolicyCorrectFindingOverlaysCI}{\textcolor{black}{33.00--37.38}}
\newcommand{\PolicyCorrectFindingOverlaysN}{\textcolor{black}{997}}
\newcommand{\PolicyContradictionOverlays}{\textcolor{black}{16.13}}
\newcommand{\PolicyContradictionOverlaysCI}{\textcolor{black}{13.66--18.60}}
\newcommand{\PolicyContradictionOverlaysN}{\textcolor{black}{699}}
\newcommand{\PolicyCoverageOverlays}{\textcolor{black}{83.75}}
\newcommand{\PolicyCoverageOverlaysCI}{\textcolor{black}{82.62--84.96}}
\newcommand{\PolicyCoverageOverlaysN}{\textcolor{black}{1200}}
\newcommand{\PolicyGradeOverlays}{\textcolor{black}{23.48}}
\newcommand{\PolicyGradeOverlaysCI}{\textcolor{black}{18.28--31.26}}
\newcommand{\PolicyGradeOverlaysN}{\textcolor{black}{472}}
\newcommand{\PublicEntityFull}{\textcolor{black}{63.70}}
\newcommand{\PublicEntityFullCI}{\textcolor{black}{60.56--66.76}}
\newcommand{\PublicEntityFullN}{\textcolor{black}{956}}
\newcommand{\PublicEntityCore}{\textcolor{black}{60.56}}
\newcommand{\PublicEntityCoreCI}{\textcolor{black}{57.39--63.68}}
\newcommand{\PublicEntityCoreN}{\textcolor{black}{956}}
\newcommand{\PublicEntityQwen}{\textcolor{black}{59.94}}
\newcommand{\PublicEntityQwenCI}{\textcolor{black}{56.75--63.06}}
\newcommand{\PublicEntityQwenN}{\textcolor{black}{956}}
\newcommand{\PublicEntityDeepSeek}{\textcolor{black}{61.09}}
\newcommand{\PublicEntityDeepSeekCI}{\textcolor{black}{57.91--64.19}}
\newcommand{\PublicEntityDeepSeekN}{\textcolor{black}{956}}
\newcommand{\PublicEntityGLM}{\textcolor{black}{52.51}}
\newcommand{\PublicEntityGLMCI}{\textcolor{black}{49.29--55.72}}
\newcommand{\PublicEntityGLMN}{\textcolor{black}{956}}
\newcommand{\PublicEntityMedGemma}{\textcolor{black}{5.23}}
\newcommand{\PublicEntityMedGemmaCI}{\textcolor{black}{3.91--6.84}}
\newcommand{\PublicEntityMedGemmaN}{\textcolor{black}{956}}
\newcommand{\PublicEntityPathAgent}{\textcolor{black}{33.58}}
\newcommand{\PublicEntityPathAgentCI}{\textcolor{black}{30.59--36.67}}
\newcommand{\PublicEntityPathAgentN}{\textcolor{black}{956}}
\newcommand{\PublicEntityPRISM}{\textcolor{black}{51.36}}
\newcommand{\PublicEntityPRISMCI}{\textcolor{black}{48.14--54.57}}
\newcommand{\PublicEntityPRISMN}{\textcolor{black}{956}}
\newcommand{\PrivateOneCancerQwen}{89.26}
\newcommand{\PrivateOneCancerQwenCI}{\textcolor{black}{82.33--94.15}}
\newcommand{\PrivateOneCancerQwenN}{121}
\newcommand{\PrivateOneCancerDeepSeek}{\textcolor{black}{90.91}}
\newcommand{\PrivateOneCancerDeepSeekCI}{\textcolor{black}{84.32--95.37}}
\newcommand{\PrivateOneCancerDeepSeekN}{121}
\newcommand{\PrivateOneCancerGLM}{\textcolor{black}{64.46}}
\newcommand{\PrivateOneCancerGLMCI}{\textcolor{black}{55.25--72.95}}
\newcommand{\PrivateOneCancerGLMN}{121}
\newcommand{\PrivateOneCancerMedGemma}{4.13}
\newcommand{\PrivateOneCancerMedGemmaCI}{\textcolor{black}{1.36--9.38}}
\newcommand{\PrivateOneCancerMedGemmaN}{121}
\newcommand{\PrivateOneCancerPathAgent}{88.43}
\newcommand{\PrivateOneCancerPathAgentCI}{\textcolor{black}{81.35--93.53}}
\newcommand{\PrivateOneCancerPathAgentN}{121}
\newcommand{\PrivateOneCancerPRISM}{\textcolor{black}{64.46}}
\newcommand{\PrivateOneCancerPRISMCI}{\textcolor{black}{55.25--72.95}}
\newcommand{\PrivateOneCancerPRISMN}{121}
\newcommand{\PrivateOneCancerCore}{\textcolor{black}{93.39}}
\newcommand{\PrivateOneCancerCoreCI}{\textcolor{black}{87.39--97.10}}
\newcommand{\PrivateOneCancerCoreN}{121}
\newcommand{\PrivateOneCancerFull}{93.39}
\newcommand{\PrivateOneCancerFullProse}{93.4}
\newcommand{\PrivateOneCancerFullCI}{\textcolor{black}{87.39--97.10}}
\newcommand{\PrivateOneCancerFullN}{121}
\newcommand{\PrivateTwoCancerQwen}{89.43}
\newcommand{\PrivateTwoCancerQwenCI}{\textcolor{black}{82.60--94.25}}
\newcommand{\PrivateTwoCancerQwenN}{123}
\newcommand{\PrivateTwoCancerDeepSeek}{\textcolor{black}{90.24}}
\newcommand{\PrivateTwoCancerDeepSeekCI}{\textcolor{black}{83.58--94.86}}
\newcommand{\PrivateTwoCancerDeepSeekN}{123}
\newcommand{\PrivateTwoCancerGLM}{\textcolor{black}{65.04}}
\newcommand{\PrivateTwoCancerGLMCI}{\textcolor{black}{55.92--73.42}}
\newcommand{\PrivateTwoCancerGLMN}{123}
\newcommand{\PrivateTwoCancerMedGemma}{4.07}
\newcommand{\PrivateTwoCancerMedGemmaCI}{\textcolor{black}{1.33--9.23}}
\newcommand{\PrivateTwoCancerMedGemmaN}{123}
\newcommand{\PrivateTwoCancerPathAgent}{88.62}
\newcommand{\PrivateTwoCancerPathAgentCI}{\textcolor{black}{81.64--93.64}}
\newcommand{\PrivateTwoCancerPathAgentN}{123}
\newcommand{\PrivateTwoCancerPRISM}{\textcolor{black}{63.41}}
\newcommand{\PrivateTwoCancerPRISMCI}{\textcolor{black}{54.25--71.91}}
\newcommand{\PrivateTwoCancerPRISMN}{123}
\newcommand{\PrivateTwoCancerCore}{\textcolor{black}{93.50}}
\newcommand{\PrivateTwoCancerCoreCI}{\textcolor{black}{87.59--97.15}}
\newcommand{\PrivateTwoCancerCoreN}{123}
\newcommand{\PrivateTwoCancerFull}{93.50}
\newcommand{\PrivateTwoCancerFullProse}{93.5}
\newcommand{\PrivateTwoCancerFullCI}{\textcolor{black}{87.59--97.15}}
\newcommand{\PrivateTwoCancerFullN}{123}
\newcommand{\PublicVsCore}{\textcolor{black}{3.99}}
\newcommand{\PublicVsCoreCI}{\textcolor{black}{2.17--5.80}}
\newcommand{\PublicVsCoreN}{1656}
\newcommand{\PublicVsQwen}{\textcolor{black}{4.65}}
\newcommand{\PublicVsQwenCI}{\textcolor{black}{2.66--6.64}}
\newcommand{\PublicVsQwenN}{1656}
\newcommand{\PublicVsDeepSeek}{\textcolor{black}{2.11}}
\newcommand{\PublicVsDeepSeekCI}{\textcolor{black}{0.00--4.35}}
\newcommand{\PublicVsDeepSeekN}{1656}
\newcommand{\PublicVsGLM}{\textcolor{black}{13.47}}
\newcommand{\PublicVsGLMCI}{\textcolor{black}{10.81--16.00}}
\newcommand{\PublicVsGLMN}{1656}
\newcommand{\PublicVsMedGemma}{\textcolor{black}{66.73}}
\newcommand{\PublicVsMedGemmaCI}{\textcolor{black}{64.43--68.90}}
\newcommand{\PublicVsMedGemmaN}{1656}
\newcommand{\PublicVsPathAgent}{32.37}
\newcommand{\PublicVsPathAgentCI}{\textcolor{black}{30.01--34.84}}
\newcommand{\PublicVsPathAgentN}{1656}
\newcommand{\PublicVsPRISM}{\textcolor{black}{19.69}}
\newcommand{\PublicVsPRISMCI}{\textcolor{black}{16.61--22.77}}
\newcommand{\PublicVsPRISMN}{1656}
\newcommand{\EvaluationVsCore}{\textcolor{black}{4.33}}
\newcommand{\EvaluationVsCoreProse}{4.3}
\newcommand{\EvaluationVsCoreCI}{\textcolor{black}{2.08--6.58}}
\newcommand{\EvaluationVsCoreN}{1200}
\newcommand{\EvaluationVsQwen}{4.42}
\newcommand{\EvaluationVsQwenProse}{4.4}
\newcommand{\EvaluationVsQwenCI}{\textcolor{black}{1.92--6.75}}
\newcommand{\EvaluationVsQwenN}{1200}
\newcommand{\EvaluationVsDeepSeek}{\textcolor{black}{0.58}}
\newcommand{\EvaluationVsDeepSeekProse}{0.6}
\newcommand{\EvaluationVsDeepSeekCI}{\textcolor{black}{-1.92 to 3.17}}
\newcommand{\EvaluationVsDeepSeekN}{1200}
\newcommand{\EvaluationVsGLM}{\textcolor{black}{10.67}}
\newcommand{\EvaluationVsGLMCI}{\textcolor{black}{7.42--13.75}}
\newcommand{\EvaluationVsGLMN}{1200}
\newcommand{\EvaluationVsMedGemma}{65.83}
\newcommand{\EvaluationVsMedGemmaCI}{\textcolor{black}{63.17--68.42}}
\newcommand{\EvaluationVsMedGemmaN}{1200}
\newcommand{\EvaluationVsPathAgent}{\textcolor{black}{26.83}}
\newcommand{\EvaluationVsPathAgentCI}{\textcolor{black}{24.08--29.50}}
\newcommand{\EvaluationVsPathAgentN}{1200}
\newcommand{\EvaluationVsPRISM}{\textcolor{black}{19.25}}
\newcommand{\EvaluationVsPRISMCI}{\textcolor{black}{15.50--22.84}}
\newcommand{\EvaluationVsPRISMN}{1200}
\newcommand{\PrivateVsCore}{\textcolor{black}{11.90}}
\newcommand{\PrivateVsCoreCI}{\textcolor{black}{4.76--19.05}}
\newcommand{\PrivateVsCoreN}{126}
\newcommand{\PrivateVsQwen}{\textcolor{black}{13.49}}
\newcommand{\PrivateVsQwenCI}{\textcolor{black}{5.56--21.43}}
\newcommand{\PrivateVsQwenN}{126}
\newcommand{\PrivateVsDeepSeek}{\textcolor{black}{9.52}}
\newcommand{\PrivateVsDeepSeekProse}{9.5}
\newcommand{\PrivateVsDeepSeekCI}{\textcolor{black}{0.79--18.25}}
\newcommand{\PrivateVsDeepSeekN}{126}
\newcommand{\PrivateVsGLM}{\textcolor{black}{24.60}}
\newcommand{\PrivateVsGLMCI}{\textcolor{black}{13.89--35.71}}
\newcommand{\PrivateVsGLMN}{126}
\newcommand{\PrivateVsMedGemma}{\textcolor{black}{66.67}}
\newcommand{\PrivateVsMedGemmaCI}{\textcolor{black}{57.94--74.60}}
\newcommand{\PrivateVsMedGemmaN}{126}
\newcommand{\PrivateVsPathAgent}{\textcolor{black}{35.71}}
\newcommand{\PrivateVsPathAgentCI}{\textcolor{black}{26.98--44.46}}
\newcommand{\PrivateVsPathAgentN}{126}
\newcommand{\PrivateVsPRISM}{\textcolor{black}{14.29}}
\newcommand{\PrivateVsPRISMCI}{\textcolor{black}{3.97--24.60}}
\newcommand{\PrivateVsPRISMN}{126}
\newcommand{\StatisticsVsHECancer}{\textcolor{black}{1.08}}
\newcommand{\StatisticsVsHECancerCI}{\textcolor{black}{-1.00 to 3.25}}
\newcommand{\StatisticsVsHECancerN}{\textcolor{black}{1200}}
\newcommand{\StatisticsVsHEDiagnostic}{\textcolor{black}{0.09}}
\newcommand{\StatisticsVsHEDiagnosticCI}{\textcolor{black}{-0.89 to 1.13}}
\newcommand{\StatisticsVsHEDiagnosticN}{\textcolor{black}{1200}}
\newcommand{\StatisticsVsHEEntity}{\textcolor{black}{2.15}}
\newcommand{\StatisticsVsHEEntityProse}{\textcolor{black}{2.2}}
\newcommand{\StatisticsVsHEEntityCI}{\textcolor{black}{-0.50 to 4.80}}
\newcommand{\StatisticsVsHEEntityN}{\textcolor{black}{604}}
\newcommand{\StatisticsVsHEFinding}{\textcolor{black}{0.01}}
\newcommand{\StatisticsVsHEFindingCI}{\textcolor{black}{-2.11 to 2.23}}
\newcommand{\StatisticsVsHEFindingN}{\textcolor{black}{997}}
\newcommand{\StatisticsVsHEGrade}{\textcolor{black}{0.80}}
\newcommand{\StatisticsVsHEGradeCI}{\textcolor{black}{-2.36 to 3.98}}
\newcommand{\StatisticsVsHEGradeN}{\textcolor{black}{472}}
\newcommand{\OverlaysVsHECancer}{\textcolor{black}{-3.50}}
\newcommand{\OverlaysVsHECancerCI}{\textcolor{black}{-6.00 to -1.17}}
\newcommand{\OverlaysVsHECancerN}{\textcolor{black}{1200}}
\newcommand{\OverlaysVsHEDiagnostic}{\textcolor{black}{-2.05}}
\newcommand{\OverlaysVsHEDiagnosticCI}{\textcolor{black}{-3.10 to -1.00}}
\newcommand{\OverlaysVsHEDiagnosticN}{\textcolor{black}{1200}}
\newcommand{\OverlaysVsHEEntity}{\textcolor{black}{-2.81}}
\newcommand{\OverlaysVsHEEntityCI}{\textcolor{black}{-6.13 to 0.33}}
\newcommand{\OverlaysVsHEEntityN}{\textcolor{black}{604}}
\newcommand{\OverlaysVsHEFinding}{\textcolor{black}{-1.30}}
\newcommand{\OverlaysVsHEFindingCI}{\textcolor{black}{-3.51 to 0.90}}
\newcommand{\OverlaysVsHEFindingN}{\textcolor{black}{997}}
\newcommand{\OverlaysVsHEGrade}{\textcolor{black}{-0.64}}
\newcommand{\OverlaysVsHEGradeCI}{\textcolor{black}{-3.62 to 2.19}}
\newcommand{\OverlaysVsHEGradeN}{\textcolor{black}{472}}
\newcommand{\FullVsStatisticsCancer}{\textcolor{black}{3.33}}
\newcommand{\FullVsStatisticsCancerCI}{\textcolor{black}{0.92--5.67}}
\newcommand{\FullVsStatisticsCancerN}{\textcolor{black}{1200}}
\newcommand{\FullVsStatisticsDiagnostic}{\textcolor{black}{2.04}}
\newcommand{\FullVsStatisticsDiagnosticCI}{\textcolor{black}{0.98--3.10}}
\newcommand{\FullVsStatisticsDiagnosticN}{\textcolor{black}{1200}}
\newcommand{\FullVsStatisticsEntity}{\textcolor{black}{1.16}}
\newcommand{\FullVsStatisticsEntityCI}{\textcolor{black}{-1.82 to 4.30}}
\newcommand{\FullVsStatisticsEntityN}{\textcolor{black}{604}}
\newcommand{\FullVsStatisticsFinding}{\textcolor{black}{5.74}}
\newcommand{\FullVsStatisticsFindingProse}{\textcolor{black}{5.7}}
\newcommand{\FullVsStatisticsFindingCI}{\textcolor{black}{3.66--7.76}}
\newcommand{\FullVsStatisticsFindingN}{\textcolor{black}{997}}
\newcommand{\FullVsStatisticsGrade}{\textcolor{black}{12.04}}
\newcommand{\FullVsStatisticsGradeProse}{\textcolor{black}{12.0}}
\newcommand{\FullVsStatisticsGradeCI}{\textcolor{black}{8.62--15.37}}
\newcommand{\FullVsStatisticsGradeN}{\textcolor{black}{472}}
\newcommand{\FullVsOverlaysCancer}{\textcolor{black}{7.92}}
\newcommand{\FullVsOverlaysCancerCI}{\textcolor{black}{5.67--10.17}}
\newcommand{\FullVsOverlaysCancerN}{\textcolor{black}{1200}}
\newcommand{\FullVsOverlaysDiagnostic}{\textcolor{black}{4.18}}
\newcommand{\FullVsOverlaysDiagnosticCI}{\textcolor{black}{3.09--5.25}}
\newcommand{\FullVsOverlaysDiagnosticN}{\textcolor{black}{1200}}
\newcommand{\FullVsOverlaysEntity}{\textcolor{black}{6.13}}
\newcommand{\FullVsOverlaysEntityCI}{\textcolor{black}{3.15--9.27}}
\newcommand{\FullVsOverlaysEntityN}{\textcolor{black}{604}}
\newcommand{\FullVsOverlaysFinding}{\textcolor{black}{7.05}}
\newcommand{\FullVsOverlaysFindingCI}{\textcolor{black}{4.80--9.18}}
\newcommand{\FullVsOverlaysFindingN}{\textcolor{black}{997}}
\newcommand{\FullVsOverlaysGrade}{\textcolor{black}{13.48}}
\newcommand{\FullVsOverlaysGradeCI}{\textcolor{black}{10.12--16.65}}
\newcommand{\FullVsOverlaysGradeN}{\textcolor{black}{472}}
\newcommand{\ZoomVsExpansionDiagnostic}{\textcolor{black}{1.17}}
\newcommand{\ZoomVsExpansionDiagnosticCI}{\textcolor{black}{-0.68 to 3.04}}
\newcommand{\ZoomVsExpansionDiagnosticN}{432}
\newcommand{\ZoomVsExpansionGrade}{\textcolor{black}{3.50}}
\newcommand{\ZoomVsExpansionGradeCI}{\textcolor{black}{-1.72 to 8.89}}
\newcommand{\ZoomVsExpansionGradeN}{\textcolor{black}{166}}
\newcommand{\KnowledgeVsGradeDiagnostic}{\textcolor{black}{2.44}}
\newcommand{\KnowledgeVsGradeDiagnosticCI}{\textcolor{black}{1.48--3.41}}
\newcommand{\KnowledgeVsGradeDiagnosticN}{\textcolor{black}{1200}}
\newcommand{\KnowledgeVsGradeGrade}{\textcolor{black}{9.07}}
\newcommand{\KnowledgeVsGradeGradeProse}{\textcolor{black}{9.1}}
\newcommand{\KnowledgeVsGradeGradeCI}{\textcolor{black}{6.09--12.00}}
\newcommand{\KnowledgeVsGradeGradeN}{\textcolor{black}{472}}
\newcommand{\FullVsCorePublicContradiction}{0.89}
\newcommand{\FullVsCorePublicContradictionCI}{-0.59 to 2.39}
\newcommand{\FullVsCorePublicContradictionN}{950}
\newcommand{\FullVsStatisticsContradiction}{\textcolor{black}{-0.92}}
\newcommand{\FullVsStatisticsContradictionCI}{\textcolor{black}{-2.89 to 1.07}}
\newcommand{\FullVsStatisticsContradictionN}{\textcolor{black}{652}}
\newcommand{\FullVsOverlaysContradiction}{\textcolor{black}{-5.16}}
\newcommand{\FullVsOverlaysContradictionCI}{\textcolor{black}{-7.81 to -2.62}}
\newcommand{\FullVsOverlaysContradictionN}{\textcolor{black}{658}}
\newcommand{\FullVsOverlaysCorrectFinding}{\textcolor{black}{6.52}}
\newcommand{\FullVsOverlaysCorrectFindingCI}{\textcolor{black}{4.45--8.55}}
\newcommand{\FullVsOverlaysCorrectFindingN}{\textcolor{black}{997}}
\newcommand{\ExpansionVsGuardCancer}{\textcolor{black}{3.47}}
\newcommand{\ExpansionVsGuardCancerCI}{\textcolor{black}{0.00--7.18}}
\newcommand{\ExpansionVsGuardCancerN}{432}
\newcommand{\ZoomVsExpansionCancer}{\textcolor{black}{1.85}}
\newcommand{\ZoomVsExpansionCancerCI}{\textcolor{black}{-1.85 to 5.32}}
\newcommand{\ZoomVsExpansionCancerN}{432}
\newcommand{\GradeEvidenceVsCoreFinding}{\textcolor{black}{-0.30}}
\newcommand{\GradeEvidenceVsCoreFindingCI}{\textcolor{black}{-2.33 to 1.65}}
\newcommand{\GradeEvidenceVsCoreFindingN}{\textcolor{black}{997}}
\newcommand{\GradeEvidenceVsCoreGrade}{\textcolor{black}{-2.35}}
\newcommand{\GradeEvidenceVsCoreGradeCI}{\textcolor{black}{-4.97 to 0.12}}
\newcommand{\GradeEvidenceVsCoreGradeN}{\textcolor{black}{472}}
\newcommand{\KnowledgeVsGradeFinding}{\textcolor{black}{2.36}}
\newcommand{\KnowledgeVsGradeFindingProse}{\textcolor{black}{2.4}}
\newcommand{\KnowledgeVsGradeFindingCI}{\textcolor{black}{0.55--4.17}}
\newcommand{\KnowledgeVsGradeFindingN}{\textcolor{black}{997}}
\newcommand{\PublicFindingVsQwen}{5.49}
\newcommand{\PublicFindingVsQwenCI}{3.69--7.29}
\newcommand{\PublicFindingVsQwenN}{1405}
\newcommand{\EvaluationFindingVsQwen}{5.75}
\newcommand{\EvaluationFindingVsQwenProse}{5.7}
\newcommand{\EvaluationFindingVsQwenCI}{3.63--7.87}
\newcommand{\EvaluationFindingVsQwenN}{997}
\newcommand{\PublicFindingVsCore}{2.60}
\newcommand{\PublicFindingVsCoreCI}{1.02--4.19}
\newcommand{\PublicFindingVsCoreN}{1405}
\newcommand{\EvaluationFindingVsCore}{2.07}
\newcommand{\EvaluationFindingVsCoreCI}{0.29--3.88}
\newcommand{\EvaluationFindingVsCoreN}{997}
\newcommand{\PublicAssertionVsQwen}{4.10}
\newcommand{\PublicAssertionVsQwenCI}{2.47--5.77}
\newcommand{\PublicAssertionVsQwenN}{1405}
\newcommand{\EvaluationAssertionVsQwen}{4.29}
\newcommand{\EvaluationAssertionVsQwenCI}{2.42--6.31}
\newcommand{\EvaluationAssertionVsQwenN}{997}
\newcommand{\PublicAssertionVsCore}{3.63}
\newcommand{\PublicAssertionVsCoreCI}{2.07--5.25}
\newcommand{\PublicAssertionVsCoreN}{1405}
\newcommand{\EvaluationAssertionVsCore}{3.22}
\newcommand{\EvaluationAssertionVsCoreCI}{1.48--4.95}
\newcommand{\EvaluationAssertionVsCoreN}{997}
\newcommand{\PublicContradictionVsQwen}{-5.79}
\newcommand{\PublicContradictionVsQwenCI}{-7.91 to -3.79}
\newcommand{\PublicContradictionVsQwenN}{938}
\newcommand{\PublicContradictionReductionVsQwen}{5.79}
\newcommand{\EvaluationContradictionVsQwen}{-5.52}
\newcommand{\EvaluationContradictionVsQwenCI}{-8.02 to -3.11}
\newcommand{\EvaluationContradictionVsQwenN}{664}
\newcommand{\EvaluationContradictionReductionVsQwen}{5.52}
\newcommand{\EvaluationContradictionReductionVsQwenProse}{5.5}
\newcommand{\PublicContradictionVsCore}{0.89}
\newcommand{\PublicContradictionVsCoreCI}{-0.59 to 2.39}
\newcommand{\PublicContradictionVsCoreN}{950}
\newcommand{\EvaluationContradictionVsCore}{1.38}
\newcommand{\EvaluationContradictionVsCoreCI}{-0.34 to 3.11}
\newcommand{\EvaluationContradictionVsCoreN}{686}
\newcommand{\ExpansionVsGuardGrade}{\textcolor{black}{9.25}}
\newcommand{\ExpansionVsGuardGradeProse}{\textcolor{black}{9.2}}
\newcommand{\ExpansionVsGuardGradeCI}{\textcolor{black}{3.84--14.84}}
\newcommand{\ExpansionVsGuardGradeN}{\textcolor{black}{166}}
\newcommand{\EvaluationNativeN}{215}
\newcommand{\EvaluationNativeHistGenFindingFOne}{17.21}
\newcommand{\EvaluationNativeHistGenFindingFOneProse}{17.2}
\newcommand{\EvaluationNativeHistGenFindingFOneCI}{13.13--21.54}
\newcommand{\EvaluationNativeHistGenFindingFOneN}{173}
\newcommand{\EvaluationNativeHistoGPTFindingFOne}{11.98}
\newcommand{\EvaluationNativeHistoGPTFindingFOneProse}{12.0}
\newcommand{\EvaluationNativeHistoGPTFindingFOneCI}{8.43--15.73}
\newcommand{\EvaluationNativeHistoGPTFindingFOneN}{173}
\newcommand{\EvaluationNativeCoreFindingFOne}{43.19}
\newcommand{\EvaluationNativeCoreFindingFOneCI}{38.01--48.50}
\newcommand{\EvaluationNativeCoreFindingFOneN}{173}
\newcommand{\EvaluationNativeFullFindingFOne}{43.26}
\newcommand{\EvaluationNativeFullFindingFOneProse}{43.3}
\newcommand{\EvaluationNativeFullFindingFOneCI}{38.01--48.68}
\newcommand{\EvaluationNativeFullFindingFOneN}{173}
\newcommand{\ReaderCompleteFullSuccess}{\textcolor{black}{62.1}}
\newcommand{\ReaderCompleteFullSuccessCI}{\textcolor{black}{54.5--69.7}}
\newcommand{\ReaderCompleteFullSuccessCount}{\textcolor{black}{164}}
\newcommand{\ReaderCompleteFullFailure}{\textcolor{black}{37.9}}
\newcommand{\ReaderCompleteFullN}{264}
\newcommand{\ReaderCompleteCoreSuccess}{\textcolor{black}{39.8}}
\newcommand{\ReaderCompleteCoreSuccessCI}{\textcolor{black}{32.6--47.0}}
\newcommand{\ReaderCompleteCoreSuccessCount}{\textcolor{black}{105}}
\newcommand{\ReaderCompleteCoreFailure}{\textcolor{black}{60.2}}
\newcommand{\ReaderCompleteCoreN}{264}
\newcommand{\ReaderCompleteQwenSuccess}{\textcolor{black}{36.0}}
\newcommand{\ReaderCompleteQwenSuccessCI}{\textcolor{black}{29.2--43.2}}
\newcommand{\ReaderCompleteQwenSuccessCount}{\textcolor{black}{95}}
\newcommand{\ReaderCompleteQwenFailure}{\textcolor{black}{64.0}}
\newcommand{\ReaderCompleteQwenN}{264}
\newcommand{\ReaderCompleteMedGemmaSuccess}{31.1}
\newcommand{\ReaderCompleteMedGemmaSuccessCI}{23.1--39.0}
\newcommand{\ReaderCompleteMedGemmaSuccessCount}{82}
\newcommand{\ReaderCompleteMedGemmaFailure}{68.9}
\newcommand{\ReaderCompleteMedGemmaN}{264}
\newcommand{\ReaderCompletePLIPSuccess}{\textcolor{black}{14.0}}
\newcommand{\ReaderCompletePLIPSuccessCI}{\textcolor{black}{9.8--18.6}}
\newcommand{\ReaderCompletePLIPSuccessCount}{\textcolor{black}{37}}
\newcommand{\ReaderCompletePLIPFailure}{\textcolor{black}{86.0}}
\newcommand{\ReaderCompletePLIPN}{264}
\newcommand{\ReaderNoMinorFull}{83.3}
\newcommand{\ReaderNoMinorFullCI}{78.8--87.9}
\newcommand{\ReaderRankFull}{1.64}
\newcommand{\ReaderNoMinorCore}{74.2}
\newcommand{\ReaderNoMinorCoreCI}{69.3--79.2}
\newcommand{\ReaderRankCore}{1.60}
\newcommand{\ReaderNoMinorQwen}{73.1}
\newcommand{\ReaderNoMinorQwenCI}{67.8--78.0}
\newcommand{\ReaderRankQwen}{1.63}
\newcommand{\ReaderNoMinorMedGemma}{87.9}
\newcommand{\ReaderNoMinorMedGemmaCI}{83.3--92.0}
\newcommand{\ReaderRankMedGemma}{2.76}
\newcommand{\ReaderNoMinorPLIP}{49.6}
\newcommand{\ReaderNoMinorPLIPCI}{45.4--53.8}
\newcommand{\ReaderRankPLIP}{3.29}

\newcommand{\ScopeCuratedVsQwen}{\textcolor{black}{4.83}}
\newcommand{\ScopeCuratedVsQwenCI}{\textcolor{black}{0.00--9.38}}
\newcommand{\ScopeOCRVsQwen}{\textcolor{black}{2.48}}
\newcommand{\ScopeOCRVsQwenCI}{\textcolor{black}{-0.83 to 5.79}}
\newcommand{\ScopeStatisticsVsHE}{\textcolor{black}{3.98}}
\newcommand{\ScopeStatisticsVsHECI}{\textcolor{black}{0.00--7.95}}
\newcommand{\ScopePrivateOneVsQwen}{11.85}
\newcommand{\ScopePrivateOneVsQwenCI}{\textcolor{black}{4.44--19.26}}
\newcommand{\ScopePrivateTwoVsQwen}{11.76}
\newcommand{\ScopePrivateTwoVsQwenCI}{\textcolor{black}{4.41--19.12}}

\newcommand{\ScopeCancerVsQwen}{\textcolor{black}{2.78}}
\newcommand{\ScopeCancerVsQwenCI}{\textcolor{black}{0.66--4.89}}
\newcommand{\ScopeCancerVsDeepSeek}{\textcolor{black}{0.42}}
\newcommand{\ScopeCancerVsDeepSeekCI}{\textcolor{black}{-1.75 to 2.66}}
\newcommand{\ScopeCancerCuratedVsQwen}{\textcolor{black}{3.29}}
\newcommand{\ScopeCancerCuratedVsQwenCI}{\textcolor{black}{-0.66 to 7.46}}
\newcommand{\ScopeCancerOCRVsQwen}{\textcolor{black}{2.58}}
\newcommand{\ScopeCancerOCRVsQwenCI}{\textcolor{black}{0.08--5.09}}
\newcommand{\ScopeCancerStatisticsVsHE}{\textcolor{black}{3.29}}
\newcommand{\ScopeCancerStatisticsVsHECI}{\textcolor{black}{0.00--6.58}}

\newcommand{\NativePublicN}{312}
\newcommand{\NativeHistGenSharedN}{314}
\newcommand{\NativePublicFullCompatibility}{79.5}
\newcommand{\NativePublicFullCompatibilityCI}{\textcolor{black}{74.6--83.8}}
\newcommand{\NativePublicFullCompatibilityN}{312}
\newcommand{\NativePublicHistGenCompatibility}{\textcolor{black}{36.2}}
\newcommand{\NativePublicHistGenCompatibilityCI}{\textcolor{black}{30.9--41.8}}
\newcommand{\NativePublicHistGenCompatibilityN}{312}
\newcommand{\NativePublicHistoGPTCompatibility}{4.2}
\newcommand{\NativePublicHistoGPTCompatibilityCI}{\textcolor{black}{2.2--7.0}}
\newcommand{\NativePublicHistoGPTCompatibilityN}{312}
\newcommand{\NativePublicVsHistGenCompatibility}{\textcolor{black}{43.3}}
\newcommand{\NativePublicVsHistGenCompatibilityCI}{\textcolor{black}{36.5--50.3}}
\newcommand{\NativePublicVsHistGenCompatibilityN}{312}
\newcommand{\NativePublicVsHistoGPTCompatibility}{75.3}
\newcommand{\NativePublicVsHistoGPTCompatibilityCI}{70.5--79.8}
\newcommand{\NativePublicVsHistoGPTCompatibilityN}{312}
\newcommand{\NativePublicFullFindingFOne}{45.7}
\newcommand{\NativePublicFullFindingFOneCI}{41.2--49.9}
\newcommand{\NativePublicFullFindingFOneN}{256}
\newcommand{\NativePublicHistGenFindingFOne}{17.2}
\newcommand{\NativePublicHistGenFindingFOneCI}{13.6--20.8}
\newcommand{\NativePublicHistGenFindingFOneN}{256}
\newcommand{\NativePublicHistoGPTFindingFOne}{12.3}
\newcommand{\NativePublicHistoGPTFindingFOneCI}{9.2--15.6}
\newcommand{\NativePublicHistoGPTFindingFOneN}{256}
\newcommand{\NativePublicVsHistGenFindingFOne}{28.5}
\newcommand{\NativePublicVsHistGenFindingFOneCI}{23.0--33.8}
\newcommand{\NativePublicVsHistGenFindingFOneN}{256}
\newcommand{\NativePublicVsHistoGPTFindingFOne}{33.4}
\newcommand{\NativePublicVsHistoGPTFindingFOneCI}{28.7--37.9}
\newcommand{\NativePublicVsHistoGPTFindingFOneN}{256}
\newcommand{\NativePublicFullAssertionCoverage}{44.9}
\newcommand{\NativePublicFullAssertionCoverageCI}{40.5--49.4}
\newcommand{\NativePublicFullAssertionCoverageN}{256}
\newcommand{\NativePublicHistGenAssertionCoverage}{21.0}
\newcommand{\NativePublicHistGenAssertionCoverageCI}{17.2--25.0}
\newcommand{\NativePublicHistGenAssertionCoverageN}{256}
\newcommand{\NativePublicHistoGPTAssertionCoverage}{10.5}
\newcommand{\NativePublicHistoGPTAssertionCoverageCI}{7.7--13.5}
\newcommand{\NativePublicHistoGPTAssertionCoverageN}{256}
\newcommand{\NativePublicVsHistGenAssertionCoverage}{23.9}
\newcommand{\NativePublicVsHistGenAssertionCoverageCI}{18.8--29.3}
\newcommand{\NativePublicVsHistGenAssertionCoverageN}{256}
\newcommand{\NativePublicVsHistoGPTAssertionCoverage}{34.4}
\newcommand{\NativePublicVsHistoGPTAssertionCoverageCI}{29.9--39.1}
\newcommand{\NativePublicVsHistoGPTAssertionCoverageN}{256}
\newcommand{\NativePublicFullCorrectCoverage}{38.2}
\newcommand{\NativePublicFullCorrectCoverageCI}{34.3--42.5}
\newcommand{\NativePublicFullCorrectCoverageN}{256}
\newcommand{\NativePublicHistGenCorrectCoverage}{13.7}
\newcommand{\NativePublicHistGenCorrectCoverageCI}{10.9--17.0}
\newcommand{\NativePublicHistGenCorrectCoverageN}{256}
\newcommand{\NativePublicHistoGPTCorrectCoverage}{9.6}
\newcommand{\NativePublicHistoGPTCorrectCoverageCI}{7.0--12.5}
\newcommand{\NativePublicHistoGPTCorrectCoverageN}{256}
\newcommand{\NativePublicVsHistGenCorrectCoverage}{24.5}
\newcommand{\NativePublicVsHistGenCorrectCoverageCI}{19.4--29.6}
\newcommand{\NativePublicVsHistGenCorrectCoverageN}{256}
\newcommand{\NativePublicVsHistoGPTCorrectCoverage}{28.6}
\newcommand{\NativePublicVsHistoGPTCorrectCoverageCI}{24.4--32.9}
\newcommand{\NativePublicVsHistoGPTCorrectCoverageN}{256}
\newcommand{\NativePublicFullContradiction}{12.7}
\newcommand{\NativePublicFullContradictionCI}{9.0--16.6}
\newcommand{\NativePublicFullContradictionN}{193}
\newcommand{\NativePublicHistGenContradiction}{32.4}
\newcommand{\NativePublicHistGenContradictionCI}{24.1--40.6}
\newcommand{\NativePublicHistGenContradictionN}{106}
\newcommand{\NativePublicHistoGPTContradiction}{8.3}
\newcommand{\NativePublicHistoGPTContradictionCI}{1.7--15.0}
\newcommand{\NativePublicHistoGPTContradictionN}{60}
\newcommand{\NativePublicVsHistGenContradiction}{-18.0}
\newcommand{\NativePublicVsHistGenContradictionCI}{-28.7 to -7.5}
\newcommand{\NativePublicVsHistGenContradictionN}{93}
\newcommand{\NativePublicVsHistoGPTContradiction}{3.0}
\newcommand{\NativePublicVsHistoGPTContradictionCI}{-3.2--8.8}
\newcommand{\NativePublicVsHistoGPTContradictionN}{56}
\newcommand{\NativePublicFullGrade}{\textcolor{black}{32.8}}
\newcommand{\NativePublicFullGradeCI}{\textcolor{black}{17.7--57.3}}
\newcommand{\NativePublicFullGradeN}{\textcolor{black}{76}}
\newcommand{\NativePublicHistGenGrade}{\textcolor{black}{23.5}}
\newcommand{\NativePublicHistGenGradeCI}{\textcolor{black}{9.4--48.8}}
\newcommand{\NativePublicHistGenGradeN}{\textcolor{black}{76}}
\newcommand{\NativePublicHistoGPTGrade}{0.0}
\newcommand{\NativePublicHistoGPTGradeCI}{\textcolor{black}{0.0--20.6}}
\newcommand{\NativePublicHistoGPTGradeN}{\textcolor{black}{76}}
\newcommand{\NativePublicVsHistGenGrade}{\textcolor{black}{9.3}}
\newcommand{\NativePublicVsHistGenGradeCI}{\textcolor{black}{-5.0--24.3}}
\newcommand{\NativePublicVsHistGenGradeN}{\textcolor{black}{76}}
\newcommand{\NativePublicVsHistoGPTGrade}{\textcolor{black}{32.8}}
\newcommand{\NativePublicVsHistoGPTGradeCI}{\textcolor{black}{23.4--42.3}}
\newcommand{\NativePublicVsHistoGPTGradeN}{\textcolor{black}{76}}
\newcommand{\NativePrivatePOneEntity}{\textcolor{black}{67.4}}
\newcommand{\NativePrivatePOneEntityCI}{\textcolor{black}{58.8--75.2}}
\newcommand{\NativePrivatePOneEntityN}{135}
\newcommand{\NativePrivatePOneFindingFOne}{76.6}
\newcommand{\NativePrivatePOneFindingFOneCI}{71.2--81.6}
\newcommand{\NativePrivatePOneFindingFOneN}{150}
\newcommand{\NativePrivatePOneAssertionCoverage}{92.7}
\newcommand{\NativePrivatePOneAssertionCoverageCI}{89.3--95.7}
\newcommand{\NativePrivatePOneAssertionCoverageN}{150}
\newcommand{\NativePrivatePOneCorrectCoverage}{75.0}
\newcommand{\NativePrivatePOneCorrectCoverageCI}{69.7--80.3}
\newcommand{\NativePrivatePOneCorrectCoverageN}{150}
\newcommand{\NativePrivatePOneContradiction}{18.7}
\newcommand{\NativePrivatePOneContradictionCI}{13.9--24.1}
\newcommand{\NativePrivatePOneContradictionN}{147}
\newcommand{\NativePrivatePOneGrade}{\textcolor{black}{27.4}}
\newcommand{\NativePrivatePOneGradeCI}{\textcolor{black}{15.8--72.4}}
\newcommand{\NativePrivatePOneGradeN}{76}
\newcommand{\NativePrivatePOneHistGenEntity}{\textcolor{black}{5.2}}
\newcommand{\NativePrivatePOneHistGenEntityCI}{\textcolor{black}{2.1--10.4}}
\newcommand{\NativePrivatePOneHistGenEntityN}{135}
\newcommand{\NativePrivatePOneHistoGPTEntity}{0.0}
\newcommand{\NativePrivatePOneHistoGPTEntityCI}{\textcolor{black}{0.0--2.7}}
\newcommand{\NativePrivatePOneHistoGPTEntityN}{135}
\newcommand{\NativePrivatePOneVsHistGenEntity}{\textcolor{black}{62.2}}
\newcommand{\NativePrivatePOneVsHistGenEntityCI}{\textcolor{black}{53.3--71.1}}
\newcommand{\NativePrivatePOneVsHistGenEntityN}{135}
\newcommand{\NativePrivatePOneVsHistoGPTEntity}{\textcolor{black}{67.4}}
\newcommand{\NativePrivatePOneVsHistoGPTEntityCI}{\textcolor{black}{59.3--75.6}}
\newcommand{\NativePrivatePOneVsHistoGPTEntityN}{135}
\newcommand{\NativePrivatePTwoEntity}{\textcolor{black}{67.6}}
\newcommand{\NativePrivatePTwoEntityCI}{\textcolor{black}{59.1--75.4}}
\newcommand{\NativePrivatePTwoEntityN}{136}
\newcommand{\NativePrivatePTwoFindingFOne}{78.9}
\newcommand{\NativePrivatePTwoFindingFOneCI}{73.7--83.6}
\newcommand{\NativePrivatePTwoFindingFOneN}{151}
\newcommand{\NativePrivatePTwoAssertionCoverage}{91.5}
\newcommand{\NativePrivatePTwoAssertionCoverageCI}{88.1--94.6}
\newcommand{\NativePrivatePTwoAssertionCoverageN}{151}
\newcommand{\NativePrivatePTwoCorrectCoverage}{77.3}
\newcommand{\NativePrivatePTwoCorrectCoverageCI}{72.1--82.3}
\newcommand{\NativePrivatePTwoCorrectCoverageN}{151}
\newcommand{\NativePrivatePTwoContradiction}{16.2}
\newcommand{\NativePrivatePTwoContradictionCI}{11.8--21.3}
\newcommand{\NativePrivatePTwoContradictionN}{148}
\newcommand{\NativePrivatePTwoGrade}{\textcolor{black}{42.2}}
\newcommand{\NativePrivatePTwoGradeCI}{\textcolor{black}{25.1--82.7}}
\newcommand{\NativePrivatePTwoGradeN}{76}
\newcommand{\NativePrivatePTwoHistGenEntity}{\textcolor{black}{5.1}}
\newcommand{\NativePrivatePTwoHistGenEntityCI}{\textcolor{black}{2.1--10.3}}
\newcommand{\NativePrivatePTwoHistGenEntityN}{136}
\newcommand{\NativePrivatePTwoHistoGPTEntity}{0.0}
\newcommand{\NativePrivatePTwoHistoGPTEntityCI}{\textcolor{black}{0.0--2.7}}
\newcommand{\NativePrivatePTwoHistoGPTEntityN}{136}
\newcommand{\NativePrivatePTwoVsHistGenEntity}{\textcolor{black}{62.5}}
\newcommand{\NativePrivatePTwoVsHistGenEntityCI}{\textcolor{black}{54.4--70.6}}
\newcommand{\NativePrivatePTwoVsHistGenEntityN}{136}
\newcommand{\NativePrivatePTwoVsHistoGPTEntity}{\textcolor{black}{67.6}}
\newcommand{\NativePrivatePTwoVsHistoGPTEntityCI}{\textcolor{black}{59.6--75.0}}
\newcommand{\NativePrivatePTwoVsHistoGPTEntityN}{136}
\newcommand{\NativePrivatePOneAvailableEntity}{\textcolor{black}{66.9}}
\newcommand{\NativePrivatePOneAvailableEntityCI}{\textcolor{black}{58.3--74.7}}
\newcommand{\NativePrivatePOneAvailableEntityN}{136}
\newcommand{\NativePrivatePOneAvailableFindingFOne}{76.0}
\newcommand{\NativePrivatePOneAvailableFindingFOneCI}{70.8--81.0}
\newcommand{\NativePrivatePOneAvailableFindingFOneN}{151}
\newcommand{\NativePrivatePOneAvailableAssertionCoverage}{92.7}
\newcommand{\NativePrivatePOneAvailableAssertionCoverageCI}{89.4--95.7}
\newcommand{\NativePrivatePOneAvailableAssertionCoverageN}{151}
\newcommand{\NativePrivatePOneAvailableCorrectCoverage}{74.5}
\newcommand{\NativePrivatePOneAvailableCorrectCoverageCI}{68.9--79.5}
\newcommand{\NativePrivatePOneAvailableCorrectCoverageN}{151}
\newcommand{\NativePrivatePOneAvailableContradiction}{19.3}
\newcommand{\NativePrivatePOneAvailableContradictionCI}{14.2--24.7}
\newcommand{\NativePrivatePOneAvailableContradictionN}{148}
\newcommand{\NativePrivatePOneAvailableGrade}{\textcolor{black}{27.4}}
\newcommand{\NativePrivatePOneAvailableGradeCI}{\textcolor{black}{15.8--72.4}}
\newcommand{\NativePrivatePOneAvailableGradeN}{76}
\newcommand{\NativePrivatePTwoAvailableEntity}{\textcolor{black}{67.2}}
\newcommand{\NativePrivatePTwoAvailableEntityCI}{\textcolor{black}{58.6--74.9}}
\newcommand{\NativePrivatePTwoAvailableEntityN}{137}
\newcommand{\NativePrivatePTwoAvailableFindingFOne}{78.4}
\newcommand{\NativePrivatePTwoAvailableFindingFOneCI}{73.2--83.2}
\newcommand{\NativePrivatePTwoAvailableFindingFOneN}{152}
\newcommand{\NativePrivatePTwoAvailableAssertionCoverage}{91.6}
\newcommand{\NativePrivatePTwoAvailableAssertionCoverageCI}{88.2--94.7}
\newcommand{\NativePrivatePTwoAvailableAssertionCoverageN}{152}
\newcommand{\NativePrivatePTwoAvailableCorrectCoverage}{76.8}
\newcommand{\NativePrivatePTwoAvailableCorrectCoverageCI}{71.4--81.9}
\newcommand{\NativePrivatePTwoAvailableCorrectCoverageN}{152}
\newcommand{\NativePrivatePTwoAvailableContradiction}{16.8}
\newcommand{\NativePrivatePTwoAvailableContradictionCI}{12.1--21.5}
\newcommand{\NativePrivatePTwoAvailableContradictionN}{149}
\newcommand{\NativePrivatePTwoAvailableGrade}{\textcolor{black}{42.2}}
\newcommand{\NativePrivatePTwoAvailableGradeCI}{\textcolor{black}{25.1--82.7}}
\newcommand{\NativePrivatePTwoAvailableGradeN}{76}

\title[PathXAI]{PathXAI: Evidence-guided structured pathology reporting by transferable tissue segmentation}

\author*[1]{\fnm{Zhaorui} \sur{Tan}}\email{tan\_zhaorui@a-star.edu.sg}
\author[1]{\fnm{Jun Aun} \sur{Tan}}
\author[2]{\fnm{Weixiang} \sur{Zhong}}
\author[2]{\fnm{Xiaoxi} \sur{Wang}}
\author[1,3]{\fnm{Tan} \sur{Pan}}
\author[1]{\fnm{Xingdong} \sur{Xue}}
\author*[4]{\fnm{Li} \sur{Liu}}\email{dscliuli@cqu.edu.cn}
\author*[5]{\fnm{Jing} \sur{Zhang}}\email{zhangjing@smmu.edu.cn}
\author*[2]{\fnm{Jing} \sur{Zhang}}\email{jzhang1989@zju.edu.cn}
\author*[1]{\fnm{Weimiao} \sur{Yu}}\email{yu\_weimiao@a-star.edu.sg}

\affil[1]{\orgname{Intelligent Digital and Molecular Pathology Lab, Bioinformatics Institute, A$^*$STAR},
\orgaddress{\country{Singapore}}}
\affil[2]{\orgname{Department of Pathology, The First Affiliated Hospital, Zhejiang University School of Medicine},
\orgaddress{\city{Hangzhou}, \country{China}}}
\affil[3]{\orgname{Artificial Intelligence Innovation and Incubation Institute, Fudan University}, \orgaddress{\city{Shanghai}, \country{China}}}
\affil[4]{\orgname{School of Big Data and Software Engineering, Chongqing University},
\orgaddress{\city{Chongqing}, \country{China}}}
\affil[5]{\orgname{Department of Pathology, Changzheng Hospital}, \orgaddress{\city{Shanghai}, \country{China}}}

\abstract{Pathology reporting translates tissue observations across regions and slides into diagnoses
and supporting morphological descriptions, \textcolor{black}{yet explicit tissue-composition measurements are rarely
used to guide H\&E-based report generation. We developed PathXAI, which pairs transferable tissue
segmentation with a pretrained vision--language model. Because supplying tissue information
directly with H\&E images gave limited benefit, PathXAI-Report uses segmentation overlays and
tissue statistics in separate steps, selects between candidate diagnoses by fixed rules,
acquires further images when evidence is insufficient and records every step for review,
without updating model weights.}
Its segmentation component exceeded baselines in mean
Dice on internal and external tests. \textcolor{black}{In automated evaluation on 1,200 public and 151 private cases,
PathXAI achieved higher agreement with reference diagnoses and documented morphology than
H\&E-based comparators. On the public cohort, it exceeded Qwen+H\&E by
\EvaluationVsQwenProse{} percentage points in cancer family match rate and
\EvaluationFindingVsQwenProse{} percentage points in finding F1.}
}
\maketitle
\relax % Keep title/abstract SyncTeX mappings in this included file.

\pdfbookmark[1]{Introduction}{introduction}
\phantomsection\label{sec:introduction}

Pathology reports translate histological observations into diagnostic
information that supports clinical decision-making. \textcolor{black}{Pathologists interpret
tissue structure across regions and often multiple slides, synthesizing findings into a
diagnosis and supporting descriptions}~\cite{tran2025histogpt}.
When examining a slide, pathologists distinguish tumour tissue from
surrounding components and assess their spatial relationships.
\textcolor{black}{This motivates report-generation systems that use explicit tissue evidence
and retain it for subsequent inspection.}

\textcolor{black}{Current methods for report generation from haematoxylin and eosin (H\&E)
images integrate visual information across tissue areas.}
\textcolor{black}{HistGen aggregates local and global image context, and HistoGPT generates skin-disease reports
from multiple whole-slide images (WSIs), linking text to image regions through saliency
maps~\cite{guo2024histgen,tran2025histogpt}. Multimodal pretraining with reports, gene expression
or clinical dialogue (mSTAR and PRISM2) and broader medical vision-language models (VLMs) such as
MedGemma further support pathology report generation~\cite{xu2025mstar,vorontsov2026prism2,sellergren2026medgemma}.
Because reporting also involves revisiting regions to resolve uncertainty, PathChat answers
questions about supplied images~\cite{lu2024pathchat}, and PathAgent selects new regions or
magnifications during inspection~\cite{Chen2025PathAgentTI}.}
\textcolor{black}{Despite these advances, the role of explicit tissue composition in guiding
report generation remains underexplored.}
In related pathology tasks, explicit modelling of cells, tissue regions and their spatial relationships improves breast-lesion
classification~\cite{Pati2021HierarchicalGR}. 
Quantifying tissue components
also provides prognostic information in colorectal
cancer~\cite{Kather2019PredictingSF}.
Together, these findings \textcolor{black}{suggest} possible advantages of incorporating explicit tissue information
into pathology report generation.

Tissue segmentation offers a way to make tissue composition and spatial organization explicit for pathology reporting.
\textcolor{black}{In radiology, MAIRA-Seg adds segmentation-mask information to a multimodal language model and
improves chest X-ray report generation~\cite{sharma2025mairaseg}.} 
Existing pathology segmentation models, including PathSegmentor and
VISTA-PATH, support diverse organ- and scale-specific
targets~\cite{chen2026pathsegmentor,liang2026vistapath}.
\textcolor{black}{Reporting, however, requires a small set of tissue categories that is consistent across regions,
slides and organs.}

\textcolor{black}{To investigate whether explicit tissue composition can guide reporting, we propose
PathXAI, which combines PathXAI-Seg for tissue segmentation with PathXAI-Report
for structured reporting across multiple slides}
(\cref{pan:overview-capabilities}).
PathXAI-Seg, trained on diverse public datasets, segments a shared set of tissue categories across datasets (\cref{tab:seg-label-mapping}).
It provides spatially aligned segmentation overlays and \textcolor{black}{the corresponding} tissue statistics for selected regions of interest (ROIs)
(\cref{pan:overview-workflow,supp:seg-inference}).
The reporting component, PathXAI-Report, uses a pretrained VLM to integrate tissue evidence with H\&E images and the known anatomical site for diagnostic comparison and targeted reinspection (\cref{pan:overview-workflow}).
\textcolor{black}{Rather than supplying all tissue evidence in one model call, PathXAI-Report assigns overlays
to candidate diagnoses and tissue statistics to an independent check, selects by fixed rules
between candidates generated with and without domain knowledge, and acquires additional regions
or zoomed views when evidence is insufficient.}
\textcolor{black}{For additional domain context, PathXAI optionally retrieves} disease definitions and morphological
descriptions from KEEP~\cite{zhou2026keep}
for comparing diagnostic hypotheses (\cref{supp:report-knowledge}).
The resulting reports combine disease classification, tumour subtype
and grade, when assessable, with descriptions of structure, cytology,
necrosis and invasion (\cref{pan:overview-workflow}).
\textcolor{black}{Every step is recorded, so each report statement can be traced to its regions, tissue evidence
and domain context}
(\cref{pan:overview-capabilities,supp:report-writing}).

We evaluated PathXAI-Seg on five internal and six external segmentation test sets.
PathXAI-Seg achieved \textcolor{black}{higher mean Dice than standard same-backbone fine-tuning} on both internal and
external tests (\cref{pan:overview-findings}). 
\textcolor{black}{We used 456 public cases for PathXAI-Report pipeline development and evaluated reports
on 1,200 separate public cases spanning 22 cancer types and 151 shared private cases.}
\textcolor{black}{Each private case had one reference report from each of two pathologists.}
\textcolor{black}{Separately}, the same two pathologists assessed preference and
correction requirements in \textcolor{black}{132 cases randomly sampled from the \textcolor{black}{PathXAI-Report pipeline development subset}} (\cref{pan:overview-findings}). 
\textcolor{black}{In both reporting cohorts, PathXAI-Report achieved higher agreement with reference
diagnoses and documented morphology than Qwen+H\&E.}
\textcolor[HTML]{000000}{Necrosis is a morphological finding represented in our segmentation labels.
In the public evaluation cohort, tissue statistics increased agreement with reference reports on
the presence or absence of necrosis from 43.1\% to 57.2\%.
Direct overlays yielded a smaller, uncertain increase to 47.7\% (\cref{tab:source_fig7:part-1}).
Diagnostic benefits of direct tissue inputs remained limited}, \textcolor{black}{whereas the complete PathXAI-Report pipeline achieved higher finding
F1 and balanced grade accuracy than either direct input.}
\textcolor{black}{Together, these findings support using tissue evidence within the reporting workflow
and offer a step towards more interpretable pathology reports with information available for review.}

\manuscriptwidefigurepart{first}{PathXAI integrates tissue segmentation with evidence-linked \textcolor[HTML]{000000}{diagnostic and morphological} reporting.}{fig:overview}{%
    \begin{minipage}{180mm}
    \manuscriptpanelblock{5mm}{54mm}{\manuscriptpanelmark{pan:overview-capabilities}}{Inputs and capabilities}{\includegraphics[width=\linewidth]{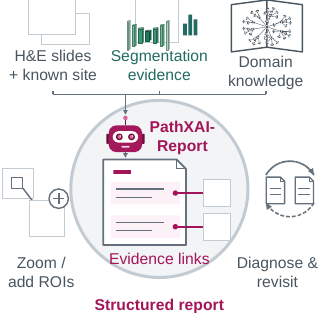}}\hspace{4mm}%
    \manuscriptpanelblock{5mm}{122mm}{\manuscriptpanelmark{pan:overview-workflow}}{Reporting with tissue evidence}{\includegraphics[width=\linewidth]{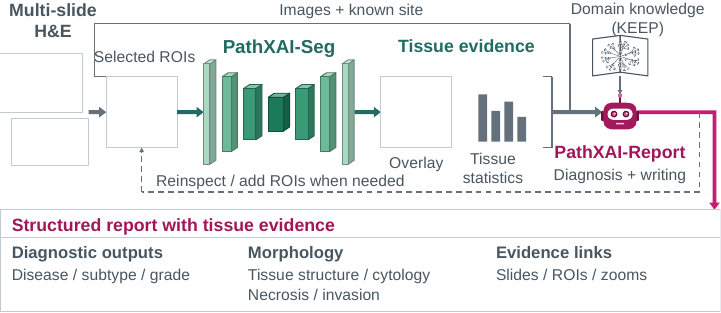}}
    \par\vspace{2mm}
    \manuscriptpanelblock{5mm}{180.0mm}{\manuscriptpanelmark{pan:overview-findings}}{Evaluation questions and findings}{\includegraphics[width=\linewidth]{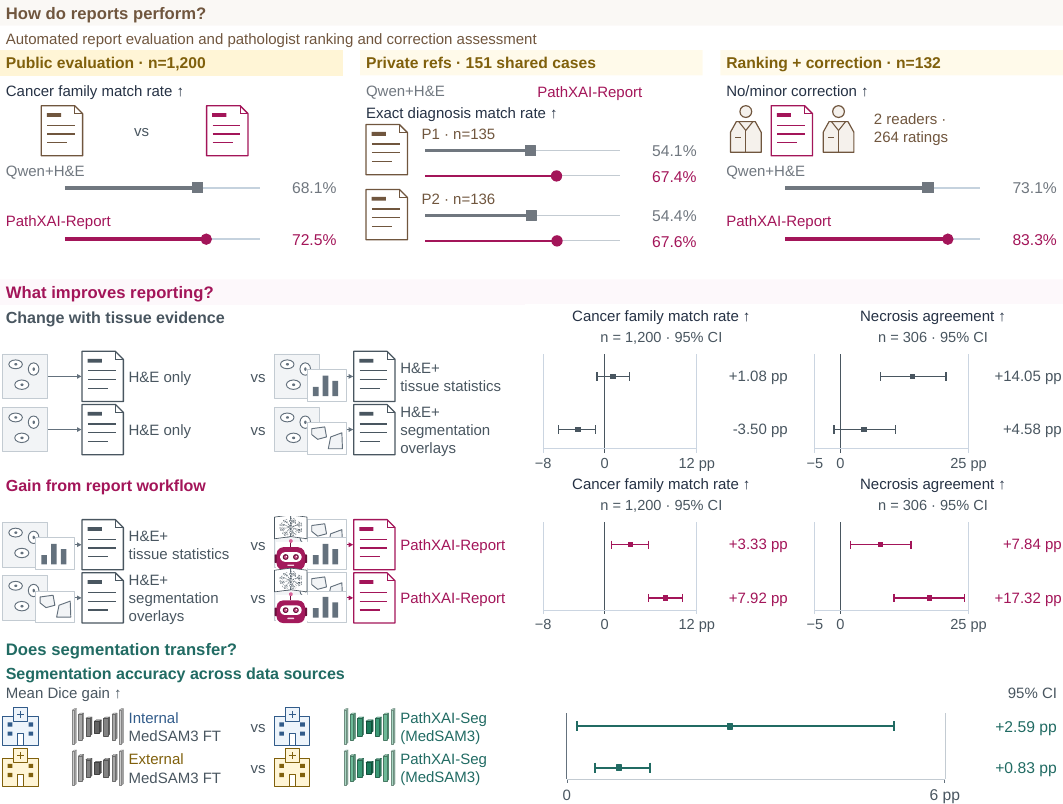}}
    \par\vspace{2mm}
    \includegraphics[width=180mm]{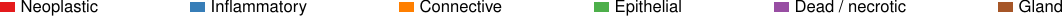}
    \end{minipage}%
  }{%
\textbf{\panelletterref{pan:overview-capabilities} Inputs and capabilities.} PathXAI-Report combines H\&E slides, known anatomical site, tissue evidence and domain knowledge for structured reporting.
\textbf{\panelletterref{pan:overview-workflow} Reporting with tissue evidence.} Segmentation maps, overlays and tissue statistics connect selected slide regions to diagnostic and morphological report fields.
\textcolor{black}{Overlays guide diagnostic candidates, tissue statistics support a separate tissue-statistics assessment, and every step is recorded and linked to report fields.}
\textbf{\panelletterref{pan:overview-findings} Evaluation questions and findings.} Automated report evaluation, pathologist assessment and segmentation comparisons are detailed in \cref{fig:public-summary,fig:private-summary,fig:seg-summary}. \textcolor{black}{The public cancer-family benchmark and tissue-input and workflow contrasts use the same 1,200-case public evaluation cohort.} \textcolor[HTML]{000000}{Necrosis agreement measures agreement with reference reports on the presence or absence of necrosis, using 306 applicable cases.} Report metrics weight applicable cases equally; P1/P2 denote the pathologists' reference sets. Gains are right minus left in percentage points (pp). Whiskers show unadjusted 95\% bootstrap confidence intervals (CIs), resampling cases for reporting and source clusters for segmentation.
(Notation: H\&E, haematoxylin and eosin; ROI, region of interest; VLM, vision-language model; $\uparrow$, higher preferred. Pink robots denote PathXAI-Report; branching networks denote domain knowledge; the dashed return denotes additional ROIs or zooming. Report fields and links are schematic; tissue colours follow the shared key.)
}

\section{Results}\label{sec:results}
\subsection{PathXAI-Report evaluation}\label{sec:results-report}

We evaluated PathXAI-Report through automated report evaluation, pathologist ranking and correction
assessment and comparisons of tissue inputs and reporting workflows.
Selected cases illustrated
links between evidence and report fields.
Study terminology is summarized in \cref{tab:glossary}.

\subsubsection{Reporting configurations and comparators}
\label{sec:results-report-design}
\textbf{PathXAI-Report configurations.}
\textcolor{black}{PathXAI-Report has two versions. PathXAI-Report (core) uses segmentation
evidence and adaptive reinspection for diagnosis. PathXAI-Report adds domain knowledge,
additional grade-focused ROIs and diagnostic constraints during writing. Both use
Qwen3.6-27B, without weight updates on either reporting cohort}
(\cref{supp:report-configurations}).

\textbf{Comparators.}
To assess gains over direct H\&E-based reporting, we compared
PathXAI-Report with general-purpose Qwen3.6-27B~\cite{qwen2026},
DeepSeek-V4.1-Flash~\cite{deepseekai2026deepseekv41flash} and
GLM-5.3-Flash~\cite{glm5team2026glm5vibecodingagentic}.
To benchmark against domain-specialized approaches, we included
MedGemma-1.5-4B-it~\cite{sellergren2026medgemma} and
PRISM2 diagnostic outputs~\cite{vorontsov2026prism2}.
For the PLIP-guided Qwen reporting baseline,
we used PLIP~\cite{huang2023plip} to select tumour-related ROIs.
We also compared the dedicated pathology report generators HistGen
and HistoGPT~\cite{guo2024histgen,tran2025histogpt}, with public and
private results reported in
\cref{fig:report-native-public,fig:report-native-private},
respectively.
Configurations and comparison protocols are detailed in
\cref{supp:report-comparators,supp:report-native-comparison}.

\subsubsection{Automated report evaluation}\label{sec:results-report-automated}
\textbf{Cohorts and reference reports.}
\textcolor{black}{Primary automated evaluation used 1,200 public cases across 22 cancer types and 14 organ systems, separate from the 456 \textcolor{black}{cases used for PathXAI-Report pipeline development}.}
\textcolor{black}{Public reference texts were OCR-processed TCGA reports}
(\cref{pan:public-summary-public-cohort}; \hyperref[sec:methods-study-design]{Methods}).
The primary private comparison included 151 breast, prostate and colorectal cases.
Two pathologists independently wrote one reference report per case, forming reference sets P1 and P2.
\textcolor{black}{Generated reports were scored against each reference set separately}
(\cref{pan:public-summary-private-cohort}; \hyperref[sec:methods-report-human]{Methods}).

\textbf{Reference agreement and evaluation scope.}
Across the private cohort, the two pathologists' reports agreed on the specific diagnosis in 122 of 127 cases assessable against both pathologists, and on grade in 45 of 75 cases with a shared grading system (\cref{pan:public-summary-reference-agreement}). Separate scoring
therefore retained differences between the two pathologists' interpretations.

\textbf{Evaluation measures.}
\textcolor{black}{We evaluated diagnostic agreement and reported morphology using five measures in both cohorts}
(\cref{pan:public-summary-public-performance,pan:public-summary-private-performance}).
\begin{itemize}
    \item \textbf{Cancer family match rate} assesses whether the generated diagnosis belongs to the target cancer family, using TCGA labels for public cases and definite cancer-family diagnoses from the pathologist references for private cases.
    \item \textbf{Exact diagnosis match rate} measures agreement with the specific reference diagnosis, including the disease entity or histological subtype, with synonymous terms treated as equivalent.
\end{itemize}
Meanwhile, for morphological evaluation:
\begin{itemize}
    \item \textbf{Finding F1} combines finding precision and recall to summarize the correctness and completeness of reported morphological findings.
    \item \textbf{Finding recall} measures the fraction of reference finding types correctly reproduced as present or absent.
    \item \textbf{Finding contradiction rate} measures the fraction of definite, comparable finding assertions that contradict the presence or absence recorded in the reference.
\end{itemize}
The three finding-based measures include only finding types explicitly documented
as present or absent in the reference. Detailed scoring rules are provided in
\hyperref[sec:methods-report-metrics]{Methods} and
\cref{tab:report-metric-overview}.

\paragraph{Diagnostic evaluation.}\label{sec:results-report-diagnostic-outputs}
\textbf{PathXAI-Report achieves the highest cancer family match rate in \textcolor{black}{both cohorts}.}
In the \textcolor{black}{1,200-case public evaluation cohort}, PathXAI-Report achieved a cancer family
match rate of \textcolor{black}{\EvaluationCancerFullProse}\%, with paired gains of
\textcolor{black}{\EvaluationVsDeepSeekProse{}}~pp over DeepSeek+H\&E and
\textcolor{black}{\EvaluationVsQwenProse{}}~pp over Qwen+H\&E
(\cref{pan:public-summary-public-performance}).
It also exceeded PathXAI-Report (core) by \textcolor{black}{\EvaluationVsCoreProse{}}~pp
(\cref{tab:report-primary-contrasts}).
In the private cohort, cancer family match rate reached
\textcolor{black}{\PrivateOneCancerFullProse}\% against P1 and \textcolor{black}{\PrivateTwoCancerFullProse}\% against P2,
exceeding Qwen+H\&E against both reference sets
(\cref{pan:public-summary-private-performance}).

\textbf{PathXAI-Report achieves the highest exact diagnosis match rate in \textcolor{black}{both cohorts}.}
In the public cohort,
PathXAI-Report achieved \textcolor{black}{\EvaluationEntityFullProse}\% agreement,
exceeding Qwen+H\&E and DeepSeek+H\&E
(\cref{pan:public-summary-public-performance}).
In the private cohort, agreement was \textcolor{black}{\PrivateOneEntityFullProse}\% for P1
and \textcolor{black}{\PrivateTwoEntityFullProse}\% for P2. Averaging the two reference-specific scores within
jointly evaluable cases yielded a paired gain over DeepSeek+H\&E of
\textcolor{black}{\PrivateVsDeepSeekProse}{}~pp
(\cref{pan:public-summary-private-performance}).
The higher agreement therefore extends beyond matching broad cancer families to naming
more specific diagnoses in both cohorts. 

\textbf{General-purpose H\&E models exceed selected specialized comparators in both cohorts.}
Qwen+H\&E, DeepSeek+H\&E and GLM+H\&E achieved higher finding F1
scores than MedGemma and PLIP-guided Qwen on public cases and against
both reference sets.
For cancer family match rate, all three models exceeded PRISM2 on public
cases, while Qwen+H\&E and DeepSeek+H\&E also exceeded PRISM2 against
both reference sets
(\cref{pan:public-summary-public-performance,pan:public-summary-private-performance}).

\paragraph{Morphological evaluation.}\label{sec:results-report-morphology}
\textbf{PathXAI-Report better reproduces documented morphology in \textcolor{black}{both cohorts}.}
On the public cohort, PathXAI-Report had the
highest finding F1, \textcolor{black}{\EvaluationFindingFullProse}\%.
Finding F1 exceeded Qwen+H\&E by \textcolor{black}{\EvaluationFindingVsQwenProse{}}~pp
and was also higher than PathXAI-Report (core)
(\cref{pan:public-summary-public-performance,tab:report-primary-contrasts}).
Finding recall reached \textcolor{black}{\EvaluationCorrectFindingFullProse}\% versus
\textcolor{black}{\EvaluationCorrectFindingQwenProse}\% for Qwen+H\&E on the same cases
(\cref{pan:public-summary-public-performance,tab:report-primary-estimates}).
Against both reference sets, PathXAI-Report likewise had the highest finding F1,
at \textcolor{black}{\PrivateOneFindingFullProse}\% for P1 and \textcolor{black}{\PrivateTwoFindingFullProse}\% for P2.
Finding recall was also highest against both reference sets
(\cref{pan:public-summary-private-performance}).

\textbf{PathXAI-Report expands finding coverage in both cohorts, with lower contradiction publicly and higher contradiction privately than Qwen+H\&E.}
Finding coverage measures the fraction of documented finding types receiving a definite,
comparable assertion. On the public cohort, PathXAI-Report had higher
finding coverage than Qwen+H\&E and \textcolor{black}{a finding contradiction rate that was
\textcolor{black}{\EvaluationContradictionReductionVsQwenProse{}}~pp lower in the paired comparison}
among cases with assessable assertions in both methods
(\cref{pan:public-summary-public-performance,tab:report-primary-contrasts}).
Against each reference set, finding coverage and finding contradiction rate were both
higher than those of Qwen+H\&E (\cref{pan:public-summary-private-performance}).
PLIP-guided Qwen had the lowest finding contradiction rate among the public-cohort
report-generating comparators, but also the lowest finding coverage and finding F1.
Its lower contradiction therefore did not translate into more complete recovery of
reference-documented findings (\cref{pan:public-summary-public-performance}).
Complete coverage and contradiction estimates are in
\cref{supp:results-public-findings,supp:results-private-full}.

\textcolor{black}{PathXAI-Report also achieved higher cancer family match rates and finding F1
than HistGen and HistoGPT on shared public evaluation cases and in the private cohort
(\cref{fig:report-native-public,fig:report-native-private}).}

\begingroup
\renewcommand{\manuscriptwidefigureplacement}{!htbp}
\manuscriptwidefigurepart{first}{Automated report evaluation on public and private cohorts.}{fig:public-summary}{%
  \begin{minipage}{180mm}\offinterlineskip
  {\setlength{\fboxsep}{0pt}\colorbox[HTML]{F2F4F6}{\parbox[c][4.2mm][c]{180mm}{\hspace*{1mm}\fontfamily{phv}\fontsize{9}{10}\selectfont\bfseries Automated report evaluation · public cohort}}}\par
  \manuscriptpanelblock{3.8mm}{180mm}{\manuscriptpanelmark{pan:public-summary-public-cohort}}{Public cohort and reference sources}{\includegraphics[width=\linewidth,trim=0mm 0.2mm 0mm 0mm,clip]{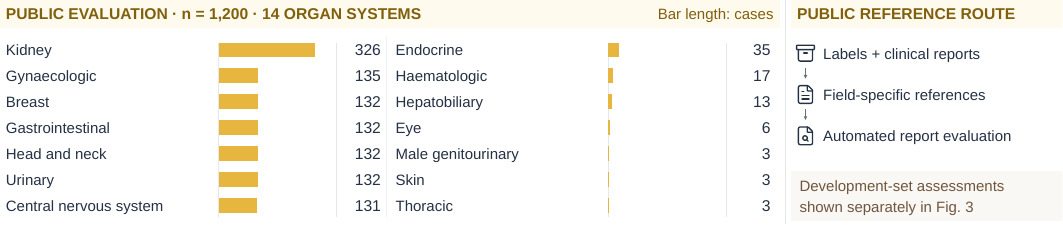}}
  \par\vspace{1.2mm}
  \manuscriptpanelblock{3.8mm}{180mm}{\manuscriptpanelmark{pan:public-summary-public-performance}}{Diagnostic outputs and morphology}{\includegraphics[width=\linewidth,trim=0mm 0.8mm 0mm 0.5mm,clip]{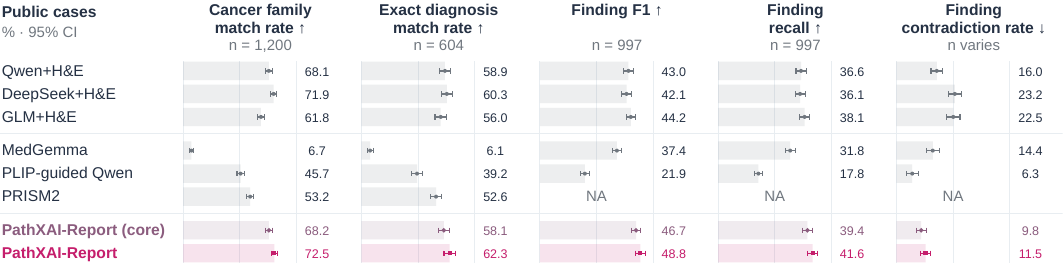}}
  \par\vspace{1.2mm}
  {\setlength{\fboxsep}{0pt}\colorbox[HTML]{F2F4F6}{\parbox[c][4.2mm][c]{180mm}{\hspace*{1mm}\fontfamily{phv}\fontsize{9}{10}\selectfont\bfseries Automated report evaluation · private cohort against pathologist-written references}}}\par
  \manuscriptpanelblock{3.8mm}{120mm}{\manuscriptpanelmark{pan:public-summary-private-cohort}}{Private cohort and pathologist references}{\includegraphics[width=\linewidth]{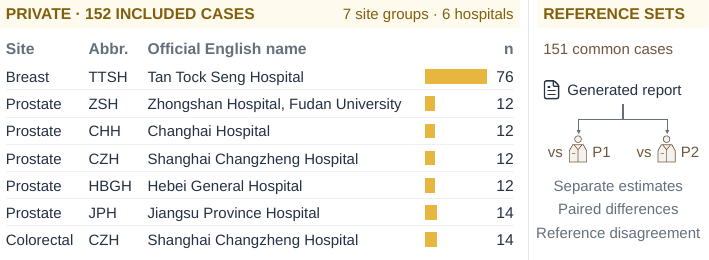}}\hfill
  \manuscriptpanelblock{3.8mm}{58mm}{\manuscriptpanelmark{pan:public-summary-reference-agreement}}{Reference agreement}{\includegraphics[width=\linewidth]{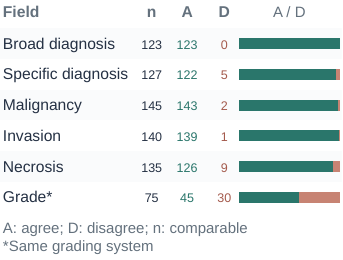}}
  \par\vspace{1.2mm}
  \manuscriptpanelblock{3.8mm}{180mm}{\manuscriptpanelmark{pan:public-summary-private-performance}}{Diagnostic outputs and morphology by reference}{\includegraphics[width=\linewidth,trim=0mm 1mm 0mm 0.3mm,clip]{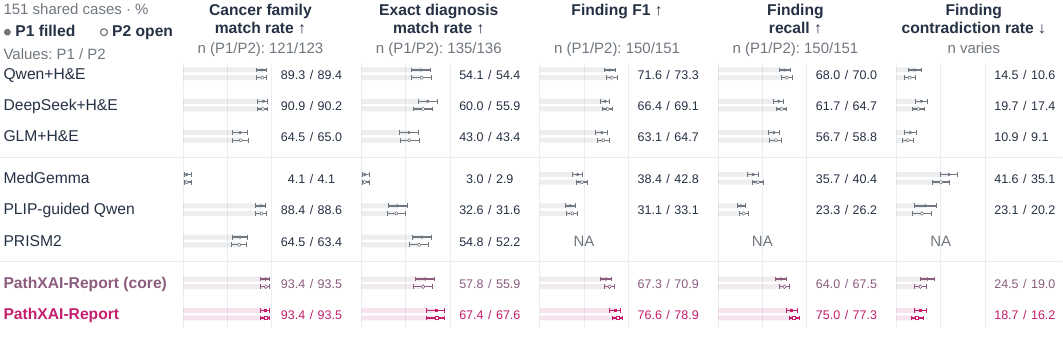}}
  \end{minipage}%
}{%
\textbf{\panelletterref{pan:public-summary-public-cohort} Public cohort and references.} \textcolor{black}{The public evaluation cohort is separate from the \textcolor{black}{PathXAI-Report pipeline development subset}.}
\textbf{\panelletterref{pan:public-summary-public-performance} Public report evaluation.} \textcolor{black}{Diagnostic and morphological outcomes on the public evaluation cohort.}
\textbf{\panelletterref{pan:public-summary-private-cohort} Private cohort and reference sets.} Each case has a separately authored report from each pathologist.
\textbf{\panelletterref{pan:public-summary-reference-agreement} Reference agreement.} Fields must be assessable in both reports; grade also requires the same grading system.
\textbf{\panelletterref{pan:public-summary-private-performance} Private report evaluation.} Methods use shared cases and are evaluated separately against each reference set.
Cancer family targets are TCGA labels publicly and definite cancer-family diagnoses privately; benign or unresolved private references are inapplicable. Applicable cases receive equal weight. Finding contradiction rate uses method-specific comparable assertions and is interpreted with finding coverage. OCR-based content analyses are exploratory. Full estimates are in Source Data. \textcolor{black}{HistGen/HistoGPT comparisons appear in \cref{fig:report-native-public,fig:report-native-private}.}
(Notation: P1/P2, pathologists' reference sets; H\&E, haematoxylin and eosin; OCR, optical character recognition; NA, not applicable. Bars show point estimates; whiskers show \textcolor{black}{pointwise 95\% confidence intervals (\cref{supp:report-statistics})}. Axis ranges vary.)
}
\endgroup

\subsubsection{Pathologist ranking and correction assessment}\label{sec:results-report-pathologist}
\textbf{Cases, readers and comparators.}
\textcolor{black}{The two pathologists who authored the private reference reports ranked reports
and rated correction requirements for 132 cases sampled randomly within cancer type
from the PathXAI-Report pipeline development subset. Each assessed MedGemma, Qwen+H\&E,
PLIP-guided Qwen, PathXAI-Report (core) and PathXAI-Report, yielding 264 case-pathologist
ratings per method (\cref{supp:rating-protocol}).}

\textbf{Assessment measures.}
Overall preference ranks ranged from 1 (best) to 5, with ties allowed.
The no/minor-correction proportion measures how often readers judged that a report required
no or only minor edits. The no/minor correction with complete fields measure also required no output failure and
explicit answers for every applicable diagnostic, grade, invasion and necrosis field.
The field check measured answer presence, not correctness; appropriate abstentions also
counted as incomplete (\cref{pan:private-summary-pathologist-assessment}).

\textbf{PathXAI-Report most often combines low correction requirements with complete answers to applicable fields among all five methods.}
PathXAI-Report met all three criteria in \ReaderCompleteFullSuccess\% of ratings
(\ReaderCompleteFullSuccessCount/\ReaderCompleteFullN),
compared with \ReaderCompleteQwenSuccess\% for Qwen+H\&E and \ReaderCompleteCoreSuccess\% for
PathXAI-Report (core), and lower proportions for MedGemma and PLIP-guided Qwen
(\cref{pan:private-summary-pathologist-assessment}). MedGemma had the highest no/minor-correction
fraction alone (\ReaderNoMinorMedGemma\%), but fewer of its reports also satisfied the output
and field-coverage checks. Full results and field definitions
are in \cref{supp:results-pathologist-assessment,supp:rating-checks,tab:source_fig10}.

\textbf{PathXAI-Report retains preference ranks similar to Qwen+H\&E and PathXAI-Report (core), with more no/minor-correction ratings.}
PathXAI-Report received no/minor-correction ratings in \ReaderNoMinorFull\% of assessments,
exceeding PathXAI-Report (core), Qwen+H\&E and PLIP-guided Qwen
(\cref{pan:private-summary-pathologist-assessment}). The difference from Qwen+H\&E arose in one reader's ratings
(\cref{supp:results-pathologist-assessment}). Mean preference ranks were similar for
PathXAI-Report, Qwen+H\&E and PathXAI-Report (core), and better than those of MedGemma and
PLIP-guided Qwen (\cref{pan:private-summary-pathologist-assessment}).

\subsubsection{\textcolor{black}{Tissue evidence helps when used within the workflow: ablation studies}}\label{sec:results-report-policy}
We compared direct tissue inputs with their use during diagnostic review and writing
on \textcolor{black}{the same 1,200 public evaluation cases as in \cref{fig:public-summary}}. The comparisons included H\&E alone, H\&E with tissue statistics,
H\&E with segmentation overlays, and PathXAI-Report
(\cref{supp:report-input-comparisons,supp:report-policy-comparisons}).
\textcolor{black}{\Cref{tab:report-component-comparisons} summarizes the components changed in each comparison.}
\textcolor{black}{We evaluated histological grading using \textbf{balanced grade accuracy}}\textcolor{black}{, which weights reference grades 1--3 equally (\hyperref[sec:methods-report-findings]{Methods}).}

\textcolor{black}{\textbf{Adding tissue evidence directly to H\&E gives limited diagnostic benefit.}}
H\&E with tissue statistics increased exact diagnosis match rate by \StatisticsVsHEEntityProse{}~pp.
\textcolor{black}{Cancer family match rate also increased slightly, although both differences were uncertain.
H\&E with segmentation overlays lowered cancer family match rate; exact diagnosis match rate
and finding F1 also had lower point estimates}
(\cref{pan:private-summary-policy-comparisons}). 

\textbf{\textcolor{black}{Assigning tissue evidence to specific steps improves morphological reporting.}}
Against direct tissue-statistics input, PathXAI-Report increased finding F1 by
\FullVsStatisticsFindingProse{}~pp, \textcolor{black}{with a smaller, uncertain increase in
exact diagnosis match rate} (\cref{pan:private-summary-policy-comparisons}). Finding contradiction rate was lower, although the difference remained uncertain. Against direct overlay input,
PathXAI-Report increased finding recall. Finding contradiction rate was lower
among cases with assessable assertions in both methods
(\cref{pan:private-summary-report-content} and \crefrange{tab:source_fig7:part-1}{tab:source_fig7:part-2}).

\textbf{\textcolor{black}{The same workflow corrects more cancer types and improves grade agreement.}}
Relative to direct overlay input, \textcolor{black}{144} reports that were incompatible with the target cancer
family became compatible under PathXAI-Report, with \textcolor{black}{49} changes in the reverse direction
(\cref{pan:private-summary-compatibility-transitions}). For three-level grading,
PathXAI-Report increased balanced grade accuracy by
\FullVsStatisticsGradeProse{}~pp over direct tissue-statistics input (\cref{pan:private-summary-policy-comparisons}). Grade-specific
predictions, including recovered unanswered fields and errors between grades, are shown in
\cref{fig:report-grade-policy}.

\textbf{\textcolor{black}{Acquiring additional regions and constraining report writing improve grade agreement.}}
Requesting additional regions when diagnostic candidates conflicted increased balanced
grade accuracy by \textcolor{black}{\ExpansionVsGuardGradeProse}{}~pp over candidate selection
alone.
Combining
knowledge-informed reasoning\textcolor{black}{, diagnostic safeguards} and diagnosis-constrained writing with grade-focused writer
images increased balanced grade accuracy by \KnowledgeVsGradeGradeProse{}~pp over the grade-image baseline (\cref{pan:private-summary-workflow-comparisons}).
Finding F1 increased by \KnowledgeVsGradeFindingProse{}~pp. \textcolor{black}{This contrast evaluates these configuration changes together.} Complete workflow estimates, including zoom-enabled
reinspection and diagnostic-composite results, are in \cref{fig:report-ablation,tab:source_supp3}.

\manuscriptwidefigurepart{first}{Pathologist ranking and correction assessment; tissue evidence and its use in reporting.}{fig:private-summary}{%
  \begingroup\setlength{\unitlength}{1mm}%
  \begin{picture}(180,195)
  \put(0,0){\includegraphics[width=180mm]{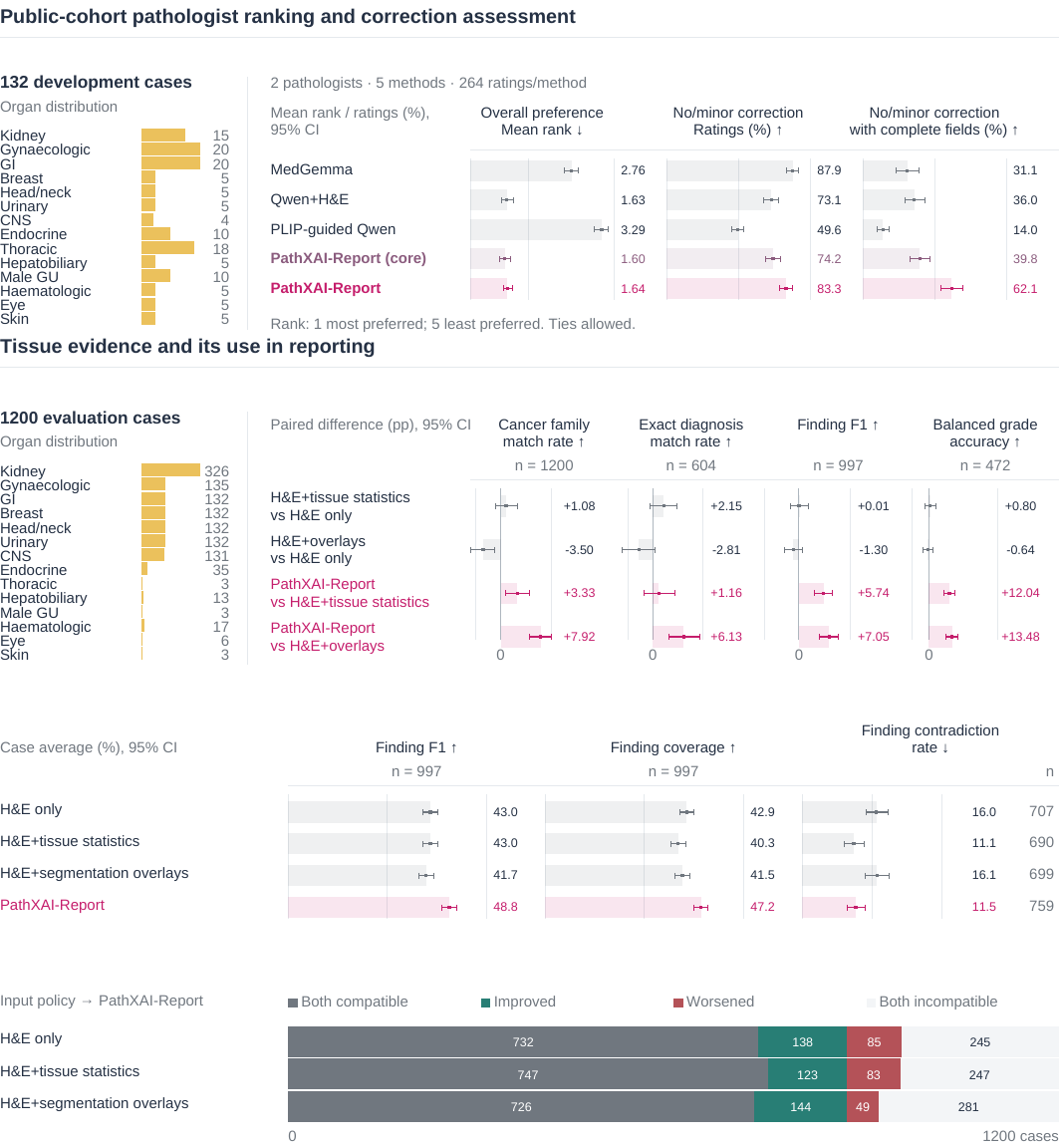}}%
  \fontfamily{phv}\fontsize{8}{9}\selectfont\bfseries
  \put(0,185){\makebox(0,0)[lb]{\manuscriptpanelmark{pan:private-summary-pathologist-assessment}\manuscriptpanellabel{pan:private-summary-preference}\manuscriptpanellabel{pan:private-summary-correction-ratings}\manuscriptpanellabel{pan:private-summary-joint-completeness}\enspace Pathologist ranking and correction assessment}}%
  \put(0,127){\makebox(0,0)[lb]{\manuscriptpanelmark{pan:private-summary-policy-comparisons}\enspace Tissue-input and report-workflow comparisons}}%
  \put(0,73){\makebox(0,0)[lb]{\manuscriptpanelmark{pan:private-summary-report-content}\enspace Report content, finding coverage and contradictions}}%
  \put(0,29){\makebox(0,0)[lb]{\manuscriptpanelmark{pan:private-summary-compatibility-transitions}\enspace Paired cancer family match transitions}}%
  \end{picture}\endgroup%
  \par\vspace{0mm}
  \manuscriptpanelblock{5mm}{180mm}{\manuscriptpanelmark{pan:private-summary-workflow-comparisons}}{Acquisition and writing policies: paired differences (pp)}{\includegraphics[width=\linewidth]{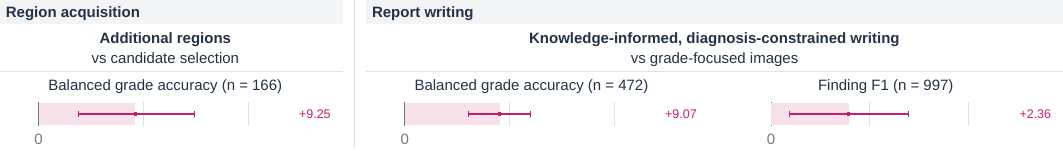}}%
}{%
\textcolor{black}{Tissue-input and writing comparisons use the 1,200 public evaluation cases in Fig.~\ref{fig:public-summary}. Pathologist ratings and region acquisition use 132 and 432 cases, respectively, from the PathXAI-Report pipeline development subset.}
\textbf{\panelletterref{pan:private-summary-pathologist-assessment} Pathologist assessment.} Preference ranks allow ties. The complete-fields measure combines no/minor correction, no output failure and definite answers in every applicable field; field completeness measures answer presence. Re-review retained the original ratings.
\textbf{\panelletterref{pan:private-summary-policy-comparisons} Tissue-input and workflow comparisons.}
\textbf{\panelletterref{pan:private-summary-report-content} Report content.}
\textbf{\panelletterref{pan:private-summary-compatibility-transitions} Cancer family match transitions.} Corrections and reversals relative to cohort targets.
\textbf{\panelletterref{pan:private-summary-workflow-comparisons} Acquisition and writing.} The writing contrast changes knowledge-informed reasoning and diagnosis constraints together. Additional policies are shown in \cref{fig:report-ablation}.
Paired differences subtract the second-named policy from the first. Whiskers show unadjusted 95\% case-bootstrap intervals, retaining both raters for \panelletterref{pan:private-summary-pathologist-assessment} and stratifying grade by reference class. Grade classes receive equal weight; other outcomes weight cases equally.
(Notation: $n$, metric-applicable cases; pp, percentage points. Faint bars start at 1 for ranks and at zero otherwise.)
}

\subsubsection{Case examples and evidence links}\label{sec:results-report-evidence-links}
\textbf{PathXAI-Report improves grading over direct Qwen+H\&E reporting in the breast example.}
In breast case C035, \textcolor{black}{PathXAI-Report} reported Grade 2, matching both pathologists,
while Qwen+H\&E reported Grade 3.
PathXAI-Report also made invasive ductal carcinoma of no special type explicit in the
diagnostic fields; Qwen+H\&E named this subtype in its final summary
(\cref{pan:case-summary-breast}).

\textbf{Recorded evidence links distinguish tissue measurements, image interpretation and diagnostic context.}
In the adrenocortical-carcinoma case TCGA-OR-A5LL
(\cref{pan:case-summary-concordant-case}), PathXAI inspected three diagnostic regions from two slides and
revisited them through two rounds of zooming. \textcolor{black}{The recorded trajectory shows how
each step followed from its inputs. Both diagnostic candidates proposed adrenocortical carcinoma,
and the selection rule retained the candidate without KEEP at equal specificity. The evidence
assessment judged the initial evidence insufficient, citing missing high-power views of mitotic
figures and of the tumour capsule; the acquisition rule therefore zoomed into R1 and R2 and then
into R3. Acquisition stopped when the diagnosis was unchanged across successive updates, and the
final tissue-statistics check raised its best-supported level from broad type to subtype.}
The resulting diagnosis matched the TCGA
cohort target. Neoplastic-tissue measurements
checked for tumour presence, while the known adrenal site and retrieved
adrenocortical-carcinoma definition supplied diagnostic context
(\cref{pan:case-summary-domain-knowledge,pan:case-summary-evidence-links}).
Architecture and cytology descriptions linked to H\&E inputs, requiring image interpretation
beyond tissue labels. Two additional H\&E fields \textcolor{black}{from a third slide} entered grade-focused writing without
segmentation masks.

\textbf{The evidence record makes conflicting regional checks and unassessable fields explicit.}
The report described
\textcolor{black}{extensive geographic necrosis characterized by large areas of eosinophilic tumor cell debris}, and its record linked this statement
to positive dead-tissue checks in \textcolor{black}{R2 and Z1} (\cref{pan:case-summary-evidence-links}). Checks in
\textcolor{black}{Z2 and Z3} were negative\textcolor{black}{, R1 and R3 were neutral}, and the overall tissue-based check for the statement was weak.
These conflicting checks expose uncertainty behind the necrosis claim. \textcolor{black}{The report marked
invasion as unassessable; the automated verification likewise marked it as not assessable
because the tumour edge and surrounding-tissue interface were not visible.}
\textcolor{black}{The trajectory shows how this field was reached: invasive features were suspected by
the candidate with KEEP in the initial diagnosis and suggested by the selected candidate after the
first zoom, whereas after the second zoom the selected candidate stated that invasion was not
assessable because no capsule or adjacent-tissue interface was visible
(\cref{pan:case-summary-concordant-case}).}
The complete image sequence and a target-discordant case are analysed in
\cref{fig:report-case-success,fig:report-case-failure}.

\begingroup
\renewcommand{\manuscriptwidefigureplacement}{!tbp}
\manuscriptwidefigurepart{first}{Case examples illustrate diagnostic reporting and links to tissue evidence.}{fig:case-summary}{%
  \begin{minipage}{180mm}
  \manuscriptpanelblock{5mm}{180mm}{\manuscriptpanelmark{pan:case-summary-breast}}{Breast case C035: grade agreement and subtype placement}{\includegraphics[width=\linewidth]{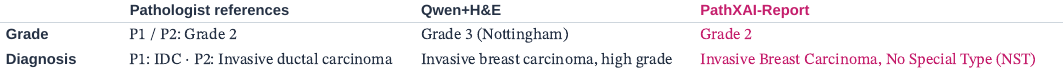}}
  \par\vspace{1mm}
  \manuscriptpanelblock{5mm}{180mm}{\manuscriptpanelmark{pan:case-summary-concordant-case}}{\textcolor{black}{Recorded reasoning trajectory: adrenocortical carcinoma (TCGA-OR-A5LL)}}{\includegraphics[width=\linewidth]{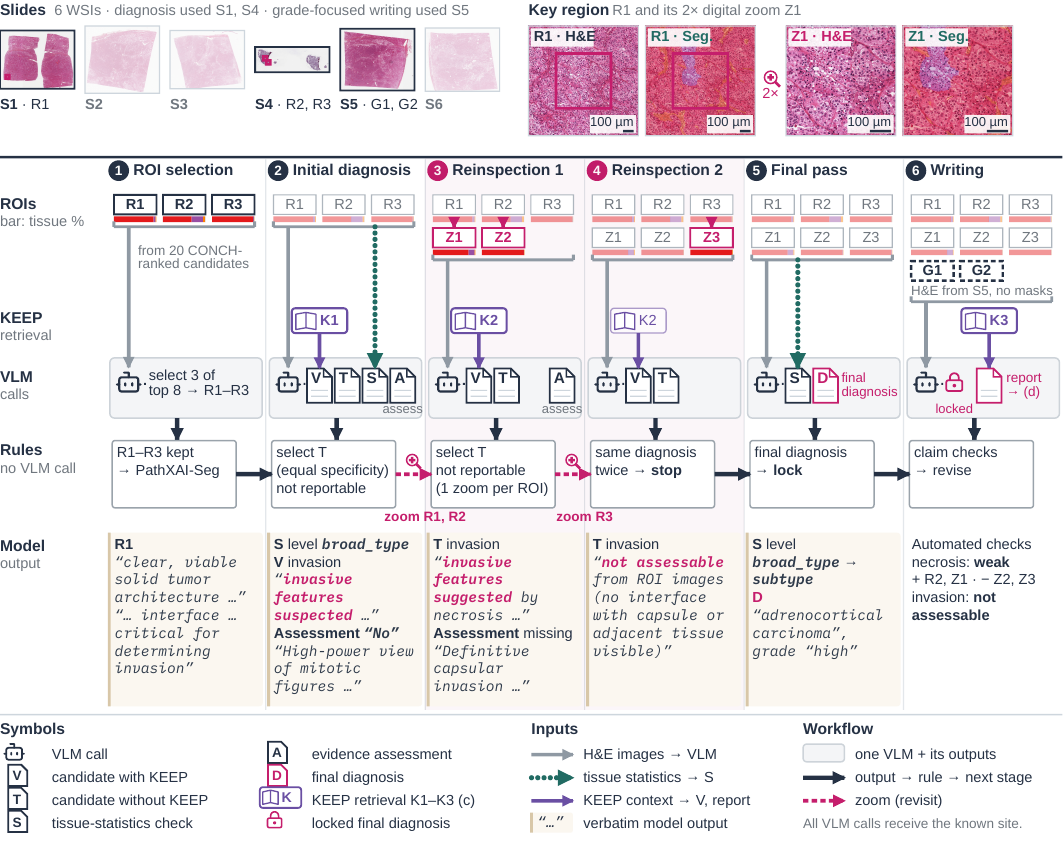}}
  \par\vspace{1mm}
  \includegraphics[width=180mm]{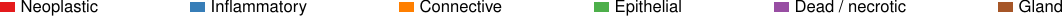}
  \par\vspace{1.5mm}
  \manuscriptpanelblock{5mm}{50mm}{\manuscriptpanelmark{pan:case-summary-domain-knowledge}}{Retrieved knowledge}{\includegraphics[width=\linewidth]{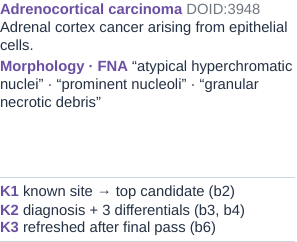}}%
  \hspace{4mm}%
  \manuscriptpanelblock{5mm}{126mm}{\manuscriptpanelmark{pan:case-summary-evidence-links}}{\textcolor{black}{Report excerpts and evidence links}}{\includegraphics[width=\linewidth]{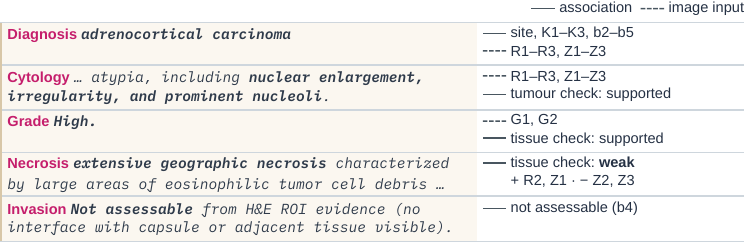}}
  \end{minipage}%
}{%
\textbf{\panelletterref{pan:case-summary-breast} Private breast report example.} Qwen+H\&E also names ductal NST in its final summary; the full comparison appears in \cref{pan:report-private-examples}.
\textbf{\panelletterref{pan:case-summary-concordant-case} \textcolor{black}{Recorded reasoning trajectory of the public adrenocortical-carcinoma example.}} \textcolor{black}{Six recorded stages of one archived Full run. Grey frames are VLM steps; arrows show the ROI images, tissue statistics and KEEP retrievals each output received, followed by the rule-based decision. Central-crop 2$\times$ digital zooms added Z1--Z3. Italic monospaced text is verbatim model output, abridged; pink highlighting added.}
\textbf{\panelletterref{pan:case-summary-domain-knowledge} Retrieved knowledge.} Paraphrased entity definition and verbatim FNA morphology fragments\textcolor{black}{; K1--K3, KEEP queries in (b)}.
\textbf{\panelletterref{pan:case-summary-evidence-links} Report excerpts and evidence links.} \textcolor{black}{Excerpts are verbatim from the report of the same run, with bold emphasis added.} Solid links denote recorded context associations or tissue checks; dashed links denote image inputs without verified claim-specific ROI attribution. Microscopy checks assess tumour presence. \textcolor{black}{The necrosis tissue check is weak (supporting R2 and Z1, contradicting Z2 and Z3) despite the automated ``supported'' status.} Case details appear in \cref{fig:report-case-success,fig:report-case-failure}.
(Notation: P1/P2, pathologists' reference sets; IDC, invasive ductal carcinoma; NST, no special type; \textcolor{black}{S1--S6, whole-slide images; R/Z/G, diagnostic ROI, zoom and grade-focused ROI;} Seg., segmentation overlay; FNA, fine-needle aspiration. Scale bars, 100~$\mu$m; tissue colours follow the shared key.)
}
\endgroup

\subsection{Transferable tissue segmentation for reporting}\label{sec:results-seg}
\textcolor{black}{We evaluated tissue segmentation across data sources and organs, comparing
same-backbone fine-tuning, pathology segmentation systems and training objectives.}

\subsubsection{Datasets and evaluation design}\label{sec:results-seg-design}
\textbf{Pixel-labelled training data.}
PathXAI-Seg uses a shared seven-label space comprising background, neoplastic tissue,
inflammatory tissue, connective tissue, epithelial tissue, dead or necrotic tissue, and gland structure.
\textcolor{black}{Training used 492 densely annotated images from five datasets:
BCSS~\cite{amgad2019bcss}, DigestPath2019~\cite{da2022digestpath},
GlaS~\cite{sirinukunwattana2017glas}, BACH whole slides~\cite{aresta2019bach}
and WSSS4LUAD~\cite{han2022wsss4luad}, covering breast, colorectal and lung tissue.}

\textbf{Auxiliary training data.}
\textcolor{black}{Auxiliary images were sampled from a pool of 344,604 images
from eighteen sources to provide image-level supervision:
CPM15~\cite{miccai2015nucleiChallenge}, CPM17~\cite{vu2019cpm},
NCT-CRC-HE-100K~\cite{Kather2019PredictingSF}, CoNIC2022~\cite{graham2024conic},
BACH microscopy images~\cite{aresta2019bach}, Janowczyk~\cite{janowczyk2016deep},
Kumar~\cite{kumar2017dataset}, Lizard~\cite{graham2021lizard},
LC25000 (LungColon)~\cite{borkowski2019lc25000}, MHIST~\cite{wei2021petri},
MoNuSAC~\cite{verma2021monusac}, NuCLS~\cite{amgad2022nucls},
Osteosarcoma~\cite{arunachalam2019osteosarcoma},
PanNuke~\cite{gamper2019pannuke,gamper2020pannukeExtension},
PatchCamelyon17 distributed through PathMMU~\cite{sun2024pathmmu},
SegPath~\cite{komura2023segpath}, SkinCancer~\cite{kriegsmann2022skinPaper}
and TNBC~\cite{naylor2019distance}.
Annotation conversion is detailed in
\cref{supp:seg-labels-inputs,fig:seg-data}.}

\textbf{Internal and external tests.}
\textcolor{black}{Testing included 124 internal samples from the five pixel-labelled
training-source datasets and 11,089 external samples across six test sets:
ACDC-LungHP~\cite{li2021acdclunghp}, AGGC2022~\cite{huo2024aggc},
METU-CCTGS~\cite{arslan2025metu}, PAIP2019~\cite{kim2021paip},
and RINGS tumour and gland segmentation~\cite{salvi2021ringsPaper}.
The two RINGS tasks used the same 500 images, counted once per task in the external
sample total. External tests used sources distinct from the training datasets and added
prostate and liver tissue, bringing the total coverage to five organs
(\cref{pan:seg-summary-data-examples}). Counting units are detailed in
\cref{supp:seg-labels-inputs,fig:seg-data}.}

\textbf{Comparators.}
We evaluated PathXAI-Seg (MedSAM3) and PathXAI-Seg (SAM2), which use
MedSAM3 and SAM2 as their respective segmentation backbones~\cite{liu2025medsam3,ravi2024sam2}.
\textcolor{black}{PathXAI-Seg (MedSAM3) was used by PathXAI-Report.}
To assess gains over standard supervised adaptation, each was compared with fine-tuning of
the same backbone on the pixel-labelled training data. Other comparators were zero-shot
MedSAM3, HoVer-Net fine-tuning~\cite{graham2019hovernet}, nnU-Net~\cite{isensee2021nnunet}
and a CONCH-based segmentation implementation, CONCHSeg~\cite{lu2024conch}.
A separate comparison used the pathology segmentation models PathSegmentor and VISTA-PATH
on three shared external datasets~\cite{chen2026pathsegmentor,liang2026vistapath}
(\cref{supp:seg-shared-comparison}).

\textbf{Evaluation metrics.}
\textcolor{black}{Dice was the primary segmentation metric~\cite{taha2015metrics}.
Internal and external mean Dice summaries gave each test set equal weight.}

\subsubsection{Segmentation transfer across data sources}\label{sec:results-seg-performance}
\textbf{PathXAI-Seg improves mean Dice over same-backbone fine-tuning.}
PathXAI-Seg (MedSAM3)
exceeded standard same-backbone fine-tuning by \textcolor{black}{2.6}~pp internally and \textcolor{black}{0.8} externally.
PathXAI-Seg (SAM2) also improved internal Dice, whereas its external gain was \textcolor{black}{0.1}~pp
(\cref{pan:seg-summary-fine-tuning-gains}). Absolute scores across
evaluation tasks are shown in \cref{pan:seg-summary-dataset-dice,tab:source_seg:part-1}.

\textbf{Both PathXAI-Seg variants exceed PathSegmentor and VISTA-PATH in mean Dice on shared external datasets.}
Across ACDC-LungHP, PAIP2019 and METU-CCTGS, paired gains in equal-dataset mean Dice
over VISTA-PATH were \textcolor{black}{2.6}~pp for PathXAI-Seg (SAM2) and
\textcolor{black}{1.6} for PathXAI-Seg (MedSAM3); both also exceeded PathSegmentor
(\cref{pan:seg-summary-shared-external-comparison}). This comparison
uses ground-truth class-presence conditioning and a shared foreground definition; its
protocol and full estimates are in \cref{supp:seg-shared-comparison,tab:source_seg:part-2}.

\textbf{Gains over same-backbone MedSAM3 fine-tuning differ by organ.}
Internal lung Dice increased from 0.449 to 0.534, whereas performance on external PAIP2019
liver histology was similar (\cref{pan:seg-summary-organ-dice}). Matched BCSS breast, ACDC-LungHP lung and PAIP2019 liver
fields show predicted tissue compartments alongside the original H\&E and ground truth
(\cref{pan:seg-summary-data-examples}); complete dataset results are in
\cref{tab:source_seg:part-4}. Selection procedures are specified in \cref{supp:seg-example-selection}.

\subsubsection{Ablation of segmentation training objectives}\label{sec:results-seg-ablations}
To assess the value of combining training objectives, we performed loss ablations for
PathXAI-Seg (MedSAM3). We compared the full joint objective with three configurations
that set the weights to zero for the auxiliary-map losses in Label-guided localization,
the scale/shape terms in Mask geometry, or spatial contrast in Feature separation
(\cref{pan:methods-training-objectives}).

{The full joint objective achieves the highest observed external mean Dice among the tested configurations.}
The spatial-contrast ablation within Feature separation (L5) showed
the largest observed decreases in mean Dice, at \textcolor{black}{0.9}~pp internally
and \textcolor{black}{0.5}~pp externally.
The Label-guided localization (L3) and Mask geometry (L2) ablations
each had external mean Dice \textcolor{black}{less than 0.1~pp below} the full joint objective,
with internal differences of \textcolor{black}{-0.7}~pp and \textcolor{black}{+0.6}~pp, respectively.
See more details in (\cref{pan:seg-summary-component-deletions,tab:source_supp2}).
Training settings are detailed in \cref{supp:seg-training-comparisons}.
\manuscriptwidefigurepart{first}{Segmentation transfer across data sources and training-objective comparisons.}{fig:seg-summary}{%
  \begin{minipage}{180mm}\offinterlineskip
  \manuscriptpanelblock{2.9mm}{180mm}{\manuscriptpanelmark{pan:seg-summary-data-examples}}{Training and test data with matched examples}{\includegraphics[width=\linewidth]{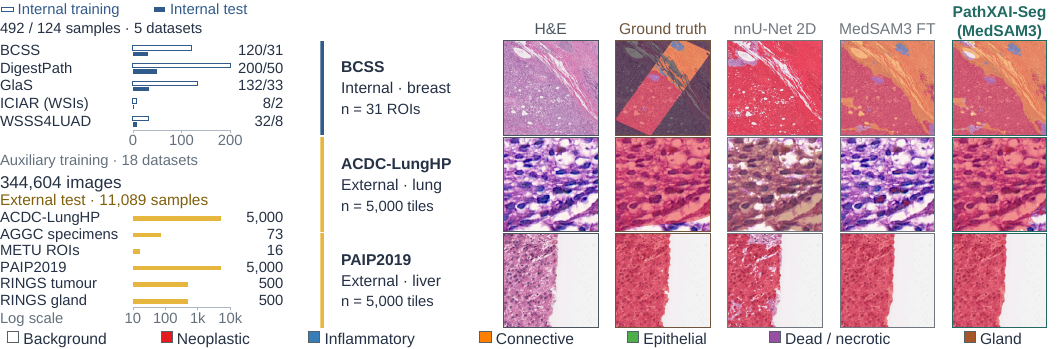}}
  \par\vspace{0.35mm}
  {\setlength{\fboxsep}{0pt}\colorbox[HTML]{F2F4F6}{\parbox[c][3.6mm][c]{180mm}{\hspace*{1mm}\fontfamily{phv}\fontsize{9}{10}\selectfont\bfseries Segmentation performance across test sources}}}\par
  \manuscriptpanelblock{2.9mm}{180mm}{\manuscriptpanelmark{pan:seg-summary-dataset-dice}}{Internal and external Dice}{\includegraphics[width=\linewidth]{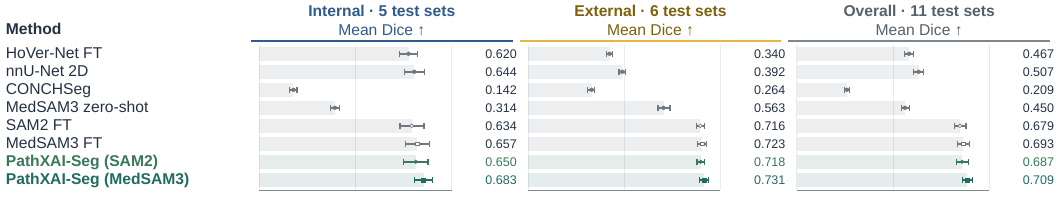}}
  \par\vspace{0.35mm}
  \manuscriptpanelblock{2.9mm}{180mm}{\manuscriptpanelmark{pan:seg-summary-shared-external-comparison}}{Shared external comparison}{\includegraphics[width=\linewidth]{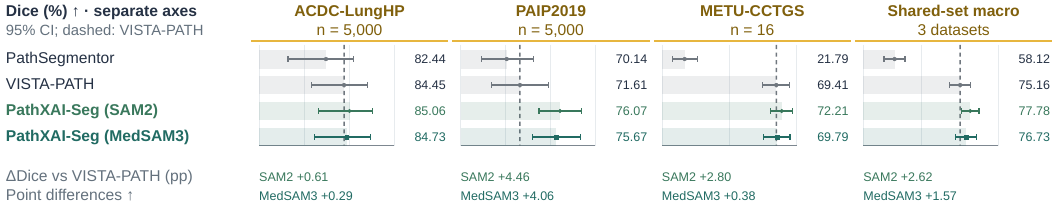}}
  \par\vspace{0.35mm}
  \manuscriptpanelblock{2.9mm}{180mm}{\manuscriptpanelmark{pan:seg-summary-organ-dice}}{PathXAI-Seg vs fine-tuning by organ}{\includegraphics[width=\linewidth]{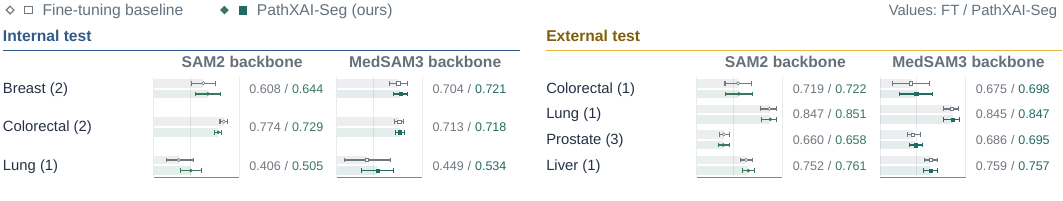}}
  \par\vspace{0.35mm}
  \manuscriptpanelblock{2.9mm}{88mm}{\manuscriptpanelmark{pan:seg-summary-fine-tuning-gains}}{PathXAI-Seg Dice gains over fine-tuning (pp)}{\includegraphics[width=\linewidth]{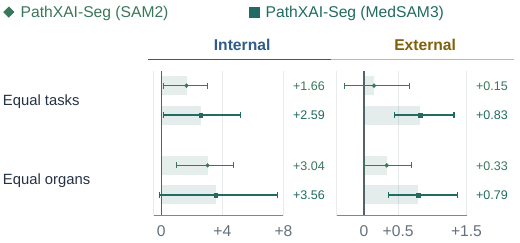}}\hspace{4mm}%
  \manuscriptpanelblock{2.9mm}{88mm}{\manuscriptpanelmark{pan:seg-summary-component-deletions}}{PathXAI-Seg objective ablations}{\includegraphics[width=\linewidth]{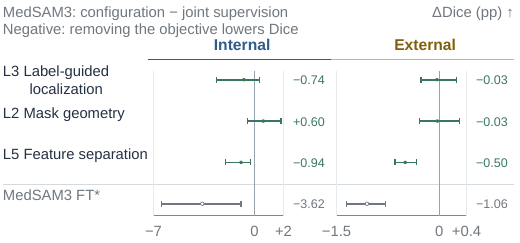}}
  \end{minipage}%
}{%
\textbf{\panelletterref{pan:seg-summary-data-examples} Data and matched examples.} Sample units follow each dataset; AGGC uses specimens. RINGS tumour and gland segmentation reuse the same images, counted for each task.
\textbf{\panelletterref{pan:seg-summary-dataset-dice} Internal and external Dice.} \textcolor{black}{Tasks} receive equal weight.
\textbf{\panelletterref{pan:seg-summary-shared-external-comparison} Shared external comparison.} Target classes follow reference annotations; VISTA-PATH receives full-field boxes (\cref{supp:seg-shared-comparison}).
\textbf{\panelletterref{pan:seg-summary-organ-dice} Dice by organ.} \textcolor{black}{PathXAI-Seg and standard fine-tuning are compared within each backbone.}
\textbf{\panelletterref{pan:seg-summary-fine-tuning-gains} Gains over matched fine-tuning.} \textcolor{black}{Paired differences are PathXAI-Seg minus same-backbone fine-tuning, using equal task or organ weights.}
\textbf{\panelletterref{pan:seg-summary-component-deletions} \textcolor{black}{PathXAI-Seg objective ablations.}} \textcolor{black}{For MedSAM3, selected weights in each labelled objective group are set to zero; differences are ablation minus joint supervision.} \textcolor{black}{The separate MedSAM3 FT* row shows fine-tuning minus joint supervision, a training-stage comparison with unmatched training budgets.} Configurations and checkpoint selection are described in \cref{supp:seg-training-comparisons}.
Whiskers show unadjusted 95\% source-cluster bootstrap intervals within each task, reflecting test-source sampling. Dataset and result details are in \cref{fig:seg-data,tab:source_supp2} and \crefrange{tab:source_seg:part-1}{tab:source_seg:part-4}.
(Notation: $n$, dataset-specific samples; FT, fine-tuning; pp, percentage points. Blue/yellow denotes internal/external tests; parentheses in \panelletterref{pan:seg-summary-organ-dice} give task counts. Dashed lines in \panelletterref{pan:seg-summary-shared-external-comparison} denote VISTA-PATH. Axis ranges vary; tissue colours follow the shared key.)
}

\section{Discussion}\label{sec:discussion}

PathXAI shows that organizing how a pretrained VLM acquires and uses tissue evidence can
improve diagnostic specificity and morphological reporting without updating the reporting
model's weights. \textcolor{black}{PathXAI-Report improved on general-purpose VLMs, which were themselves competitive with
selected pathology-specialized comparators.} Higher agreement against both pathologists' reference sets in
the private cohort extends these findings beyond TCGA-derived references
(\cref{fig:public-summary}).

The tissue-input comparisons help explain where these gains arise. \textcolor{black}{Adding tissue statistics
to H\&E produced small, uncertain diagnostic gains, while the complete workflow increased finding F1
and balanced grade accuracy over direct tissue-statistics input}
(\cref{pan:private-summary-policy-comparisons}). H\&E interpretation remains essential for
architecture and cytology. The comparisons therefore support using tissue measurements to guide
diagnostic review and preserving the resulting decisions during writing
(\cref{pan:private-summary-workflow-comparisons}).

Report usefulness depends on both coverage and correctness. PathXAI-Report more often
combined no/minor correction with complete fields, a distinction that preference ranks or
correction ratings alone did not capture (\cref{pan:private-summary-pathologist-assessment}).
The completeness component checks answer presence, so appropriate abstentions count as incomplete.
Broader finding coverage accompanied lower contradiction than Qwen+H\&E in public cases but higher
contradiction against both private reference sets. Grade gains also included recovered answers
alongside errors between classes. These findings favour writing policies that seek evidence
for required fields and preserve uncertainty when that evidence is insufficient.

Automated assessment also depends on the information documented in the reference reports.
For morphology, an undocumented finding is not a confirmed negative.
\textcolor{black}{Predicting all morphological findings as present yielded high finding F1 in the public
cohort: most documented reference findings were positive, and undocumented finding types were unscored
(\cref{tab:finding-fixed-denominator,tab:finding-reference-polarity}).}
These considerations
make reference agreement one component of report evaluation alongside completeness and reader
assessment.

Evidence links complement these assessments by exposing the information associated with
report fields. H\&E images supply morphology, tissue measurements quantify composition, and
retrieved disease descriptions provide interpretive context. Distinguishing these roles helps
reviewers separate observations of the case from general disease knowledge. \textcolor{black}{The record stores
examined regions, the rule selecting the diagnosis, reasons for requesting further images and
statements qualified during revision, enabling errors to be traced to specific steps.} The necrosis
example retains conflicting regional checks alongside the generated statement
(\cref{pan:case-summary-evidence-links}), illustrating how the record can identify claims
needing reassessment. Whether each linked ROI supports its associated claim still requires
independent evaluation. \textcolor{black}{The record exposes workflow decisions, not the VLM's internal
computation, and whether it shortens review or improves error detection requires a usability
study with pathologists.}

A shared tissue vocabulary provides a consistent basis for these measurements. PathXAI-Seg
combines dense masks and auxiliary image-level targets under common labels, allowing images
without tissue-region masks to contribute to training. Its external gains varied by backbone
and organ (\cref{fig:seg-summary}). \textcolor{black}{Building on advances represented by CONCH~\cite{lu2024conch},
PathSegmentor~\cite{chen2026pathsegmentor} and VISTA-PATH~\cite{liang2026vistapath}, future segmentation
foundation models could use shared labels to integrate heterogeneous annotations, and should be
evaluated for how segmentation errors affect downstream report statements.}

\textcolor{black}{The earlier evidence-organization
and acquisition comparisons used \textcolor{black}{cases used for PathXAI-Report pipeline development}; the tissue-input and writing comparisons
used the 1,200-case public evaluation cohort, which, like the private cohort, did not contribute
to PathXAI-Report pipeline development.} 
Segmentation scores included reference-present classes, and all models in the shared
external comparison received class identities without spatial locations.
\textcolor{black}{External foreground evaluation covered tumour-region masks and gland structures
across four organs, with a broader foreground definition for METU-CCTGS~\cite{arslan2025metu}.} Foundation-model
pretraining overlap remains unknown. The reader study used two pathologists \textcolor{black}{and cases from the \textcolor{black}{PathXAI-Report pipeline development subset}}, with prior ratings
visible during re-review. Further evaluation should therefore apply fixed policies to new
cases and assess reports with additional readers in a fresh, blinded study.

The evaluated workflow used self-hosted Qwen inference without study-specific VLM
fine-tuning. Prospective studies in hospital workflows should test
whether the gains in diagnostic agreement and documented morphology translate into shorter
correction times and fewer residual errors while preserving diagnostic safety. PathXAI thus
offers a route to improving reporting with existing models by making tissue evidence useful
during generation and available for subsequent review.

\FloatBarrier
\section{Methods}\label{sec:methods}
\subsection{Study design, ethics and data governance}\label{sec:methods-study-design}
\textbf{\textcolor{black}{Reporting data preparation.}}
\textcolor{black}{TCGA slides were linked by case to OCR-processed pathology-report text
released by WsiCaption~\cite{chen2024wsicaption}. We restructured these texts into a common
format using Qwen3.6-27B (\cref{supp:report-cohorts,fig:report-data}).
Private cases were randomly sampled within disease categories from six hospitals
(\cref{tab:private_hospital_sources}). Automated report evaluation used cases as the analysis unit.
Reporting cohort sizes and the pathologist-assessment sample size were determined by case
availability and pathologist review capacity, without a prospective sample-size calculation.}

\phantomsection\label{sec:methods-seg-partition}
\textbf{\textcolor{black}{Segmentation data partitioning.}}
\textcolor{black}{Public training sources used fixed per-dataset 80:20 sample partitions
(seed 42); held-out pixel-labelled samples formed the internal test set. These sample partitions
do not establish patient-level separation (\cref{supp:seg-labels-inputs}). Internal and external
test data were used only for evaluation.}

\textbf{\textcolor{black}{Ethics approval.}}
\textcolor{black}{Approval} for the prostate and colorectal studies was obtained at Shanghai Changzheng Hospital
(approval nos. 2024SL136 and CHEC2024-421, respectively), and for the breast study at
Tan Tock Seng Hospital (DSRB 2025-1447). Approval 2024SL136 covered prostate data from all five
contributing hospitals. Informed consent was waived for all three studies.

\textbf{\textcolor{black}{Data governance and reporting.}}
\textcolor{black}{Data from China were processed on servers in China, and data from Singapore
on servers in Singapore. Private reference reports and ratings remain restricted; the private
dataset will not be publicly released. Figures contain no private-case images.
Item-level reporting checklists with manuscript locations accompany the submission.}

\subsection{PathXAI-Report}
PathXAI-Report combines H\&E images and anatomical site information with tissue evidence from
PathXAI-Seg to guide diagnostic comparison, reinspection and structured report generation
(\cref{fig:methods-summary}). The segmentation model provides tissue label maps, aligned
overlays and tissue-class fractions for selected regions; its architecture, training and inference
are described in \hyperref[sec:methods-seg]{PathXAI-Seg}.
\textcolor{black}{Adding these outputs directly to the VLM input was not sufficient: tissue statistics gave small,
uncertain diagnostic gains and overlays lowered cancer family match rate
(\hyperref[sec:results-report-policy]{Results}). PathXAI-Report therefore assigns tissue evidence to
specific steps through four mechanisms. (i) Overlays and tissue statistics are used separately:
overlays guide diagnostic candidates, whereas statistics support an independent check.
(ii) A rule-based selector chooses between two candidate diagnoses, generated with and without
domain knowledge. (iii) An evidence assessment decides whether the current images suffice and,
if not, which views to acquire. (iv) Every step is recorded and linked to the report fields it supports.}
\manuscriptwidefigurepart{first}{PathXAI architecture and report generation with tissue evidence.}{fig:methods-summary}{%
  \begin{minipage}{180mm}\offinterlineskip\vspace*{0.2mm}
\newcommand{\figsixpanelblock}[5]{%
  \begin{minipage}[t]{#2}\offinterlineskip
  \parbox[t][#1][t]{\linewidth}{{\fontencoding{T1}\fontfamily{phv}\fontsize{9}{10}\selectfont\textbf{#3}\hspace{1.5mm}\textbf{#4}}}\par
  #5\end{minipage}%
}

  {\setlength{\fboxsep}{0pt}\colorbox[HTML]{EFF7F4}{\parbox[c][3.6mm][c]{180mm}{\hspace*{1mm}\color[HTML]{216B63}\fontfamily{phv}\fontsize{9}{10}\selectfont\bfseries PathXAI-Seg: architecture and training}}}\par
  \begin{overpic}[width=180mm]{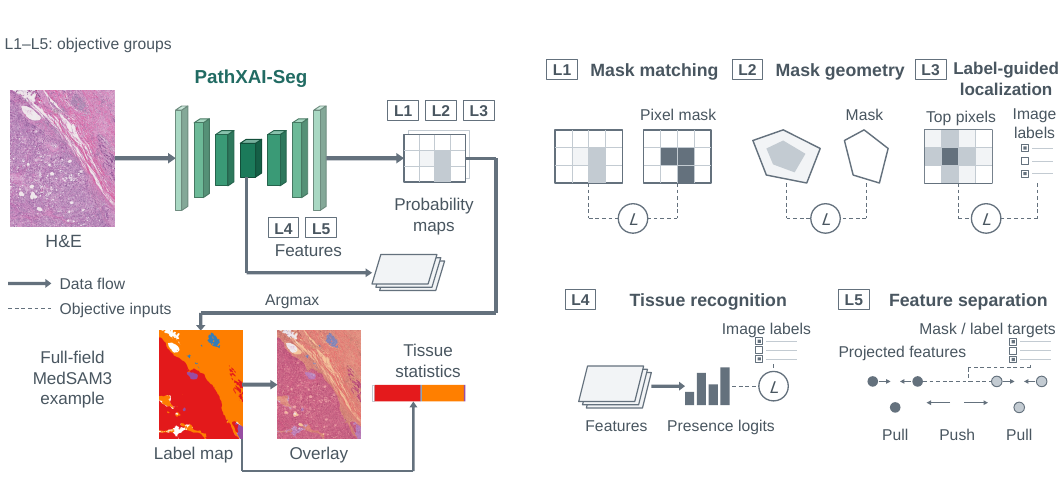}
    \put(0.22,43.55){{\color[HTML]{253144}\fontencoding{T1}\fontfamily{phv}\fontsize{9}{10}\selectfont\bfseries\manuscriptpanelmark{pan:methods-seg-model}\hspace{1.5mm}Segmentation and tissue evidence}}
    \put(50,43.55){{\color[HTML]{253144}\fontencoding{T1}\fontfamily{phv}\fontsize{9}{10}\selectfont\bfseries\manuscriptpanelmark{pan:methods-training-objectives}\hspace{1.5mm}Training objectives}}
  \end{overpic}
  \par\vspace{0.4mm}
  \includegraphics[width=180mm]{figures/f007.pdf}
  \par\vspace{0.6mm}
  {\setlength{\fboxsep}{0pt}\colorbox[HTML]{FCF1F7}{\parbox[c][3.6mm][c]{180mm}{\hspace*{1mm}\color[HTML]{A31559}\fontfamily{phv}\fontsize{9}{10}\selectfont\bfseries PathXAI-Report: overview and key mechanisms}}}\par
  \vspace{0.4mm}
  \parbox[c][6.2mm][c]{180mm}{{\fontencoding{T1}\fontfamily{phv}\fontsize{9}{10}\selectfont\textbf{\manuscriptpanelmark{pan:methods-report-overview}}\hspace{1.5mm}\textbf{Report generation overview}\hfill{\fontsize{8}{9}\selectfont\textbf{V}\hspace{1mm}Candidate with KEEP\hspace{3mm}\textbf{T}\hspace{1mm}Candidate without KEEP\hspace{3mm}\textbf{S}\hspace{1mm}Tissue-statistics assessment\hspace{3mm}}\raisebox{-1.3mm}[0pt][0pt]{\includegraphics[width=6mm,height=6mm]{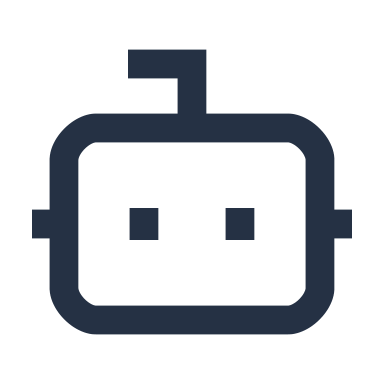}}\hspace{1mm}{\fontsize{8}{9}\selectfont VLM}\hspace{1mm}}}\par
  \includegraphics[width=180mm]{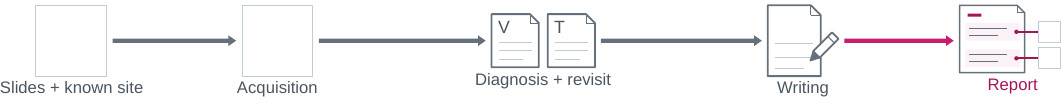}
  \par\vspace{0.4mm}
  \figsixpanelblock{3.6mm}{108mm}{\manuscriptpanelmark{pan:methods-evidence-candidates}}{Evidence acquisition and diagnostic candidates}{\includegraphics[width=\linewidth]{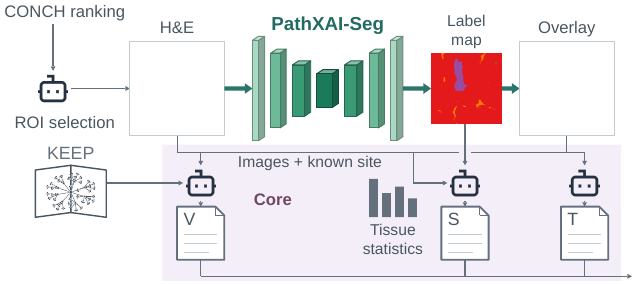}}\hspace{4mm}%
  \figsixpanelblock{3.6mm}{68mm}{\manuscriptpanelmark{pan:methods-diagnosis-revisit}}{Diagnosis selection and revisiting}{\includegraphics[width=\linewidth]{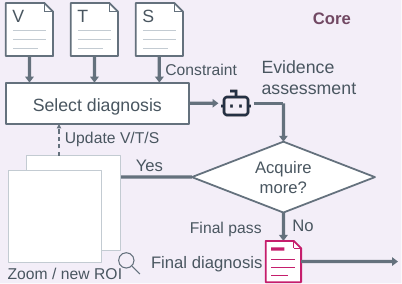}}
  \par\vspace{0.4mm}
  \figsixpanelblock{3.6mm}{180mm}{\manuscriptpanelmark{pan:methods-constrained-writing}}{Constrained writing and structured report}{\includegraphics[width=\linewidth]{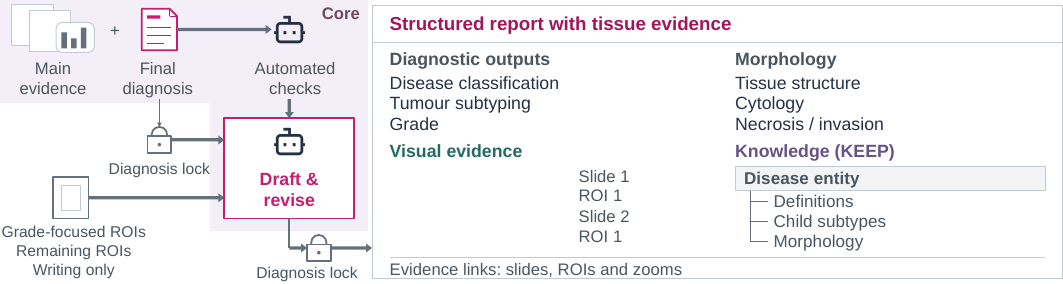}}
  \end{minipage}%
}{%
\textbf{\panelletterref{pan:methods-seg-model} Tissue segmentation.} SAM2 and MedSAM3 implementations; examples use MedSAM3.
\textbf{\panelletterref{pan:methods-training-objectives} Training objectives.} L3 uses image labels; L5 spatial contrast uses masks or image-label-constrained pseudo-labels.
\textbf{\panelletterref{pan:methods-report-overview}--\panelletterref{pan:methods-constrained-writing} Reporting workflow, diagnostic candidates, reinspection and constrained writing.} \textcolor{black}{V denotes the candidate augmented with KEEP in the complete configuration; T denotes the candidate without KEEP. Both use the same \textcolor{black}{H\&E--overlay composites}, site and VLM. Core omits KEEP. S \textcolor{black}{denotes} the tissue-statistics assessment\textcolor{black}{, which uses} images, site and tissue statistics for broad-lineage and subtype-sufficiency checks. V, T and S enter diagnosis selection in (e), where S supplies the subtype constraint; the final diagnosis constrains report writing in (f).} PathXAI-Report extends invasion safeguards and adds writer-only grade ROIs and diagnosis locking. \textcolor{black}{Grade-focused ROIs are selected from the remaining ranked candidates. The lock preserves the final diagnostic fields during writing and revision.} \textcolor{black}{Evidence assessment triggers zooms or new ROIs, which update V, T and S.} \textcolor{black}{Automated claim checks precede drafting and revision.} Configuration details are in \cref{supp:report-configurations}; architecture and workflow details are in \cref{fig:seg-model-sam2,fig:seg-model,fig:report-model}.
(Notation: VLM, vision-language model; ROI, region of interest; $L$, loss. White denotes background; tensors, features and report fields are schematic. Light purple marks core operations; locks mark diagnosis-constrained writing. Arrows show data flow; fine L5 arrows, attraction/repulsion; headless dashes, objective inputs; dashed arrows, revisiting. \textcolor{black}{Robots denote VLM roles.})
}

\subsubsection{Reporting configurations and inference}\label{sec:methods-report-inference}
Inference used the fixed \texttt{Qwen3.6-27B-Q6\_K.gguf} model on a researcher-operated,
self-hosted service. Core and complete configurations used the same settings across public and
private cohorts, with temperature 0.2 and a maximum of
4,096 generated tokens per call, including report writing. \textcolor{black}{Thinking mode was disabled
for all evaluated VLMs.} Images were resized to a maximum
edge of 768 pixels and transmitted as JPEGs at quality 75.

Both configurations construct an evidence graph, perform automated claim checks and write
the report. The diagnostic, acquisition and writing stages below specify their shared procedures
and the additional safeguards in the complete configuration. Selected ROIs may differ between
configurations, so the complete-versus-core comparison includes changes to both evidence inputs
and workflow.
Predefined rules select between candidate diagnoses, retrieve disease
information and choose new regions or zoomed views, without additional
VLM calls.

\textcolor{black}{PathXAI-Report pipeline development and debugging used only 456 public cases, which also supported the earlier evidence-organization and acquisition comparisons} (\cref{supp:report-policy-comparisons}).
\textcolor{black}{Primary report evaluation used the separate 1,200-case public evaluation cohort and the 151 shared private cases; combined public results are in \cref{supp:results-public-full}.}
Stage inputs, message assembly, prompt templates and output formats are provided in
\cref{supp:report-configurations}; the machine-readable templates accompany the supplementary material.

\subsubsection{ROI ranking and initial selection}\label{sec:methods-report-acquisition}
The inputs are H\&E slides for each case and supplied site descriptions
(\cref{pan:overview-workflow,pan:methods-evidence-candidates}).
CONCH~\cite{lu2024conch} encodes tissue tiles and diagnostic text prompts to
prioritize candidate regions before segmentation. The configured tile size is
1,024 pixels at 0.5~$\mu$m per pixel. For each tile, let $p_j$ and $n_k$ denote
cosine similarities to the positive and negative prompts, respectively. Its
ranking score is
\[
s = 0.70\max_j p_j + 0.30\operatorname{mean}_j p_j - 0.35\max_k n_k.
\]
The positive and negative query topics are listed in \cref{supp:roi-ranking}.
Candidates from the input slides are pooled by score, retaining up to 20 per
case. Ranked tiles identify the centres of larger diagnostic ROIs, extracted at slide level 0
with a default side length of 4,096 pixels.

The VLM reviews images of the leading eight candidates and selects up to
three initial diagnostic ROIs. Selection follows the returned candidate
indices; if none is valid, the highest-ranked candidates provide the fallback.
The selected H\&E regions then receive PathXAI-Seg (MedSAM3) predictions.

\subsubsection{\textcolor{black}{Tissue evidence in two forms}}\label{sec:methods-report-evidence}
\textcolor{black}{Each segmented ROI provides tissue evidence in two forms that answer different questions.
An overlay shows where each tissue component lies on the same pixels the VLM inspects. Each
H\&E ROI is paired with its overlay as an H\&E--overlay composite, the visual input to both
diagnostic candidates (\cref{pan:methods-evidence-candidates}).}
The VLM assesses architecture
and cytology from the images, which contain information beyond the seven segmentation classes.
\textcolor{black}{Tissue statistics state how much of each component is present, which is difficult to estimate
reliably from the image alone. They are used in a separate tissue-statistics assessment, S.}
\textcolor{black}{S assesses} broad lineage and subtype sufficiency using ROI images,
anatomical context and regional tissue fractions. \textcolor{black}{The subtype assessment subsequently
constrains the selected candidate, and both preliminary records enter the evidence graph.}
\textcolor{black}{The two forms are kept apart:}
\textcolor{black}{neither candidate prompt includes numeric tissue summaries or the preliminary assessment text}\textcolor{black}{,
so S checks the candidates independently rather than steering how they are generated.}

\subsubsection{Diagnostic candidates and selection}\label{sec:methods-report-diagnosis}
Two \textcolor{black}{VLM} calls generate diagnostic candidates V and T from the
\textcolor{black}{same H\&E--overlay composites} and anatomical context. \textcolor{black}{In the complete
configuration, V also receives KEEP context and T receives none; core omits KEEP. Retrieved
knowledge can make a diagnosis more specific but can also pull it towards a retrieved entity,
so T serves as a knowledge-free reference for V.}
In the complete configuration, the V path also receives disease definitions and child-subtype
candidates retrieved from the KEEP disease graph and pathology semantic groups~\cite{zhou2026keep}.
The initial query uses specimen site; subsequent queries use the preliminary subtype or primary
diagnosis and up to three differential diagnoses. Context is refreshed when reinspection changes
the hypothesis. \textcolor{black}{Retrieved material provides disease context}; matching and ranking rules are given in
\cref{supp:report-knowledge}.

The selector compares candidate validity and diagnostic specificity. A candidate is valid
if its diagnostic text is non-empty and contains no explicit non-diagnostic failure marker.
For valid candidates, specificity is 3 for a non-generic subtype; otherwise 2 for a
non-generic primary diagnosis not ending in ``not otherwise specified'' (NOS); otherwise 1 for a broad type other
than generic neoplasm; otherwise 0. Invalid candidates receive $-1$.

If only one candidate is valid, it is selected; if neither is valid, V is returned as a fallback.
When both are valid, the core selector
retains V for a non-generic tumour-family disagreement, an invasive/in-situ conflict, or when T is less specific than an
already specific V (V specificity score at least 2 and T lower); otherwise T is selected, including
at equal specificity. The complete configuration adds invasion-state safeguards. It retains V
when V indicates invasion and T indicates in-situ disease, ambiguous in-situ disease or
unassessable invasion. It also retains V when V indicates in-situ or ambiguous in-situ disease
and T indicates invasion. Other cases use the core selection rules.
\textcolor{black}{Because selection follows fixed rules, the rule applied in each round is recorded with both
candidates and can be reproduced (\hyperref[sec:methods-report-trace]{Traceable reasoning record}).}

\subsubsection{\textcolor{black}{Evidence assessment and reinspection}}\label{sec:methods-report-reinspection}
\textcolor{black}{The initial ROIs may miss tumour or the features needed for subtyping and grading. Much as a
pathologist returns to the slide, an evidence assessment after each selection decides whether
more images are needed.} The \textcolor{black}{assessment} prompt asks whether tumour tissue and the required architecture, cytology,
invasion and necrosis are visible. The model returns sufficiency (yes, partially or no),
self-reported diagnostic confidence (high, medium or low), separate lists of missing image and non-image evidence, suggested
additional ROIs, and a Boolean reportability decision. The reinspection logic in
\cref{pan:report-model-reinspection} uses this Boolean decision to trigger further acquisition.

The missing-evidence categories determine the next acquisition action\textcolor{black}{, so images are acquired
for a stated gap rather than uniformly}. Eligible current ROIs
are preferred for zoom when image gaps mention cytology, nuclear detail or mitotic features;
other image gaps trigger additional regions, and non-image-only gaps do not trigger image acquisition.
A failed or exhausted zoom falls back to new regions. A zoom uses the aligned parent mask
crop when available and is re-segmented otherwise; a new ROI receives a new PathXAI-Seg prediction.
Digital zoom enlarges a crop within the parent ROI without increasing optical sampling.
New ROIs are chosen from the unselected candidates.
The first addition has the highest remaining score; subsequent additions
maximize minimum coordinate distance from the regions selected in that
acquisition step.
\textcolor{black}{New images update V, T and S, and selection is repeated.}

While the decision remains false, the system acquires more evidence until it reaches
a round or ROI limit, has no unused acquisition action, or obtains the same diagnosis signature
in two successive updates. \textcolor{black}{The signature comprises broad type, primary diagnosis, subtype and malignancy after case and punctuation normalization.} Reinspection allowed up to two
ROIs per round, four rounds, one zoom level and eleven final ROIs per case. Acquisition can stop with evidence
gaps unresolved; \textcolor{black}{a final diagnostic pass then produces the final diagnosis used for report writing.}
\textcolor{black}{Additional acquisition details are given in \cref{supp:report-acquisition}.}

\subsubsection{Diagnosis-constrained report generation}\label{sec:methods-report-writing}
Before drafting, both configurations associate claims extracted from the diagnostic fields with
image inputs, tissue statistics and contextual nodes in an evidence graph. Automated assessments of these
claims guide subsequent report revision. Heuristic checks assign an initial support status;
a valid VLM assessment for the same claim replaces that overall status while preserving the
tissue-rule subcheck. Missing or failed VLM assessments retain the heuristic result. Known-site
statements stay supported, and explicit ``not assessable'' statements cannot be relabelled weak,
unsupported or contradicted. These automated checks have not been independently validated by
pathologists.

The writer first produces a structured draft with sections for specimen/site, diagnosis,
lesion description, microscopic findings, invasion/necrosis/margins, ancillary studies,
stage/grade/prognostic features and a summary. Drafts missing required headings or minimum
content are retried once, followed by a diagnosis-preserving fallback if needed. Revision then
uses the claim assessments to state supported claims affirmatively, qualify weak claims,
remove unsupported or contradicted claims and explicitly mark unassessable fields. An invalid
revision leaves the draft unchanged.

In the complete configuration, report generation follows the \textcolor{black}{final diagnostic pass}. A diagnosis
lock preserves diagnostic identity, subtype and invasion from the \textcolor{black}{final diagnosis}
(\textcolor{black}{\cref{pan:methods-constrained-writing}}), with anatomical site retained as input context\textcolor{black}{,
so writing cannot change the diagnosis reached through selection and reinspection}.
A final consistency step
preserves the locked diagnostic fields after revision.
To supplement the writing evidence, a grade-focused selector excludes the diagnostic ROIs and
ranks the remaining candidates using existing scores and grade-related metadata. Its default
budget is two additional H\&E ROIs, configurable up to four; the illustrated complete run used
two. These images receive no additional segmentation and enter writing only, supporting grade,
necrosis and cytological description without revising site, primary diagnosis, subtype or
invasion. If the \textcolor{black}{final diagnosis} already specifies a grade, that grade is also retained;
the auxiliary images can inform its accompanying description. Ranking and duplicate-exclusion
rules are specified in \cref{supp:report-writing}.

\subsubsection{\textcolor{black}{Traceable reasoning record}}\label{sec:methods-report-trace}
\textcolor{black}{PathXAI-Report stores the intermediate results of every step, not only the final report.
For each case, the record contains the candidate and selected ROIs and zooms with slide
coordinates; the images, overlays and tissue statistics supplied to each call; the V, T and S
outputs, the selected diagnosis and the selection rule applied in each round; each evidence
assessment and the acquisition action it triggered; and the extracted claims with their
evidence links and support states before and after revision. Each returned VLM output is stored
with its prompt, generation settings and hashes of the encoded images.}
The resulting report is displayed alongside slides, ROI/zoom
relationships, overlays and tissue statistics in the viewer, which also shows acquisition steps,
image inputs and automated checks associated with each report field
(\cref{fig:report-interface,fig:report-evidence-interface,fig:report-reasoning-interface,fig:report-verification-interface,supp:report-interface}).
\textcolor{black}{A reader can thus follow any report statement back to its supporting regions, tissue evidence
and reasoning steps. The record makes each workflow decision open to inspection; it does not
explain the VLM's internal computation.}

\FloatBarrier
\subsection{Evaluation and statistical analysis of reporting}\label{sec:methods-report-evaluation}
\subsubsection{Pathologist reference-report authoring}\label{sec:methods-report-human}
Weixiang Zhong and Xiaoxi Wang are attending pathologists at The First Affiliated Hospital,
Zhejiang University School of Medicine, Hangzhou, China; both hold doctoral degrees. They had
16 years of experience in gastrointestinal and breast pathology and 12 years in central nervous
system pathology, respectively. Each authored private reference reports and independently
assessed public reports in the ranking and correction assessment described below.

The two pathologists' reports form separate reference sets, P1 and P2.
Pathologists could leave individual reference fields blank at their discretion.
Generated reports were evaluated separately against each reference set, using only
cases with assessable reference information for each metric.
For every private case, each pathologist authored a reference report from slides and
anatomical site information, which was provided at the pathologists' request.
They did not have access to the original clinical reports or diagnoses.
The Chinese reference reports were translated into English using Qwen and checked against
the originals by both pathologists. The English references were finalized before automatic scoring
and were not subsequently revised in response to model outputs or scores. The WSI/ROI workspace
provided structured fields and narrative sections
(\cref{fig:physician-authoring,supp:reference-authoring}).

For agreement between the pathologists' reports for each case (\cref{pan:public-summary-reference-agreement}),
we compared broad diagnosis, specific diagnosis, malignancy, invasion, necrosis and grade using
the terminology and polarity rules in \hyperref[sec:methods-report-extraction]{Report text extraction}. Each field included only
cases with definite values in both pathologists' reports; grade additionally required the same grading
system. We counted matching and differing values separately and divided each count by the
number of comparable cases to obtain the displayed proportions.
\textcolor{black}{Chance-adjusted agreement is reported in \cref{tab:private-reference-agreement}.}

\subsubsection{Automated report evaluation}\label{sec:methods-report-automated}
Text-overlap metrics such as ROUGE~\cite{lin2004rouge} do not reliably distinguish a correct diagnosis
from a similarly worded report with a clinically important error~\cite{tran2025histogpt}. In chest X-ray reporting, entity-and-relation measures and
clinical-error measures aligned more closely with radiologist assessments than text-overlap
scores~\cite{yu2023evaluation,ostmeier2024green}. We therefore designed a pathology-specific
evaluation framework that separately assesses diagnostic identity, morphological agreement and
completeness, and opposing assertions. Shared terminology,
synonym and polarity rules allow these comparisons across differently worded reports, while
reference-based applicability avoids treating undocumented findings as negative observations.

\paragraph{Report text extraction}\label{sec:methods-report-extraction}
All methods used the same terminology and extraction rules for final reports. Scoring used final
text, reference fields and anatomical context, without internal reasoning, segmentation predictions
or method identity.

The study lexicon contains 84 entities and 236 canonical or synonymous
expressions. Recognized diagnostic names, abbreviations and synonyms map to canonical entities;
synonym equivalence, parent-child relations and TCGA-family membership are separate mappings.

Diagnostic extraction prioritized recognized diagnosis sections. If none was present, it used
final-summary and unsectioned text. Explicit subtype fields in other eligible sections contributed
only their stated subtype. Recognized history, differential-diagnosis, recommendation and
ancillary-study sections were excluded. \textcolor{black}{Final diagnostic addenda were included;
gross-description and frozen-section text did not supply the final diagnosis.}
A longer child entity could refine a nested parent label;
unrelated affirmed entities remained conflicting. Negated, uncertain and alternative diagnoses
did not receive credit for an exact diagnosis match. Anatomical context could resolve an unqualified entity but
could not supply an unreported histological subtype.

Finding extraction uses 20 concepts covering invasion and its specific routes, necrosis,
metastasis, cytological features, tissue architecture, stromal changes, inflammatory infiltrate
and calcification. Case-insensitive, finding-specific patterns normalize lexical variants and
assign present, absent, uncertain or conflicting states. Polarity is assigned to individual
findings within clauses, including shared negative predicates in coordinated lists. Requests to
assess a finding are not observations. Only explicitly documented positive or negative reference
findings are assessable; reference silence is not a negative finding. Structured pathologist
reference fields retain the recorded judgement. The full concept list and scope-specific
extraction rules are given in \cref{supp:report-omissions}.

A sensitivity analysis also included final-summary diagnoses when a diagnosis section was present.
The extraction rules have not been independently validated across methods and report formats.
Detailed text normalization, polarity rules, reference-target handling and the sensitivity analysis
are specified in \cref{supp:report-scoring}; the machine-readable terminology and matching rules
\textcolor{black}{are included in the released evaluation code (see Code availability)}. Public content analyses based on OCR-derived
references were exploratory.

\paragraph{Primary evaluation measures}\label{sec:methods-report-metrics}
All primary measures are reported as percentages, averaging equally over applicable cases.
\Cref{tab:report-metric-overview}
summarizes the primary and additional measures.

\textbf{Cancer family match rate.}
This is the fraction of applicable cases in which the generated diagnosis belongs to the
target cancer family. Public targets come from TCGA disease-family labels. Private targets
are derived separately from each pathologist's definite cancer diagnosis using the same
family-membership rules; benign or unresolved private references are inapplicable.
A diagnosis can match the cancer family without identifying the exact histological subtype.
Let $I$ be the cases with an applicable cancer-family target and $s_i$ indicate
whether the generated diagnosis matches that family. The reported percentage is
$100|I|^{-1}\sum_{i\in I}s_i$. An applicable case with a missing, uncertain,
conflicting or incompatible prediction contributes zero. \textcolor{black}{Cases with inapplicable references are excluded from $I$.}

\textbf{Exact diagnosis match rate.}
This is the percentage of cases with a definite, specific reference diagnosis in which the
generated diagnosis maps to the same canonical disease entity or subtype. For example,
``conventional renal cell carcinoma'' and ``clear cell renal cell carcinoma'' map to the same
entity, whereas generic ``renal cell carcinoma'' does not identify this specific subtype. Public targets come from the case reports,
and private targets come from each pathologist's report separately.
We divide the number of matching diagnoses by the number of cases with an applicable reference
and multiply by 100. Missing, uncertain, conflicting or incompatible predictions contribute zero;
cases without an applicable reference diagnosis are excluded.

\textbf{Finding F1.}
This score balances how often definite finding statements agree with the reference
(precision) and how many reference findings are correctly reproduced (recall), using their
harmonic mean. The same finding-specific rules map recognized lexical variants, such as
``mitoses'' and ``mitotic figures'', to one concept in generated and reference narratives.
Scoring includes only finding types explicitly documented as present or absent
in the reference, such as invasion, necrosis or glandular architecture. Both agreeing that a
finding is present and agreeing that it is absent count as matches. For each case, the score
is twice the number of matching finding types divided by the sum of the documented reference
types and the types addressed by a definite generated assertion. Missing, uncertain or
conflicting statements provide no match or definite assertion. A case with documented
reference findings but no definite generated assertions scores zero; a case with no documented
reference findings is excluded. We average the applicable case scores and multiply by 100.
For example, two matches among three assertions against four documented reference findings
give a case score of $4/7$.

\Needspace{8\baselineskip}
\textbf{Finding recall.}
This measures the completeness of reference-matching morphology. For each case, it is the
number of documented reference finding types correctly reproduced as present or absent,
divided by the total number of documented reference finding types. It is the recall component
of finding F1. Missing, uncertain, conflicting or opposite statements receive no credit;
findings not documented in the reference do not enter the denominator. A case with no documented
reference findings is excluded. The reported percentage is 100 times the mean of the applicable
case-level fractions, with equal weight for each applicable case.

\textbf{Finding contradiction rate.}
This measures how often a definite generated statement opposes the reference among finding
types addressed by both reports. For each case, we divide the number of opposite present/absent
statements by the number of reference-documented finding types assigned a definite present/absent
assertion in the generated report. Missing, uncertain or conflicting assertions do not enter this
denominator, and cases with no comparable definite assertions are excluded. The reported percentage
is 100 times the mean of the applicable case-level fractions. Because omitting findings can reduce
opportunities for contradiction, we interpret this measure alongside finding recall
and finding coverage.

\textbf{Finding coverage.}
This additional measure describes how much of the documented morphology a report addresses,
whether correctly or incorrectly. For each case, we divide the number of reference-documented
finding types addressed by a definite generated present/absent assertion by the total number
of documented reference finding types. Missing, uncertain or conflicting statements do not
count as definite assertions. Cases with documented findings but no definite assertions score
zero; cases with no documented reference findings are excluded. The reported percentage is
100 times the mean of the applicable case-level fractions.

\Needspace{7\baselineskip}
For the four morphology measures, let $D_i$ be the number of finding types documented as
present or absent in case $i$'s reference, $A_i$ the number of these types addressed by a definite generated
assertion, $C_i$ the number with matching polarity and $E_i$ the number with opposite polarity,
so that $A_i=C_i+E_i$. The case-level formulas, with the applicability rules above, are
\begin{equation}
\mathrm{F1}_i=\frac{2C_i}{D_i+A_i},\qquad
\mathrm{Recall}_i=\frac{C_i}{D_i},\qquad
\mathrm{Contradiction}_i=\frac{E_i}{A_i},\qquad
\mathrm{Coverage}_i=\frac{A_i}{D_i}.
\end{equation}

\paragraph{Additional grade and coverage measures}\label{sec:methods-report-findings}
\textbf{Balanced grade accuracy.}
This is the mean of the correct-prediction fractions within each represented reference-grade
class, giving each grade equal weight. Thus, a common grade does not dominate the score simply
because it occurs more often. An explicit overall grade takes precedence over incidental
adjectives; isolated nuclear morphology does not establish overall grade. Grading systems are
evaluated separately. Missing or incompatible predictions are incorrect when the reference
grade is assessable, and unassessable references are excluded. For the set $G$ of represented
reference grades and cases $I_g$ with reference grade $g$, balanced grade accuracy is
\begin{equation}
B_{\mathrm{grade}}=\frac{1}{|G|}\sum_{g\in G}\frac{1}{|I_g|}
\sum_{i\in I_g}\mathbf 1(\widehat g_i=g).
\end{equation}
We multiply this score by 100 for percentage reporting. Confusion counts and an
always-highest-grade control complement this measure (\cref{supp:report-grade-findings}).

\textbf{Report field completeness.}
Report field completeness measured whether a report gave definite answers for each applicable
field among broad diagnosis, specific diagnosis, malignancy, grade, invasion and necrosis. A reference
field was applicable only when definite and non-null; the specific diagnosis field additionally required a
specific disease entity or histological subtype. Missing, uncertain or conflicting generated fields were uncovered. Answers
did not have to be correct or use a compatible grading system to count towards coverage.
This measure follows the rationale for explicit elements in synoptic reporting~\cite{schaad2024synoptic}
and differs from finding recall, which requires agreement with documented morphology.
For applicable fields $K_i$, this coverage is
$|K_i|^{-1}\sum_{k\in K_i}\mathbf 1(\text{generated field }k\text{ is definite})$.
A case with no applicable fields is excluded. We average the applicable case scores and
multiply by 100.

Definitions of the additional diagnostic, grade, critical-finding and wording-overlap measures
reported in the supplementary analyses are provided in
\cref{supp:report-scoring,supp:report-grade-findings,supp:report-omissions,supp:report-wording}.

\paragraph{Report uncertainty and paired comparisons}\label{sec:methods-report-statistics}
Case-level scores were averaged equally across applicable cases, without cancer-type weighting.
Balanced grade accuracy instead weighted the represented reference grades equally.
\textcolor{black}{Pointwise 95\% confidence intervals used exact binomial limits for binary
proportions and conservative simultaneous binomial limits for balanced and ordinal grade
scores and joint-reference binary means (\cref{supp:report-statistics}). Other case-average measures and paired differences
used 4,000 case-bootstrap draws (seed 20260831), with resampling within reference-grade
classes for paired balanced grade accuracy. Bootstrap limits were the 2.5th and 97.5th
percentiles. These intervals were not adjusted across comparisons.}
\textcolor{black}{An additional sensitivity analysis applied Holm adjustment to the 126 paired
comparisons of PathXAI-Report with the displayed automated-report comparators
(\cref{supp:results-report-multiplicity}); test definitions are in \cref{supp:report-statistics}.}
\textcolor{black}{Cancer-type- and organ-stratified intervals are reported in
\cref{supp:results-report-stratification}.}

Primary comparisons used cases with reports available from every included method or
configuration; supplementary private estimates also used all available reports.
\textcolor{black}{A sensitivity analysis retained all eligible cases and assigned zero to
unavailable reports for diagnostic match rates, finding F1 and finding recall
(\cref{supp:results-report-availability}).}
Paired contrasts used within-case score differences on cases applicable to both methods,
averaged equally across cases for ordinary measures and first within reference-grade classes
then equally across classes for balanced grade accuracy. Each private
reference set determined which cases were eligible for scoring. Paired comparisons of finding contradiction rate required at least one comparable assertion from each method in each retained case;
the methods need not assert the same findings. For joint-reference comparisons, we
averaged the two reference-specific scores within each case where both fields were applicable before resampling cases.
Patient-level linkage was unavailable in the private cohort, so these intervals assumed independence
between cases.

\textcolor{black}{A sensitivity analysis excluded identifiable TCGA cases shared with segmentation
training pools (\cref{supp:results-report-training-overlap}).}

Case-eligibility rules, bootstrap settings and reference-grade limitations are in
\cref{supp:report-statistics}.

\subsubsection{Pathologist ranking and correction assessment}\label{sec:methods-report-ranking}
\paragraph{\textcolor{black}{Presentation and re-review}}
The two pathologists initially assessed reports from MedGemma, Qwen+H\&E,
PLIP-guided Qwen and PathXAI-Report (core) in a four-report comparison. PathXAI-Report was
then assessed alongside these reports on a 1--5 rank scale, with the existing scores available
for revision. All five reports underwent \textcolor{black}{re-review under anonymous method labels},
with a separate key linking labels to methods. \textcolor{black}{Report presentation order was randomized using a key specific to each case,
pathologist and round, and remained fixed within each assignment.} Prior overall ranks and
correction ratings remained visible after reordering. Both pathologists retained those values
unchanged. Each rating was counted once, and ratings were analysed separately without
consensus adjudication. \textcolor{black}{Readers had no access to reference reports or TCGA
diagnoses during either assessment round.}

\paragraph{Ratings and statistical analysis}
Overall ranks ranged from 1 (best) to 5 and allowed ties. Correction categories were none,
minor, major or unusable. Checkbox concerns recorded unsupported major claims and visible
omissions as reader selections, not adjudicated events. Section-specific rankings were excluded
from the primary analysis because prior section scores could be reused. Rare concern flags
were reported descriptively. Interface controls and concern-flag interpretation are described in
\cref{fig:physician-ranking,supp:rating-protocol}.
Ranks, correction categories and reader concern flags were analysed separately.
For $N=132$ cases and $R=2$ readers, the mean rank for method $m$ is
$(NR)^{-1}\sum_{i=1}^{N}\sum_{r=1}^{R}q_{irm}$, where $q_{irm}$ is the assigned rank
and 1 denotes the most preferred report. Each correction-category proportion is
the number of ratings in that category divided by $NR=264$. The no/minor-correction
proportion combines the first two categories. Concern-flag proportions use the
same denominator. Percentages multiply these proportions by 100.
Case-cluster resampling of pathologist ratings retained both observed readers and
all methods together; these intervals describe uncertainty for the evaluated cases and fixed readers. The bootstrap draw count and interval construction follow the automatic-report
analysis described above.

\paragraph{Report-content checks}\label{sec:methods-report-checks}
\textcolor{black}{The no/minor correction with complete fields criterion was defined before
pathologist assessment.} We applied two content checks to reports rated as requiring no or only minor correction.
The first flagged empty reports, empty recognized Diagnosis/Primary Diagnosis sections, or
explicit full-field unavailable results such as ``not specified'' or ``No diagnosis generated'',
using identical rules across methods. Generic diagnoses remained nonempty; unrecognized
headings required format review. The second required a definite answer for each
applicable field among broad diagnosis, specific diagnosis, malignancy, grade, invasion and necrosis.
A field was applicable when it could be assessed in the reference report; missing, uncertain or
conflicting answers failed this check. A major/unusable rating, output failure or failure of any
applicable field check counted as joint failure, with overlaps counted once. All 264 ratings per method
remained in the denominator. These automated checks left the pathologists' ratings unchanged.
For the no/minor correction with complete fields measure, each case-reader pair
contributes one only if all three requirements hold, and zero otherwise; the
mean of these indicators is reported as a percentage. Joint failure is its complement.
The field checks assessed whether the fields had definite answers, not whether those answers were correct
(\cref{supp:rating-checks,tab:source_fig10}).

\subsection{PathXAI-Seg}\label{sec:methods-seg}
Study-specific training is confined to the segmentation component,
trained on a single NVIDIA GeForce RTX 4090 GPU.
\subsubsection{Tissue labels and image preparation}\label{sec:methods-seg-inputs}
Dataset-specific annotations are aligned to the shared seven-label space introduced in
\hyperref[sec:results-seg-design]{Results}, using the conversion rules in
\cref{supp:seg-labels-inputs,tab:seg-label-mapping}. Pixel masks define tissue location,
while auxiliary image-level targets extend supervision to images without tissue-region masks.
Auxiliary targets derive from nuclear annotations, marker masks and image classes, including
dataset-specific proxies for tissue identity.

Unannotated pixels are distinguished from known background. Zero-valued pixels in partially
annotated BCSS images are excluded from the supervised Dice and binary cross-entropy (BCE)
losses and from the known-background penalty. Images are magnification-aligned to a
40-times-equivalent scale, sampled into a $4,096\times4,096$-pixel field and resized to
$1,024\times1,024$ pixels for SAM2 or $1,008\times1,008$ pixels for MedSAM3.

\subsubsection{Backbone architectures}\label{sec:methods-seg-architecture}
We implemented the shared tissue representation with SAM2 and MedSAM3 using visual low-rank
adaptation~\cite{ravi2024sam2,liu2025medsam3,hu2021lora}.

PathXAI-Seg (SAM2) retains the pretrained Hiera-Large visual encoder~\cite{ryali2023hiera} and original neck, inserts vision
low-rank adapters, and adds a custom top-down feature pyramid network (FPN)~\cite{lin2017fpn} that maps multiscale features to tissue logits
(\cref{pan:seg-model-sam2-architecture}). The original encoder and neck weights remain fixed. The semantic FPN
and vision adapters are trainable throughout training. The visual adapters have rank 8 and
scaling parameter 16. This implementation does not use a
language encoder or text-initialized class embeddings. Each neck feature map is projected to
256 channels by a $1\times1$ convolution. Starting at the coarsest level, features are bilinearly
upsampled, added to the next lateral projection and processed by a $3\times3$ convolution,
16-group normalization and GELU. A final $1\times1$ convolution produces seven tissue logits.
Auxiliary tasks use the original neck feature maps.

PathXAI-Seg (MedSAM3)~\textcolor{black}{\cite{liu2025medsam3}} uses the SAM3 image architecture~\cite{carion2025sam3} and prompt-conditioned decoding
(\cref{pan:seg-model-architecture}). Vision adapters have rank 8, scaling parameter 16 and dropout 0.05.
Original visual weights, the language encoder, multimodal transformer and mask decoder remain fixed.
Tissue text prompts initialize class embeddings, which are updated during supervised adaptation and fixed
in the two continuation stages. Auxiliary tasks use the
last visual-trunk feature map before its neck and decoder. MedSAM3's spatial projection maps
1,024 channels through 256 channels to 128-dimensional features. Its tissue-classifier weights
supply detached global prototypes, which are distinct from the text-initialized class embeddings.

\subsubsection{Training loss definitions}\label{sec:methods-seg-losses}
Training proceeds through supervised adaptation, joint supervision and focal refinement. Supervised adaptation
learns tissue regions from pixel masks using a Dice-based objective~\cite{milletari2016vnet}
and binary cross-entropy. Joint
supervision interleaves pixel-labelled and image-labelled samples through five functional groups.
Mask matching fits annotated tissue probabilities; Mask geometry constrains area, boundaries and
smoothness; Label-guided localization uses image-level presence information on spatial probability
maps; Tissue recognition predicts tissue presence from pooled features; and Feature separation
pulls same-tissue features together while separating different tissues. Each sample contributes only
the objectives supported by its annotation type. Focal refinement retains these groups and
reweights the pixel cross-entropy term toward difficult predictions~\cite{lin2017focal}.

The five objective groups in \cref{pan:methods-training-objectives} combine the terms
defined below. For a tissue class $c$, $p_{ic}$ denotes its predicted probability at
pixel $i$, $g_{ic}\in\{0,1\}$ its reference mask and $m_i$ the indicator that the pixel
has an assessable annotation. The set $\Omega=\{i:m_i=1\}$ contains supervised pixels,
and $B$ contains known-background pixels. Unannotated BCSS pixels are excluded from
$\Omega$ and $B$. Pixel-mask losses are averaged over classes present in the reference.

\textbf{L1, Mask matching.}
We use a masked Dice objective~\cite{milletari2016vnet} with a linear denominator and
unit smoothing, together with binary cross-entropy (BCE),
\begin{equation}
\begin{aligned}
\mathcal L_{\mathrm{Dice},c}
&=1-\frac{2\sum_i p_{ic}g_{ic}+1}{\sum_i m_i p_{ic}+\sum_i g_{ic}+1},\\
\mathcal L_{\mathrm{BCE},c}
&=-\frac{1}{|\Omega|}\sum_{i\in\Omega}
\left[g_{ic}\log p_{ic}+(1-g_{ic})\log(1-p_{ic})\right].
\end{aligned}
\end{equation}
Probabilities entering logarithms are clamped to $[10^{-6},1-10^{-6}]$.
Focal refinement replaces BCE with focal BCE~\cite{lin2017focal},
\begin{equation}
\mathcal L_{\mathrm{focal},c}
=-\frac{1}{|\Omega|}\sum_{i\in\Omega}(1-p_{t,ic})^\gamma\log p_{t,ic},
\qquad p_{t,ic}=g_{ic}p_{ic}+(1-g_{ic})(1-p_{ic}),\quad\gamma=1.
\end{equation}
Where enabled and $B$ is nonempty, the known-background penalty is
$\mathcal L_{\mathrm{bg},c}=|B|^{-1}\sum_{i\in B}p_{ic}^2$.
MedSAM3 also uses a probability-margin term
$\mathcal L_{\mathrm{margin},c}=\max\{0,0.1-(\bar p_{+,c}-\bar p_{-,c})\}$,
where $\bar p_{+,c}$ and $\bar p_{-,c}$ are mean probabilities over target pixels and
labelled non-target pixels, respectively. This term is omitted if either set is empty.

\textbf{L2, Mask geometry.}
Let $a_c$ and $a_c^*$ be the mean predicted probability and mean binary target over
the complete image grid, each clamped below at $10^{-6}$. Scale regularization is
\begin{equation}
\mathcal L_{\mathrm{scale},c}=h(\log a_c-\log a_c^*),\qquad
h(u)=\begin{cases}\tfrac12 u^2,&|u|<1,\\|u|-\tfrac12,&|u|\geq1,\end{cases}
\end{equation}
using smooth L1 loss~\cite{girshick2015fast}. Define $b(x)$ as the difference between
$5\times5$ max-pool dilation and erosion, with stride 1 and padding 2, clipped to
$[0,1]$. Shape regularization combines boundary overlap with anisotropic
total variation~\cite{beck2009tv},
\begin{equation}
\begin{aligned}
\mathcal L_{\mathrm{shape},c}
&=1-\frac{2\sum_i b(p_c)_i b(g_c)_i+1}
{\sum_i b(p_c)_i+\sum_i b(g_c)_i+1}+0.05\,\mathrm{TV}(p_c),\\
\mathrm{TV}(p_c)
&=\operatorname{mean}_{i\sim_h j}|p_{ic}-p_{jc}|
+\operatorname{mean}_{i\sim_v j}|p_{ic}-p_{jc}|.
\end{aligned}
\end{equation}
The two means cover horizontally and vertically adjacent pixels. Both geometry
terms are zero when the target contains fewer than four pixels. These terms are computed over
the full image grid.

\textbf{L3, Label-guided localization.}
\textcolor{black}{For an auxiliary probability map with $N$ pixels, including padded locations,} let $T_c$ contain the
$\max(1,\lfloor0.02N\rfloor)$ largest probabilities for non-background tissue class
$c$, and let $y_c$ indicate whether its image-level target is positive. Within a
multiple-instance supervision framework~\cite{pinheiro2015mil}, we use
\begin{equation}
\begin{alignedat}{2}
q_c&=\frac{1}{|T_c|}\sum_{i\in T_c}p_{ic},
&\qquad\mathcal L_{\mathrm{MIL}}
&=-\frac{1}{C}\sum_{c=1}^{C}
\left[y_c\log\widetilde q_c+(1-y_c)\log(1-\widetilde q_c)\right],\\
\mathcal L_{\mathrm{absent}}
&=\frac{\sum_c(1-y_c)q_c}{\sum_c(1-y_c)},
&\qquad\mathcal L_{\mathrm{entropy}}
&=-\frac{1}{NC}\sum_{i,c}
\left[p_{ic}\log p_{ic}+(1-p_{ic})\log(1-p_{ic})\right].
\end{alignedat}
\end{equation}
Here $C$ is the number of non-background tissue classes and $\widetilde q_c$ is
$q_c$ clamped to $[10^{-4},1-10^{-4}]$ for BCE. The absent-class term is zero if no
class is labelled absent. Entropy minimization~\cite{grandvalet2004entropy} acts on
the probability maps. These terms use binary presence derived from the supplied
tissue targets.

\textbf{L4, Tissue recognition.}
Pooled image features feed a tissue classifier with sigmoid outputs $r_k$ and
supplied targets $t_k$. For the set $K$ of known targets, its loss is
\begin{equation}
\mathcal L_{\mathrm{tissue}}
=-\frac{1}{|K|}\sum_{k\in K}
\left[t_k\log r_k+(1-t_k)\log(1-r_k)\right].
\end{equation}
An empty known-target set contributes zero. MedSAM3 additionally predicts organ
identity with softmax probabilities $o_j$ and cross-entropy
$\mathcal L_{\mathrm{organ}}=-\log o_{j^*}$ for the target organ $j^*$.
\textcolor{black}{Known-label masks, pseudo-target rules and backbone-specific gradient routes are specified in \cref{supp:seg-training-comparisons}.}

\textbf{L5, Feature separation.}
Spatial contrast samples up to 128 target-class feature locations and 128 other
labelled locations per tissue class, requiring at least four of each before
sampling. Let $T$ and $O$ be these sets, $s_{ij}=\cos(f_i,f_j)$ their cosine
similarities and $w_i$ a location's confidence weight. The spatial loss is
\begin{equation}
\mathcal L_{\mathrm{spatial},c}
=\frac{\sum_{i,j\in T,\,i\ne j}w_iw_j(1-s_{ij})}
{\max(\sum_{i,j\in T,\,i\ne j}w_iw_j,10^{-6})}
+\frac{\sum_{i\in T,j\in O}w_iw_j\max(0,s_{ij}/0.07)}
{\max(\sum_{i\in T,j\in O}w_iw_j,10^{-6})}.
\end{equation}
Spatial contrast losses are summed over tissue classes meeting the sampling requirements above,
whereas mask matching and geometry losses are averaged over reference-present classes.
Reference-labelled locations have unit weight. Auxiliary images use
image-label-constrained pseudo-labels. Tissue probabilities are clamped to
$[10^{-6},1-10^{-6}]$, converted to logits and normalized across classes by
softmax. Each pixel receives the highest-probability class; when positive image
labels exist, assignments to absent classes are excluded. Confidence is
$1-H/\log C$, where $H$ is the softmax entropy. Values at or below 0.2 receive
zero weight. Label maps use nearest-neighbour resizing and confidence maps use
bilinear resizing. Confidence weights retain their probability-map gradients.

For MedSAM3 auxiliary images, a global feature-prototype term uses the
mean-over-positives log-softmax form of supervised contrastive
learning~\cite{khosla2020supcon},
\begin{equation}
\mathcal L_{\mathrm{prototype}}
=\log\sum_k\exp(z_k)-\frac{1}{|P|}\sum_{k\in P}z_k,
\qquad z_k=\cos(f,v_k)/0.2.
\end{equation}
Here $f$ is the pooled image feature, $v_k$ is the detached tissue-classifier
prototype and $P$ contains classes with positive image-level targets. The term
is omitted when $P$ is empty; its gradient updates the image feature, not the
prototype weights. Each training sample contributes the weighted sum of its
applicable terms, \textcolor{black}{with gradient routes and coefficients specified in \cref{supp:seg-training-comparisons}.}

\subsubsection{Optimization, sampling and checkpoint selection}\label{sec:methods-seg-optimization}
\textcolor{black}{Supervised adaptation, joint supervision and focal refinement use backbone-specific
optimization schedules, with the selected checkpoint from each phase initializing the next.
Both backbones use AdamW~\cite{loshchilov2019adamw}.}

\textbf{\textcolor{black}{MedSAM3.}}
\textcolor{black}{MedSAM3 starts from its pretrained initialization and uses 5,000, 2,500 and 2,500
configured iterations for supervised adaptation, joint supervision and focal refinement,
respectively. Joint supervision and focal refinement balance pixel-labelled sampling across
tissue classes and datasets and interleave auxiliary images every sixth loader position.}

\textbf{\textcolor{black}{SAM2.}}
\textcolor{black}{SAM2 starts from a supervised fine-tuning run configured for 10,000 microsteps.
The three subsequent phases use 5,000, 2,500 and 2,500 microsteps, respectively, with
auxiliary images scheduled every sixth sampling step.}

\textcolor{black}{For both backbones, we retained the checkpoint with the best training-set
performance. Training comparisons examine selected checkpoints, objective configurations and
refinement branches with differing optimization budgets. Learning rates, gradient accumulation,
sampling weights, crop and augmentation settings, and run-specific comparisons are given in
\cref{supp:seg-training-comparisons,tab:seg-optimization-settings}.}

\subsubsection{Tissue-evidence outputs}\label{sec:methods-seg-inference}
For SAM2, each foreground tissue logit $z_c$ gives $p_c=\operatorname{sigmoid}(z_c)$,
for \textcolor{black}{$c\in\{1,\ldots,6\}$}. The background score is \textcolor{black}{$p_0=1-\max_{c\in\{1,\ldots,6\}}p_c$};
the background-channel logit is not used to construct this score. MedSAM3 combines
class-conditioned query scores and mask probabilities. Larger fields use windowed inference
with overlap-aware aggregation. After spatial resizing and aggregation,
classwise argmax produces the predicted tissue label map. The same coordinate system is retained for
aligned segmentation overlays and tissue statistics (\cref{pan:seg-model-inference-evidence}). Tissue-class
fractions use all pixels in the ROI mask as their denominator, while foreground fraction is reported
separately. \textcolor{black}{These fractions describe predicted area under the dataset-specific label mappings.
The output labels are mutually exclusive at each pixel, although gland structures can contain
neoplastic or non-neoplastic epithelium (\cref{tab:seg-label-mapping}).} \textcolor{black}{Auxiliary recognition and feature-separation heads are used during training. PathXAI-Report receives the tissue label maps, overlays and tissue fractions.}

\subsection{Evaluation and statistical analysis of segmentation}\label{sec:methods-seg-validation}
\subsubsection{Comparator implementations}
\textcolor{black}{The nnU-Net model uses the v2 2D RGB configuration. HoVer-Net uses the official
PanNuke-pretrained fast encoder.}

CONCHSeg is a study-specific CONCH-based tile classifier. The evaluated configuration freezes
CONCH and learns a neighbour aggregator, class-specific affine transformations of text
embeddings and a logit scale. It broadcasts each tile's class logits across its pixels,
without a pixel decoder or cell adapter. This comparator therefore evaluates tile-level
classification transferred to the pixel grid. \textcolor{black}{Training budgets and checkpoint
selection for these comparators are given in \cref{supp:seg-validation-protocol}.}

\subsubsection{Segmentation measures and aggregation}
Dice was the primary segmentation measure; intersection over union (IoU) was retained as a
secondary measure~\cite{maierhein2024metrics,taha2015metrics}.
Only tissue classes present in the reference contributed; known background was included
according to each task's annotation protocol.

For a reference-present class, true-positive (TP) pixels belong to that class in
both masks, false-positive (FP) pixels occur only in the prediction, and
false-negative (FN) pixels occur only in the reference. We calculated
\begin{equation}
\mathrm{Dice}=\frac{2\mathrm{TP}+\epsilon}
{2\mathrm{TP}+\mathrm{FP}+\mathrm{FN}+\epsilon},\qquad
\mathrm{IoU}=\frac{\mathrm{TP}+\epsilon}
{\mathrm{TP}+\mathrm{FP}+\mathrm{FN}+\epsilon},\quad\epsilon=10^{-8}.
\end{equation}
Absent reference classes do not contribute correct empty-mask scores. For BCSS,
class 0 is excluded from the scored class set, but zero-labelled pixels remain
non-target pixels when scoring foreground classes and can contribute false
positives. Other tasks retain background where specified by their annotation protocols.

Each task mean averaged image-class scores, or specimen-class scores for AGGC. The internal,
external and overall summaries in \cref{pan:seg-summary-dataset-dice} gave equal weight to
five, six and eleven tasks, respectively. RINGS tumour and gland segmentation each contributed
one task mean.
Organ scores first averaged task means within each organ (\cref{pan:seg-summary-organ-dice});
equal-organ summaries then gave each organ equal weight. Paired gains in
\cref{pan:seg-summary-fine-tuning-gains} show both task- and organ-weighted summaries.
The objective comparisons in \cref{pan:seg-summary-component-deletions} used equal-task means.
Dice differences were expressed in pp, so a difference of 0.01 corresponds
to 1~pp. The shared external comparison used the same overlap definition but averaged
its three datasets equally (\cref{pan:seg-summary-shared-external-comparison}).

Specifically, for task $t$, let $\mathcal P_t$ contain the assessable sample-class
pairs and $d_{tj}$ denote each pair's Dice score. Task and equal-task summaries are
\begin{equation}
D_t=\frac{1}{|\mathcal P_t|}\sum_{j\in\mathcal P_t}d_{tj},\qquad
D_{\mathrm{task}}=\frac{1}{T}\sum_{t=1}^{T}D_t.
\end{equation}
For $O$ organs, with task set $\mathcal T_o$ for organ $o$, the equal-organ summary is
$D_{\mathrm{organ}}=O^{-1}\sum_{o=1}^{O}|\mathcal T_o|^{-1}
\sum_{t\in\mathcal T_o}D_t$. Paired gains apply these same weights to
$100(d^{\mathrm{PathXAI}}-d^{\mathrm{comparator}})$, giving differences in pp.

\subsubsection{Statistical analysis}
All eight methods used
pointwise 95\% percentile source-cluster bootstrap intervals with 10,000 shared draws within
each task (seed 20260922). \textcolor{black}{Resampling units were available source clusters or
specimens; each sampled unit retained all its class scores.} The same draws
were used across methods for absolute estimates and paired differences, including equal-organ gains.
Case/source pairing and a sensitivity analysis for the shared RINGS images are described in
\cref{supp:seg-validation-protocol}. These intervals describe test-sample uncertainty,
not variation across independently trained models.

\textcolor{black}{Checkpoint selection using training-set performance is described in}
\hyperref[sec:methods-seg-optimization]{Optimization, sampling and checkpoint selection};
run-specific details are in \cref{supp:seg-training-comparisons}.

\subsubsection{Shared external comparison}
The shared external datasets were selected using the evaluated methods' published training
datasets. All methods received
the class IDs for tissues present in the annotations, without their locations. VISTA used full-field boxes and
aligned mask assembly. The released systems were evaluated without matching their training
budgets (\cref{supp:seg-shared-comparison}). Qualitative-example selection procedures are
specified in \cref{supp:seg-example-selection}.

\section*{AI assistance}
OpenAI Codex assisted with manuscript \textcolor{black}{polishing}, refinement of selected
plotting scripts and review of the research code.
\textcolor{black}{Reference-text consistency checks used GPT-6.1 through OpenAI Codex.}

\section*{Data availability}\phantomsection\label{sec:data-availability}
\textbf{Report datasets.}
Public TCGA slides are available from the National Cancer Institute Genomic Data Commons
(\url{https://portal.gdc.cancer.gov/}). \textcolor{black}{Public reference reports used
WsiCaption's OCR-extracted TCGA report texts (\url{https://github.com/cpystan/Wsi-Caption})~\cite{chen2024wsicaption},
restructured as described in \cref{supp:report-cohorts}.}
The manuscript includes selected de-identified
report excerpts; private clinical images, complete reports and case-level data will not
be publicly released.

\textbf{Segmentation datasets.}
The sources used for pixel-labelled training, auxiliary training and external
evaluation are listed below. Original access and usage terms apply.
\textcolor{black}{We thank the PathSegmentor authors for compiling and organizing public
pathology datasets in PathSeg~\cite{chen2026pathsegmentor}.
Access links for shared datasets follow their paper and acquisition guide
(\url{https://github.com/zhi-xuan-chen/PathSegmentor}); links for additional datasets identify
their original providers or the distribution used.}

\phantomsection\label{sec:seg-data-availability}
\textit{Pixel-labelled sources.}
BCSS (\textcolor{black}{\url{https://github.com/PathologyDataScience/BCSS}})~\cite{amgad2019bcss},
DigestPath2019 (\url{https://digestpath2019.grand-challenge.org/})~\cite{da2022digestpath},
GlaS (\textcolor{black}{Kaggle distribution: \url{https://www.kaggle.com/datasets/sani84/glasmiccai2015-gland-segmentation}})~\cite{sirinukunwattana2017glas},
BACH whole slides (ICIAR2018) (\url{https://zenodo.org/records/3632035})~\cite{aresta2019bach,polonia2019bach}, and
WSSS4LUAD (\url{https://wsss4luad.grand-challenge.org/})~\cite{han2022wsss4luad}.

\textit{Auxiliary sources.}
CPM15 (\textcolor{black}{\url{https://drive.google.com/drive/folders/11ko-GcDsPpA9GBHuCtl_jNzWQl6qY_-I}})~\cite{miccai2015nucleiChallenge},
CPM17 (\textcolor{black}{\url{https://drive.google.com/drive/folders/1sJ4nmkif6j4s2FOGj8j6i_Ye7z9w0TfA}})~\cite{vu2019cpm},
NCT-CRC-HE-100K (CRC100K) (\url{https://zenodo.org/records/1214456})~\cite{Kather2019PredictingSF,kather2018nctcrc100k},
CoNIC2022 (\url{https://conic-challenge.grand-challenge.org/})~\cite{graham2024conic},
BACH microscopy images (ICIAR2018) (\url{https://zenodo.org/records/3632035})~\cite{aresta2019bach,polonia2019bach},
Janowczyk (\url{https://andrewjanowczyk.com/use-case-1-nuclei-segmentation/})~\cite{janowczyk2016deep},
Kumar (\textcolor{black}{\url{https://drive.google.com/drive/folders/1bI3RyshWej9c4YoRW-_q7lh7FOFDFUrJ}})~\cite{kumar2017dataset},
Lizard (\textcolor{black}{Kaggle distribution: \url{https://www.kaggle.com/datasets/aadimator/lizard-dataset}})~\cite{graham2021lizard},
LC25000 (LungColon) (\url{https://github.com/tampapath/lung_colon_image_set})~\cite{borkowski2019lc25000},
MHIST (\url{https://bmirds.github.io/MHIST/})~\cite{wei2021petri},
MoNuSAC (\url{https://monusac-2020.grand-challenge.org/Data/})~\cite{verma2021monusac},
NuCLS (\textcolor{black}{\url{https://sites.google.com/view/nucls/home}})~\cite{amgad2022nucls},
Osteosarcoma (\url{https://www.cancerimagingarchive.net/collection/osteosarcoma-tumor-assessment/})~\cite{arunachalam2019osteosarcoma,leavey2019osteosarcoma},
PanNuke (\url{https://warwick.ac.uk/fac/cross_fac/tia/data/pannuke/})~\cite{gamper2019pannuke,gamper2020pannukeExtension},
PatchCamelyon17 (\textcolor{black}{PathMMU \texttt{images.zip}: \url{https://huggingface.co/datasets/jamessyx/PathMMU}})~\cite{sun2024pathmmu},
SegPath (\url{https://dakomura.github.io/SegPath/})~\cite{komura2023segpath},
SkinCancer (\url{https://doi.org/10.11588/data/7QCR8S})~\cite{kriegsmann2022skinPaper,kriegsmann2023skinDataset} and
TNBC (\textcolor{black}{\url{https://peterjacknaylor.github.io/data/}})~\cite{naylor2019distance,naylor2018tnbc}.

\textit{External sources.}
ACDC-LungHP (\url{https://acdc-lunghp.grand-challenge.org/DATA/})~\cite{li2021acdclunghp},
AGGC2022 (\url{https://aggc22.grand-challenge.org/AGGC22/})~\cite{huo2024aggc,yu2022aggc},
METU-CCTGS (\url{https://eakbas.github.io/metu-cctgs/})~\cite{arslan2025metu},
PAIP2019 (\url{https://paip2019.grand-challenge.org/})~\cite{kim2021paip}, and
RINGS (\url{https://data.mendeley.com/datasets/h8bdwrtnr5/1}) (tumour and gland tasks)~\cite{salvi2021ringsPaper,salvi2021ringsDataset}.
Links identify data providers or public distributions; their access and usage terms apply.

\textbf{Domain-knowledge data.}
KEEP is available from its provider (\url{https://huggingface.co/datasets/Loie/KEEP_dataset})
after accepting its access conditions.

\textcolor{black}{Aggregate numerical results underlying the figures and tables are available at
\url{https://github.com/zhaorui-tan/PathXAI/tree/main/source_data}.
Public case and segmentation sample lists are in the same repository's \texttt{data\_lists} directory.}

\section*{Code availability}
\textcolor{black}{Inference code, including a Jupyter notebook, agent prompts and an evidence-linked
HTML report viewer, and two pathologist tools (reference-report authoring, and report ranking
and correction assessment) are available at \url{https://github.com/zhaorui-tan/PathXAI}.}
\textcolor{black}{The repository also provides the report-scoring code and final public case lists.
Adapter weights for PathXAI-Seg (MedSAM3) are available at
\url{https://huggingface.co/RayTan/PathXAI-Seg}, subject to access approval.
SAM3 base weights and CONCH weights are obtained separately from their providers.}
\textcolor{black}{Original PathXAI code is licensed under Creative Commons
Attribution-NonCommercial 4.0 International (CC BY-NC 4.0).
Model weights and third-party components retain their respective licenses.}

\section*{Acknowledgements}
For report evaluation, we acknowledge the TCGA Research Network for the public
slides and pathology reports and the contributing hospitals for the private cohort.

For tissue segmentation, we thank the investigators and dataset providers who made the
pixel-labelled, auxiliary-training and external-evaluation datasets available
(\hyperref[sec:seg-data-availability]{Data availability}).

\section*{Funding}
This work was supported by the Industrial Alignment Funding Pre-Positioning Programme
(IAF-PP Grant H25J6a0011), awarded to Weimiao Yu and his iDMP lab.
The funder had no role in study design, data collection and analysis, manuscript preparation
or the decision to submit for publication.
\section*{Author contributions}
Zhaorui Tan conceived and developed the method, designed the experiments, collected the data,
implemented the software, conducted the experiments, analysed the results and wrote the
manuscript. Jun Aun Tan assisted with VLM
deployment and inference integration and proofread the manuscript. Tan Pan contributed to
discussions of the study concept. Xingdong Xue developed
the data backend and report-evaluation platform. Weixiang Zhong and Xiaoxi Wang authored
the reference pathology reports for the private cohort and evaluated reports produced by
the compared systems. Weimiao Yu, Li Liu, Jing Zhang (Changzheng Hospital) and Jing Zhang
(The First Affiliated Hospital, Zhejiang University School of Medicine) supervised the study.

\section*{Competing interests}
The authors declare no competing interests.

\FloatBarrier
\begingroup
\setlength{\bibsep}{4pt}
\bibliography{references}
\endgroup

\clearpage
\crefalias{figure}{suppfigure}
\crefalias{subfigure}{suppfigurepanel}
\renewcommand{\figurename}{Supplementary Figure}
\renewcommand{\thefigure}{\arabic{figure}}
\renewcommand{\theHfigure}{supplementary.\arabic{figure}}
\setcounter{figure}{0}
\crefalias{table}{supptable}
\renewcommand{\tablename}{Supplementary Table}
\renewcommand{\thetable}{\arabic{table}}
\renewcommand{\theHtable}{supplementary.\arabic{table}}
\setcounter{equation}{0}
\renewcommand{\theequation}{S\arabic{equation}}
\renewcommand{\theHequation}{supplementary.\arabic{equation}}
\section*{Supplementary Information}
\setcounter{table}{0}
This supplement describes tissue segmentation, multi-slide reporting and their evaluation.

\subsection*{Contents}
\begingroup
\setlength{\tabcolsep}{0pt}
\renewcommand{\arraystretch}{1.12}
\newcommand{\suppcontentsentry}[2]{%
  \hyperref[#1]{\ref*{#1}. #2} & \pageref{#1}\\[2pt]}
\noindent\begin{tabular}{@{}p{0.93\linewidth}>{\raggedleft\arraybackslash}p{0.07\linewidth}@{}}
\textbf{Supplementary Methods} & \textbf{Page}\\[3pt]
\suppcontentsentry{supp:segmentation-methods}{Segmentation data, models, training and validation}
\suppcontentsentry{supp:report-generation}{Cohorts and multi-slide report generation}
\suppcontentsentry{supp:report-evaluation}{Report comparisons, scoring and statistical analysis}
\suppcontentsentry{supp:physician-tools}{Pathologist reference-report authoring and ranking and correction assessment}
\suppcontentsentry{supp:report-interface}{Evidence-linked report viewer}
\addlinespace[5pt]
\textbf{Supplementary Results} & \\[3pt]
\suppcontentsentry{supp:public-report-results}{Automated report evaluation on the public cohort}
\suppcontentsentry{supp:private-report-results}{Automated report evaluation on the private cohort}
\suppcontentsentry{supp:pathologist-results}{Pathologist ranking and correction assessment and report-content checks}
\suppcontentsentry{supp:report-policy-results}{Tissue evidence and its use in reporting}
\suppcontentsentry{supp:report-scope-results}{Sensitivity to diagnostic extraction scope}
\suppcontentsentry{supp:public-case-results}{Evidence-linked public cases and failure analysis}
\suppcontentsentry{supp:segmentation-results}{Segmentation transfer and training-objective comparisons}
\addlinespace[5pt]
\textbf{Reading guide} & \\[3pt]
\hyperref[supp:visual-conventions]{Colour and symbol conventions} & \pageref{supp:visual-conventions}\\[2pt]
\hyperref[supp:terminology]{Terminology and operational definitions} & \pageref{supp:terminology}\\
\end{tabular}
\endgroup
\par\medskip
\newcommand{\privateDataReleaseBoundary}{%
The private-cohort tables report aggregate results, not case-level clinical reports or ratings.
The private dataset is not publicly available; private-case images are not included as
illustrative examples. Selected de-identified textual comparisons do not release the
underlying reports or case-level evaluation files.}
\section*{Supplementary Methods}
\supplementaryinfosection{supp:figure-details}
The five sections cover segmentation, report generation, automated evaluation,
pathologist ranking and correction assessment and the evidence-linked viewer. Data preparation and
implementation details precede the associated validation procedures.

\renewcommand{\floatpagefraction}{0.96}
\supplementarymethodsection{Segmentation data, models, training and validation}{supp:segmentation-methods}
Performance and training comparisons appear in \cref{supp:segmentation-results}.

\supplementarymethodsubsection{Datasets, label harmonization and inputs}{supp:seg-labels-inputs}
The per-dataset partition rule is specified in
\textcolor{black}{\hyperref[sec:methods-seg-partition]{Methods, Segmentation data partitioning}}.
\textcolor{black}{External datasets provided independent evaluation for both backbones and were not used for training or checkpoint selection.}
The training and test data are summarized in
\cref{pan:seg-summary-data-examples}, with counts for each dataset in \cref{fig:seg-data}.
Counts use the dataset-specific units of images, tiles, slides or specimens. The \textcolor{black}{six external test sets} include RINGS tumour and RINGS gland,
which use the same 500 images. Each RINGS \textcolor{black}{test set} contributes 500
to the external total of 11,089 samples, and AGGC counts specimens.
Some eligible training samples may not be drawn within the fixed number of training steps.
\textcolor{black}{The accompanying Source Data list sample IDs, their datasets and training/test assignments.}
\textcolor{black}{Among samples with recoverable TCGA patient identifiers, training pools and
the internal test set shared four patients, although their sample identifiers did not overlap.}

In the organ-composition matrix (\cref{pan:seg-data-organ-composition}), circle area represents each organ's share of samples; mixed-organ
and unclassified samples remain separate. The tissue-label matrix (\cref{pan:seg-data-label-prevalence}) covers dense training masks
and internal and external test annotations. For each tissue label, prevalence is the mean proportion
of samples containing that label across these datasets with suitable annotations. Measured zeros are
included; datasets without suitable annotations are excluded. Labels can co-occur, so these
summaries describe neither tissue area nor disease prevalence. 

Dataset-specific annotations are mapped to the shared seven-label vocabulary
(\cref{pan:seg-model-tissue-vocabulary,pan:seg-model-input-harmonization}). 
RINGS images are paired by filename with the supplied manual tumour and gland masks;
nonzero mask pixels map to neoplastic and gland foreground, respectively.
Input magnification, crop dimensions, backbone-specific resizing and the handling of unannotated
pixels are specified in \hyperref[sec:methods-seg-inputs]{Methods, Tissue labels and image preparation}.

Public dataset links are listed in \hyperref[sec:seg-data-availability]{Data availability}.
\noindent\textbf{Annotation conversion and eligibility.}
\Cref{tab:seg-label-mapping} specifies the dataset-specific conversions and the classes
supervised by the tissue-classification loss.
\textcolor{black}{The seven labels form an operational annotation vocabulary. Gland denotes a
structure, while neoplastic and epithelial describe tissue identity; these biological concepts
can overlap. Source annotations determine the supervised label, so a gland mask contributes
gland supervision without establishing whether its epithelium is neoplastic. The final argmax
map assigns one label per pixel, and its class fractions describe the predicted partition
under these conversion rules.}
Only the five pixel-labelled sources supply tissue-region masks. Nuclei annotations and
SegPath marker masks supply auxiliary image-level targets, not tissue-region segmentation targets.
Typed nuclei and marker annotations supply category-specific targets; untyped nuclear foreground
and some image classes supply proxy targets.
For mask-derived targets, MedSAM3 retains the fraction of crop pixels in each tissue class;
nuclei-derived targets and image-class targets are binary. SAM2 binarizes these tissue targets
at greater than zero. Classification BCE supervises only the classes listed in the final column;
all other classes, including background, are unknown for that loss. The separate auxiliary-map
objectives use the tissue vector without this mask (\cref{supp:seg-training-objectives}).

\begingroup
\manuscripttablefont
\setlength{\tabcolsep}{4pt}
\renewcommand{\arraystretch}{1.06}
\setlength{\LTcapwidth}{\linewidth}
\begin{longtable}{@{}>{\raggedright\arraybackslash}p{0.21\linewidth}>{\raggedright\arraybackslash}p{0.64\linewidth}>{\raggedright\arraybackslash}p{\dimexpr0.15\linewidth-4\tabcolsep\relax}@{}}
\caption{Dataset-specific training-annotation mappings.
Target IDs are 0, background; 1, neoplastic; 2, inflammatory; 3, connective;
4, epithelial; 5, dead/necrotic; and 6, gland. Arrows map source labels to target IDs.
An empty target means no positive tissue label, not exclusion of the image.
The last column lists tissue classes eligible for classification BCE.}
\label{tab:seg-label-mapping}\\
\toprule
Source & Conversion & Supervised classes \\
\midrule
\endfirsthead
\multicolumn{3}{@{}l}{\manuscripttablecontinued{}}\\
\toprule
Source & Conversion & Supervised classes \\
\midrule
\endhead
\midrule
\multicolumn{3}{r@{}}{Continued on next page}\\
\endfoot
\bottomrule
\endlastfoot
\multicolumn{3}{@{}l}{\textbf{Pixel-labelled sources}}\\
BCSS & Source IDs $1,19,20\to1$; $3,10,11\to2$; $2,14,16,18\to3$;
$8\to4$; $4\to5$; $13\to6$; $0,5,6,7,9,12,15,17,21\to0$.
Remapped zero pixels are excluded from supervised Dice/BCE and the known-background penalty.
& 1--6 \\
GlaS & Annotated gland objects $\to6$; remaining pixels $\to0$.
The source train, test-a and test-b images are pooled before the study partition.
& 6 \\
DigestPath2019 & Grayscale mask $>127\to1$, otherwise $\to0$.
Only paired positive images with nonempty foreground are retained; the negative-image directory is excluded.
& 1 \\
WSSS4LUAD & Source IDs $0\to1$, $1\to3$, $2\to4$; other values $\to0$.
Images and masks come from the source validation subset before the study partition.
& 1, 3, 4 \\
BACH whole slides & Preprocessed integer mask IDs are retained without further remapping.
\textcolor{black}{Consistency with the XML-to-mask conversion script has not been verified.}
& 1, 4 \\
\addlinespace[3pt]
\multicolumn{3}{@{}l}{\textbf{Auxiliary image-level sources}}\\
NuCLS & \textcolor{black}{Images and masks are from EvaluationSet.}
Red-channel IDs $1,2\to1$; $3\to3$; $4$--$8\to2$.
IDs 0, 9 and $\geq253$ provide no positive tissue target; only image-mask pairs are loaded.
& 1, 2, 3 \\
PanNuke & \textcolor{black}{Images and masks are from Part 1.}
Positive pixels in source channels $0,1,2,3,4$ imply tissue IDs $1,2,3,5,4$, respectively.
& 1--5 \\
CoNIC2022; Lizard & Nucleus IDs $1,3,4,5\to2$; $2\to4$; $6\to3$.
Any mapped nucleus establishes image-level presence.
& 2, 3, 4 \\
MoNuSAC & Epithelial $\to4$; lymphocyte, macrophage and neutrophil $\to2$.
Ambiguous/artifact and unrecognized annotations provide no positive target.
& 2, 4 \\
Kumar; CPM15; CPM17; TNBC & Binary nucleus foreground is assigned to ID 1 before deriving presence.
This loader proxy does not distinguish the cell types in untyped nucleus annotations.
TNBC uses a grayscale threshold of 127; the other sources use positive instance IDs.
& 1 \\
Janowczyk & Nuclear mask pixels $>127\to4$ for image-level presence.
& 4 \\
SegPath & panCK $\to4$; CD3/CD20, CD45RB, MIST1 and MNDA $\to2$;
aSMA and ERG $\to3$. CD235a and unknown markers yield an empty target.
Presence targets derive from positive marker-mask pixels.
& 2, 3, 4 \\
NCT-CRC-HE-100K & TUM $\to1$; LYM $\to2$; MUS and STR $\to3$;
NORM $\to4$; DEB $\to5$. ADI, BACK and MUC yield empty targets but remain eligible images.
& 1--5 \\
BACH microscopy & Benign, InSitu and Invasive $\to1$; Normal $\to4$.
& 1, 4 \\
MHIST & SSA $\to1$; HP $\to4$.
& 1, 4 \\
PatchCamelyon17; LungColon & The loader assigns ID 1 to every loaded image;
it does not read a per-image tumour/non-tumour class. \textcolor{black}{LungColon includes
the normal-tissue directories \texttt{colon\_n} and \texttt{lung\_n}.}
& 1 \\
SkinCancer & All loaded images receive ID 1, including images in the source's
12 non-tumour categories and four tumour categories. Source class labels are not used.
& 1 \\
Osteosarcoma & Viable $\to1$; Non-Viable-Tumor $\to5$; Non-Tumor yields an empty target.
& 1, 5 \\
\end{longtable}
\endgroup

Zero-valued mask pixels in GlaS, DigestPath2019, WSSS4LUAD and the preprocessed BACH
whole slides are treated as known negatives for their annotated tissue classes; they do not
establish absence of all other biological tissues. Unlisted source values default to zero
in lookup-table conversions. Nucleus or marker annotations assigned zero contribute no
positive tissue target. In particular, an empty auxiliary target is distinct from a missing
annotation. Eligible classes still receive zero targets in classification BCE.

\noindent\textbf{\textcolor{black}{Auxiliary proxy labels.}}\label{supp:skin-auxiliary-labels}
\textcolor{black}{SkinCancer and LungColon assign neoplastic presence to images from both tumour
and non-tumour source categories; BACH microscopy also maps its Benign category to neoplastic.
These proxy labels may introduce noise into tissue recognition and auxiliary-map supervision.
Their effect on final segmentation performance was not quantified separately.}
\textcolor{black}{For SkinCancer,} the tile-level training and auxiliary holdout partitions share source
cases and therefore do not provide case-independent validation. SkinCancer is not included
in the final pixel-level segmentation test sets.

\manuscriptnativefigure{\textbf{Segmentation datasets, organ composition and tissue labels.}}{fig:seg-data}{%
    \begin{minipage}{180mm}
    \manuscriptpanelblock{5mm}{180.0mm}{\manuscriptpanelmark{pan:seg-data-datasets}}{Segmentation training and evaluation data}{\includegraphics[width=\linewidth]{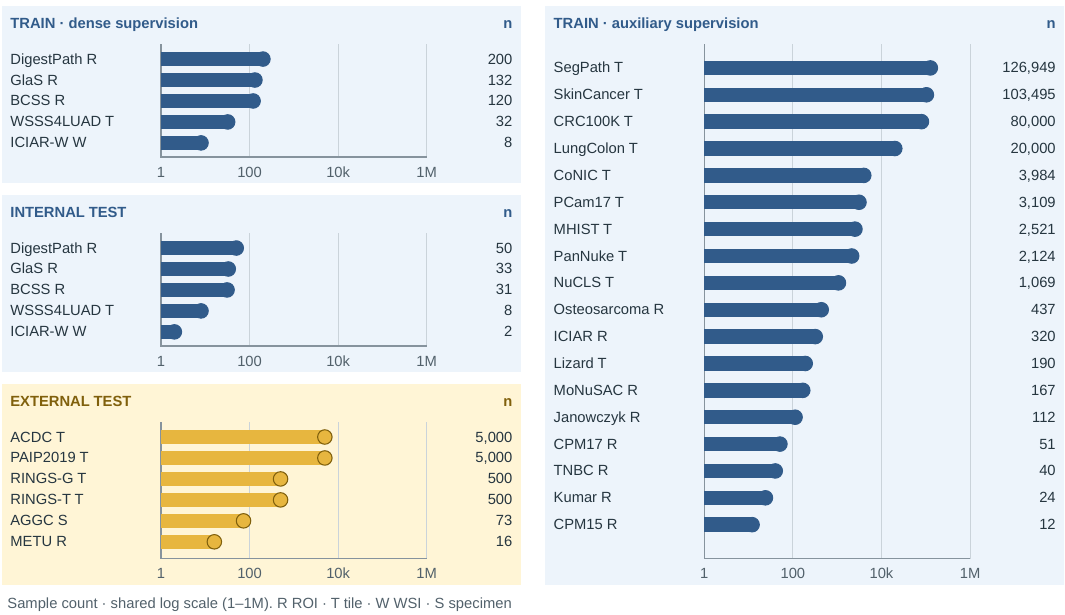}}
    \par\vspace{2mm}
    \manuscriptpanelblock{5mm}{104.0mm}{\manuscriptpanelmark{pan:seg-data-organ-composition}}{Organ sample composition}{\includegraphics[width=\linewidth]{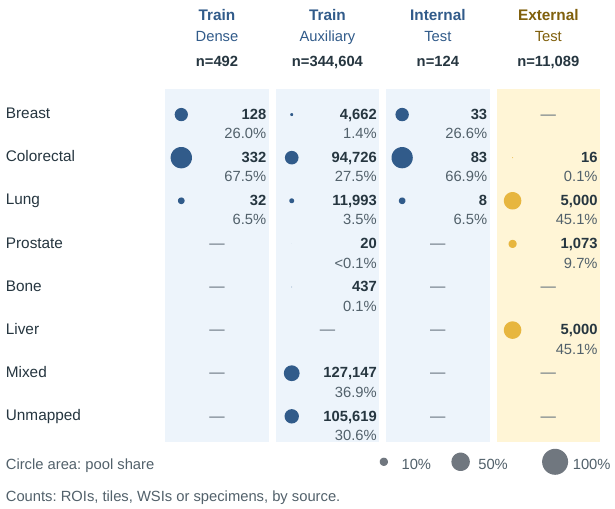}}\hfill\manuscriptpanelblock{5mm}{72.0mm}{\manuscriptpanelmark{pan:seg-data-label-prevalence}}{Tissue-label prevalence}{\includegraphics[width=\linewidth]{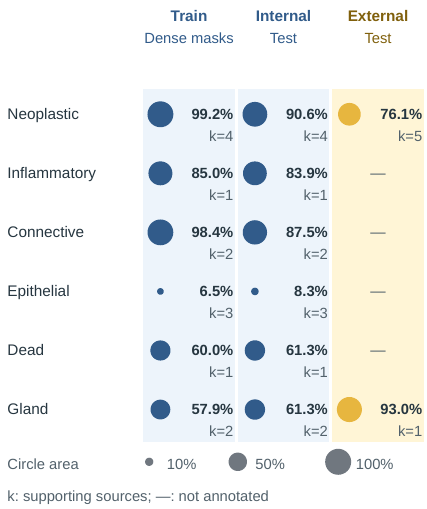}}
    \end{minipage}%
  }{%
\textbf{\panelletterref{pan:seg-data-datasets} Data sources.} Counts use the labelled dataset-specific units on a logarithmic axis. RINGS tumour and gland segmentation reuse the same images, counted for each task; AGGC counts specimens.
\textbf{\panelletterref{pan:seg-data-organ-composition} Organ composition.} Circle area represents each organ's share of samples; mixed and unspecified sources remain separate.
\textbf{\panelletterref{pan:seg-data-label-prevalence} Tissue-label prevalence.} Circle area represents the proportion of samples containing each label, averaged equally across datasets with suitable annotations. Labels can co-occur; prevalence describes label presence, not tissue area.
(Notation: ROI, region of interest; WSI, whole-slide image. A dash denotes unavailable annotations, not zero prevalence. Blue denotes training and internal tests; yellow denotes external tests.)
}

\supplementarymethodsubsection{Architecture, training diagrams and tissue-evidence outputs}{supp:seg-architecture}\label{supp:seg-inference}
\label{supp:seg-training-objectives}
Backbone construction, parameter freezing and auxiliary feature routes are specified in
\hyperref[sec:methods-seg-architecture]{Methods, Backbone architectures}
and \hyperref[sec:methods-seg-losses]{Training loss definitions}.

\noindent\textbf{Diagram conventions.}
In \cref{fig:seg-model,fig:seg-model-sam2}, flames mark optimized parameters, padlocks mark fixed
parameters and numerals below flames identify stage-specific optimization. Stage colours indicate
when an objective is introduced or modified; tissue colours identify biological classes. Forward
inputs are distinguished from supervision and initialization references. Spatial and global
feature separation are terms within one functional group; feature and probability glyphs are
schematic.

\noindent\textbf{Tissue-evidence outputs.}
The probability-to-label conversion and ROI-fraction calculation are specified in
\hyperref[sec:methods-seg-inference]{Methods, Tissue-evidence outputs}.
\Cref{pan:seg-model-inference-evidence} shows the aligned outputs; the detailed reporting
inputs and identifier handling are described in \cref{supp:input-identifiers}.

\manuscriptnativefigure{\textbf{PathXAI-Seg (MedSAM3) architecture and training.}}{fig:seg-model}{%
    \begin{minipage}{180mm}
    \manuscriptpanelblock{5mm}{88.0mm}{\manuscriptpanelmark{pan:seg-model-tissue-vocabulary}}{A shared tissue vocabulary}{\includegraphics[width=\linewidth]{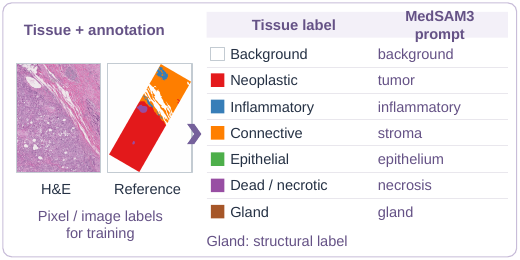}}\hfill\manuscriptpanelblock{5mm}{88.0mm}{\manuscriptpanelmark{pan:seg-model-input-harmonization}}{Harmonizing heterogeneous image inputs}{\includegraphics[width=\linewidth]{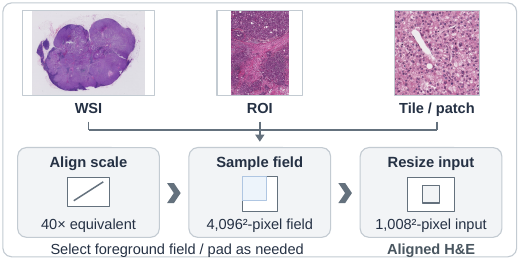}}
    \par\vspace{0.8mm}
    \manuscriptpanelblock{5mm}{180.0mm}{\manuscriptpanelmark{pan:seg-model-architecture}}{PathXAI-Seg (MedSAM3) architecture}{\includegraphics[width=\linewidth]{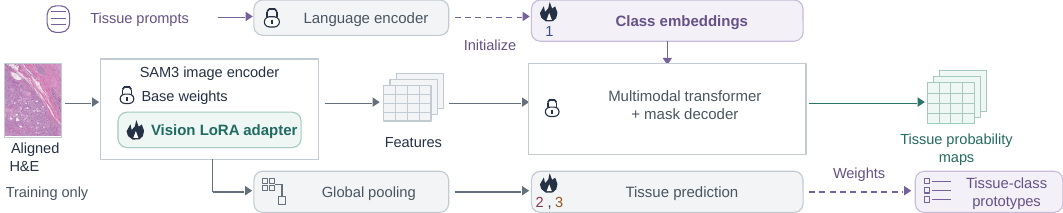}}
    \par\vspace{1mm}
    \includegraphics[width=180mm]{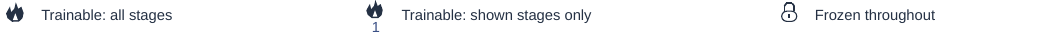}
    \par\vspace{1mm}
    \manuscriptpanelblock{5mm}{180.0mm}{\manuscriptpanelmark{pan:seg-model-training-objectives}}{Stage-wise training objectives}{\includegraphics[width=\linewidth]{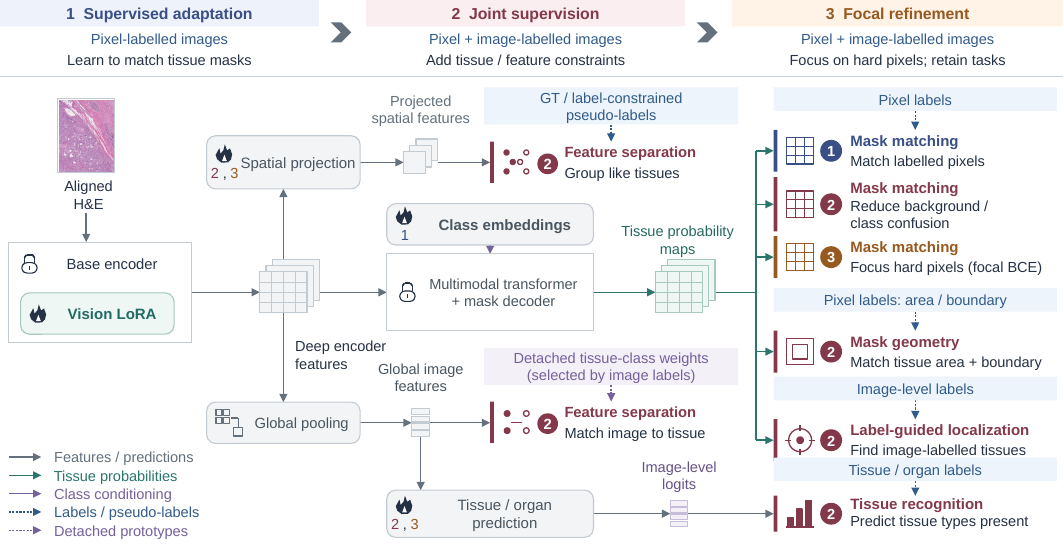}}
    \par\vspace{0.8mm}
    \manuscriptpanelblock{5mm}{180.0mm}{\manuscriptpanelmark{pan:seg-model-inference-evidence}}{Inspectable spatial evidence at inference}{\includegraphics[width=\linewidth]{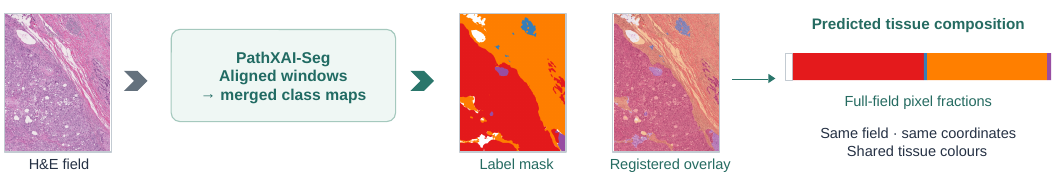}}
    \end{minipage}%
  }{%
\textbf{\panelletterref{pan:seg-model-tissue-vocabulary},\panelletterref{pan:seg-model-input-harmonization} Tissue labels and input harmonization.} Gland is a structural label; white BCSS mask pixels are unannotated.
\textbf{\panelletterref{pan:seg-model-architecture} Architecture.}
\textbf{\panelletterref{pan:seg-model-training-objectives} Training stages.}
\textbf{\panelletterref{pan:seg-model-inference-evidence} Tissue evidence outputs.} Tissue fractions use all ROI pixels; foreground is reported separately. Implementation is described in \crefrange{supp:seg-architecture}{supp:seg-inference}.
(Notation: ROI, region of interest. Feature stacks and maps are schematic. Solid arrows carry features, predictions or conditioning; dashed arrows carry labels (blue) or detached prototypes (purple), not gradients. The key defines parameter states and training stages.)
}

\manuscriptnativefigure{\textbf{PathXAI-Seg (SAM2) architecture and training.}}{fig:seg-model-sam2}{%
    \begin{minipage}{180mm}
    \manuscriptpanelblock{5mm}{180.0mm}{\manuscriptpanelmark{pan:seg-model-sam2-architecture}}{SAM2 semantic segmentation architecture}{\includegraphics[width=\linewidth]{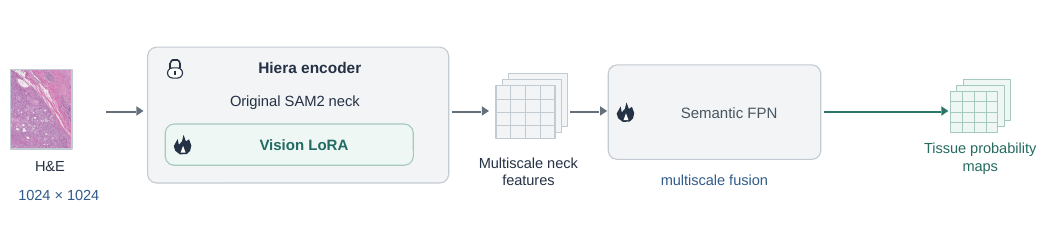}}
    \par\vspace{1mm}
    \includegraphics[width=180mm]{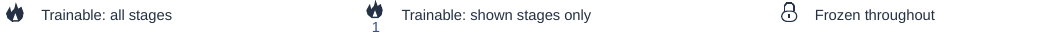}
    \par\vspace{2mm}
    \manuscriptpanelblock{5mm}{180.0mm}{\manuscriptpanelmark{pan:seg-model-sam2-training-objectives}}{Stage-wise training objectives}{\includegraphics[width=\linewidth]{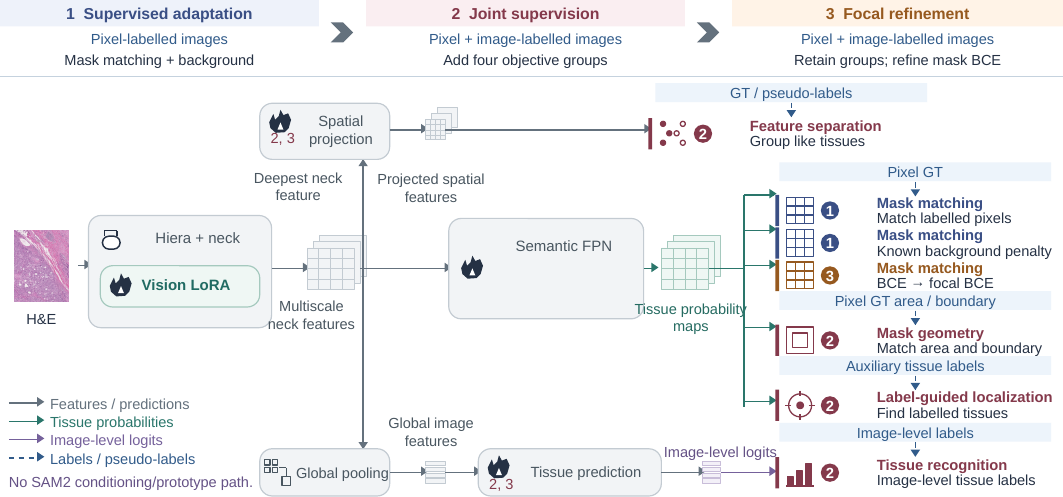}}
    \end{minipage}%
  }{%
\textbf{\panelletterref{pan:seg-model-sam2-architecture} Architecture.} Hiera and its neck retain frozen weights; visual LoRA and the semantic FPN are trained. Multiscale neck features produce tissue probability maps.
\textbf{\panelletterref{pan:seg-model-sam2-training-objectives} Training objectives.} Supervised adaptation, joint supervision and focal refinement use the objectives supported by each sample's labels. The deepest neck feature supplies spatial projection and pooled tissue prediction; image-level logits represent tissue presence. Objective-specific heads train during joint supervision and focal refinement. SAM2 uses task-adapted initialization and a distinct training budget; shared vocabulary and evidence outputs appear in \cref{fig:seg-model}.
(Notation: LoRA, low-rank adaptation; FPN, feature pyramid network; BCE, binary cross-entropy; H\&E, haematoxylin and eosin. Flames denote trainable parameters; numerals restrict training to the indicated stages. Locks denote frozen parameters. Colours mark objective introduction or modification. Features and tissue maps are schematic.)
}

\supplementarymethodsubsection{Training budgets, checkpoint selection and comparisons}{supp:seg-training-comparisons}
\textcolor{black}{The training phases, budgets and checkpoint-selection procedure are summarized in
\hyperref[sec:methods-seg-optimization]{Methods, Optimization, sampling and checkpoint selection}.
\Cref{tab:seg-optimization-settings} gives the optimizer settings. The sampling,
augmentation and objective implementation details below specify these training procedures.}

\begingroup
\begin{table}[!htbp]
\color{black}
\manuscripttablefont
\setlength{\tabcolsep}{4pt}
\renewcommand{\arraystretch}{1.12}
\caption{Segmentation optimization settings.
Both backbones use AdamW with $\beta_1=0.9$, $\beta_2=0.999$, $\epsilon=10^{-8}$,
weight decay 0.01 and random seed 42. Configured phase budgets are given in
\hyperref[sec:methods-seg-optimization]{Methods}.}
\label{tab:seg-optimization-settings}
\begin{tabular}{@{}>{\raggedright\arraybackslash}p{0.18\linewidth}>{\raggedright\arraybackslash}p{0.40\linewidth}>{\raggedright\arraybackslash}p{\dimexpr0.42\linewidth-4\tabcolsep\relax}@{}}
\toprule
Setting & MedSAM3 & SAM2 \\
\midrule
Initial learning rates
& Supervised adaptation: visual adapters and class embeddings, $10^{-4}$.
Joint supervision and focal refinement: visual adapters and configured auxiliary-parameter
group, $5\times10^{-6}$; class embeddings, 0.
& Initial fine-tuning: visual adapters and decoder, $10^{-4}$.
Subsequent supervised adaptation: $10^{-6}$; joint supervision and focal refinement:
$5\times10^{-6}$. Auxiliary-head rate, $5\times10^{-5}$. \\
Cosine schedule
& Adaptation: horizon 5,000 optimizer updates, clipped at 1\% of each initial rate.
Continuation: horizon 2,500 optimizer updates, minimum 0.
& For $N$ configured microsteps, horizon $\lceil N/8\rceil$ optimizer updates;
minimum rate is 1\% of the smaller initial adapter and decoder rates. \\
Gradient accumulation
& Loader positions count all samples, including skipped samples with no loss. At positions
8, 16, $\ldots$ that contribute a loss, accumulated gradients are updated. Losses are not
divided by eight. At epoch end, remaining gradients are updated if the loss-bearing sample
count is not divisible by eight.
& Eight microsteps per update; each loss is divided by eight. A final partial window
is updated at the last microstep. \\
Scheduler updates
& Once per optimizer-update boundary. The training iteration count advances per
loss-bearing sample, separately from the scheduler counter.
& Once per optimizer update, including the final partial window. \\
Mixed precision
& BF16 autocast with gradient scaling.
& CUDA autocast with gradient scaling. \\
Gradient clipping
& Segmentation and classification parameter groups clipped separately to norm 1.
& Decoder plus visual adapters, and auxiliary heads, clipped separately to norm 1. \\
Evaluation scheduling
& Every 1,000 adaptation iterations and 500 continuation iterations.
& Configured for 1,000 initial fine-tuning microsteps and 500 subsequent microsteps;
evaluations occur at coinciding optimizer updates and at the final microstep. \\
\bottomrule
\end{tabular}
\end{table}
\endgroup

\textcolor{black}{For MedSAM3, the iteration count advances for samples contributing a loss;
the scheduler advances at optimizer updates.}
{\color{black}
\noindent\textbf{MedSAM3 sampling.}
During joint supervision and focal refinement, pixel-labelled images are indexed by the tissue classes they
contain. Each image-class pair is sampled with replacement with weight
$1/(D_c n_{dc})$, where $D_c$ is the number of datasets containing class $c$ and $n_{dc}$
is the number of pairs for that class in dataset $d$. The 492 pixel-labelled images yield
842 pairs. A separate shuffled loader supplies auxiliary images, replacing the dense sample
at every sixth loader position, starting with the first position of each epoch. Both loaders use one image per batch and four workers.

\noindent\textbf{MedSAM3 augmentation.}
Images are rescaled to a 40-times-equivalent magnification using bilinear interpolation;
masks use nearest-neighbour interpolation. Images smaller than the $4,096\times4,096$ crop
are zero-padded on the right and bottom. Dense continuation crops are centred on a randomly
chosen target-class pixel, with random cropping when the target class is unavailable.
Auxiliary crops are centred on an annotated foreground pixel when available and otherwise
sampled randomly. Training applies horizontal flipping with probability 0.5, independent
brightness, contrast and saturation factors sampled uniformly from 0.8--1.2, and hue shifts
from $-0.05$ to 0.05. Evaluation uses centre crops without augmentation.
For auxiliary tissue recognition and global prototype contrast, an additional view uses a
random crop with side lengths scaled by a uniform factor of 0.7--1, horizontal flipping
with probability 0.5, brightness, contrast and saturation factors of 0.6--1.4, and hue
shifts from $-0.07$ to 0.07. Auxiliary probability-map objectives use the original loader crop.
Padding is retained in the images used for tissue recognition and in the probability maps
used for multiple-instance supervision, absent-class suppression and entropy.
These auxiliary objectives have no spatial content mask.
}

\noindent\textbf{\textcolor{black}{SAM2 sampling.}}
\textcolor{black}{The foreground-crop probability is 0.5.}
The SAM2 implementation draws training samples
with replacement. Its initial weight is 0.7 times the class/dataset-balancing
weight plus 0.3 times the sum, over declared tissue classes, of
$1/\max(\overline{\mathrm{Dice}}_{\mathrm{class,dataset}}+0.1,0.05)$;
missing class/dataset scores default to 0.5. Auxiliary images use a shuffled loader.

{\color{black}
\noindent\textbf{Gradient routes.}\label{sec:methods-seg-gradients}
In both continuation stages, visual adapters and auxiliary task parameters are trainable;
the SAM2 semantic FPN remains trainable and MedSAM3 class embeddings remain fixed.
Tissue recognition uses pooled features; spatial contrast uses pixel labels or image-label-constrained
pseudo-labels (\cref{pan:seg-model-training-objectives,pan:seg-model-sam2-training-objectives}). SAM2 tissue recognition updates its
task predictor and visual adapters. In MedSAM3, tissue recognition on pixel-labelled images uses
detached features; on auxiliary images, tissue-presence prediction can update visual adapters,
whereas organ prediction uses detached features. Spatial feature contrast updates the projection and
visual adapters in both architectures. Pseudo-label identities are non-differentiable, but their
confidence weights retain a probability-map gradient path.

\textbf{Objective coefficients.}
For MedSAM3, supervised adaptation uses Dice and BCE weights of 1 each and no auxiliary objectives.
Joint supervision and focal refinement retain these weights and enable a known-background
probability penalty with weight 1 on fully annotated sources. Area and shape regularizers each
have weight 0.001; the classification coefficient is 1; image-label multiple-instance, absent-class
and entropy terms have weights 0.1, 0.01 and 0.01. Spatial contrast has weight 0.005 for mask-labelled images and half this weight for
image-label-constrained pseudo-labels. The probability-margin term has weight 0.0001 and margin 0.1; the global feature-prototype
term on auxiliary images has weight 0.03. Cell-density conditioning, organ-prompt bias,
gland-lumen, density-compatibility, separate background-suppression and BYOL terms are inactive;
auxiliary tissue classification remains active. Fixed class embeddings receive no
prompt-orthogonality updates in the continuation phases.

The classification coefficient is multiplied by task-specific factors. MedSAM3 uses 0.5 for
tissue-presence BCE and 1 for known-organ cross-entropy. On pixel-labelled images with an
unknown or mixed organ, confident tissue pseudo-targets use a factor of 0.3; predicted-organ
pseudo-targets use 0.5. Organ supervision on auxiliary images begins after 500 iterations;
the pixel-labelled route has no corresponding warmup gate. Tissue pseudo-targets retain
probabilities above 0.7 among supervised classes; organ pseudo-targets use the most probable
class. Tissue BCE applies the supplied known-label mask. The auxiliary multiple-instance,
absent-class and global-prototype terms derive binary presence from positive tissue targets,
without applying that BCE mask.

SAM2 uses the same continuation coefficients for scale, shape, tissue recognition,
multiple-instance supervision, absent-class suppression, entropy and spatial contrast.
Its known-background penalty has weight 1 in all three phases, where annotations are complete;
it has neither the MedSAM3 global prototype term nor language-initialized class embeddings.
For partial annotations, the Dice/BCE exclusion of unannotated pixels does not
imply that every geometry or auxiliary objective uses the same masking rule.
}

\noindent\textbf{\textcolor{black}{Checkpoint selection.}}
\textcolor{black}{For both backbones, we retained the checkpoint with the best training-set performance; no separate validation set was used. The 124 internal test samples and external datasets were used only for evaluation.}
\textcolor{black}{The selected final MedSAM3 checkpoint is at iteration 2,500, with optimizer
counter 352 and learning rate $4.7594\times10^{-6}$. The initial SAM2 fine-tuning checkpoint
is at microstep 7,000 (875 optimizer updates), and its final checkpoint is at microstep 2,500
(313 optimizer updates).}

\noindent\textbf{Training stages and refinement branches.}
The MedSAM3 comparisons include successive training checkpoints and refinement branches
initialized from a common checkpoint (\cref{tab:source_supp2}). The refinement branches are
separate from the final sequential checkpoint. Each training run is represented by its selected
checkpoint. Successive stages include additional training and are not budget-matched
single-objective interventions.

\noindent\textbf{Objective-weight configurations.}
Joint-supervision comparisons change selected objective weights
(\cref{pan:seg-summary-component-deletions,tab:source_supp2}). The label-guided-localization
deletion sets the auxiliary-map multiple-instance, absent-class and entropy loss weights to
zero; it does not remove auxiliary images or all auxiliary recognition and contrastive tasks.
The scale/shape deletion zeros the two regularizer weights; the intra-image-contrast variant
zeros its spatial-contrast weight. In the available reference implementation, this weight also
gates the auxiliary-map multiple-instance, absent-class and entropy terms. Their activation
in the evaluated runs requires confirmation, so this comparison is interpreted at the
configuration level.

\supplementarymethodsubsection{Validation estimands, pairing and uncertainty}{supp:seg-validation-protocol}
Baseline implementations, Dice/IoU definitions and task- and organ-weighted calculations are specified in
\hyperref[sec:methods-seg-validation]{Methods, Evaluation and statistical analysis of segmentation}.
Test strata are breast, colorectal and lung internally, and colorectal, lung, prostate and liver
externally; auxiliary training includes additional organs. METU-CCTGS is assigned to
colorectal and \textcolor{black}{AGGC and RINGS to prostate}.
\textcolor{black}{External foreground annotations cover tumour regions in ACDC-LungHP, PAIP2019,
AGGC and RINGS tumour, and gland structures in RINGS gland. METU-CCTGS uses the broader
foreground mapping described below. Thus, external pixel-level validation covers lung,
liver, prostate and colorectal tissue; inflammatory, connective, epithelial and dead/necrotic
outputs are evaluated only where internal reference annotations are available.}

Scores from different methods are paired by sample ID, using filenames and fixed ordering where
needed. Resampling uses available source clusters or specimens.
Dataset-specific units, pairing rules and cluster counts are provided in Source Data.
The shared-draw bootstrap and paired-gain calculations are specified in
\hyperref[sec:methods-seg-validation]{Methods, Evaluation and statistical analysis of segmentation}.

\textcolor{black}{Estimates from small numbers of sources, particularly the two BACH whole slides
and eight RINGS source clusters, are interpreted descriptively because resampling can capture
only limited between-source variation.}

\textcolor{black}{For the main comparison, class-specific Dice and IoU average the available
scores for each reference-present class within each dataset. To compare samples with one
or multiple scored reference classes, we first average class scores within each sample
and then average samples within the group (\cref{tab:seg-main-classes}). Source Data also
provides foreground-only sample means, excluding background and samples with no scored
foreground class. Samples are images or ROIs, except for AGGC specimen aggregates. These descriptive summaries retain the original predictions and label
mappings; the primary sample-class-weighted estimates are unchanged.}

\textcolor{black}{nnU-Net uses the 100-epoch 2D RGB trainer, fold 0 and its final checkpoint.
HoVer-Net fine-tunes its encoder and semantic feature-pyramid head for five epochs using
cross-entropy, AdamW, batch size 4 and learning rate $10^{-4}$; validation Dice selects epoch 2.
CONCHSeg uses patch-classification cross-entropy, AdamW and a cosine learning-rate schedule,
with batch size 50, learning rate 0.01 and weight decay 0.01 over 10,000 training steps.
Its checkpoint maximizes validation Dice, evaluated every 500 steps.}

RINGS tumour and gland are separate tasks in the equal-task macro but reuse all 500 images
and eight source clusters. The primary task-stratified bootstrap resamples them separately.
A sensitivity analysis coupled draws from the shared sources of the two tasks;
its results are reported in \cref{supp:results-seg-validation}.

\supplementarymethodsubsection{Shared external comparison with pathology segmentation systems}{supp:seg-shared-comparison}
The common evaluation set for PathSegmentor, VISTA-PATH and both PathXAI-Seg variants was fixed
before inspecting baseline accuracy. It contains 5,000 ACDC-LungHP images, 5,000 PAIP2019
images and 16 METU-CCTGS images/ROIs, with 132, 50 and 16 source clusters, respectively. Disclosed
training datasets determined eligibility, so RINGS was excluded because it appears among
PathSegmentor's training datasets, and AGGC because it appears in the VISTA training table. This dataset-level
check does not establish absence from all inherited foundation-model pretraining. The comparison
evaluates the released systems with their existing training budgets.

All four configurations use the same class-presence condition, label mapping, field-of-view
convention and metric definitions. Reference annotations supply present class IDs, not spatial pixel prompts; these
tasks map to IDs 0 and 1 within the unchanged seven-label vocabulary. VISTA receives a box spanning
the full padded input, $[0,0,512,512]$. \textcolor{black}{Each VISTA class query returns foreground
and background probabilities. Whenever background is requested, the task's foreground classes
are queried even if absent from that image; their minimum native background probability
supplies the background channel. Requested foreground scores populate their corresponding
channels, and unrequested class channels are zero.} Native model input sizes
are retained (PathSegmentor, 1,024; VISTA, 512 pixels). Scores are mapped back to the image grid
and averaged where predictions overlap; each pixel is assigned the class with the highest score.
For VISTA, this procedure differs from the official pipeline, which uses Otsu thresholding and
overwrites masks in area order, and from its native learned no-box condition.

Each configuration is evaluated on the same 12,448 reference-present image-class pairs. Dataset means retain
the original weighting, including background when present; the shared-set macro equally weights
the three datasets. Pointwise 95\% percentile CIs use 10,000 source-cluster bootstrap draws (seed
20260818), with common draws across methods for paired differences. Intervals are not adjusted for
multiple comparisons. METU retains the original grayscale-nonzero foreground, including normal
mucosa; this evaluation therefore measures more than tumour segmentation. IoU, paired contrasts and descriptive
class-set strata are retained with the source tables. The descriptive strata do not carry subgroup
significance tests.

\textcolor{black}{Class-specific summaries average scores over images containing the respective
reference class. We also summarize images containing both foreground and background, averaging the two
class scores within each dataset and then weighting datasets equally
(\cref{tab:seg-shared-class}). These summaries retain the original predictions and label mappings.}

The annotations in \cref{pan:seg-summary-shared-external-comparison} show differences between
point estimates; they do not provide confidence intervals or significance tests.
This class-presence-conditioned benchmark differs from reporting, where tissue composition is
unknown. Report evaluation uses PathXAI-Seg outputs in the shared tissue vocabulary.
Segmentation transfer and reporting-policy comparisons are evaluated separately; the effect
of segmentation accuracy on report performance was not isolated.

\supplementarymethodsubsection{Selection of qualitative examples}{supp:seg-example-selection}
For PAIP2019, three neoplastic-present candidates were identified using
per-image scores. These were the candidate with the highest PathXAI-Seg Dice and the two with the largest gains over the stronger of
nnU-Net and MedSAM3 fine-tuning. Each model was evaluated once per candidate on the same full field.
The displayed candidate had the highest full-field PathXAI-Seg mean Dice over reference-present
classes, including background. The displayed MedSAM3 masks
are separate inference outputs and have no per-image quantitative scores attached.
Dataset-level comparisons use the complete test sets (\crefrange{tab:source_seg:part-1}{tab:source_seg:part-4}).
The physical pixel size of these displayed crops could not be verified through their complete
resampling history, so no micrometre scale is assigned to them.

\supplementarymethodsection{Cohorts and multi-slide report generation}{supp:report-generation}
Comparisons and scoring are specified in \cref{supp:report-evaluation}; results are reported in
\cref{supp:public-report-results,supp:private-report-results,supp:report-policy-results}.

{\color{black}
\supplementarymethodsubsection{Cohorts and reference fields}{supp:report-cohorts}
TCGA slides, generated reports and references were linked by case (\cref{fig:report-data}).
Both public reference sets used OCR-processed TCGA pathology-report text released by
WsiCaption~\cite{chen2024wsicaption}, which we restructured into a common format using
Qwen3.6-27B. We initially selected 1,708 public cases, comprising a 508-case initial subset
and a 1,200-case expansion. Manual verification identified 51 cases whose supplied
site descriptions contained disease names. These cases were excluded from all public report
evaluations and their overlapping subsets before calculating results and confidence intervals.
One further case was excluded because its source WSI was corrupt.
The final public benchmark comprised 1,656 cases, including the 456-case \textcolor{black}{PathXAI-Report pipeline development subset}
and the 1,200-case \textcolor{black}{public evaluation cohort} (\cref{tab:site-prior-sensitivity}). The \textcolor{black}{PathXAI-Report pipeline development subset} was randomly sampled.
Both subsets share the report source and Qwen restructuring step.
}

The excluded cases comprised 17 each from the cholangiocarcinoma,
pheochromocytoma/paraganglioma and sarcoma groups. Their supplied descriptions included
``bile duct / cholangiocarcinoma tissue'', ``adrenal medulla / paraganglia / pheochromocytoma tissue''
and ``soft tissue / sarcoma''. The exclusion also applies to the overlapping policy and
pathologist-assessment subsets.

Public cases required a case-matched local whole-slide image (WSI) and readable reference text.
Automated screening required at least 100 characters, an alphabetic fraction of at least 40\%,
and non-duplicate normalized reference text. Slide checks required readable thumbnails and
sampled regions, micrometre-per-pixel values between 0.1 and 1.0 and no more than 2\% axis
anisotropy. Eligible cases were shuffled within cancer type and selected by cycling through
cancer types.
The same screening and sampling approach was used across the public cohort; the 1,200-case set used seed
20260920 \textcolor{black}{and did not overlap with the \textcolor{black}{PathXAI-Report pipeline development subset}}.

The public cohort was sampled from TCGA cancer projects and was not
designed to represent the distribution of benign, borderline and
malignant diagnoses encountered in routine pathology practice.

\paragraph{Reference-report processing}
The supplied OCR text was reorganized into eight report sections using Qwen3.6-27B.
\textcolor{black}{Reference processing included consistency checks against the source OCR text using OpenAI Codex (GPT-6.1).}
Diagnostic entities, grade and morphological findings were
extracted using the normalization and applicability rules in
\cref{supp:report-scoring,supp:report-grade-findings}.

\paragraph{Private cohort sources and eligibility}
The private cohort comprises 152 cases from six hospitals, with 76 breast,
62 prostate and 14 colorectal cases.
Each case has one reference report from each of two pathologists; applicable case counts are
reported separately for each metric in \crefrange{tab:source_fig9:part-1}{tab:source_fig9:part-11}.

Private cases were randomly sampled within disease categories from each contributing hospital cohort.
Breast cases were collected in 2026, and prostate and colorectal cases were collected in 2020--2025.
Detailed hospital-specific eligibility criteria were not available
to the study team. Patient demographics, including age and sex, were not collected or analysed
as study variables.

\begingroup
\manuscripttablefont
\renewcommand{\arraystretch}{1.08}
\setlength{\LTcapwidth}{\linewidth}
\setlength{\LTpre}{7pt}\setlength{\LTpost}{10pt}
\setlength{\tabcolsep}{4pt}
\begin{longtable}{@{}p{21mm}p{17mm}p{108mm}r@{}}
\caption{Private cases by anatomical site and source hospital. Hospital names
link to official institutional, university or government sources.}
\label{tab:private_hospital_sources}\\
\toprule
Site & Abbrev. & Hospital & Cases\\
\midrule
\endfirsthead
\multicolumn{4}{l}{\manuscripttablecontinued{}}\\
\toprule
Site & Abbrev. & Hospital & Cases\\
\midrule
\endhead
\midrule
\multicolumn{4}{r}{Continued on next page}\\
\endfoot
\bottomrule
\endlastfoot
Breast & TTSH & \href{https://www.ttsh.com.sg/}{Tan Tock Seng Hospital} & 76 \\
Prostate & ZSH & \href{https://shmc.fudan.edu.cn/eng/2094/list.htm}{Zhongshan Hospital, Fudan University} & 12 \\
Prostate & CHH & \href{https://english.shanghai.gov.cn/en-Latest-WhatsNew/20250530/18ca7fc1c4a94930b49d77bde8f50805.html}{Changhai Hospital} & 12 \\
Prostate & CZH & \href{https://english.shanghai.gov.cn/en-Latest-WhatsNew/20250822/e4ef0658065b44dfaaed6522f6b02660.html}{Shanghai Changzheng Hospital} & 12 \\
Prostate & HBGH & \href{https://www.hbpphosp.cn/}{Hebei General Hospital} & 12 \\
Prostate & JPH & \href{https://english.njmu.edu.cn/_t965/6080/list.htm}{Jiangsu Province Hospital} & 14 \\
Colorectal & CZH & \href{https://english.shanghai.gov.cn/en-Latest-WhatsNew/20250822/e4ef0658065b44dfaaed6522f6b02660.html}{Shanghai Changzheng Hospital} & 14 \\
\midrule
Total & & Six hospitals; three anatomical sites & 152\\
\end{longtable}
\endgroup

ZSH, CHH, CZH, HBGH and JPH are abbreviations defined for this manuscript;
TTSH is the hospital's established abbreviation. Jiangsu Province Hospital
is also the First Affiliated Hospital of Nanjing Medical University.
Hospital assignments were supplied by the investigators and matched to the
evaluated cases. \textcolor{black}{The accompanying Source Data provide aggregate site/hospital counts and official-name source links.}

\paragraph{Cohort summaries and Source Data}
\textcolor{black}{Slide counts summarize available WSIs linked to each public case; private slide linkage is unavailable.} Field availability is calculated across the full cohort separately
for each reference set. Cancer-type target availability in the public cohort follows TCGA labels;
availability of a specific diagnosis in the private cohort is determined separately for each reference set.

The accompanying Source Data list case IDs and the cases included in the input-policy,
HistGen/HistoGPT and pathologist comparisons, whose subsets overlap.
The \texttt{public\_cohorts} directory contains case membership and method availability,
and \texttt{aggregate\_results} contains the numerical results and their table index.
Private records remain restricted;
\cref{tab:private_hospital_sources} gives their aggregate source composition.

\textcolor{black}{Public demographic characteristics were retrieved from the
\href{https://docs.gdc.cancer.gov/API/Users_Guide/Search_and_Retrieval/}{NCI Genomic Data Commons}
and linked by TCGA case identifier. \Cref{tab:public-cohort-metadata} summarizes age,
sex at birth, race and ethnicity, together with available slide types classified from
\href{https://docs.gdc.cancer.gov/Encyclopedia/pages/TCGA_Barcode/}{TCGA barcodes}.}

\Needspace{0.64\textheight}
\begingroup
\color{black}
\manuscripttablefont
\setlength{\LTcapwidth}{\linewidth}
\setlength{\tabcolsep}{4pt}
\renewcommand{\arraystretch}{1.05}
\begin{longtable}{@{}p{0.51\linewidth}*{2}{>{\centering\arraybackslash}p{\dimexpr(0.49\linewidth-4\tabcolsep)/2\relax}}@{}}
\caption{Public cohort characteristics and available slide types. Values are counts (\%) unless indicated. Clinical characteristics use GDC metadata linked by case ID; slide percentages use the recoverable slide inventory. IQR, interquartile range.}\label{tab:public-cohort-metadata}\\
\toprule
Characteristic & \textcolor{black}{PathXAI-Report pipeline development subset} & Evaluation cohort\\
\midrule
\endfirsthead
\multicolumn{3}{@{}l}{\manuscripttablecontinued{}}\\[3pt]
\toprule
Characteristic & \textcolor{black}{PathXAI-Report pipeline development subset} & Evaluation cohort\\
\midrule
\endhead
\midrule\multicolumn{3}{r@{}}{Continued on next page}\\\endfoot
\bottomrule\endlastfoot
Cases & 456 & 1,200\\
\addlinespace[3pt]\multicolumn{3}{@{}l}{\textbf{Age at GDC index (years)}}\\*
Available & 453 (99.3\%) & 1,198 (99.8\%)\\
Median (IQR) & 59.0 (50.0--67.0) & 58.0 (47.0--68.0)\\
Missing & 3 (0.7\%) & 2 (0.2\%)\\
Age privacy flag & 5 (1.1\%) & 12 (1.0\%)\\
\addlinespace[3pt]\multicolumn{3}{@{}l}{\textbf{Sex at birth}}\\*
Female & 204 (44.7\%) & 569 (47.4\%)\\
Male & 252 (55.3\%) & 631 (52.6\%)\\
\addlinespace[3pt]\multicolumn{3}{@{}l}{\textbf{Race recorded by GDC}}\\*
White & 316 (69.3\%) & 912 (76.0\%)\\
Black or African American & 44 (9.6\%) & 107 (8.9\%)\\
Asian & 40 (8.8\%) & 96 (8.0\%)\\
American Indian or Alaska Native & 0 (0.0\%) & 5 (0.4\%)\\
Unknown / not reported & 56 (12.3\%) & 80 (6.7\%)\\
\addlinespace[3pt]\multicolumn{3}{@{}l}{\textbf{Ethnicity recorded by GDC}}\\*
Hispanic or Latino & 15 (3.3\%) & 67 (5.6\%)\\
Not Hispanic or Latino & 308 (67.5\%) & 899 (74.9\%)\\
Unknown / not reported & 133 (29.2\%) & 234 (19.5\%)\\
\addlinespace[3pt]\multicolumn{3}{@{}l}{\textbf{Available slide inventory}}\\*
Slides with recoverable identifiers & 934 & 3,121\\
DX (diagnostic) & 405 (43.4\%) & 1,337 (42.8\%)\\
TS (top) & 374 (40.0\%) & 1,339 (42.9\%)\\
BS (bottom) & 150 (16.1\%) & 433 (13.9\%)\\
MS (middle) & 5 (0.5\%) & 12 (0.4\%)\\
\end{longtable}
Age at index is the age on the reference date recorded by GDC; this date is diagnosis for
1,655 cases and unspecified for one. Released age values, including those flagged as modified
for privacy, are retained; the privacy flag is unreported for one evaluation case.
Age summaries exclude missing values. Race and ethnicity are
reported as recorded by GDC, with unknown and unreported values grouped in the table.
Slide codes are derived from TCGA barcodes. \textcolor{black}{Slide counts for the PathXAI-Report pipeline development subset} differ between this
inventory (934) and the earlier aggregate (937); the three-slide discrepancy remains
unresolved. Counts describe available slides, not the images supplied
to each reporting run. Source Data provide per-case metadata and slide codes.
\endgroup

{\color{black}
\paragraph{Overlap with segmentation training data}
We matched TCGA patient identifiers in the segmentation training sample lists to the
public report cases. A sensitivity analysis excluded all matching cases from the public
evaluation cohort, applying the same exclusion to every reporting method. Reference
applicability and the paired-case bootstrap procedure were unchanged
(\cref{supp:results-report-training-overlap}). The matching covers identifiable TCGA
cases; sources without recoverable patient identifiers and undisclosed foundation-model
pretraining data could not be assessed.
}

\manuscriptnativefigure{\textbf{Reporting cohorts and evaluation design.}}{fig:report-data}{%
    \begin{minipage}{180mm}
    \manuscriptpanelblock{5mm}{180.0mm}{\manuscriptpanelmark{pan:report-data-cohorts}}{External cohorts and evaluation subsets}{\includegraphics[width=\linewidth]{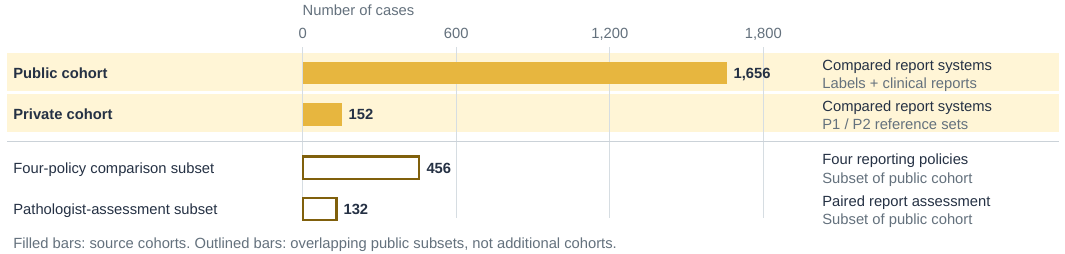}}
    \par\vspace{2mm}
    \begin{minipage}[t]{88mm}\vspace{0pt}\manuscriptpanelblock{5mm}{88.0mm}{\manuscriptpanelmark{pan:report-data-organ-source-composition}}{Organ and source composition}{\includegraphics[width=\linewidth]{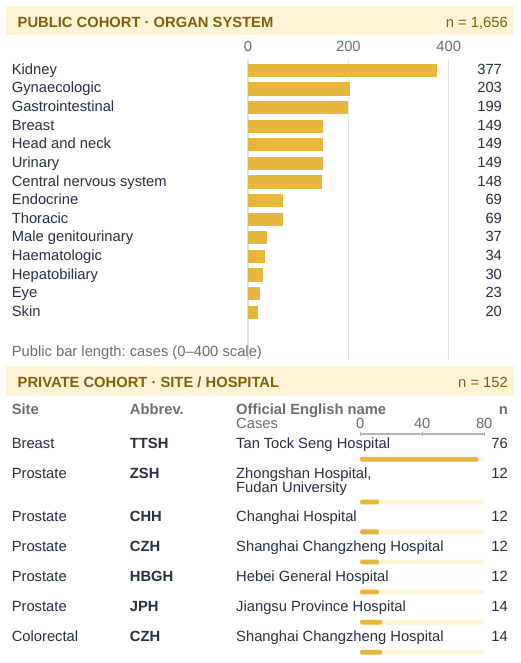}}\end{minipage}\hfill\begin{minipage}[t]{88mm}\vspace{0pt}\manuscriptpanelblock{5mm}{88.0mm}{\manuscriptpanelmark{pan:report-data-slides-per-case}}{Available slides per case}{\includegraphics[width=\linewidth]{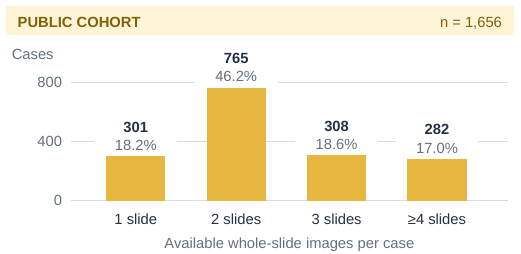}}\par\vspace{2mm}\manuscriptpanelblock{5mm}{88.0mm}{\manuscriptpanelmark{pan:report-data-reference-fields}}{Assessable reference fields}{\includegraphics[width=\linewidth]{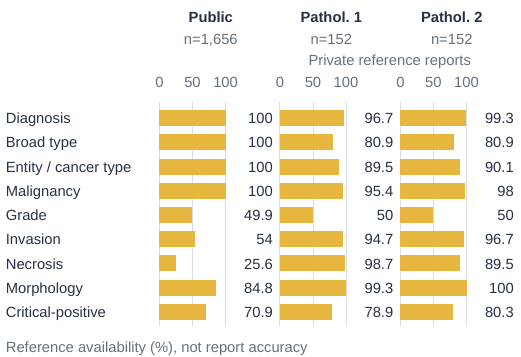}}\end{minipage}
    \par\vspace{2mm}
    \manuscriptpanelblock{5mm}{180.0mm}{\manuscriptpanelmark{pan:report-data-evaluation-design}}{Reference comparisons and pathologist ranking and correction assessment}{\includegraphics[width=\linewidth]{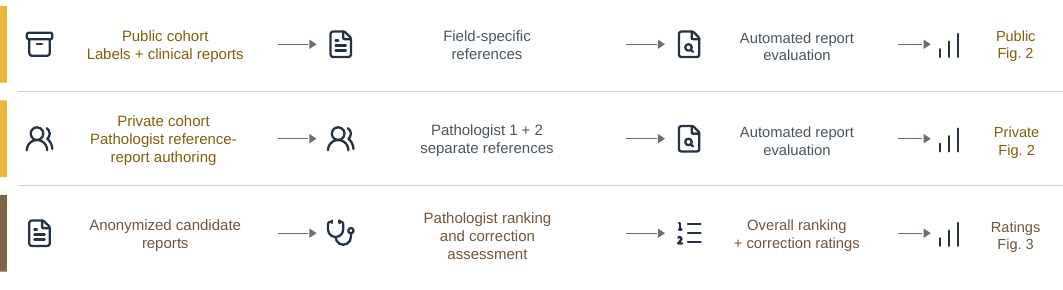}}
    \end{minipage}%
  }{%
\textbf{\panelletterref{pan:report-data-cohorts} Cohorts and overlapping evaluation subsets.}
\textbf{\panelletterref{pan:report-data-organ-source-composition} Organ and hospital composition.}
\textbf{\panelletterref{pan:report-data-slides-per-case} Available slides per case.}
\textbf{\panelletterref{pan:report-data-reference-fields} Reference-field applicability.} Public cancer-family targets and private specific diagnoses define diagnostic applicability. Critical-positive measures require a documented positive finding; private reference sets are evaluated separately.
\textbf{\panelletterref{pan:report-data-evaluation-design} Evaluation tasks.} Reference-based automated scoring and separate pathologist ranking and correction assessment. Cohort definitions are in \cref{supp:report-cohorts,tab:private_hospital_sources}.
(Notation: WSI, whole-slide image; P1/P2, pathologists' reference sets; OCR, optical character recognition. Outlined bars identify overlapping subsets; counts describe cohort composition.)
}

\supplementarymethodsubsection{Reporting configurations, inputs and prompts}{supp:report-configurations}\label{supp:input-identifiers}
Core and complete configurations, their inference settings and the diagnostic selector are specified in
\hyperref[sec:methods-report-inference]{Methods, Reporting configurations and inference}
and \hyperref[sec:methods-report-diagnosis]{Diagnostic candidates and selection}.
\textcolor{black}{\Cref{fig:report-model} details candidate selection and reinspection; the overall workflow is shown in \cref{fig:methods-summary}.} Input-policy variants are specified separately in
\cref{supp:report-input-comparisons,supp:report-policy-comparisons}.

\manuscriptnativefigure{\textbf{\textcolor{black}{Diagnostic candidate selection and reinspection rules.}}}{fig:report-model}{%
    \begin{minipage}{180mm}
    \manuscriptpanelblock{4mm}{180.0mm}{\manuscriptpanelmark{pan:report-model-diagnosis-selection}}{V/T candidate selection}{\includegraphics[width=\linewidth]{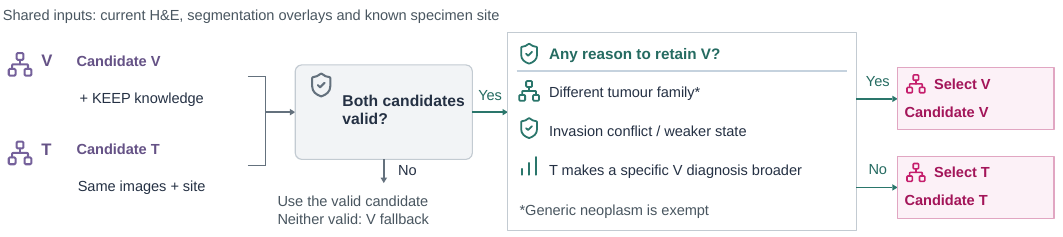}}
    \par\vspace{2.5mm}
    \hfill\manuscriptpanelblock{4mm}{88.0mm}{\manuscriptpanelmark{pan:report-model-reinspection}}{Reinspection and stopping rules}{\includegraphics[width=\linewidth]{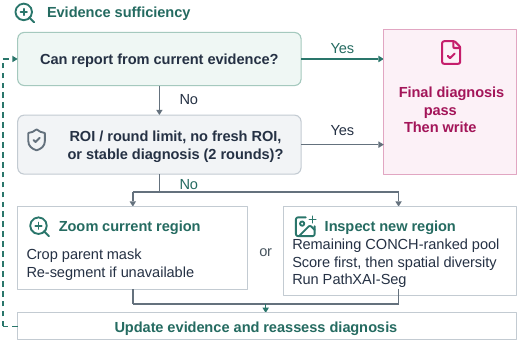}}\hfill\mbox{}
    \end{minipage}%
  }{%
\textcolor{black}{\textbf{\panelletterref{pan:report-model-diagnosis-selection} Candidate selection.} Validity, tumour-family, invasion-state and specificity checks select between V and T. The selected candidate is subsequently constrained by the subtype-sufficiency assessment (S; \cref{pan:methods-diagnosis-revisit}).
\textbf{\panelletterref{pan:report-model-reinspection} Reinspection.} The Boolean reportability decision, acquisition limits, availability of fresh ROIs and diagnosis stability govern further acquisition. All exits proceed to a final diagnostic pass.
Detailed rules are given in \hyperref[sec:methods-report-diagnosis]{Methods, Diagnostic candidates and selection} and \hyperref[sec:methods-report-reinspection]{\textcolor{black}{Evidence assessment and reinspection}}. V/T denote diagnostic candidates; ROI, region of interest.}
}

\paragraph{Image inputs and identifiers}
PathXAI-Report receives H\&E images, aligned overlays and tissue statistics, with known-site context and no hidden
segmentation features or embeddings (\cref{supp:seg-inference}). Anatomical site was also provided to the pathologists,
at their request, when they authored the private reference reports
(\hyperref[sec:methods-report-human]{Methods, Pathologist reference-report authoring}). Screening of the supplied site descriptions is described in
\cref{supp:report-cohorts}.

\textcolor{black}{For a square diagnostic ROI, the H\&E image and aligned overlay are each
resized to $1,024\times1,024$ pixels and placed side by side, with H\&E on the left.
The resulting $2,048\times1,052$-pixel composite includes a 28-pixel label strip.
The entire composite is then resized to a maximum edge of 768 pixels, leaving approximately
$384\times384$ pixels per image panel; a standalone square H\&E image uses
$768\times768$ pixels. Parent and zoom composites are initially saved at JPEG qualities
85 and 88, respectively, and the inference client transmits them at quality 75.}

\textcolor{black}{When the evidence assessment requests additional image detail, zoom can
extract the central $2,048\times2,048$ pixels of a $4,096\times4,096$-pixel parent ROI,
subject to the acquisition limits. The crop uses the source slide at the parent's level
or the stored parent image, with the corresponding parent mask cropped to match.
It is resized and paired with its overlay using the same transmitted image budget,
so the smaller field occupies the same panel size. For level-0 parent ROIs, physical
field width depends on the slide calibration $m$ in $\mu$m per pixel, giving $4,096m$ for the parent
and $2,048m$ for the zoom. The $0.5$~$\mu$m-per-pixel setting in ROI ranking applies to
CONCH tiles.}

PathXAI-Report's candidate-ROI descriptions identify images by within-case ROI
order and omit file paths and case or slide identifiers. 
The inference client reads images locally and sends them as inline image data to the configured self-hosted inference service, without exposing local file paths or providing filesystem tools to the model.
Auxiliary writer and claim-check inputs nevertheless retain ROI identifiers containing TCGA
identifiers and coordinates in the public evaluation.
Prompts additionally prohibit diagnostic inference from identifiers or
coordinates.
\textcolor{black}{Reference reports are used only for subsequent evaluation.}

\textcolor{black}{In the public-comparison implementations, Qwen+H\&E and MedGemma also use
ordinal ROI descriptions for report writing, but their claim-check graphs can retain raw
ROI identifiers. PLIP-guided Qwen includes the final 40 characters of each selected ROI
identifier in its textual feature description. DeepSeek+H\&E and GLM+H\&E omit explicit
case identifiers, cohort codes and file paths from prompt text; PRISM2 uses an anonymous
slide name and a site-conditioned question. HistGen and HistoGPT generate reports from
image features without case identifiers in their generation inputs.}

\paragraph{Prompt stages and message assembly}\label{supp:agent-prompts}
The templates below reproduce the instructions and requested output formats of the documented
core and complete implementations. Angle-bracket placeholders identify dynamically inserted
content; JSON braces and example values belong to the original templates. The machine-readable
supplement \texttt{report\_prompts.json} retains the prompt functions, source expressions,
output definitions and parsing rules. \textcolor{black}{Complete case-specific request payloads
were not retained for the original experiments.}

Each call supplies its instructions as user content, without a separate system message.
The request places the prompt text before the current images, encoded as inline JPEGs in the
order supplied by the stage. Earlier outputs enter through explicit fields such as the current
diagnosis, evidence gaps and claim assessments; these two configurations do not use the optional
conversation-history wrapper. Known site bypasses the site-prediction call. The optional
morphology reader is disabled in both configurations, so its description fields are empty;
the shared templates retain their generic morphology headings and instructions.

\begingroup\small
\setlength{\tabcolsep}{4pt}\renewcommand{\arraystretch}{1.12}
\noindent\begin{tabular}{@{}p{0.19\linewidth}p{0.36\linewidth}p{0.39\linewidth}@{}}
\toprule
Stage & Principal inputs & Requested output \\
\midrule
ROI selection & Ranked candidate images and candidate list & Selected indices and reasons \\
Broad lineage & Current ROI images, site and available tissue/context summaries & Lineage hypotheses and discriminating features \\
Subtype sufficiency & Current evidence, broad hypotheses and subtype candidates & Supported diagnostic level and missing evidence \\
Diagnostic candidates & \textcolor{black}{H\&E-overlay images and site; complete V also receives KEEP} & Diagnosis, morphology, ancillary observations and recommendations \\
Evidence sufficiency & Current ROI images, diagnosis and site & Reportability and image/non-image evidence gaps \\
Claim assessment & Claims, evidence graph, tissue summaries and ROI images & Per-claim support status, cited evidence and reason \\
Report writing & Diagnostic state, evidence gaps, ROI images and writing context & Eight-section report \\
Report revision & Draft, diagnostic state and claim assessments & Revised report with the same section structure \\
\bottomrule
\end{tabular}\par
\endgroup

Both configurations use these stages. The complete configuration adds retrieved KEEP context
to \textcolor{black}{candidate V}, invasion safeguards in candidate selection, a locked diagnostic
state for writing, and auxiliary grade-focused images supplied only to the writer.
Broad-lineage and subtype-sufficiency calls run once on the current evidence before the V/T
candidate pair; the broad-lineage output feeds the subtype-sufficiency call. After candidate
selection, a subtype assessment whose supported level is below subtype can broaden an
insufficiently supported diagnosis and add its unresolved discriminators to the evidence-gap fields.
Both preliminary outputs also enter the evidence graph. Reinspection repeats this sequence
and sufficiency assessment with updated evidence.
Candidate selection, knowledge retrieval, acquisition decisions and final lock enforcement are
implemented in code; their rules are described in Methods.

\paragraph{Dynamic input fields}
The candidate and selected-ROI lists use one-based image order, crop dimensions, ranking scores
and available role information. Site placeholders contain the supplied anatomical site or its
structured record. Tissue-summary placeholders contain ROI foreground fractions and named
non-background tissue fractions. The compact formatter includes up to eight ROI summaries,
retaining class fractions above 0.5\%; the tissue-evidence policies in
\cref{supp:report-policy-comparisons} define the inputs to this formatter. \textcolor{black}{This eight-entry
limit applies to the text summary; acquisition can retain up to eleven diagnostic ROIs,
with additional writer-only images supplied separately.}
Retrieved-knowledge placeholders contain disease definitions and subtype candidates. In the
diagnosis template, these comprise at most three definitions, each truncated to 300 characters,
and eight candidate names with up to three synonyms each. Diagnostic-state, sufficiency, graph
and claim placeholders are JSON records produced by the preceding stages. Writer-only ROI
identifiers and their role policy enter the auxiliary block. The following templates show
which fields actually enter each prompt; unused optional blocks contribute no description.

\paragraph{Initial ROI selection}
The candidate list follows the attached image order. The selection budget is three, from the
leading eight candidates described in Methods. Returned indices are one-based.
\begin{agentprompt}
\noindent /no\char95\relax\allowbreak{}think\par
\noindent You are selecting WSI ROI evidence for pathology report generation.\par
\vspace{0.45\baselineskip}
\noindent You will see candidate ROI images in the same order as the list below.\par
\noindent Choose up to \char60\relax{}maximum\char95\relax\allowbreak{}selected\char95\relax\allowbreak{}rois\char62\relax{} ROIs that best support diagnosis and report generation.\par
\vspace{0.45\baselineskip}
\noindent Selection criteria:\par
\noindent - Must contain viable lesional/tumor tissue when possible.\par
\noindent - Prefer ROIs with diagnostic architecture, cytology, necrosis, invasion interface, or representative tumor.\par
\noindent - Avoid mostly blank/background/artifact/normal tissue.\par
\noindent - Select fewer than \char60\relax{}maximum\char95\relax\allowbreak{}selected\char95\relax\allowbreak{}rois\char62\relax{} if fewer are truly useful.\par
\noindent - Do not infer site or diagnosis from filenames, case IDs, directory names, or coordinates.\par
\vspace{0.45\baselineskip}
\noindent Candidate list:\par
\noindent \char60\relax{}candidate\char95\relax\allowbreak{}roi\char95\relax\allowbreak{}list\char62\relax{}\par
\vspace{0.45\baselineskip}
\noindent Return ONLY JSON:\par
\noindent \char123\relax{}\par
\noindent \hspace*{1.0em}"selected\char95\relax\allowbreak{}roi\char95\relax\allowbreak{}indices": [1, 2],\par
\noindent \hspace*{1.0em}"reason": "why these ROIs are enough or why more would be needed",\par
\noindent \hspace*{1.0em}"roi\char95\relax\allowbreak{}scores": [\par
\noindent \hspace*{2.0em}\char123\relax{}"roi\char95\relax\allowbreak{}index": 1, "diagnostic\char95\relax\allowbreak{}value": 0-5, "tumor\char95\relax\allowbreak{}likelihood": 0-5, "tissue\char95\relax\allowbreak{}quality": 0-5, "reason": "..."\char125\relax{}\par
\noindent \hspace*{1.0em}]\par
\noindent \char125\relax{}\par
\end{agentprompt}

\paragraph{Broad-lineage reasoning}
This stage proposes broad histogenetic categories and the features needed to distinguish them.
The output remains separate from the subsequent specific diagnosis.
\begin{agentprompt}
\noindent /no\char95\relax\allowbreak{}think\par
\noindent Classify the histogenetic BROAD LINEAGE from the ROI morphology.\par
\vspace{0.45\baselineskip}
\noindent **CRITICAL: best\char95\relax\allowbreak{}broad\char95\relax\allowbreak{}type MUST be exactly ONE of these labels:**\par
\noindent \hspace*{1.0em}carcinoma | sarcoma | lymphoma | melanoma | glioma |\par
\noindent \hspace*{1.0em}neuroendocrine neoplasm | germ cell tumor | mesothelioma |\par
\noindent \hspace*{1.0em}carcinosarcoma | thymic epithelial neoplasm | neoplasm\par
\vspace{0.45\baselineskip}
\noindent **DO NOT** put a specific diagnosis here (e.g. "Adrenocortical Carcinoma" is NOT\par
\noindent a broad type \mbox{\normalfont\textemdash} its broad type is "carcinoma"). If you cannot narrow below\par
\noindent "neoplasm", say "neoplasm".\par
\vspace{0.45\baselineskip}
\noindent Rules:\par
\noindent - Use only visible ROI morphology, segmentation evidence, morphology evidence text,\par
\noindent \hspace*{1.0em}known/predicted site context, and KEEP background candidate knowledge.\par
\noindent - KEEP is background knowledge, not patient-specific evidence.\par
\noindent - Morphology evidence text is a non-binding descriptive clue, not ground truth.\par
\noindent - Segmentation fractions are localized tissue-composition / ROI-quality evidence,\par
\noindent \hspace*{1.0em}not lineage labels. They may support tumor amount, necrosis, stromal or\par
\noindent \hspace*{1.0em}epithelial-labeled tissue presence, but must not override broad lineage from\par
\noindent \hspace*{1.0em}H\&E morphology.\par
\noindent - Low or absent segmentation fractions are not negative exclusion evidence for a\par
\noindent \hspace*{1.0em}lineage or subtype.\par
\noindent - Do not use filenames, case IDs, TCGA codes, directory names, or coordinates.\par
\noindent - If known site is high confidence, site-specific primary lineages remain the\par
\noindent \hspace*{1.0em}default primary space unless multiple ROIs directly support another lineage.\par
\noindent \char60\relax{}optional\char95\relax\allowbreak{}morphology\char95\relax\allowbreak{}rules\char62\relax{}\par
\vspace{0.45\baselineskip}
\noindent Site:\par
\noindent \char60\relax{}known\char95\relax\allowbreak{}site\char95\relax\allowbreak{}json\char62\relax{}\par
\vspace{0.45\baselineskip}
\noindent Segmentation:\par
\noindent \char60\relax{}segmentation\char95\relax\allowbreak{}summary\char95\relax\allowbreak{}json\char62\relax{}\par
\vspace{0.45\baselineskip}
\noindent \char60\relax{}optional\char95\relax\allowbreak{}morphology\char95\relax\allowbreak{}label\char62\relax{}:\par
\noindent \char60\relax{}optional\char95\relax\allowbreak{}morphology\char95\relax\allowbreak{}text\char62\relax{}\par
\vspace{0.45\baselineskip}
\noindent KEEP context:\par
\noindent \char60\relax{}retrieved\char95\relax\allowbreak{}knowledge\char95\relax\allowbreak{}json\char62\relax{}\par
\vspace{0.45\baselineskip}
\noindent Selected ROI evidence:\par
\noindent \char60\relax{}selected\char95\relax\allowbreak{}roi\char95\relax\allowbreak{}list\char62\relax{}\par
\vspace{0.45\baselineskip}
\noindent Return ONLY JSON:\par
\noindent \char123\relax{}\par
\noindent \hspace*{1.0em}"hypotheses": [\par
\noindent \hspace*{2.0em}\char123\relax{}"broad\char95\relax\allowbreak{}type": "one of the allowed labels above", "confidence": "high/medium/low", "supporting\char95\relax\allowbreak{}features": ["..."], "main\char95\relax\allowbreak{}mimics": ["..."], "missing\char95\relax\allowbreak{}discriminators": ["..."]\char125\relax{}\par
\noindent \hspace*{1.0em}],\par
\noindent \hspace*{1.0em}"best\char95\relax\allowbreak{}broad\char95\relax\allowbreak{}type": "one of: carcinoma | sarcoma | lymphoma | melanoma | glioma | neuroendocrine neoplasm | germ cell tumor | mesothelioma | carcinosarcoma | thymic epithelial neoplasm | neoplasm",\par
\noindent \hspace*{1.0em}"confidence": "high/medium/low",\par
\noindent \hspace*{1.0em}"reason": "..."\par
\noindent \char125\relax{}\par
\end{agentprompt}

\paragraph{Subtype sufficiency}
This stage assesses the candidate entities and separates image-actionable gaps from unavailable
ancillary or clinical information. An absent current diagnosis is inserted as an empty object.
\begin{agentprompt}
\noindent /no\char95\relax\allowbreak{}think\par
\noindent Judge whether current ROI evidence is sufficient to distinguish site-specific subtype candidates.\par
\vspace{0.45\baselineskip}
\noindent Rules:\par
\noindent - Evaluate subtype candidates one by one.\par
\noindent - Use KEEP candidates as a candidate list only; do not treat them as patient-specific evidence.\par
\noindent - Treat morphology evidence suspected diagnoses as non-binding clues.\par
\noindent - D1 segmentation classes are tissue-pattern evidence, not lineage labels.\par
\noindent - Segmentation stats may support tumor richness, necrosis amount, tissue-class\par
\noindent \hspace*{1.0em}composition, and ROI quality only.\par
\noindent - Do NOT use segmentation stats to decide exact subtype, invasive vs in-situ\par
\noindent \hspace*{1.0em}status, histologic grade, or malignancy unless direct H\&E morphology also\par
\noindent \hspace*{1.0em}supports that statement.\par
\noindent - Do NOT infer DCIS/in-situ disease, invasion, or subtype from epithelial/glandular\par
\noindent \hspace*{1.0em}fractions, central necrosis, low stromal fraction, or absent segmentation\par
\noindent \hspace*{1.0em}classes alone.\par
\noindent - Low or absent segmentation fractions are not negative exclusion evidence.\par
\noindent - If broad lineage/site is supported but subtype discriminators are missing,\par
\noindent \hspace*{1.0em}recommend a broad/NOS/not-further-classifiable diagnosis.\par
\noindent - Do NOT demote a visually supported likely subtype only because ancillary\par
\noindent \hspace*{1.0em}confirmation, molecular tests, grade, stage, margins, or complete scoring\par
\noindent \hspace*{1.0em}systems are unavailable. Those are limitations/prognostic gaps, not always\par
\noindent \hspace*{1.0em}reasons to erase the best-supported diagnostic entity.\par
\noindent - Use "possible\char95\relax\allowbreak{}but\char95\relax\allowbreak{}insufficient" when a candidate has some supporting visual\par
\noindent \hspace*{1.0em}features but unresolved differential diagnosis remains; use "supported" when\par
\noindent \hspace*{1.0em}current visual+morphology evidence supports it as the best working diagnosis,\par
\noindent \hspace*{1.0em}even if ancillary confirmation is still recommended.\par
\noindent - Distinguish image-actionable missing evidence from nonimage metadata.\par
\noindent - Do not use filenames, case IDs, TCGA codes, directory names, or coordinates.\par
\noindent - `recommended\char95\relax\allowbreak{}primary\char95\relax\allowbreak{}diagnosis` must contain only the diagnostic entity, not\par
\noindent \hspace*{1.0em}phrases like "not further classifiable from ROI evidence".\par
\noindent \char60\relax{}optional\char95\relax\allowbreak{}morphology\char95\relax\allowbreak{}rules\char62\relax{}\par
\vspace{0.45\baselineskip}
\noindent Site: \char60\relax{}known\char95\relax\allowbreak{}site\char62\relax{}\par
\vspace{0.45\baselineskip}
\noindent Broad hypotheses:\par
\noindent \char60\relax{}broad\char95\relax\allowbreak{}hypotheses\char95\relax\allowbreak{}json\char62\relax{}\par
\vspace{0.45\baselineskip}
\noindent Current diagnosis, if any:\par
\noindent \char60\relax{}current\char95\relax\allowbreak{}diagnosis\char95\relax\allowbreak{}json\char62\relax{}\par
\vspace{0.45\baselineskip}
\noindent Site-specific subtype candidates:\par
\noindent \char60\relax{}subtype\char95\relax\allowbreak{}candidates\char95\relax\allowbreak{}json\char62\relax{}\par
\vspace{0.45\baselineskip}
\noindent Segmentation:\par
\noindent \char60\relax{}segmentation\char95\relax\allowbreak{}summary\char95\relax\allowbreak{}json\char62\relax{}\par
\vspace{0.45\baselineskip}
\noindent \char60\relax{}optional\char95\relax\allowbreak{}morphology\char95\relax\allowbreak{}label\char62\relax{}:\par
\noindent \char60\relax{}optional\char95\relax\allowbreak{}morphology\char95\relax\allowbreak{}text\char62\relax{}\par
\vspace{0.45\baselineskip}
\noindent Selected ROI evidence:\par
\noindent \char60\relax{}selected\char95\relax\allowbreak{}roi\char95\relax\allowbreak{}list\char62\relax{}\par
\vspace{0.45\baselineskip}
\noindent Return ONLY JSON:\par
\noindent \char123\relax{}\par
\noindent \hspace*{1.0em}"best\char95\relax\allowbreak{}supported\char95\relax\allowbreak{}level": "broad\char95\relax\allowbreak{}type/subtype/not\char95\relax\allowbreak{}assessable",\par
\noindent \hspace*{1.0em}"recommended\char95\relax\allowbreak{}primary\char95\relax\allowbreak{}diagnosis": "...",\par
\noindent \hspace*{1.0em}"confidence": "high/medium/low",\par
\noindent \hspace*{1.0em}"subtype\char95\relax\allowbreak{}assessments": [\par
\noindent \hspace*{2.0em}\char123\relax{}\par
\noindent \hspace*{3.0em}"subtype": "...",\par
\noindent \hspace*{3.0em}"status": "supported/possible\char95\relax\allowbreak{}but\char95\relax\allowbreak{}insufficient/unsupported/not\char95\relax\allowbreak{}assessable",\par
\noindent \hspace*{3.0em}"supporting\char95\relax\allowbreak{}features": ["..."],\par
\noindent \hspace*{3.0em}"missing\char95\relax\allowbreak{}discriminators": ["..."],\par
\noindent \hspace*{3.0em}"needed\char95\relax\allowbreak{}image\char95\relax\allowbreak{}evidence": ["..."],\par
\noindent \hspace*{3.0em}"needed\char95\relax\allowbreak{}nonimage\char95\relax\allowbreak{}metadata": ["..."]\par
\noindent \hspace*{2.0em}\char125\relax{}\par
\noindent \hspace*{1.0em}],\par
\noindent \hspace*{1.0em}"image\char95\relax\allowbreak{}actionable\char95\relax\allowbreak{}gaps": ["..."],\par
\noindent \hspace*{1.0em}"nonimage\char95\relax\allowbreak{}metadata\char95\relax\allowbreak{}gaps": ["..."],\par
\noindent \hspace*{1.0em}"reason": "..."\par
\noindent \char125\relax{}\par
\end{agentprompt}

\paragraph{Diagnostic candidates}
The same template generates the V and T candidate records from the current H\&E-overlay
composites. Its internal two-step instruction requests a fresh broad-lineage judgement and
subtype refinement; it does not insert the earlier broad-lineage or subtype-sufficiency records
as text. Numeric tissue summaries enter the preliminary reasoning, claim-assessment and writing
prompts. The diagnostic template below does not serialize its \texttt{segmentation\_context}
argument. Optional morphology text and the chat-history override block are empty in these
configurations. Retrieved definitions and candidate names enter the complete configuration's
V call; the T call receives no retrieved KEEP text. \textcolor{black}{In the core configuration,
V and T are repeated calls with identical prompt and image inputs.}
\begin{agentprompt}
\noindent /no\char95\relax\allowbreak{}think\par
\noindent You are a pathologist making a diagnosis from H\&E WSI ROI images.\par
\vspace{0.45\baselineskip}
\noindent Specimen site: \char60\relax{}known\char95\relax\allowbreak{}site\char62\relax{}\par
\vspace{0.45\baselineskip}
\noindent You have ROI images showing representative tumor tissue.\par
\vspace{0.45\baselineskip}
\noindent REASON IN TWO STEPS:\par
\vspace{0.45\baselineskip}
\noindent STEP 1 \mbox{\normalfont\textemdash} BROAD LINEAGE (from ROI images ONLY):\par
\noindent Look at the ROI images. Determine the broad histogenetic lineage:\par
\noindent carcinoma / sarcoma / lymphoma / melanoma / glioma / germ cell tumor / \par
\noindent neuroendocrine neoplasm / mesothelioma / other neoplasm\par
\vspace{0.45\baselineskip}
\noindent Base this decision on the ROI MORPHOLOGY alone. Do not let any text context\par
\noindent below influence this step. The images have the final say.\par
\vspace{0.45\baselineskip}
\noindent STEP 2 \mbox{\normalfont\textemdash} SUBTYPE REFINEMENT (from text context + broad from Step 1):\par
\noindent With the broad lineage locked from Step 1, use the context below to refine\par
\noindent the specific subtype, malignancy, and grade. The context is supplemental \mbox{\normalfont\textemdash}\par
\noindent only commit to a subtype when the morphology supports it.\par
\vspace{0.45\baselineskip}
\noindent \char60\relax{}optional\char95\relax\allowbreak{}morphology\char95\relax\allowbreak{}block\char62\relax{}\par
\noindent \char60\relax{}retrieved\char95\relax\allowbreak{}disease\char95\relax\allowbreak{}definitions\char62\relax{}\par
\noindent \char60\relax{}retrieved\char95\relax\allowbreak{}subtype\char95\relax\allowbreak{}candidates\char62\relax{}\par
\vspace{0.45\baselineskip}
\noindent Critical evidence-boundary rules:\par
\noindent - These are H\&E ROI images unless explicitly stated otherwise.\par
\noindent - Do NOT report actual IHC/molecular positivity or negativity from H\&E.\par
\noindent - Morphology evidence and KEEP are supporting context, not ground truth.\par
\noindent - Segmentation context, when provided, is support for localized tissue\par
\noindent \hspace*{1.0em}composition / tumor burden / necrosis / ROI quality only. It cannot determine\par
\noindent \hspace*{1.0em}the primary diagnosis, subtype, invasive-vs-in-situ status, grade, or\par
\noindent \hspace*{1.0em}malignancy by itself.\par
\noindent - Do not change an invasive-vs-in-situ call based only on segmentation stats.\par
\noindent \hspace*{1.0em}If the invasion interface or in-situ architecture is not directly visible in\par
\noindent \hspace*{1.0em}H\&E morphology, keep that field uncertain and do not let it replace the\par
\noindent \hspace*{1.0em}primary diagnosis.\par
\noindent - Low or absent segmentation fractions are not negative exclusion evidence\par
\noindent \hspace*{1.0em}against a cancer type or subtype.\par
\noindent - If ROI evidence does not support a specific subtype, leave subtype empty.\par
\vspace{0.45\baselineskip}
\noindent Candidate ROIs:\par
\noindent \char60\relax{}candidate\char95\relax\allowbreak{}roi\char95\relax\allowbreak{}list\char62\relax{}\par
\vspace{0.45\baselineskip}
\noindent \char60\relax{}chat\char95\relax\allowbreak{}history\char95\relax\allowbreak{}override\char62\relax{}\par
\noindent Return ONLY JSON:\par
\noindent \char123\relax{}\par
\noindent \hspace*{1.0em}"broad\char95\relax\allowbreak{}type": "from Step 1: carcinoma/sarcoma/lymphoma/melanoma/glioma/germ cell tumor/neuroendocrine neoplasm/mesothelioma/other",\par
\noindent \hspace*{1.0em}"subtype": "from Step 2: specific entity, or '' if not classifiable",\par
\noindent \hspace*{1.0em}"malignancy": "benign/borderline/malignant",\par
\noindent \hspace*{1.0em}"grade": "histologic grade if assessable, else ''",\par
\noindent \hspace*{1.0em}"primary\char95\relax\allowbreak{}diagnosis": "the most specific defensible diagnosis",\par
\noindent \hspace*{1.0em}"confidence": "high/medium/low",\par
\noindent \hspace*{1.0em}"differential": ["alternative 1", "alternative 2"],\par
\noindent \hspace*{1.0em}"tumor\char95\relax\allowbreak{}architecture": "growth pattern description",\par
\noindent \hspace*{1.0em}"cytology": "nuclear and cellular features",\par
\noindent \hspace*{1.0em}"necrosis": "present/absent with description",\par
\noindent \hspace*{1.0em}"invasion": "invasive features if visible",\par
\noindent \hspace*{1.0em}"stroma\char95\relax\allowbreak{}inflammation": "stromal and inflammatory features",\par
\noindent \hspace*{1.0em}"ihc\char95\relax\allowbreak{}observed": "actual IHC/molecular if provided; else 'not assessable from H\&E images'",\par
\noindent \hspace*{1.0em}"ihc\char95\relax\allowbreak{}molecular": "same as ihc\char95\relax\allowbreak{}observed; do not place recommended markers here",\par
\noindent \hspace*{1.0em}"missing\char95\relax\allowbreak{}evidence": ["what image evidence is missing"],\par
\noindent \hspace*{1.0em}"recommended\char95\relax\allowbreak{}ancillary\char95\relax\allowbreak{}studies": ["recommended IHC/molecular/clinical studies"]\par
\noindent \char125\relax{}\par
\end{agentprompt}

\paragraph{Evidence sufficiency}
The Boolean reportability field controls the acquisition loop described in Methods.
\textcolor{black}{The model also returns self-reported confidence as high, medium or low.}
\begin{agentprompt}
\noindent /no\char95\relax\allowbreak{}think\par
\noindent Assess whether the current ROI evidence is sufficient for a confident pathology diagnosis.\par
\vspace{0.45\baselineskip}
\noindent Current diagnosis: \char60\relax{}primary\char95\relax\allowbreak{}diagnosis\char62\relax{} (site: \char60\relax{}known\char95\relax\allowbreak{}site\char62\relax{})\par
\vspace{0.45\baselineskip}
\noindent Evidence ROIs:\par
\noindent \char60\relax{}selected\char95\relax\allowbreak{}roi\char95\relax\allowbreak{}list\char62\relax{}\par
\vspace{0.45\baselineskip}
\noindent Evaluate:\par
\noindent 1. Is there enough tumor tissue to characterize the lesion?\par
\noindent 2. Are key diagnostic features (architecture, cytology, invasion, necrosis) visible?\par
\noindent 3. What image-observable evidence is missing?\par
\noindent 4. What non-image information (IHC, molecular, clinical) is needed?\par
\vspace{0.45\baselineskip}
\noindent Return ONLY JSON:\par
\noindent \char123\relax{}\par
\noindent \hspace*{1.0em}"sufficient\char95\relax\allowbreak{}for\char95\relax\allowbreak{}diagnosis": "Yes/Partially/No",\par
\noindent \hspace*{1.0em}"diagnostic\char95\relax\allowbreak{}confidence": "high/medium/low",\par
\noindent \hspace*{1.0em}"missing\char95\relax\allowbreak{}image\char95\relax\allowbreak{}evidence": ["..."],\par
\noindent \hspace*{1.0em}"missing\char95\relax\allowbreak{}nonimage\char95\relax\allowbreak{}metadata": ["..."],\par
\noindent \hspace*{1.0em}"recommended\char95\relax\allowbreak{}additional\char95\relax\allowbreak{}rois": ["type of ROI that would help"],\par
\noindent \hspace*{1.0em}"can\char95\relax\allowbreak{}confidently\char95\relax\allowbreak{}report": true/false\par
\noindent \char125\relax{}\par
\end{agentprompt}

\paragraph{Claim assessment}
The model receives current ROI images with the claims and their evidence graph. The returned
assessment is merged with the tissue-rule checks as described in
\hyperref[sec:methods-report-writing]{Methods, Diagnosis-constrained report generation}.
\begin{agentprompt}
\noindent /no\char95\relax\allowbreak{}think\par
\noindent Verify pathology claims against the provided evidence graph and ROI images.\par
\vspace{0.45\baselineskip}
\noindent Statuses:\par
\noindent - supported: directly supported by visual ROI evidence, morphology text, or segmentation stats.\par
\noindent - weak: plausible but not directly proven.\par
\noindent - unsupported: no available evidence supports the claim.\par
\noindent - contradicted: available evidence suggests the claim is false.\par
\noindent - not\char95\relax\allowbreak{}assessable: cannot be assessed from H\&E ROI evidence.\par
\vspace{0.45\baselineskip}
\noindent Rules:\par
\noindent - Do not use filenames, case IDs, TCGA codes, directory names, or coordinates.\par
\noindent - KEEP is background knowledge, not patient-specific evidence.\par
\noindent - IHC/molecular/stage/gross/margins are not assessable unless explicitly supplied.\par
\noindent - Cite evidence using ROI ids, evidence::morphology, evidence::knowledge, evidence::subtype\char95\relax\allowbreak{}sufficiency, or evidence::sufficiency.\par
\vspace{0.45\baselineskip}
\noindent Claims:\par
\noindent \char60\relax{}claims\char95\relax\allowbreak{}json\char62\relax{}\par
\vspace{0.45\baselineskip}
\noindent Evidence graph:\par
\noindent \char60\relax{}evidence\char95\relax\allowbreak{}graph\char95\relax\allowbreak{}json\char62\relax{}\par
\vspace{0.45\baselineskip}
\noindent Segmentation:\par
\noindent \char60\relax{}segmentation\char95\relax\allowbreak{}context\char95\relax\allowbreak{}json\char62\relax{}\par
\vspace{0.45\baselineskip}
\noindent Return ONLY JSON:\par
\noindent \char123\relax{}\par
\noindent \hspace*{1.0em}"claims": [\par
\noindent \hspace*{2.0em}\char123\relax{}"claim\char95\relax\allowbreak{}id": "C001", "status": "supported/weak/unsupported/contradicted/not\char95\relax\allowbreak{}assessable", "evidence": ["..."], "reason": "..."\char125\relax{}\par
\noindent \hspace*{1.0em}],\par
\noindent \hspace*{1.0em}"unsupported\char95\relax\allowbreak{}summary": ["..."],\par
\noindent \hspace*{1.0em}"missing\char95\relax\allowbreak{}evidence": ["..."]\par
\noindent \char125\relax{}\par
\end{agentprompt}

\paragraph{Structured report writing}
The writer receives the selected image inputs and the diagnostic state. Non-empty tier and
differential fields are inserted after the primary diagnosis; the optional staged-diagnosis
instruction explicitly repeats that diagnosis. The complete configuration additionally inserts
the diagnosis-lock and auxiliary-policy blocks shown below. A structurally invalid draft is
retried once with the same prompt and images.
\begin{agentprompt}
\noindent /no\char95\relax\allowbreak{}think\par
\noindent Generate a structured pathology report from the provided ROI evidence.\par
\vspace{0.45\baselineskip}
\noindent Specimen site: \char60\relax{}known\char95\relax\allowbreak{}site\char62\relax{}\par
\noindent Primary diagnosis: \char60\relax{}primary\char95\relax\allowbreak{}diagnosis\char62\relax{}\char60\relax{}staged\char95\relax\allowbreak{}diagnosis\char95\relax\allowbreak{}instruction\char62\relax{}\char60\relax{}diagnosis\char95\relax\allowbreak{}tiers\char62\relax{}\char60\relax{}differential\char95\relax\allowbreak{}diagnoses\char62\relax{}\par
\noindent \char60\relax{}diagnosis\char95\relax\allowbreak{}lock\char95\relax\allowbreak{}block\char62\relax{}\par
\vspace{0.45\baselineskip}
\noindent Evidence gaps identified:\par
\noindent - Missing image evidence: \char60\relax{}missing\char95\relax\allowbreak{}image\char95\relax\allowbreak{}evidence\char95\relax\allowbreak{}json\char62\relax{}\par
\noindent - Missing non-image metadata: \char60\relax{}missing\char95\relax\allowbreak{}nonimage\char95\relax\allowbreak{}metadata\char95\relax\allowbreak{}json\char62\relax{}\par
\noindent \char60\relax{}optional\char95\relax\allowbreak{}morphology\char95\relax\allowbreak{}block\char62\relax{}\char60\relax{}segmentation\char95\relax\allowbreak{}summary\char95\relax\allowbreak{}block\char62\relax{}\char60\relax{}auxiliary\char95\relax\allowbreak{}roi\char95\relax\allowbreak{}policy\char95\relax\allowbreak{}block\char62\relax{}\par
\vspace{0.45\baselineskip}
\noindent Rules:\par
\noindent 1. Only report what is visible in the ROI images or stated above.\par
\noindent 2. Do NOT fabricate IHC results, stage, margin status, or gross measurements.\par
\noindent 3. If IHC/molecular data were not explicitly provided, write actual IHC results as\par
\noindent \hspace*{1.5em}"Not assessable from H\&E images"; recommended stains may be listed separately.\par
\noindent 4. If the evidence sufficiency says the case cannot be confidently reported,\par
\noindent \hspace*{1.5em}do not automatically erase the best-supported primary diagnosis. Use the\par
\noindent \hspace*{1.5em}staged primary diagnosis above when it is supported by morphology,\par
\noindent \hspace*{1.5em}segmentation, subtype-sufficiency, or claim verification. Put uncertainty\par
\noindent \hspace*{1.5em}about subtype, malignancy, IHC, stage, margins, or grade in the appropriate\par
\noindent \hspace*{1.5em}sections rather than appending it to the Diagnosis section.\par
\noindent \hspace*{2.0em}5. The Diagnosis section must contain ONLY the primary diagnosis name. Put\par
\noindent \hspace*{3.5em}malignancy, grade, subtype uncertainty, and limitations in the Stage/Grade or\par
\noindent \hspace*{3.5em}Final Summary sections. Do not merge differential diagnoses into Diagnosis.\par
\noindent \hspace*{2.0em}6. Be specific about what IS visible \mbox{\normalfont\textemdash} use precise histomorphological terminology.\par
\noindent \hspace*{2.0em}7. Use segmentation evidence only as localized support for tissue composition,\par
\noindent \hspace*{3.5em}tumor-rich regions, necrosis, epithelial-labeled/stromal tissue classes, or\par
\noindent \hspace*{3.5em}ROI quality. Do not infer IHC, molecular status, margins, stage, clinical\par
\noindent \hspace*{3.5em}outcomes, exact subtype, grade, malignancy, or invasive-vs-in-situ status\par
\noindent \hspace*{3.5em}from segmentation.\par
\noindent \hspace*{2.0em}8. Do not amplify segmentation/reader evidence. If necrosis is described as\par
\noindent \hspace*{3.5em}focal, trace, limited, uncertain, or low-fraction, do NOT call it extensive,\par
\noindent \hspace*{3.5em}geographic, widespread, or high-grade necrosis.\par
\noindent \hspace*{2.0em}9. The Diagnosis section must stay aligned with the staged Primary diagnosis\par
\noindent \hspace*{3.5em}above. Do not replace it with a different entity because segmentation\par
\noindent \hspace*{3.5em}stats mention epithelial-labeled tissue, necrosis, tumor burden, or\par
\noindent \hspace*{3.5em}low/absent tissue classes.\par
\noindent \hspace*{2.0em}10. If a LOCKED FINAL DIAGNOSIS STATE is provided, copy its diagnosis,\par
\noindent \hspace*{4.0em}subtype, malignancy, grade, invasion, and necrosis conclusions into the\par
\noindent \hspace*{4.0em}appropriate report sections. Segmentation or ROI text may support or\par
\noindent \hspace*{4.0em}limit those fields but may not overwrite them.\par
\noindent \hspace*{2.0em}11. If auxiliary grade-focused ROIs are listed, treat them as secondary\par
\noindent \hspace*{4.0em}grading evidence only. They are not subtype-selection evidence.\par
\noindent \hspace*{2.0em}12. If a LOCKED REPORT GRADE FIELD is provided, the Stage/Grade section must\par
\noindent \hspace*{4.0em}verbalize that exact grade. Do not contradict it with "grade not\par
\noindent \hspace*{4.0em}assessable"; reserve not-assessable language for stage, margins,\par
\noindent \hspace*{4.0em}molecular/IHC, or other unavailable prognostic fields.\par
\vspace{0.45\baselineskip}
\noindent ROI evidence:\par
\noindent \char60\relax{}selected\char95\relax\allowbreak{}roi\char95\relax\allowbreak{}list\char62\relax{}\par
\vspace{0.45\baselineskip}
\noindent Return the report in this EXACT format (8 sections, each on a new line after the heading):\par
\vspace{0.45\baselineskip}
\noindent Specimen/Site:\par
\noindent Diagnosis:\par
\noindent Tumor/Lesion Description:\par
\noindent Key Microscopic Findings:\par
\noindent Invasion/Necrosis/Margins:\par
\noindent Immunohistochemistry/Ancillary Studies:\par
\noindent Stage/Grade/Prognostic Features:\par
\noindent Final Summary:\par
\end{agentprompt}

\Needspace{8\baselineskip}
\noindent\textbf{Diagnosis-lock block.}
The inserted JSON carries primary diagnosis, broad type, subtype, malignancy, grade, invasion
and necrosis from the diagnostic state supplied to the writer.
\begin{agentprompt}
\noindent LOCKED FINAL DIAGNOSIS STATE \mbox{\normalfont\textemdash} do not revise these fields. The report writer must verbalize this state, not perform a new diagnosis:\par
\noindent \char60\relax{}locked\char95\relax\allowbreak{}diagnosis\char95\relax\allowbreak{}json\char62\relax{}\par
\end{agentprompt}

\Needspace{8\baselineskip}
\noindent\textbf{Tissue-summary block.}
The compact JSON representation is truncated to 2,500 characters for writing.
\begin{agentprompt}
\noindent Compact segmentation evidence (localized ROI tissue support; do not use this alone as a cancer-type label):\par
\noindent \char60\relax{}compact\char95\relax\allowbreak{}segmentation\char95\relax\allowbreak{}json\char62\relax{}\par
\end{agentprompt}

\Needspace{10\baselineskip}
\noindent\textbf{Auxiliary-image role block.}
The inserted record identifies the auxiliary ROIs, their count and their writing-only role;
its serialized text is truncated to 1,800 characters.
\begin{agentprompt}
\noindent Auxiliary evidence role policy:\par
\noindent \char60\relax{}auxiliary\char95\relax\allowbreak{}roi\char95\relax\allowbreak{}policy\char95\relax\allowbreak{}json\char62\relax{}\par
\noindent Use auxiliary grade-focused ROIs only for grade, necrosis, cytologic atypia, mitotic activity, and ROI-level quality support. Do not use these auxiliary ROIs to change the primary diagnosis, broad lineage, histologic subtype, specimen site, or invasive-vs-in-situ status unless the main diagnostic ROI set and staged diagnosis already support that change.\par
\end{agentprompt}

\paragraph{Claim-guided report revision}
Revision is a text-only call. The report-heading placeholder expands to the same eight headings
in the writing template. A failed or structurally invalid revision leaves the draft unchanged;
the complete configuration then enforces the locked diagnostic fields in code.
\begin{agentprompt}
\noindent /no\char95\relax\allowbreak{}think\par
\noindent Rewrite this structured pathology report so every statement obeys claim verification.\par
\vspace{0.45\baselineskip}
\noindent Diagnosis payload:\par
\noindent \char60\relax{}diagnosis\char95\relax\allowbreak{}payload\char95\relax\allowbreak{}json\char62\relax{}\par
\vspace{0.45\baselineskip}
\noindent Evidence sufficiency:\par
\noindent \char60\relax{}sufficiency\char95\relax\allowbreak{}json\char62\relax{}\par
\vspace{0.45\baselineskip}
\noindent Claim verification:\par
\noindent \char60\relax{}claims\char95\relax\allowbreak{}json\char62\relax{}\par
\vspace{0.45\baselineskip}
\noindent Draft report:\par
\noindent ---\par
\noindent \char60\relax{}draft\char95\relax\allowbreak{}report\char62\relax{}\par
\noindent ---\par
\vspace{0.45\baselineskip}
\noindent Rules:\par
\noindent - Supported claims may be written affirmatively.\par
\noindent - Weak claims must be hedged as possible/suggestive.\par
\noindent - Unsupported or contradicted claims must be removed.\par
\noindent - Not-assessable claims must be written as "not assessable from H\&E ROI evidence".\par
\noindent - Do not downgrade malignancy to not-assessable when the diagnosis payload\par
\noindent \hspace*{1.0em}explicitly states benign, borderline, or malignant. In that case only grade,\par
\noindent \hspace*{1.0em}Weiss score, IHC, molecular, stage, margins, and other unavailable fields are\par
\noindent \hspace*{1.0em}not assessable.\par
\noindent - Do not invent IHC, molecular, stage, gross size, margin distance, or lymph-node count.\par
\noindent - The Diagnosis section must contain only the main diagnosis entity, not limitations,\par
\noindent \hspace*{1.0em}differential diagnoses, or qualifiers.\par
\noindent - Keep exactly these sections:\par
\noindent \char60\relax{}eight\char95\relax\allowbreak{}report\char95\relax\allowbreak{}headings\char62\relax{}\par
\vspace{0.45\baselineskip}
\noindent Return only the rewritten report text.\par
\end{agentprompt}

\supplementarymethodsubsection{ROI candidate ranking and initial selection}{supp:roi-ranking}
CONCH tile size, scoring, candidate pooling and initial VLM selection are specified in
\hyperref[sec:methods-report-acquisition]{Methods, ROI ranking and initial selection}.
Positive prompts describe malignant tumour, invasive carcinoma, cancer cells,
cellular atypia and high tumour cellularity, tumour necrosis, and invasion at a
tissue interface. Negative prompts describe benign tissue, blank glass,
adipose tissue without tumour, and poor-quality or out-of-focus tissue.
These diagnostic prompts are separate from site-conditioned KEEP queries.

\supplementarymethodsubsection{Domain-knowledge retrieval and diagnostic context}{supp:report-knowledge}
The query sequence and V-path routing are specified in
\hyperref[sec:methods-report-diagnosis]{Methods, Diagnostic candidates and selection}.
Disease and semantic-group entries are ranked by text-token overlap and site matching;
candidate subtypes are ranked by frequency in the matching semantic groups. Disease names
are matched by name, synonym and then token overlap to obtain definitions and child subtypes.
Morphology descriptions are taken from semantic-group captions, prioritizing specimen-site matches.
Retrieved associations are retained with diagnostic decisions.

\supplementarymethodsubsection{Multi-slide acquisition}{supp:report-acquisition}
\textcolor{black}{Candidate-selection rules, reportability checks and acquisition limits are illustrated in \cref{fig:report-model} and specified in \hyperref[sec:methods-report-reinspection]{Methods, \textcolor{black}{Evidence assessment and reinspection}}. The full public-case acquisition sequence appears in \cref{fig:report-case-success}.}

\supplementarymethodsubsection{Diagnosis-constrained writing and evidence associations}{supp:report-writing}
The diagnosis lock, report structure, retry rules and claim-assessment precedence are specified in
\hyperref[sec:methods-report-writing]{Methods, Diagnosis-constrained report generation}.

\noindent\textbf{Writer-only region selection.}
The auxiliary selector excludes the diagnostic ROIs and reuses a matching cached selection
when available. Otherwise it ranks the remaining candidates by their existing score plus
keyword weights from ROI metadata. Grade, mitotic activity, necrosis and atypia terms receive
bonuses, whereas background, artefact or normal-tissue terms incur a penalty. Selected regions retain their candidate images, coordinates, image levels and available
metadata, without an additional segmentation step. Duplicate exclusion uses ROI identifiers
or, when absent, coordinate-and-level keys; it does not require spatial non-overlap.

\noindent\textbf{Evidence associations.}
An associated input image does not necessarily locate the evidence for a particular phrase. A tissue-fraction
check can support tumour presence without identifying a
histological subtype or architecture. A final automated ``supported'' status can coexist with a weak or conflicting
segmentation subcheck, as illustrated in \cref{pan:case-summary-evidence-links}. The viewer retains
these distinct levels of information for inspection (\cref{supp:report-interface}).

\supplementarymethodsection{Report comparisons, scoring and statistical analysis}{supp:report-evaluation}
The comparisons use the inputs and references described in
\cref{supp:report-cohorts}. \textcolor{black}{Tissue-input and writing-policy comparisons use the same 1,200 public evaluation cases as the main analysis; inference failures count as incorrect. Each metric includes only cases with the required reference information.}

\textcolor{black}{PathXAI-Report pipeline development and debugging used only 456 public cases};
the \textcolor{black}{1,200-case public evaluation cohort} and private cohort were not used for \textcolor{black}{PathXAI-Report pipeline development}.
\textcolor{black}{Earlier evidence-organization and acquisition comparisons use \textcolor{black}{cases used for PathXAI-Report pipeline development}.}

\supplementarymethodsubsection{Comparator models and implementation}{supp:report-comparators}
The benchmark includes general-purpose VLM baselines (Qwen+H\&E, DeepSeek+H\&E and GLM+H\&E),
medically pretrained MedGemma, PLIP-guided Qwen (a single-pass adaptation of PathAgent), PRISM2 and the PathXAI
configurations (\cref{pan:report-public-systems,supp:report-configurations}).
\textcolor{black}{\Cref{tab:report-comparator-inputs} summarizes their input conditions.
Qwen+H\&E, DeepSeek+H\&E, GLM+H\&E, MedGemma, PLIP-guided Qwen and PRISM2
received known anatomical site, as did PathXAI-Report. These comparisons assess reporting
systems with shared access to anatomical context. HistGen and HistoGPT used their
released image-feature pipelines.}
\begin{supplementarytables}
\color{black}
\begin{longtable}{@{}>{\raggedright\arraybackslash}p{0.22\linewidth}>{\raggedright\arraybackslash}p{0.36\linewidth}>{\raggedright\arraybackslash}p{0.36\linewidth}@{}}
\caption{Inputs and output scope of the report comparators. Known site is supplied as anatomical context where indicated. Model settings and native-model decoding are described in \cref{supp:report-comparators,supp:report-native-comparison}.}\label{tab:report-comparator-inputs}\\
\toprule Method & Image evidence & Additional context and output \\ \midrule
\endfirsthead
\multicolumn{3}{@{}l}{\manuscripttablecontinued{}}\\[3pt]
\toprule Method & Image evidence & Additional context and output \\ \midrule
\endhead
\midrule\multicolumn{3}{r@{}}{Continued on next page}\\\endfoot
\bottomrule\endlastfoot
Qwen+H\&E; MedGemma & Selected H\&E ROIs. & Known site; generated report.\\[4pt]
DeepSeek+H\&E; GLM+H\&E & One to three fixed initial H\&E ROIs, shared across these models; maximum image edge 768 pixels. & Known site; diagnostic call followed by report generation.\\[4pt]
PLIP-guided Qwen & PLIP ranks and classifies three central 224-by-224-pixel crops from saved ROIs. & Crop descriptions and known site enter Qwen; single-pass report generation.\\[4pt]
PRISM2 & One selected WSI per case, preferentially a diagnostic slide; WSI features computed by the released model. & Known site; one diagnostic line, without grade or morphology reporting.\\[4pt]
HistGen & Features from all listed slides, concatenated per case. & No separate site prompt; native report decoder.\\[4pt]
HistoGPT & Features from all listed slides, concatenated per case. & ``Final diagnosis:'' prompt without a separate site field; native report generation without GPT refinement.\\[4pt]
PathXAI-Report (core) & H\&E and segmentation overlays; diagnostic regions can be revisited or expanded. & Known site and tissue summaries at the stages specified in Methods; generated report.\\[4pt]
PathXAI-Report & Core image evidence plus grade-focused H\&E regions supplied only to writing. & Known site, tissue summaries, KEEP context and diagnosis-constrained writing.\\[4pt]
\end{longtable}
\end{supplementarytables}

\paragraph{VLM deployment and image inputs}
The deployed models include Qwen3.6-27B~\cite{qwen2026} in Q6\_K GGUF format and
MedGemma-1.5-4B-it~\cite{sellergren2026medgemma} in BF16 GGUF format.
The DeepSeek+H\&E and GLM+H\&E baselines use the deployment identifiers
\texttt{deepseek-v4.1-flash} and \texttt{glm-5-3-flash}
(described by the providers as \href{https://api-docs.deepseek.com/news/news260910/}{DeepSeek-V4.1-Flash}
and \href{https://autoclaw.z.ai/blog/model/glm-5.3-flash/}{GLM-5.3-Flash}).
Qwen and MedGemma inference used researcher-operated, self-hosted services; the GLM+H\&E
baseline used the A*STAR A*RAC deployment. The public DeepSeek+H\&E and GLM+H\&E baselines use
one to three H\&E ROIs, converted to RGB JPEGs with maximum edge 768 pixels and quality 75.
A diagnostic call precedes report generation; temperature is 0.2, the output limit is 16,384
tokens, and each task permits up to three attempts. Exact converted checkpoints and immutable
remote-model revisions were not established for every evaluated report.

\paragraph{PLIP-guided Qwen}
PLIP-guided Qwen
uses PLIP image-text representations~\cite{huang2023plip}. Crops are ranked by cosine similarity
to a configured query, and top-ranked crops are classified against tissue and cell descriptions.
The implementation uses central 224-by-224-pixel crops from the saved level-0 ROIs and retains
three crops. Their descriptions and known site feed a single-pass Qwen report generator with
temperature 0.15 and a maximum of 1,024 generated tokens. This differs from
the original PathAgent's iterative WSI navigation, morphology perception and reasoning for visual
question answering~\cite{Chen2025PathAgentTI}.

\paragraph{PRISM2}
PRISM2~\cite{vorontsov2026prism2} contributes the
diagnostic outputs collected in this study. Its prompt requests one diagnostic line, excluding
grade, stage, differential diagnoses and explanatory text, with a maximum of 100 generated tokens.

\supplementarymethodsubsection{HistGen and HistoGPT comparison}{supp:report-native-comparison}
We additionally compared HistGen~\cite{guo2024histgen} and HistoGPT
reports with PathXAI-Report (core) and the complete configuration using the
same report-scoring protocol.
These comparisons are presented separately
because their training and input conditions differ from the main
benchmark.

\paragraph{Model inputs and decoding}
For each case, both implementations concatenate
features from all listed slides in their input order before generating one report.
HistGen uses 1,024-dimensional DINOv2-CPath features from non-overlapping 512-by-512-pixel
level-0 tissue tiles, with FP32 inference. Decoding uses beam size 3, maximum length 100,
region size 96 and 512 prototypes. HistoGPT uses fourfold slide downscaling and 256-pixel
patches, resized to 224-by-224 pixels for its CTransPath encoder. A local implementation of
its upstream sampler uses the prompt ``Final diagnosis:'', temperature 0.7, top-$k$ 40,
top-$p$ 0.95 and maximum length 256, without GPT refinement. Both native-model runners use
seed 42.

\paragraph{Training-overlap exclusion and shared public cases}
HistGen patient splits are available from the
\href{https://huggingface.co/datasets/david4real/HistGen}{authors' data release}.
To limit potential training overlap, we
excluded every public case matching a patient in the training or validation arrays of
either \texttt{annotation.json} or \texttt{annotation712\_update.json}.
Membership was determined from the top-level arrays; test-set membership alone did not exclude a case.
The exact split used to
train the released checkpoint is not independently established. Of \textcolor{black}{1,656} public
cases, \textcolor{black}{1,340} were excluded by this rule and 316 remained eligible; 314 had reports available for HistGen and both
PathXAI configurations, and 312 had reports from all four methods, including HistoGPT. HistoGPT was developed for dermatopathology, so
comparisons on non-skin cases are out of domain.

\paragraph{Private-case comparison}
Private reports were analysed
separately against each pathologist's reference. The primary comparison comprised
151 cases shared by all reporting methods. A sensitivity analysis used all available reports
from the 152-case cohort. Scoring and resampling follow \cref{supp:report-scoring,supp:report-statistics}.

\supplementarymethodsubsection{Four-policy input comparison}{supp:report-input-comparisons}
\Cref{pan:private-summary-policy-comparisons,pan:private-summary-report-content,pan:private-summary-compatibility-transitions} and \cref{fig:report-input} compare H\&E alone, H\&E with tissue statistics, H\&E with segmentation
overlays, and PathXAI-Report on \textcolor{black}{the same 1,200 public evaluation cases used in the main comparison. The writing-policy comparison uses these cases.} Reports and references are matched by case and evaluated with the same metric-specific
weighting rules (\cref{supp:report-statistics}). \textcolor{black}{A \textcolor{black}{PathXAI-Report pipeline development subset} sensitivity analysis of tissue-statistics availability is retained in \cref{tab:tissue-context-sensitivity}.} The
three direct-input policies use known-site context without knowledge retrieval or reinspection.
PathXAI-Report adds knowledge-guided reasoning, evidence-driven reinspection and constrained writing
to segmentation evidence.

The two input contrasts are tissue statistics minus H\&E and segmentation overlays minus H\&E.
The two workflow contrasts are PathXAI-Report minus tissue statistics and PathXAI-Report minus
segmentation overlays.
The four transition counts in \cref{pan:private-summary-compatibility-transitions} describe paired cancer family match rate in the same shared public cases. They measure
changes in diagnosis compatibility, not whole-report correctness.

\supplementarymethodsubsection{Reporting-policy comparisons}{supp:report-policy-comparisons}
Final report texts from eight configurations were assessed with the same scoring rules
used for the public-cohort analysis (\cref{pan:report-ablation-acquisition,pan:report-ablation-writing-constraints}; \cref{fig:report-ablation}). Three
sets of comparisons address evidence organization, active evidence acquisition and writing
\textcolor{black}{constraints. Evidence-organization and acquisition comparisons use 42 and 432 paired cases, respectively, \textcolor{black}{from the 456-case PathXAI-Report pipeline development subset}; writing constraints are evaluated on the 1,200 public evaluation cases.}
\textcolor{black}{After source-data exclusions, the archived evidence-organization comparison
contained 45 cases with saved outputs from all three configurations. Three cases had
dry-run placeholder text in at least one configuration, leaving 42 cases with generated
reports. The acquisition comparison similarly retained 432 of 456 cases after excluding
24 with at least one placeholder.}
For each metric, applicable cases are determined by the available reference fields.

\textcolor{black}{\Cref{tab:report-component-comparisons} links each component comparison to its
changed inputs or rules and corresponding results. The workflow contrasts are also shown in
\cref{pan:report-ablation-policy-comparisons}. Comparisons that change several inputs or rules evaluate those changes together.} Writer-only
grade ROIs provide additional H\&E regions for writing without entering the
diagnostic branches. The final writing comparison also changes knowledge routing, \textcolor{black}{invasion-state
selection safeguards} and diagnosis-constrained writing. Supplying these additional images differs from supplying tissue
statistics only during writing.

Positive tissue-statistics input supplies the foreground fraction and retains neoplastic
fractions of at least 5\% and dead/necrotic fractions of at least 10\% for up to eight ROIs.
These fractions use all ROI mask pixels as the denominator. Other tissue classes and lower
fractions are omitted from the supplied evidence. The prompt limits their use to
tumour amount, substantial necrosis and ROI usability, without inferring subtype, grade or invasion.
Conflict-triggered expansion uses at most two acquisition rounds and rolls back diagnoses that
become generic or change tumour lineage; these rules apply throughout this comparison.
\begin{supplementarytables}
\color{black}
\begin{longtable}{@{}>{\raggedright\arraybackslash}p{0.16\linewidth}>{\raggedright\arraybackslash}p{0.27\linewidth}>{\raggedright\arraybackslash}p{0.34\linewidth}>{\raggedright\arraybackslash}p{0.14\linewidth}@{}}
\caption{Report-component comparisons, grouped by cohort. The public evaluation group uses the same 1,200 cases as \cref{fig:public-summary}. The development group uses paired subsets of the separate 456 cases used for PathXAI-Report pipeline development. Its smaller case counts reflect the coverage of generated reports across the compared configurations after excluding dry-run placeholders (\cref{supp:report-policy-comparisons}). All configurations receive known-site context. The cases column gives the paired comparison set; each metric's applicable count further depends on the available reference fields. Each contrast is the first configuration minus the second.}\label{tab:report-component-comparisons}\\
\toprule Component & Compared configurations & Changed inputs or rules & Cases and results \\ \midrule
\endfirsthead
\multicolumn{4}{@{}l}{\manuscripttablecontinued{}}\\[3pt]
\toprule Component & Compared configurations & Changed inputs or rules & Cases and results \\ \midrule
\endhead
\midrule\multicolumn{4}{r@{}}{Continued on next page}\\\endfoot
\bottomrule\endlastfoot
\multicolumn{4}{@{}l}{\textbf{Public evaluation cohort (1,200 cases)}}\\[4pt]
Direct tissue input & H\&E+tissue statistics versus H\&E alone; H\&E+overlays versus H\&E alone & Supplies numeric tissue summaries or segmentation overlays for direct reporting, without KEEP retrieval or reinspection. & \textcolor{black}{1,200}; \cref{pan:private-summary-policy-comparisons}\\[5pt]
Complete workflow & PathXAI-Report versus direct tissue-statistics input; PathXAI-Report versus direct overlay input & Adds diagnostic review, reinspection and constrained writing to tissue evidence. These are comparisons of complete workflows. & \textcolor{black}{1,200}; \cref{pan:private-summary-policy-comparisons,pan:private-summary-report-content}\\[5pt]
Grade-focused images & Report-only grade evidence versus PathXAI-Report (core) & Adds auxiliary H\&E regions to writing only; diagnostic candidates do not receive these regions. & \textcolor{black}{1,200}; \cref{pan:report-ablation-writing-constraints}\\[5pt]
Knowledge and diagnostic constraints & PathXAI-Report versus report-only grade evidence & Adds KEEP routing, invasion-state selection safeguards and diagnosis-locked writing. The comparison evaluates these changes together, including any resulting changes in acquired evidence. & \textcolor{black}{1,200}; \cref{pan:report-ablation-writing-constraints}\\[5pt]
\midrule
\multicolumn{4}{@{}l}{\textbf{PathXAI-Report pipeline development subset (456 cases)}}\\[4pt]
Evidence placement & Tissue statistics at writing only versus positive tissue statistics & Restricts tissue summaries to writing and removes the reference configuration's positive-only filter. Both placement and filtering change. & 42; \cref{pan:report-ablation-evidence-organization}\\[5pt]
Candidate selection & Guarded dual-path reasoning versus positive tissue statistics & Adds two diagnostic candidates and guarded selection; also removes the positive-only tissue-summary filter. & 42; \cref{pan:report-ablation-evidence-organization}\\[5pt]
Additional regions & Conflict-triggered expansion versus guarded dual-path reasoning & Adds up to two acquisition rounds when candidates conflict, with rollback of generic or changed-lineage diagnoses. & 432; \cref{pan:report-ablation-acquisition}\\[5pt]
Zoom-enabled review & Zoom-enabled reinspection versus conflict-triggered expansion & Uses the reinspection policy with zooms and inherited tissue statistics. & 432; \cref{pan:report-ablation-acquisition}\\[5pt]
\end{longtable}
\end{supplementarytables}

\FloatBarrier

\supplementarymethodsubsection{Evaluation measures and terminology}{supp:report-metric-overview}
\Cref{tab:report-metric-overview} provides a guide to the primary and supporting measures, their preferred directions
and their definitions. Automated reference agreement and pathologist ranking and correction assessment
are distinct outcomes.
\begin{table}[!htbp]
\centering
\caption{What the report evaluation measures. The five measures in \cref{fig:public-summary} appear first, in the same order. Arrows indicate the preferred direction.
Automated measures use only cases with applicable reference information; private
references are evaluated separately. Methods and Supplementary Methods provide
extraction rules, formulas, denominators and uncertainty estimates.}
\label{tab:report-metric-overview}
\manuscripttablefont
\setlength{\tabcolsep}{4pt}
\renewcommand{\arraystretch}{1.03}
\begin{tabular}{@{}>{\raggedright\arraybackslash}p{0.28\linewidth}>{\raggedright\arraybackslash}p{0.58\linewidth}>{\raggedright\arraybackslash}p{\dimexpr0.14\linewidth-4\tabcolsep\relax}@{}}
\toprule
Measure & What it asks & Details \\
\midrule
\multicolumn{3}{@{}l}{\textbf{Five measures shown in the automated report evaluation figure}}\\
\addlinespace[3pt]
Cancer family match rate $\uparrow$ &
What fraction of diagnoses belong to the target cancer family? Public targets use TCGA labels; private targets use each pathologist's definite cancer diagnosis. Benign or unresolved private references are inapplicable. &
\hyperref[sec:methods-report-metrics]{Methods}\\
\addlinespace[3pt]
Exact diagnosis match rate $\uparrow$ &
What fraction of diagnoses match the specific entity or subtype in the case's reference report? Synonyms count as matches; related subtypes do not. A definite, specific reference diagnosis is required. &
\hyperref[sec:methods-report-metrics]{Methods}\\
\addlinespace[3pt]
Finding F1 $\uparrow$ &
How well does the report balance the correctness of its finding statements and recovery of documented reference findings? It combines precision and recall; matching presence and matching absence both count. Undocumented finding types are not scored. &
\hyperref[sec:methods-report-metrics]{Methods}\\
\addlinespace[3pt]
Finding recall $\uparrow$ &
What fraction of documented reference finding types is correctly reproduced as present or absent? Missing, uncertain, conflicting and opposite statements receive no credit. &
\hyperref[sec:methods-report-metrics]{Methods}\\
\addlinespace[3pt]
Finding contradiction rate $\downarrow$ &
Among documented finding types that the report definitely calls present or absent, what fraction disagrees with the reference? Undefined when none is addressed; interpreted alongside coverage because omissions are not contradictions. &
\hyperref[sec:methods-report-metrics]{Methods}\\
\midrule
\multicolumn{3}{@{}l}{\textbf{Additional diagnostic, grade and coverage measures}}\\
\addlinespace[3pt]
Broad diagnostic agreement; malignancy agreement $\uparrow$ &
Does the report match the reference's broad diagnostic category and malignancy status, respectively? &
\hyperref[supp:report-scoring]{SI Methods}\\
\addlinespace[3pt]
Diagnostic composite $\uparrow$ &
What is the weighted agreement across broad diagnosis, specific diagnosis and malignancy? Weights are renormalized over applicable fields; grade is evaluated separately. In diagnostic composite (cohort target), TCGA cancer family match rate replaces the specific diagnosis component. &
\hyperref[supp:report-scoring]{SI Methods}\\
\addlinespace[3pt]
Balanced grade accuracy $\uparrow$ &
How often does the grade match the reference when each represented grade is weighted equally? Correct fractions are calculated within grades and then averaged. Grading systems are evaluated separately. &
\hyperref[sec:methods-report-findings]{Methods}\\
\addlinespace[3pt]
Compatible grade coverage $\uparrow$ &
Does the report supply a valid grade in the reference's grading system, irrespective of whether its value is correct? &
\hyperref[supp:report-grade-findings]{SI Methods}\\
\addlinespace[3pt]
Invasion agreement; necrosis agreement $\uparrow$ &
Does the report agree with the reference on whether invasion or necrosis is present or absent? &
\hyperref[supp:report-omissions]{SI Methods}\\
\addlinespace[3pt]
Critical-positive recall $\uparrow$ &
What fraction of critical findings documented as present is also reported as present? &
\hyperref[supp:report-omissions]{SI Methods}\\
\addlinespace[3pt]
Report field completeness $\uparrow$ &
Which of the applicable broad diagnosis, specific diagnosis, malignancy, grade, invasion and necrosis fields have definite answers? Incorrect answers also count towards coverage. &
\hyperref[sec:methods-report-findings]{Methods}\\
\addlinespace[3pt]
Finding coverage $\uparrow$ &
What fraction of documented reference finding types does the report definitely address as present or absent, whether correct or incorrect? &
\hyperref[sec:methods-report-metrics]{Methods}\\
\addlinespace[3pt]
ROUGE-1 F1; ROUGE-2 F1 $\uparrow$ &
\textcolor{black}{How much unigram and bigram wording overlaps the narrative reference?} &
\hyperref[supp:report-wording]{SI Methods}\\
\midrule
\multicolumn{3}{@{}l}{\textbf{Pathologist ranking and correction assessment and supporting checks}}\\
\addlinespace[3pt]
Mean overall rank $\downarrow$ &
Which complete report does each pathologist prefer? Rank 1 is best; ties are retained. &
\hyperref[sec:methods-report-ranking]{Methods}\\
\addlinespace[3pt]
Correction ratings &
Does the report require no, minor or major correction, or is it unusable? We report category proportions, no/minor $\uparrow$ and major/unusable $\downarrow$. &
\hyperref[sec:methods-report-ranking]{Methods}\\
\addlinespace[3pt]
Concern flags $\downarrow$ &
Did the pathologist flag an unsupported major claim or a major visible omission? These are reader-selected concerns. &
\hyperref[sec:methods-report-ranking]{Methods}\\
\addlinespace[3pt]
No/minor correction with complete fields $\uparrow$; joint failure $\downarrow$ &
Does the report need no or only minor correction, pass the output check and give definite answers for all applicable fields? Failing any requirement counts as joint failure. The original ratings remain unchanged. &
\hyperref[sec:methods-report-checks]{Methods}\\
\bottomrule
\end{tabular}
\end{table}
\FloatBarrier
\Needspace{14\baselineskip}
\paragraph{Study terminology and operational definitions}
\phantomsection\label{supp:terminology}

\begingroup
\manuscripttablefont
\renewcommand{\arraystretch}{1.0}
\setlength{\extrarowheight}{0pt}
\setlength{\tabcolsep}{0pt}
\setlength{\LTcapwidth}{\linewidth}
\begin{longtable}{@{}>{\raggedright\arraybackslash\bfseries}p{0.30\linewidth}@{\hspace{6mm}}>{\raggedright\arraybackslash}p{\dimexpr0.70\linewidth-6mm\relax}@{}}
\caption{Study terminology and operational definitions.}\label{tab:glossary}\\
\toprule
\textbf{Term} & \textbf{Definition} \\
\midrule
\endfirsthead
\multicolumn{2}{@{}l}{\manuscripttablecontinued{}}\\[3pt]
\toprule
\textbf{Term} & \textbf{Definition} \\
\midrule
\endhead
\midrule
\multicolumn{2}{r@{}}{Continued on next page}\\
\endfoot
\bottomrule
\endlastfoot
\rowcolor{black!7}\multicolumn{2}{@{}l}{\strut\textbf{System and modules}}\\*[2pt]
PathXAI & The integrated study system, comprising PathXAI-Seg and PathXAI-Report. \\[2pt]
Structured report with tissue evidence & A structured pathology report with diagnostic and descriptive sections and recorded associations to slide, ROI, tissue and contextual evidence. \\[2pt]
PathXAI-Seg & The tissue-segmentation module. It produces predicted tissue label maps, aligned segmentation overlays, and segmentation-derived tissue statistics. \\[2pt]
PathXAI-Report & The complete report-generation module, including evidence acquisition, diagnostic reasoning, and structured report writing. \\[2pt]
PathXAI-Report (core) & The report-module comparator with the core dual-path workflow. Relative to the complete configuration, it lacks KEEP retrieval and auxiliary grade-focused writer inputs and uses a different candidate-selection policy without diagnosis locking. \\[2pt]
\addlinespace
\rowcolor{black!7}\multicolumn{2}{@{}l}{\strut\textbf{Segmentation outputs and evidence}}\\*[2pt]
Predicted tissue label map & A categorical mask assigning a predicted tissue class to each pixel in a region of interest (ROI). \\[2pt]
Segmentation overlay & The visual output formed by registering and displaying a predicted tissue label map over its corresponding H\&E image. \\[2pt]
Tissue statistics & Per-ROI summary of tissue fractions and the separately reported foreground fraction, calculated from the predicted tissue label map. \\[2pt]
Tissue fractions (\%) & For each non-background class, 100 times the predicted class-pixel count divided by all ROI mask pixels. Fractions are not renormalized over foreground. \\[2pt]
Foreground fraction (\%) & 100 times the non-background pixel count divided by all ROI mask pixels; reported separately from tissue fractions. \\[2pt]
Tissue evidence & Segmentation-derived tissue label maps, aligned overlays and tissue statistics used to support diagnostic reasoning and reporting. \\[2pt]
\addlinespace
\rowcolor{black!7}\multicolumn{2}{@{}l}{\strut\textbf{Image regions and acquisition}}\\*[2pt]
Region of interest (ROI) & A selected image region used in report generation. Diagnostic ROIs can include aligned segmentation evidence; grade-focused writer ROIs supply H\&E only and do not enter diagnostic reasoning. \\[2pt]
Evidence acquisition / reinspection & Acquisition comprises initial ROI selection and reinspection. Reinspection digitally enlarges a current ROI or acquires a new region. Digital zoom retains the parent relationship without increasing optical sampling; new diagnostic ROIs receive region-specific segmentation. \\[2pt]
Whole-slide image (WSI) & A digitized histology slide; model evidence is selected from regions of these slides. \\[2pt]
Haematoxylin and eosin (H\&E) & The histological stain used for the image inputs and visual evidence. \\[2pt]
\addlinespace
\rowcolor{black!7}\multicolumn{2}{@{}l}{\strut\textbf{Evaluation cohorts and measures}}\\*[2pt]
Public report evaluation & Cancer family match rate uses TCGA disease-family targets. Exact diagnosis match rate uses patient-specific reports, separately for \textcolor{black}{the \textcolor{black}{PathXAI-Report pipeline development subset} and public evaluation cohort}. Finding and grade measures use their applicable references; ordinary scores weight cases equally. Reporting cases did not update model weights; the 456 \textcolor{black}{\textcolor{black}{cases used for PathXAI-Report pipeline development}} were used for \textcolor{black}{PathXAI-Report pipeline development}. \\[2pt]
Private report evaluation & Private cases evaluated separately against each pathologist's reference set for exact diagnosis match rate, using canonicalized subtype when available and otherwise primary diagnosis, including benign entities. Applicable cases are weighted equally for each reference set. \\[2pt]
Balanced grade accuracy & Within each named grading system, exact grade recall averages cases equally within each reference grade, then weights the represented grades equally. Systems are reported separately; incompatible or missing predictions score as incorrect when the grade reference is applicable. \\[2pt]
Report field completeness & Fraction of applicable fields with definite answers for broad diagnosis, specific diagnosis, malignancy, grade, invasion and necrosis. Incorrect answers and grades from incompatible systems also count towards coverage. \\[2pt]
Compatible grade coverage & Fraction of applicable cases with a valid grade in the reference grading system, whether or not it agrees with the reference. \\[2pt]
Reference entity availability & \textcolor{black}{Fraction of references containing a definite, specific diagnostic entity.} \\[2pt]
Finding coverage & The proportion of documented binary reference findings with a comparable, explicit present/absent assertion; uncertainty, conflict and omission are not assertions. \\[2pt]
Finding recall & The proportion of documented binary reference findings with an explicit assertion agreeing in polarity. Unrecorded findings do not count as negative references. \\[2pt]
Finding contradiction rate & The proportion of comparable explicit binary assertions with polarity opposite to the reference (opposite/asserted, E/A). It is undefined when no comparable assertion is made and is reported with finding coverage and finding recall because it is conditional on output. \\[2pt]
Finding F1 & Harmonic mean of reference-conditioned precision $C/A$ and recall $C/D$, where $D$ counts documented findings, $A$ comparable assertions of those finding types and $C$ polarity agreements. The direct formula $2C/(D+A)$ gives zero when $A=0$, $D>0$. Unrecorded finding types are outside this measure. \\[2pt]
Segmentation internal / external evaluation & Internal tests use held-out samples from the training datasets. The \textcolor{black}{six external test sets} are excluded from study training; RINGS tumour and gland segmentation share images. Neither test set establishes patient disjointness or absence from foundation-model pretraining. \\[2pt]
Pathologist reference-report authoring / ranking and correction assessment & Each pathologist authors one reference report per private case, forming reference sets P1 and P2. Ranking and correction assessment compares generated public-case reports by overall preference and correction requirements. Weixiang Zhong and Xiaoxi Wang performed both tasks. \\*[2pt]
Diagnostic composite & The study-defined normalized weighted agreement across broad diagnosis, specific diagnosis and malignancy. Diagnostic composite (cohort target) substitutes TCGA cancer family match rate for the specific diagnosis component. Both are distinct from cancer family match rate alone. \\[2pt]
\end{longtable}
\endgroup
\supplementarymethodsubsection{Diagnostic extraction, targets and additional measures}{supp:report-scoring}
The common extraction pipeline is described in
\hyperref[sec:methods-report-extraction]{Methods, Report text extraction}; the primary measures are defined in
\hyperref[sec:methods-report-metrics]{Primary evaluation measures}.
The lexicon includes full names and abbreviations. For example, ``conventional renal cell
carcinoma'', ``clear cell RCC'' and ``ccRCC'' map to clear cell renal cell carcinoma;
``invasive carcinoma of no special type'' and ``IDC'' map to invasive ductal carcinoma.
\textcolor{black}{The study-specific terminology, organ-context and TCGA-family tables, matching rules and executable implementation are available in the
\href{https://github.com/zhaorui-tan/PathXAI/tree/main/evaluation}{code repository} (Code availability).}

\paragraph{Terminology and text normalization}
Entity matching lowercases text, retains letters a--z and digits, and replaces other characters with spaces.
\textcolor{black}{Assertion status is evaluated for each disease mention using the punctuation-preserving
text. Semicolons and contrasting clauses delimit local negation and uncertainty; differential-list
qualifiers apply to subsequent alternatives until a new diagnostic assertion. Thus, ``cannot exclude''
is uncertain, and a negative morphological observation does not negate a separately stated diagnosis.
Uncertainty in a separately named grade, stage or finding field does not invalidate an affirmed entity.
Recognized diagnostic terms split across lines are rejoined. Limited spelling tolerance applies to
section headings, without correcting disease words. Associated ductal carcinoma in situ can be
retained as a component of an invasive breast diagnosis; unrelated affirmed entities remain conflicting.}

\paragraph{Reference targets and applicability}
Exact diagnosis match rate requires a definite, specific diagnosis in the patient's reference report. Private
references use the structured subtype field, falling back to primary diagnosis only when the subtype
field is empty; an explicitly unknown subtype is not replaced. Public targets for this measure come
from the case report. TCGA cancer family match rate instead tests membership in the project's
accepted disease family, allowing multiple entities for heterogeneous projects. The cohort label
does not supply a missing patient-level subtype. Public broad-diagnosis and malignancy targets come
from TCGA, while findings and grade come from the case report.

\textcolor{black}{Among cases with an applicable reference diagnosis, we partitioned
exact-diagnosis outcomes into matched diagnoses,
definite non-specific diagnoses, definite different specific diagnoses and cases without
a definite recognized diagnosis. The last group includes uncertain, conflicting,
organ-incompatible, unrecognized and negated extractions. We separately counted definite
specific predictions in cases without an applicable reference diagnosis (Source Data,
\texttt{diagnostic\_outcomes}). These extraction categories require case review to distinguish
correct broader diagnoses, abstentions and clinically incorrect diagnoses.}

{\color{black}
\textbf{Constant-diagnosis control.}
To quantify performance achievable with one fixed diagnosis, we considered the unique
accepted entity names in the cancer-family table and selected the one maximizing family
matches on the 456-case \textcolor{black}{PathXAI-Report pipeline development subset}, resolving ties alphabetically.
This selected squamous cell carcinoma. We then applied this diagnosis to every public
evaluation case using the unchanged family-matching rules, without images or anatomical-site
inputs. Paired differences compare each reporting method with this fixed control on the
same cases, using 4,000 case-bootstrap draws conditional on the selected diagnosis
(\cref{tab:fixed-diagnosis-control}).
}

Private cancer family match rate derives its target from each pathologist's frozen diagnosis
extraction. An asserted reference must satisfy one of the existing invasive-breast carcinoma,
prostate adenocarcinoma or colorectal adenocarcinoma family definitions. The same membership
function then scores the generated diagnosis, including the rule that an in-situ diagnosis alone
does not satisfy an invasive-breast target. Anatomical site alone does not establish a target.
Benign, missing, uncertain or insufficiently specific references are inapplicable; missing,
uncertain or incompatible generated diagnoses score zero on applicable references.
The 151 shared private cases yield 121 applicable references for P1 and 123 for P2.
The family definitions and extracted diagnoses are unchanged; this endpoint applies the family
rules to the diagnoses in each pathologist's reference report.

\paragraph{Additional diagnostic measures}
\textbf{Broad diagnostic agreement.}
This is the percentage of applicable cases in which the generated diagnosis matches the
reference's broad diagnostic category. Public targets come from TCGA labels, and private
targets come from each pathologist's report. We divide the number of category matches by the
number of cases with a definite reference category and multiply by 100. Missing, uncertain,
conflicting or mismatched predictions receive no credit; cases without a definite reference
category are excluded.

\textbf{Malignancy agreement.}
This is the percentage of applicable cases in which the generated report matches the reference's
malignancy status, classified as benign, borderline or malignant. Public targets come from TCGA
labels, and private targets come from each pathologist's report. We divide the number of status
matches by the number of cases with a definite reference status and multiply by 100. Missing,
uncertain, conflicting or mismatched predictions receive no credit; cases without a definite
reference status are excluded.

\textbf{Diagnostic composite.}
This combines broad diagnostic agreement, exact diagnosis match rate and malignancy
agreement into one case-level score, with weights of 0.35, 0.35 and 0.15, respectively.
Weights are renormalized over reference-applicable fields. For case $i$,
\begin{equation}
S_i=\frac{\sum_{k\in K}w_k a_{ik}s_{ik}}{\sum_{k\in K}w_k a_{ik}},
\qquad K=\{\text{broad diagnosis, specific diagnosis, malignancy}\},
\end{equation}
where $a_{ik}$ indicates reference applicability and $s_{ik}$ is categorical agreement.
A zero denominator makes the case inapplicable. We average applicable case scores and multiply
by 100.
\textcolor{black}{When all three fields are applicable, the normalized weights are
$7/17$, $7/17$ and $3/17$, giving the two diagnostic levels equal weight and each
more weight than malignancy. A sensitivity analysis assigns equal weight to each
applicable component, retaining the same fields, case denominators and paired
bootstrap draws (\cref{tab:diagnostic-weight-sensitivity}).}

\textbf{Diagnostic composite (cohort target).}
This variant substitutes TCGA cancer family match rate for exact diagnosis match rate
in the diagnostic composite. It uses the same component weights, renormalization and
case-average calculation, and requires an applicable TCGA cancer-family target.

\paragraph{Sensitivity to diagnostic-section scope}
An exploratory sensitivity analysis adds text under recognized Final Summary or Conclusion
headings to diagnosis text before calculating exact diagnosis match rate and TCGA cancer family match rate (\crefrange{tab:report-scope-sensitivity:part-1}{tab:report-scope-sensitivity:part-2} and \crefrange{tab:cohort-scope-sensitivity:part-1}{tab:cohort-scope-sensitivity:part-2}). It retains the same terminology, polarity rules,
explicit subtype fields and reference targets. Reports without a recognized diagnosis section
retain the primary fallback rule. The section parser is unchanged; an unrecognized heading
does not end the preceding section. We analyse \textcolor{black}{the \textcolor{black}{PathXAI-Report pipeline development subset} and public evaluation cohort}
separately, retain the 151 shared private cases and compare reporting policies on the \textcolor{black}{456-case \textcolor{black}{PathXAI-Report pipeline development subset}}. Paired changes use 4,000 shared case-bootstrap draws, restricted to applicable
references for exact diagnosis match rate. Cancer family match rate uses all cases in
each public cohort and the same disease-family membership rules, including the breast
in-situ and lower-grade glioma overrides. Its pooled and source-specific analyses share
cases; the reporting-policy set is also a subset of the public cohort.

\supplementarymethodsubsection{Grade and structured-field coverage}{supp:report-grade-findings}
Balanced grade accuracy and report field completeness are defined in
\hyperref[sec:methods-report-findings]{Methods, Additional grade and coverage measures}.
The following grade measures use the same grade-extraction and reference-applicability rules.
Diagnosis-only PRISM2 outputs have inapplicable content and grade metrics.

\textbf{Case-average exact grade agreement.}
This is the percentage of applicable cases whose predicted grade exactly matches the reference
grade. It uses the same grade-extraction and applicability rules as balanced grade
accuracy, but weights cases equally. We divide the number of exact matches by the number of
cases with an assessable reference grade and multiply by 100; missing or incompatible
predictions contribute zero.

\textbf{Compatible grade coverage.}
This measures whether a report supplies a valid grade from the reference grading system,
regardless of whether the value matches. Each case with an assessable reference grade receives
one if the generated grade is valid within that system and zero otherwise. Cases without an
assessable reference grade are excluded. The reported percentage is 100 times the mean of
these case-level indicators. A grade from another system can cover the structured field while
failing compatible-grade coverage and exact agreement.

{\color{black}
\textbf{Conditional grade accuracy.}
This is the percentage of exact grade matches among reports supplying a definite grade
in the reference grading system. Its denominator is the number of covered cases defined
above; it is undefined when no case is covered. We report it alongside compatible grade
coverage and balanced grade accuracy, separately for each grading system. Methods can
answer different subsets of cases, so conditional accuracy alone does not provide a
paired comparison. An always-Grade-2 control is additionally evaluated within the
three-tier histological grading system.
}

{\color{black}
\textbf{Linear weighted grade $\kappa$.}
This measures ordinal agreement beyond that expected from the reference and predicted
grade distributions, using only definite predictions in the reference grading system.
With disagreement weights $w_{jk}=|j-k|/(g_{\max}-g_{\min})$ and the permitted grade limits
$g_{\min}$ and $g_{\max}$ defined below, we calculate
$\kappa_w=1-\sum_{jk}w_{jk}p_{jk}/\sum_{jk}w_{jk}p_{j\cdot}p_{\cdot k}$,
where $p_{jk}$ is the proportion of answered cases with reference grade $j$ and predicted
grade $k$, and the marginal proportions define expected disagreement.
Grading systems are analysed separately. We use 4,000 bootstrap draws of answered case
pairs and report percentile intervals over draws with defined $\kappa_w$; the statistic is
undefined with no compatible answers or zero expected disagreement. Source Data give the
number of defined draws and results by cancer type or organ. Answering subsets differ
between methods, so we report $\kappa_w$ alongside coverage and overall accuracy.
}

\textbf{Ordinal grade agreement.}
This gives partial credit according to how close the predicted grade is to the reference grade.
For valid, compatible grades, the case score is
$1-|\widehat g_i-g_i|/(g_{\max}-g_{\min})$; a missing or incompatible prediction scores zero.
Here $g_{\min}$ and $g_{\max}$ are the fixed permitted limits of the reference grading system,
\textcolor{black}{namely 1--3 for histological or three-tier nuclear grade; 1--4 for four-tier nuclear,
CNS WHO, Fuhrman or renal WHO/ISUP grade; 1--5 for prostate ISUP grade group;
and 2--10 for Gleason score.} Within each grading system, we average scores across
cases with assessable reference grades and multiply by 100.

\textcolor{black}{The same extraction rules are applied to references and generated reports.
Anatomical context distinguishes renal WHO/ISUP grades from prostate ISUP grade groups
and excludes grades attributed to a different primary lesion. Named systems are not
converted into one another. Numeric or descriptive ranges remain uncertain, and
unassessable or conflicting grades do not supply a definite answer. An explicit grade
can be retained when a hedged repetition states the same value. If both Gleason score
and prostate ISUP grade group are recorded, both are retained and predictions are
assessed in the reference's named system; the reference uses the explicitly recorded
grade group when both are present.}

\textbf{Grade macro-F1.}
This balances precision and recall for each represented reference-grade class, then gives
each class equal weight. It averages $2TP_g/(2TP_g+FP_g+FN_g)$ over the set $G$ of represented reference grades, where
$TP_g$, $FP_g$ and $FN_g$ are case counts of true-positive, false-positive and false-negative
predictions for grade $g$, and multiplies the result by 100. Grading systems are evaluated
separately using cases with assessable reference grades. Missing or incompatible predictions
contribute a false negative for the reference grade; predicted grades absent from the reference
do not add classes to $G$.

\paragraph{Confusion matrices and reference-defined control}
Confusion matrices show raw counts and the number of cases in each reference-grade class.
Missing and incompatible predictions appear outside the valid-grade cells; shading shows
proportions within each reference-grade row. A
reference-defined always-highest-grade control predicts the highest value in that system's range
for every applicable case. Its balanced agreement is $1/K$ for $K$ represented reference classes
when the highest class is represented, and zero otherwise. A single represented class cannot
establish discrimination. The text examples accompanying \cref{pan:report-public-grade-results}
illustrate diagnostic specificity and a non-high-grade reference.

\supplementarymethodsubsection{Finding agreement, finding coverage and contradiction}{supp:report-omissions}
The finding vocabulary comprises 20 binary concepts, namely invasion, necrosis, lymphovascular invasion,
perineural invasion, metastasis, pleomorphism, mitoses, glandular architecture, papillary
architecture, solid architecture, nested architecture, trabecular architecture, cribriform
architecture, spindle cells, clear cytoplasm, hyperchromasia, prominent nucleoli, desmoplasia,
inflammatory infiltrate and calcification. Case-insensitive,
finding-specific regular expressions map recognized word forms and alternative expressions to
these concepts. For example, ``necrosis'' and ``necrotic'' map to necrosis; ``mitoses'', ``mitotic
figures'' and ``mitotic activity'' map to mitoses; and ``glandular'', ``gland formation'' and
``gland forming'' map to glandular architecture.

A negative statement about a specific subtype, extent or location does not establish absence
of the broader finding. Unavailable invasion-depth or interface measurements are not evidence
that invasion is absent. General tissue invasion is assessed separately from lymphovascular,
perineural and capsular invasion; inflammatory-cell infiltration does not count as tumour
invasion. General polarity rules are described in
\hyperref[sec:methods-report-extraction]{Methods, Report text extraction}; missing-reference rules,
case-level formulas and a worked example are given in
\hyperref[sec:methods-report-metrics]{Primary evaluation measures}.

A prediction for an undocumented finding remains unscored. These measures assess agreement with extracted reference
facts, not why a report omits a finding, whether every reference finding is visible in the
evaluated tissue or whether an unrecorded claim is clinically false. Errors in the reference
text or automated extraction can affect these scores. Finding-level measures
remain distinct from structured report field completeness (\cref{supp:report-grade-findings}).

{\color{black}
\textbf{Contradictions per documented finding.}
To compare methods on a reference-defined denominator, we additionally calculate $E_i/D_i$
for all cases with $D_i>0$, using the counts defined in Methods. This equals finding coverage
minus finding recall. Cases with no comparable generated assertion contribute zero, so this
measure is interpreted alongside coverage and recall. We report both the equal-case mean
and the pooled fraction $\sum_i E_i/\sum_i D_i$. A paired shared-finding analysis restricts each
case to reference-documented types asserted by both compared methods, calculates their
opposite-polarity fractions on those identical types and averages the difference across
cases with at least one shared type. Intervals use 4,000 paired case-bootstrap resamples.

\textbf{Fixed-polarity controls.}
We compare the generated reports with three controls that assert all 20 finding types as
present, absent or the type-specific majority polarity in the 456-case \textcolor{black}{PathXAI-Report pipeline development subset}.
The majority rule uses only documented references from the \textcolor{black}{PathXAI-Report pipeline development subset}, defaults to present for a tie
and is fixed for the public evaluation and private cohorts. These controls use neither images
nor case-specific reference information. They are scored with the same reference-applicability
rules; assertions on undocumented types remain unscored. Reference polarity counts and
results are given in \cref{tab:finding-fixed-denominator,tab:finding-matched-contrasts,tab:finding-reference-polarity}.
}

\textbf{Invasion agreement.}
This is the percentage of cases with definite reference invasion status in which the generated
report agrees that invasion is present or absent. Missing, uncertain, conflicting or opposite
predictions score zero; cases without a definite reference status are excluded. We average
these case-level match indicators and multiply by 100.

\textbf{Necrosis agreement.}
This uses the same case-level matching and aggregation rules as invasion agreement, applied
to necrosis status. It includes only cases with definite reference necrosis status and gives
no credit to missing, uncertain, conflicting or opposite predictions.

\textbf{Critical-positive recall.}
This measures how many documented positive critical findings a report reproduces. For each
case, we divide the number of correctly asserted positive findings among invasion, necrosis,
lymphovascular invasion, perineural invasion and metastasis by the number documented as positive
in the reference. Cases without a positive reference finding in this set are excluded. We average
the applicable case scores and multiply by 100.

\supplementarymethodsubsection{Wording-overlap measures}{supp:report-wording}
\textbf{ROUGE-1 F1.}
This auxiliary measure quantifies single-word overlap with the narrative reference~\cite{lin2004rouge}.
After text normalization, let $n_R(u)$ and $n_P(u)$ be the counts of word $u$ in the reference
and prediction. The case score is
$2\sum_u\min\{n_R(u),n_P(u)\}/[\sum_u n_R(u)+\sum_u n_P(u)]$,
with zero when both counts are zero. Cases without a nonempty narrative reference and
diagnosis-only outputs are inapplicable. We average applicable case scores and multiply by 100.

\textbf{ROUGE-2 F1.}
This auxiliary measure quantifies overlap in consecutive word pairs (bigrams). It uses the
ROUGE-1 F1 calculation with bigram counts in place of single-word counts, the same applicability
rules and the same case-average percentage reporting.

\supplementarymethodsubsection{Case weighting, confidence intervals and paired contrasts}{supp:report-statistics}
Each public TCGA case represents one patient; private analyses retain the case as their sampling unit.
Score aggregation for the report benchmark and reporting-policy comparisons
used Python 3.9.6 and NumPy 2.0.2. The HistGen and HistoGPT comparisons used the same
aggregation and resampling functions with Python 3.13.12 and NumPy 2.4.4.
Case weighting, paired applicability, interval construction and reader clustering are defined in
\hyperref[sec:methods-report-statistics]{Methods, Report uncertainty and paired comparisons}.
\textcolor{black}{Binary proportions used two-sided Clopper--Pearson limits. Balanced grade
accuracy used the mean of the lower and upper binomial limits calculated separately for
$K$ represented reference grades, each at confidence level $1-0.05/K$.
For ordinal grade agreement, the fixed grading-system limits define
$J=g_{\max}-g_{\min}$. Each case score $X_i$ satisfies
$X_i=J^{-1}\sum_{j=1}^{J}\mathbf{1}(X_i\geq j/J)$.
We computed Clopper--Pearson limits for all $J$ threshold proportions at confidence
level $1-0.05/J$ and averaged their lower and upper bounds. This construction was applied
to every ordinal grade estimate. Joint-reference binary means used the same approach
across references, retaining the jointly applicable cases. Bonferroni adjustment within
each mean gives at least 95\% coverage under independent case sampling; independence
between thresholds or references is not required. Absent reference-grade classes are
not imputed. Interval calculations used SciPy 1.13.1.}

\textcolor{black}{For paired differences, percentile bootstrap intervals were reported when
the observed differences varied across cases, or within at least one reference-grade
class for balanced grade accuracy. When all resampled differences were constant, we
retained the point estimate and marked the bootstrap interval as unavailable with
$^{\dagger}$. Reference-only availability differences have no model-comparison interval.}

\textcolor{black}{An exploratory multiplicity sensitivity covered PathXAI-Report versus every
other comparator on the five main automated report measures: 32 comparisons on the
1,200 public evaluation cases, 10 on the 215 shared public cases for HistGen and HistoGPT,
and 42 for each private reference set on the 151 shared cases. PRISM2 contributed only
the two diagnostic measures. All 126 comparisons, including null and adverse differences,
formed one Holm adjustment family. Two-sided exact McNemar tests used discordant pairs
for cancer family match rate and exact diagnosis match rate. Two-sided paired $t$ tests
used case-level differences for finding F1, finding recall and finding contradiction rate.
The latter tests assume independent cases and use a $t$ approximation for the mean
difference; sparse comparable assertions can limit this approximation. The analysis used
SciPy 1.13.1 and retained the original jointly applicable cases. Source Data gives paired
counts, nonzero differences, variance, and raw and adjusted $P$ values. This sensitivity
does not adjust the original bootstrap intervals or the separate supplementary analyses.}

\textcolor{black}{An exploratory stratified bootstrap retained the same 126 comparisons and
jointly applicable cases. We resampled paired differences within public TCGA cancer types
or private organ groups, keeping each group's paired count fixed in 4,000 draws
(seed 20261105). Each replicate averaged all sampled case differences equally; percentile
limits therefore condition on the observed paired-applicable composition. Singleton and
constant groups contributed their fixed observed totals; an interval was unavailable if
no group had variation. P1 and P2 were analysed separately. These pointwise intervals
describe case-level uncertainty within the recorded groups; hospital clustering was not
estimated because a case-to-hospital mapping was unavailable. Subgroup means and counts
are descriptive (\cref{supp:results-report-stratification}).}

\textcolor{black}{The report-availability sensitivity used the full eligible cohorts, retaining
reference-based applicability and assigning zero to unavailable outputs for cancer family
match rate, exact diagnosis match rate, finding F1 and finding recall. We compared these
estimates with shared-case and available-output analyses. Absolute binary proportions used
Wilson score intervals; finding means and paired differences used the same case-bootstrap procedure.
Finding contradiction rate requires definite assertions and was not imputed for unavailable
outputs (\cref{supp:results-report-availability}).}

All public report analyses excluded the same 51 cases whose supplied site descriptions contained
disease names. Each comparison used the remaining eligible cases shared by its methods,
with metric-specific applicability determined after this exclusion
(\cref{tab:site-prior-sensitivity}).

Each result table records metric-specific applicable denominators; separate P1 and P2 estimates
use their corresponding reference fields. Human-rating checks and their paired denominators
are specified in \cref{supp:rating-checks}.

\supplementarymethodsection{Pathologist reference-report authoring and ranking and correction assessment}{supp:physician-tools}

\paragraph{Research tools and data separation}
The WSI service supplies image tiles; annotations and
exports remain in the deployment environment. Reference reports, anonymous presentation
assignments and analytic method decoding have separate roles. Restricted slides, credentials,
reader-assignment keys and case-level study records are separated from software and aggregate
results. Whether analysts were blinded to method identities during automatic scoring and
statistical analysis could not be confirmed.

\supplementarymethodsubsection{Pathologist reference-report authoring}{supp:reference-authoring}
The reference-authoring protocol, translation verification and timing are specified in
\hyperref[sec:methods-report-human]{Methods, Pathologist reference-report authoring}.
The two separately authored references per private case yield 304 reports for 152 cases.
We also compare reference fields that are assessable in both reports.

The authoring workspace combines a WSI viewer, rectangular diagnostic ROI annotations,
structured fields and free-text report sections (\cref{fig:physician-authoring}). It displays
a neutral study label and known site, retaining filenames, paths and diagnosis-derived cohort
labels on the server. ROIs can be linked to a diagnostic purpose, coordinates and a free-text
note. Structured fields cover primary diagnosis, broad tumour type, subtype, malignancy,
invasion, grade, necrosis, confidence and unassessable findings. Narrative fields retain diagnosis,
microscopic findings, evidence or rationale, limitations and final summary.

The interface supports drafts, recovery, submission locking and corrections with version
history. The translation procedure uses temperature 0 and reuses translations of identical strings. The prompt contains field text without case
identifiers or system-generated reports and requests preservation of diagnostic specificity,
negation, uncertainty and numbers. Automatic checks flag changed numbers, untranslated text,
lost question marks and potentially altered negation or uncertainty.

\manuscriptnativefigure{\textbf{Pathologist reference-report authoring interface.}}{fig:physician-authoring}{%
    \begin{minipage}{180mm}
    \includegraphics[width=180mm]{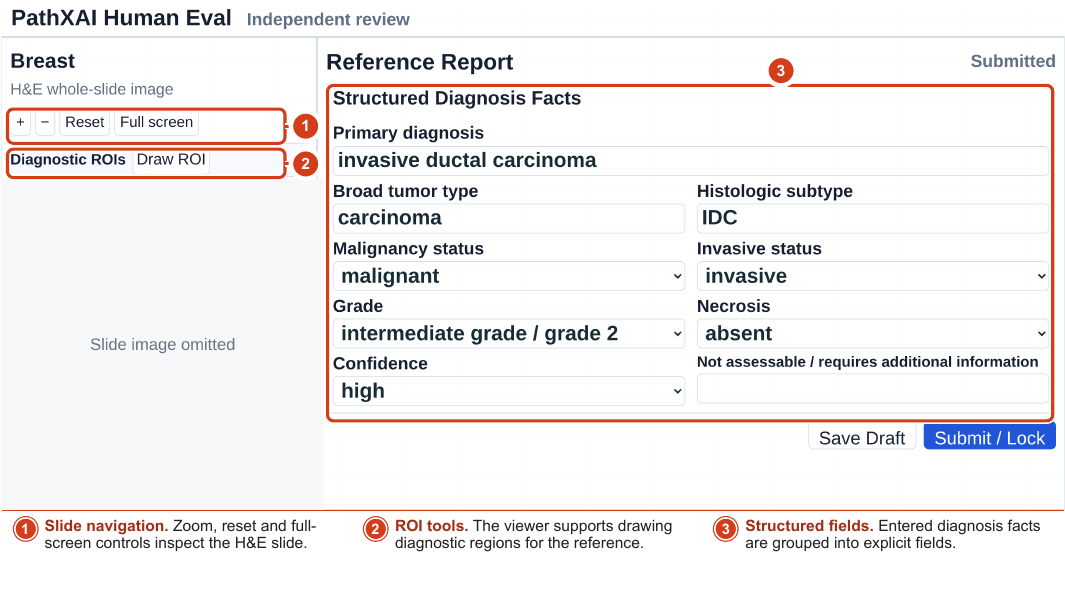}
    \end{minipage}%
  }{%
Structured report fields and diagnostic ROI annotations accompany slide navigation. Draft, submission-lock and correction controls are shown; the slide viewport and thumbnail are omitted. The report is illustrative; the authoring protocol is in \cref{supp:reference-authoring}.
(Notation: WSI, whole-slide image; ROI, region of interest; ``Independent review'', interface mode label.)
}

\Needspace{12\baselineskip}
\supplementarymethodsubsection{Pathologist ranking and correction assessment}{supp:rating-protocol}
Reader qualifications are given in \hyperref[sec:methods-report-human]{Methods, Pathologist reference-report authoring}.
\textcolor{black}{Cases were randomly sampled within cancer type. Randomized report presentation and re-review
under anonymous method labels are described in}
\hyperref[sec:methods-report-ranking]{Pathologist ranking and correction assessment}.
The 132 evaluable cases were drawn from the \textcolor{black}{456}-case \textcolor{black}{\textcolor{black}{PathXAI-Report pipeline development subset}}; corrupted source
data were ineligible. PRISM2 was excluded from report ranking because it supplied diagnosis-only
outputs. The interface retains separate overall, section-specific, correction and
concern fields; the primary analysis uses the overall ranks, correction categories and concern flags described in Methods.
If neither reader flagged a concern, resampling
also yields zero counts; this does not establish that the underlying risk is absent.

\paragraph{\textcolor{black}{Rating guidance}}
\textcolor{black}{The interface instructed readers to inspect the WSI and assess each report on its
own merits. No correction meant clinically acceptable as written; minor correction meant
clinically usable after a limited edit; major correction required changes to the diagnosis
or management-relevant content. Unusable ratings were left to each pathologist's judgment. Readers were
instructed to select concern flags only when a major problem was present.}

\paragraph{\textcolor{black}{Agreement between readers}}
\textcolor{black}{For each method, we measured exact agreement and linear weighted Cohen's
$\kappa$ across the ordered correction categories (none, minor, major and unusable).
The disagreement weight was $|j-k|/3$ for categories indexed from 0 to 3.
We also calculated Kendall's $\tau_b$ between the two readers' five-method preference
rankings within each case, accounting for ties, and averaged the estimable case values.
Cases in which either reader tied all five methods had undefined $\tau_b$ and were
excluded only from this mean. Both analyses used 4,000 case-bootstrap replicates
(seed 20260831), retaining both readers and all methods within each draw, with
percentile 95\% intervals. Results describe the two observed readers
(\cref{tab:reader-agreement}).}

\Cref{fig:physician-ranking} shows the anonymous comparison controls.

\manuscriptnativefigure{\textbf{\textcolor{black}{Report-ranking and correction interface with anonymous method labels.}}}{fig:physician-ranking}{%
    \begin{minipage}{180mm}
    \includegraphics[width=180mm]{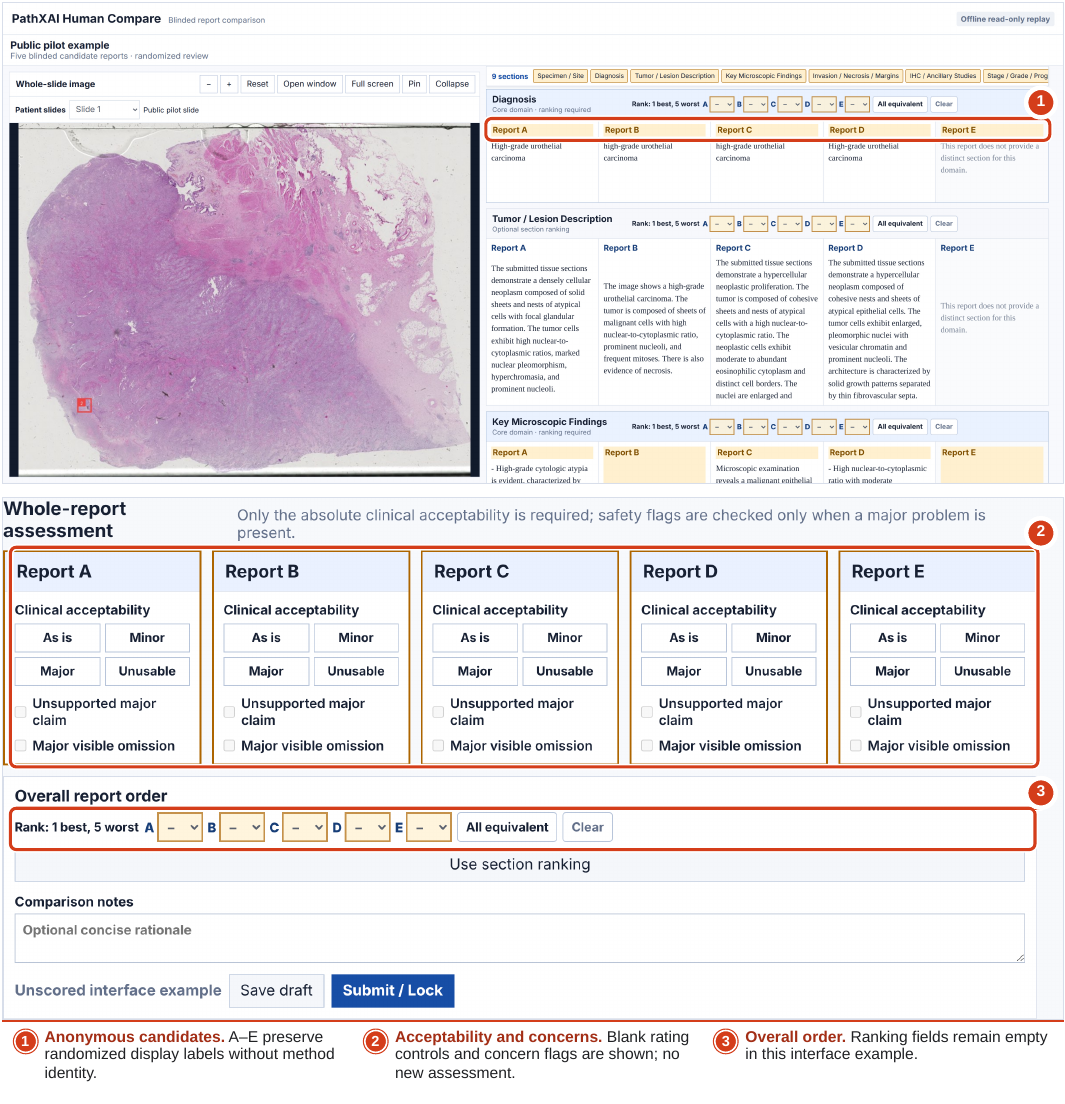}
    \end{minipage}%
  }{%
Anonymous method labels A--E accompany section-aligned reports and ranking, correction and concern controls. \textcolor{black}{Prior ratings remained visible during re-review.} Procedures are in \cref{supp:rating-protocol}.
}

\supplementarymethodsubsection{Output and structured-field checks}{supp:rating-checks}
The output-failure screen and no/minor correction with complete fields measure are defined in
\hyperref[sec:methods-report-checks]{Methods, Report-content checks}.
An abstention fails the field-completeness check even when clinically appropriate.
\Cref{pan:report-physician-correction-ratings,tab:source_fig10} retain the original rating
categories and the cumulative checks. Each report's check results are paired with its original
ratings; no case or reader rating is removed by these screens.

\supplementarymethodsection{Evidence-linked report viewer}{supp:report-interface}

\textcolor{black}{The Result, Evidence, Reasoning and Verification views show selected report fields,
spatial inputs, acquisition steps and automated claim assessments from one randomly selected public case.}
Identifiers are replaced by neutral labels; callouts and clarified interface labels are added.
Evaluation references and ratings are omitted.
Interface usability and review time were not formally evaluated.

\manuscriptnativefigure{\textbf{Result view of the report viewer.}}{fig:report-interface}{%
    \begin{minipage}{180mm}
    \includegraphics[width=180mm]{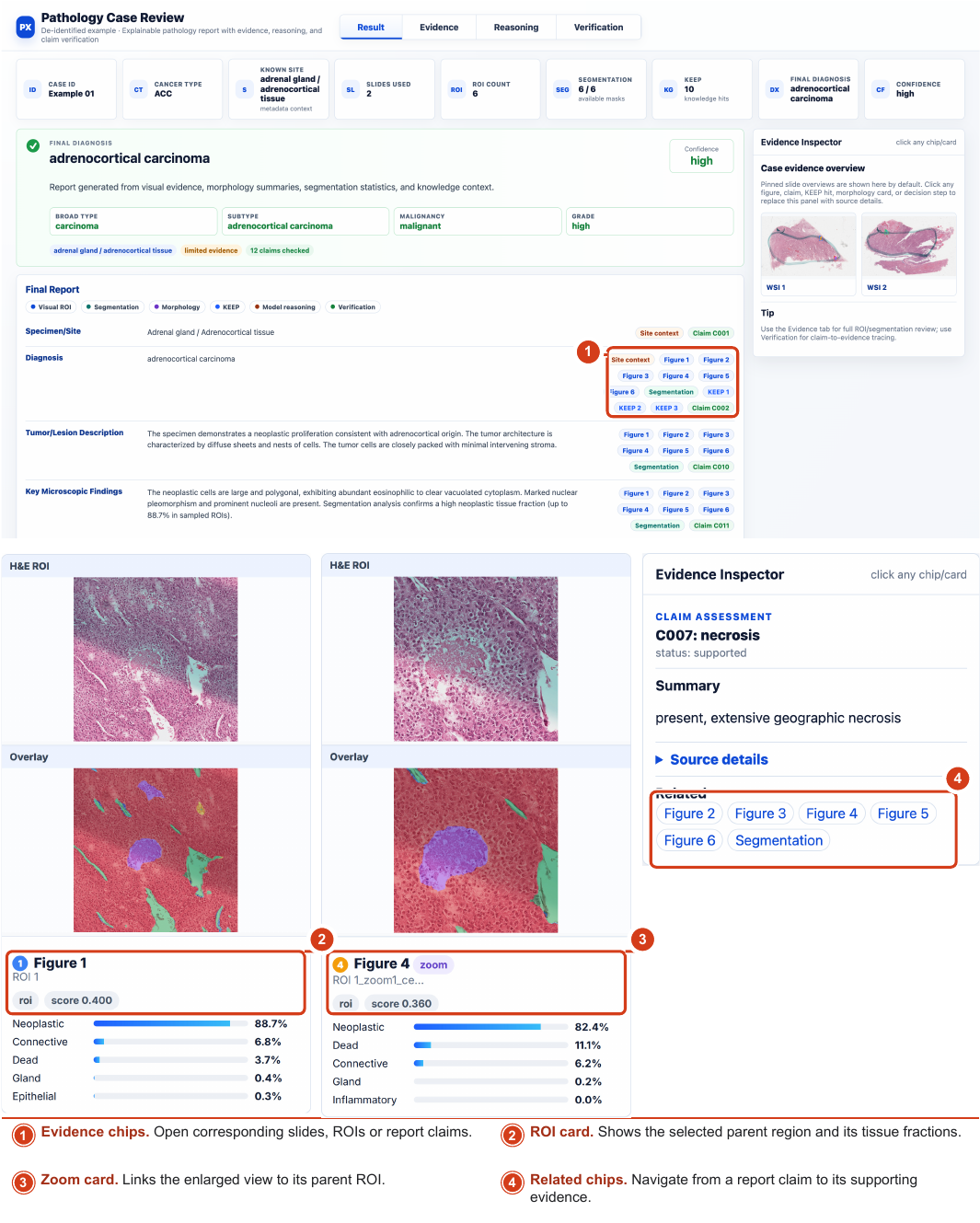}
    \end{minipage}%
  }{%
The generated report links to slide regions, H\&E images, segmentation overlays and tissue statistics. The inspector displays a linked necrosis statement and its automated assessment. \textcolor{black}{Links follow the associations stored in the evidence graph.}
(Notation: H\&E, haematoxylin and eosin.)
}

\manuscriptnativefigure{\textbf{Evidence view of the report viewer.}}{fig:report-evidence-interface}{%
    \begin{minipage}{180mm}
    \manuscriptpanelblock{5mm}{180.0mm}{\manuscriptpanelmark{pan:report-evidence-interface-evidence-tab}}{Evidence tab}{\includegraphics[width=\linewidth]{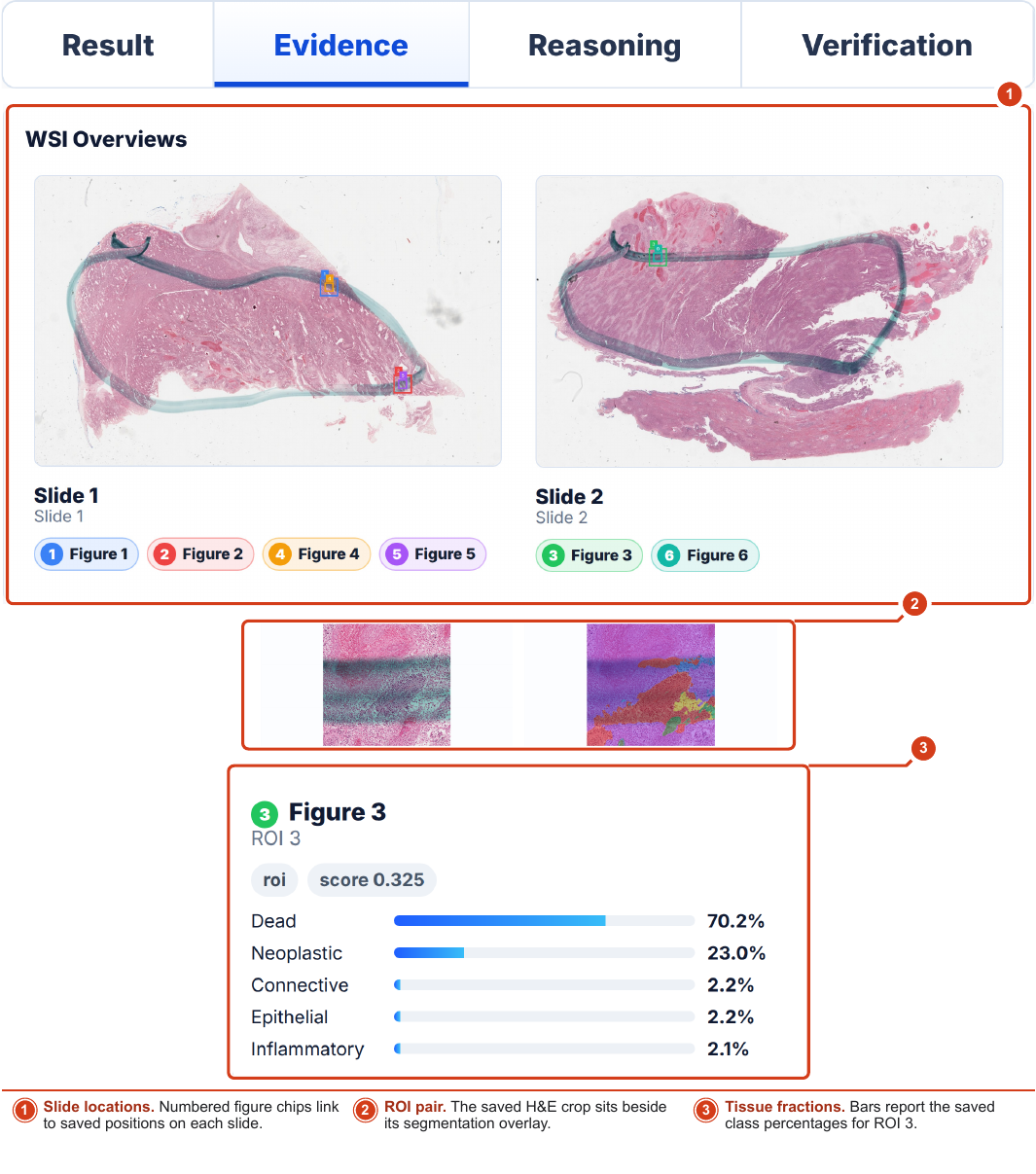}}
    \end{minipage}%
  }{%
\textbf{\panelletterref{pan:report-evidence-interface-evidence-tab}} links indexed slide regions to H\&E images, segmentation overlays and tissue statistics for the case in \cref{fig:report-interface}. The selected tissue statistics also appear in the necrosis assessment in \cref{fig:report-verification-interface}.
(Notation: ROI, region of interest; H\&E, haematoxylin and eosin. Interface colours differ from the shared main-figure palette.)
}

\manuscriptnativefigure{\textbf{Reasoning view of the report viewer.}}{fig:report-reasoning-interface}{%
    \begin{minipage}{180mm}
    \manuscriptpanelblock{5mm}{180.0mm}{\manuscriptpanelmark{pan:report-reasoning-interface-reasoning-tab}}{Reasoning tab}{\includegraphics[width=\linewidth]{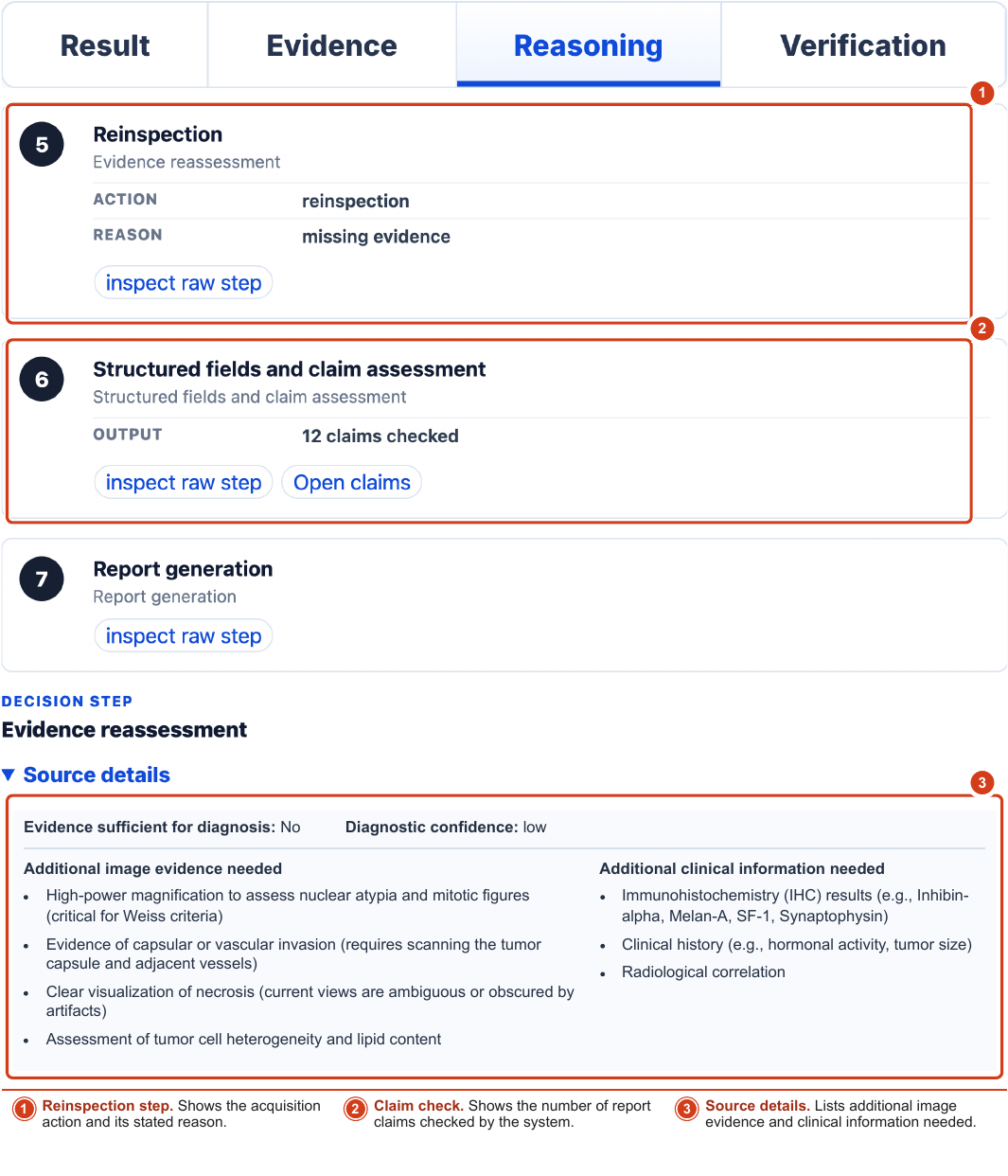}}
    \end{minipage}%
  }{%
\textbf{\panelletterref{pan:report-reasoning-interface-reasoning-tab}} shows reinspection, claim checks and report-generation steps. The expanded inspector records model assessments of evidence sufficiency and diagnostic confidence, together with missing image or clinical information.
}
\manuscriptnativefigure{\textbf{Verification view of the report viewer.}}{fig:report-verification-interface}{%
    \begin{minipage}{180mm}
    \manuscriptpanelblock{5mm}{180.0mm}{\manuscriptpanelmark{pan:report-verification-interface-verification-tab}}{Verification tab}{\includegraphics[width=\linewidth]{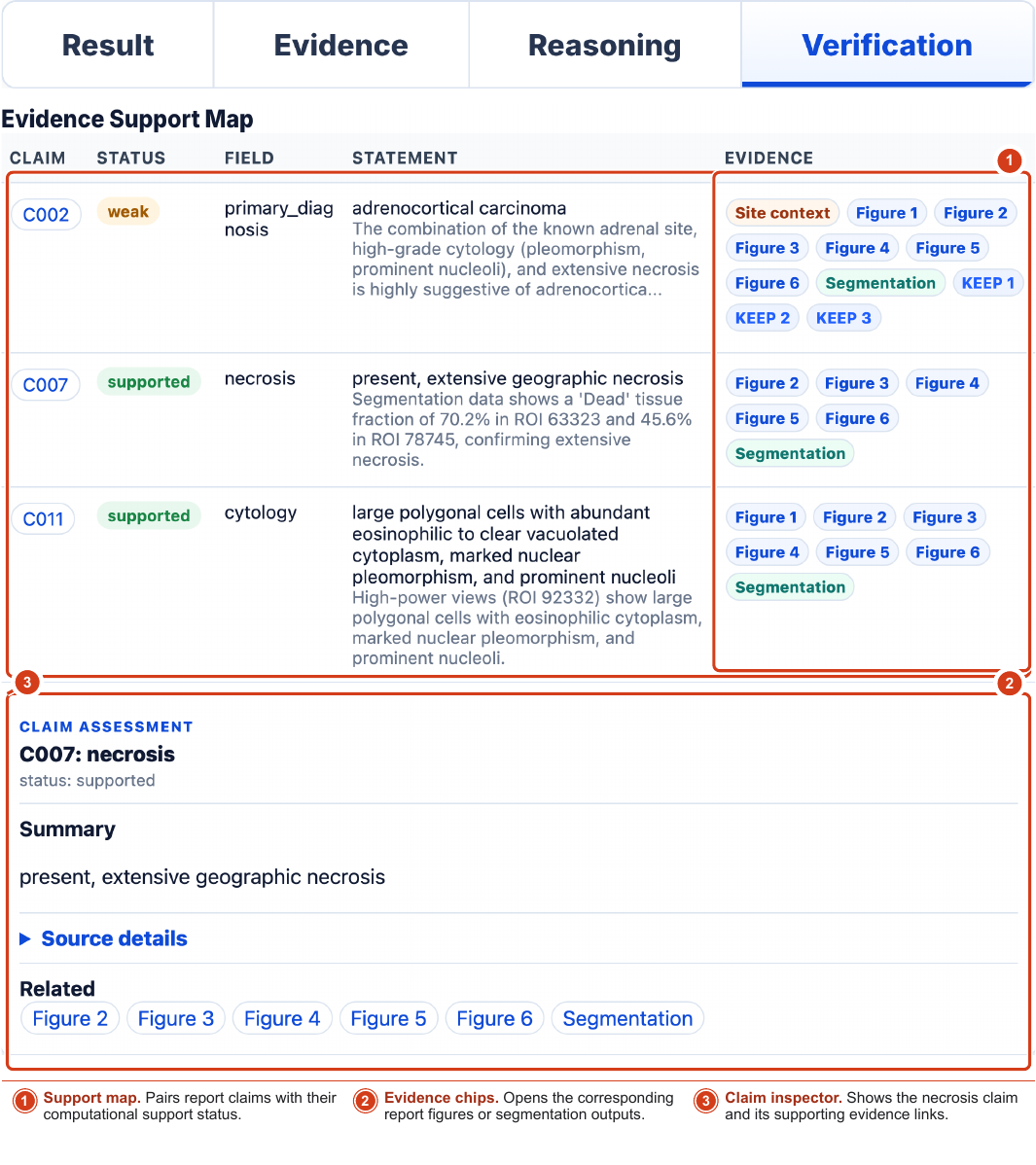}}
    \end{minipage}%
  }{%
\textbf{\panelletterref{pan:report-verification-interface-verification-tab}} shows selected diagnosis, necrosis and cytology claims with their automated support states and evidence links. The expanded necrosis inspector connects to the tissue evidence in \cref{fig:report-evidence-interface}. The diagnosis retains its weaker support state; intervening rows are omitted for print.
(Notation: C-prefixed codes, viewer-local claim identifiers; supported/weak, automated assessment labels.)
}

\FloatBarrier
\renewcommand{\floatpagefraction}{0.85}
\FloatBarrier
\section*{Supplementary Results}
Tables report pointwise 95\% confidence intervals and sample sizes for each metric. Source Data provide full-precision estimates. Rates and Dice are percentages, paired differences are expressed in pp, and preference rank uses its 1--5 scale. NA denotes unavailable or inapplicable values. Metric definitions and weighting are given in \cref{tab:report-metric-overview}.

Arrows indicate whether higher or lower metric values are preferred. Paired differences subtract the comparator method's score from the target method's score.

\ifdefined\privateDataReleaseBoundary\privateDataReleaseBoundary\par\fi

\supplementaryresultsection{Automated report evaluation on the public cohort}{supp:public-report-results}
\label{supp:results-public-full}
\textcolor{black}{Primary evaluation-cohort estimates are given in \cref{tab:report-primary-estimates,tab:report-primary-contrasts,tab:report-primary-native}.}
\begin{supplementarytables}
\begingroup\color{black}
\begin{longtable}{@{}>{\raggedright\arraybackslash}p{0.20\linewidth}>{\centering\arraybackslash}p{0.136\linewidth}>{\centering\arraybackslash}p{0.136\linewidth}>{\centering\arraybackslash}p{0.136\linewidth}>{\centering\arraybackslash}p{0.136\linewidth}>{\centering\arraybackslash}p{0.136\linewidth}@{}}
\caption{Primary public report evaluation on 1,200 cases (\cref{pan:public-summary-public-performance}). Percentages with \textcolor{black}{pointwise 95\% confidence intervals} and applicable case counts. Finding contradiction rate uses method-specific comparable assertions. \textcolor{black}{Interval construction follows \cref{supp:report-statistics}.}}\label{tab:report-primary-estimates}\\
\toprule Method & Cancer family match rate & Exact diagnosis match rate & Finding F1 & Finding recall & Finding contradiction rate \\ \midrule
\endfirsthead
\multicolumn{6}{@{}l}{\manuscripttablecontinued{}}\\[3pt]
\toprule Method & Cancer family match rate & Exact diagnosis match rate & Finding F1 & Finding recall & Finding contradiction rate \\ \midrule
\endhead
\midrule\multicolumn{6}{r@{}}{Continued on next page}\\\endfoot
\bottomrule\endlastfoot
Qwen+H\&E  & \begin{tabular}[t]{@{}c@{}}68.1\\{}[65.36, 70.72]\\n=1200\end{tabular} & \begin{tabular}[t]{@{}c@{}}58.9\\{}[54.90, 62.89]\\n=604\end{tabular} & \begin{tabular}[t]{@{}c@{}}43.0\\{}[40.63, 45.42]\\n=997\end{tabular}  & \begin{tabular}[t]{@{}c@{}}36.6\\{}[34.34, 38.88]\\n=997\end{tabular}  & \begin{tabular}[t]{@{}c@{}}16.0\\{}[13.72, 18.42]\\n=707\end{tabular}\\[3pt]
DeepSeek+H\&E  & \begin{tabular}[t]{@{}c@{}}71.9\\{}[69.28, 74.44]\\n=1200\end{tabular} & \begin{tabular}[t]{@{}c@{}}60.3\\{}[56.24, 64.19]\\n=604\end{tabular} & \begin{tabular}[t]{@{}c@{}}42.1\\{}[39.82, 44.43]\\n=997\end{tabular}  & \begin{tabular}[t]{@{}c@{}}36.1\\{}[33.98, 38.40]\\n=997\end{tabular}  & \begin{tabular}[t]{@{}c@{}}23.2\\{}[20.57, 25.84]\\n=738\end{tabular}\\[3pt]
GLM+H\&E  & \begin{tabular}[t]{@{}c@{}}61.8\\{}[59.02, 64.59]\\n=1200\end{tabular} & \begin{tabular}[t]{@{}c@{}}56.0\\{}[51.90, 59.96]\\n=604\end{tabular} & \begin{tabular}[t]{@{}c@{}}44.2\\{}[41.85, 46.47]\\n=997\end{tabular}  & \begin{tabular}[t]{@{}c@{}}38.1\\{}[35.86, 40.42]\\n=997\end{tabular}  & \begin{tabular}[t]{@{}c@{}}22.5\\{}[19.96, 25.22]\\n=751\end{tabular}\\[3pt]
MedGemma  & \begin{tabular}[t]{@{}c@{}}6.7\\{}[5.32, 8.23]\\n=1200\end{tabular} & \begin{tabular}[t]{@{}c@{}}6.1\\{}[4.35, 8.35]\\n=604\end{tabular} & \begin{tabular}[t]{@{}c@{}}37.4\\{}[35.12, 39.70]\\n=997\end{tabular}  & \begin{tabular}[t]{@{}c@{}}31.8\\{}[29.54, 33.92]\\n=997\end{tabular}  & \begin{tabular}[t]{@{}c@{}}14.4\\{}[11.90, 17.02]\\n=610\end{tabular}\\[3pt]
PLIP-guided Qwen  & \begin{tabular}[t]{@{}c@{}}45.7\\{}[42.82, 48.53]\\n=1200\end{tabular} & \begin{tabular}[t]{@{}c@{}}39.2\\{}[35.32, 43.26]\\n=604\end{tabular} & \begin{tabular}[t]{@{}c@{}}21.9\\{}[19.84, 24.04]\\n=997\end{tabular}  & \begin{tabular}[t]{@{}c@{}}17.8\\{}[15.97, 19.65]\\n=997\end{tabular}  & \begin{tabular}[t]{@{}c@{}}6.3\\{}[3.99, 8.83]\\n=351\end{tabular}\\
\end{longtable}
\begin{longtable}{@{}>{\raggedright\arraybackslash}p{0.20\linewidth}>{\centering\arraybackslash}p{0.136\linewidth}>{\centering\arraybackslash}p{0.136\linewidth}>{\centering\arraybackslash}p{0.136\linewidth}>{\centering\arraybackslash}p{0.136\linewidth}>{\centering\arraybackslash}p{0.136\linewidth}@{}}
\caption{HistGen and HistoGPT comparison on 215 shared cases from the public evaluation cohort. Released HistGen training and validation patients are excluded. Percentages with unadjusted 95\% case-bootstrap intervals and applicable case counts. Finding contradiction rate uses method-specific comparable assertions.}\label{tab:report-primary-native}\\
\toprule Method & Cancer family match rate & Exact diagnosis match rate & Finding F1 & Finding recall & Finding contradiction rate \\ \midrule
\endfirsthead
\multicolumn{6}{@{}l}{\manuscripttablecontinued{}}\\[3pt]
\toprule Method & Cancer family match rate & Exact diagnosis match rate & Finding F1 & Finding recall & Finding contradiction rate \\ \midrule
\endhead
\midrule\multicolumn{6}{r@{}}{Continued on next page}\\\endfoot
\bottomrule\endlastfoot
HistGen  & \begin{tabular}[t]{@{}c@{}}39.1\\{}[32.51, 45.94]\\n=215\end{tabular} & \begin{tabular}[t]{@{}c@{}}42.5\\{}[30.97, 54.59]\\n=73\end{tabular} & \begin{tabular}[t]{@{}c@{}}17.2\\{}[13.13, 21.54]\\n=173\end{tabular}  & \begin{tabular}[t]{@{}c@{}}13.6\\{}[10.12, 17.33]\\n=173\end{tabular}  & \begin{tabular}[t]{@{}c@{}}32.2\\{}[22.97, 41.67]\\n=74\end{tabular}\\[3pt]
HistoGPT  & \begin{tabular}[t]{@{}c@{}}5.1\\{}[2.58, 8.97]\\n=215\end{tabular} & \begin{tabular}[t]{@{}c@{}}1.4\\{}[0.03, 7.40]\\n=73\end{tabular} & \begin{tabular}[t]{@{}c@{}}12.0\\{}[8.43, 15.73]\\n=173\end{tabular}  & \begin{tabular}[t]{@{}c@{}}8.8\\{}[6.15, 11.84]\\n=173\end{tabular}  & \begin{tabular}[t]{@{}c@{}}9.3\\{}[2.33, 18.60]\\n=43\end{tabular}\\[3pt]
PathXAI-Report (core)  & \begin{tabular}[t]{@{}c@{}}73.5\\{}[67.06, 79.26]\\n=215\end{tabular} & \begin{tabular}[t]{@{}c@{}}58.9\\{}[46.77, 70.29]\\n=73\end{tabular} & \begin{tabular}[t]{@{}c@{}}43.2\\{}[38.01, 48.50]\\n=173\end{tabular}  & \begin{tabular}[t]{@{}c@{}}35.8\\{}[31.06, 40.86]\\n=173\end{tabular}  & \begin{tabular}[t]{@{}c@{}}12.5\\{}[8.13, 17.66]\\n=126\end{tabular}\\[3pt]
PathXAI-Report  & \begin{tabular}[t]{@{}c@{}}77.7\\{}[71.51, 83.06]\\n=215\end{tabular} & \begin{tabular}[t]{@{}c@{}}63.0\\{}[50.91, 74.03]\\n=73\end{tabular} & \begin{tabular}[t]{@{}c@{}}43.3\\{}[38.01, 48.68]\\n=173\end{tabular}  & \begin{tabular}[t]{@{}c@{}}36.0\\{}[31.12, 41.04]\\n=173\end{tabular}  & \begin{tabular}[t]{@{}c@{}}12.7\\{}[8.12, 17.87]\\n=127\end{tabular}\\
\end{longtable}
\begin{longtable}{@{}>{\raggedright\arraybackslash}p{0.20\linewidth}>{\centering\arraybackslash}p{0.136\linewidth}>{\centering\arraybackslash}p{0.136\linewidth}>{\centering\arraybackslash}p{0.136\linewidth}>{\centering\arraybackslash}p{0.136\linewidth}>{\centering\arraybackslash}p{0.136\linewidth}@{}}
\caption{Paired differences on the public evaluation cohort. Each estimate is PathXAI-Report minus the listed comparator, in percentage points, with an unadjusted 95\% case-bootstrap interval. Applicable counts refer to cases assessable in both methods.}\label{tab:report-primary-contrasts}\\
\toprule Method & Cancer family match rate & Exact diagnosis match rate & Finding F1 & Finding recall & Finding contradiction rate \\ \midrule
\endfirsthead
\multicolumn{6}{@{}l}{\manuscripttablecontinued{}}\\[3pt]
\toprule Method & Cancer family match rate & Exact diagnosis match rate & Finding F1 & Finding recall & Finding contradiction rate \\ \midrule
\endhead
\midrule\multicolumn{6}{r@{}}{Continued on next page}\\\endfoot
\bottomrule\endlastfoot
Qwen+H\&E & \begin{tabular}[t]{@{}c@{}}+4.42\\{}[1.92, \textcolor{black}{6.75}]\\n=1200\end{tabular} & \begin{tabular}[t]{@{}c@{}}\textcolor{black}{+3.31}\\{}[\textcolor{black}{0.00}, \textcolor{black}{6.62}]\\n=\textcolor{black}{604}\end{tabular} & \begin{tabular}[t]{@{}c@{}}+5.75\\{}[3.63, 7.87]\\n=997\end{tabular} & \begin{tabular}[t]{@{}c@{}}+5.07\\{}[3.04, 7.03]\\n=997\end{tabular} & \begin{tabular}[t]{@{}c@{}}-5.52\\{}[-8.02, -3.11]\\n=664\end{tabular}\\[3pt]
DeepSeek+H\&E & \begin{tabular}[t]{@{}c@{}}\textcolor{black}{+0.58}\\{}[\textcolor{black}{-1.92}, \textcolor{black}{3.17}]\\n=1200\end{tabular} & \begin{tabular}[t]{@{}c@{}}\textcolor{black}{+1.99}\\{}[\textcolor{black}{-1.82}, \textcolor{black}{5.79}]\\n=\textcolor{black}{604}\end{tabular} & \begin{tabular}[t]{@{}c@{}}+6.67\\{}[4.29, 9.27]\\n=997\end{tabular} & \begin{tabular}[t]{@{}c@{}}+5.49\\{}[3.11, 7.78]\\n=997\end{tabular} & \begin{tabular}[t]{@{}c@{}}-9.19\\{}[-12.11, -6.17]\\n=634\end{tabular}\\[3pt]
GLM+H\&E & \begin{tabular}[t]{@{}c@{}}\textcolor{black}{+10.67}\\{}[\textcolor{black}{7.42}, \textcolor{black}{13.75}]\\n=1200\end{tabular} & \begin{tabular}[t]{@{}c@{}}\textcolor{black}{+6.29}\\{}[\textcolor{black}{2.15}, \textcolor{black}{10.60}]\\n=\textcolor{black}{604}\end{tabular} & \begin{tabular}[t]{@{}c@{}}+4.60\\{}[1.89, 7.15]\\n=997\end{tabular} & \begin{tabular}[t]{@{}c@{}}+3.50\\{}[1.06, 5.96]\\n=997\end{tabular} & \begin{tabular}[t]{@{}c@{}}-10.01\\{}[-13.07, -7.04]\\n=644\end{tabular}\\[3pt]
MedGemma & \begin{tabular}[t]{@{}c@{}}+65.83\\{}[\textcolor{black}{63.17}, 68.42]\\n=1200\end{tabular} & \begin{tabular}[t]{@{}c@{}}\textcolor{black}{+56.13}\\{}[\textcolor{black}{52.15}, \textcolor{black}{60.10}]\\n=\textcolor{black}{604}\end{tabular} & \begin{tabular}[t]{@{}c@{}}+11.33\\{}[8.75, 13.89]\\n=997\end{tabular} & \begin{tabular}[t]{@{}c@{}}+9.86\\{}[7.46, 12.35]\\n=997\end{tabular} & \begin{tabular}[t]{@{}c@{}}-1.08\\{}[-3.97, 1.75]\\n=546\end{tabular}\\
\end{longtable}
\endgroup
\end{supplementarytables}
{\color{black}
\paragraph{Constant-diagnosis control}
Always predicting squamous cell carcinoma, selected on the \textcolor{black}{PathXAI-Report pipeline development subset}, gave a
cancer family match rate of 32.8\% in the public evaluation cohort, compared with 72.5\%
for PathXAI-Report. The paired difference was 39.75~pp
(\cref{tab:fixed-diagnosis-control}).
\begin{supplementarytables}
\begin{longtable}{@{}p{.27\linewidth}*{2}{>{\centering\arraybackslash}p{.19\linewidth}}>{\centering\arraybackslash}p{.27\linewidth}@{}}
\caption{\textcolor{black}{Cancer family match rate with a constant-diagnosis control. The control always predicts squamous cell carcinoma, selected using only the 456-case \textcolor{black}{PathXAI-Report pipeline development subset} and then fixed for all 1,200 public evaluation cases. The same cancer-family rules score controls and generated diagnoses. Differences subtract the control from each method in the evaluation cohort; brackets give unadjusted 95\% paired-case bootstrap intervals. No images or anatomical-site inputs determine the constant diagnosis. \textcolor{black}{Results from the PathXAI-Report pipeline development subset} are descriptive in-sample values.}}\label{tab:fixed-diagnosis-control}\\
\toprule \textcolor{black}{Method} & \textcolor{black}{PathXAI-Report pipeline development match rate (\%)} & \textcolor{black}{Evaluation match rate (\%)} & \textcolor{black}{Evaluation difference (pp) [95\% CI]}\\\midrule
\endfirsthead
\multicolumn{4}{@{}l}{\manuscripttablecontinued{}}\\
\toprule \textcolor{black}{Method} & \textcolor{black}{PathXAI-Report pipeline development match rate (\%)} & \textcolor{black}{Evaluation match rate (\%)} & \textcolor{black}{Evaluation difference (pp) [95\% CI]}\\\midrule
\endhead
\bottomrule\endlastfoot
\textcolor{black}{Constant diagnosis} & \textcolor{black}{14.9} & \textcolor{black}{32.8} & \textcolor{black}{---}\\[3pt]
\textcolor{black}{Qwen+H\&E} & \textcolor{black}{71.5} & \textcolor{black}{68.1} & \textcolor{black}{+35.33 [+32.17, +38.50]}\\[3pt]
\textcolor{black}{DeepSeek+H\&E} & \textcolor{black}{70.6} & \textcolor{black}{71.9} & \textcolor{black}{+39.17 [+36.17, +42.17]}\\[3pt]
\textcolor{black}{GLM+H\&E} & \textcolor{black}{55.9} & \textcolor{black}{61.8} & \textcolor{black}{+29.08 [+25.67, +32.42]}\\[3pt]
\textcolor{black}{MedGemma} & \textcolor{black}{7.7} & \textcolor{black}{6.7} & \textcolor{black}{-26.08 [-29.00, -22.92]}\\[3pt]
\textcolor{black}{PLIP-guided Qwen} & \textcolor{black}{29.8} & \textcolor{black}{45.7} & \textcolor{black}{+12.92 [+9.33, +16.58]}\\[3pt]
\textcolor{black}{PRISM2} & \textcolor{black}{55.9} & \textcolor{black}{53.2} & \textcolor{black}{+20.50 [+16.75, +24.25]}\\[3pt]
\textcolor{black}{PathXAI-Report (core)} & \textcolor{black}{73.7} & \textcolor{black}{68.2} & \textcolor{black}{+35.42 [+32.25, +38.50]}\\[3pt]
\textcolor{black}{PathXAI-Report} & \textcolor{black}{76.8} & \textcolor{black}{72.5} & \textcolor{black}{+39.75 [+36.42, +43.00]}\\[3pt]
\end{longtable}
\end{supplementarytables}
}
\Cref{fig:report-public} summarizes the \textcolor{black}{1,656}-case public comparison.
\Crefrange{tab:source_fig8:part-1}{tab:source_fig8:part-7} give complete estimates, including reference-specific grading systems,
ordinal grade agreement, compatible grade coverage and critical-positive recall.

\supplementaryresultsubsection{Diagnostic specificity and grading}{supp:results-public-specificity}
\textbf{PathXAI-Report achieves a higher exact diagnosis match rate in both public reference subsets.}
Exact diagnosis match rate requires a match to the specific diagnosis in each reference
report.
The \textcolor{black}{combined public estimate} averages the
case-level scores across \PublicEntityFullN{} applicable cases, comprising
\CuratedEntityFullN{} \textcolor{black}{in the \textcolor{black}{PathXAI-Report pipeline development subset}} and \OCREntityFullN{} \textcolor{black}{in the public evaluation cohort}.
Each applicable case receives equal weight, and 95\% intervals use 4,000 case-bootstrap
resamples with the same procedure as the other public outcomes. The \textcolor{black}{public evaluation cohort} was not used for \textcolor{black}{PathXAI-Report pipeline development}
(\cref{supp:report-policy-comparisons}). With the primary extraction rule,
PathXAI-Report had the highest exact diagnosis match rate against case-specific public references, reaching \CuratedEntityFull\% in \CuratedEntityFullN{} applicable cases \textcolor{black}{in the \textcolor{black}{PathXAI-Report pipeline development subset}}
and \OCREntityFull\% in \OCREntityFullN{} \textcolor{black}{in the public evaluation cohort}, compared with
\CuratedEntityQwen\% and \OCREntityQwen\% for Qwen+H\&E (\cref{tab:report-scope-sensitivity:part-1}).
\textcolor{black}{The selected renal case TCGA-B0-5080 illustrates the cancer-family comparison.} PathXAI-Report named
clear cell renal cell carcinoma, matching the cohort target, whereas Qwen+H\&E reported a renal
neoplasm not otherwise specified (\cref{pan:report-public-examples}).
Sensitivity to including final-summary diagnoses is assessed in
\cref{supp:results-report-scope}.

\textcolor{black}{On the 604 applicable public evaluation cases, the diagnosis-outcome
counts for PathXAI-Report versus Qwen+H\&E were 376 versus 356 matches, 67 versus 97
definite non-specific diagnoses, 149 versus 128 different specific diagnoses and 12
versus 23 cases without a definite recognized diagnosis. Thus, more matching diagnoses
coexisted with more different specific diagnoses. Among the remaining 596 cases without
an applicable reference diagnosis, specific predictions occurred in 474 and 402 cases,
respectively; their exact agreement could not be assessed. Complete counts for all
methods and cohorts are provided in Source Data (\texttt{diagnostic\_outcomes}).}

\textbf{Histological-grade (1--3) agreement exceeds Qwen+H\&E and an always-Grade-3 control.}
Balanced grade accuracy gives each reference-grade class equal weight, so
frequent grades do not dominate the score. Histological-grade (1--3) agreement was \PublicGradeFull\%
(95\% CI, \PublicGradeFullCI; n=\PublicGradeFullN), compared with
\PublicGradeQwen\% for Qwen+H\&E, 33.3\% for the always-Grade-3 control and
\PublicGradeDeepSeek\% for DeepSeek+H\&E
(\cref{pan:report-public-grade-results}). The textual example TCGA-2H-A9GM contrasts
moderate with high-grade wording. The reference
describes moderately differentiated adenocarcinoma arising in Barrett's epithelium;
PathXAI-Report gives moderate differentiation and PLIP-guided Qwen high grade.
\textcolor{black}{For other public grading systems, agreement was 0\% for Fuhrman (n=118), three-level nuclear grade (n=12) and four-level nuclear grade (n=13), and 23.6\% for WHO grade (n=28; \cref{tab:source_fig8:part-5}).}

\manuscriptnativefigure{\textbf{Automated report evaluation on the public cohort.}}{fig:report-public}{%
    \begin{minipage}{180mm}
    \manuscriptpanelblock{5mm}{180.0mm}{\manuscriptpanelmark{pan:report-public-systems}}{Report systems and output scope}{\includegraphics[width=\linewidth]{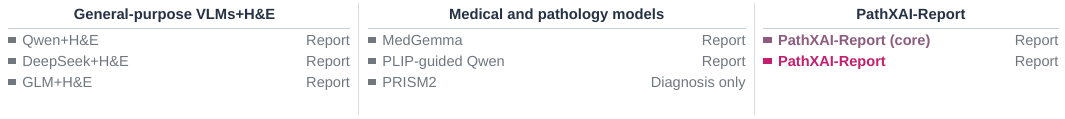}}
    \par\vspace{2mm}
    \manuscriptpanelblock{5mm}{180.0mm}{\manuscriptpanelmark{pan:report-public-compatibility-gains}}{Cancer family match rate and paired gains}{\includegraphics[width=\linewidth]{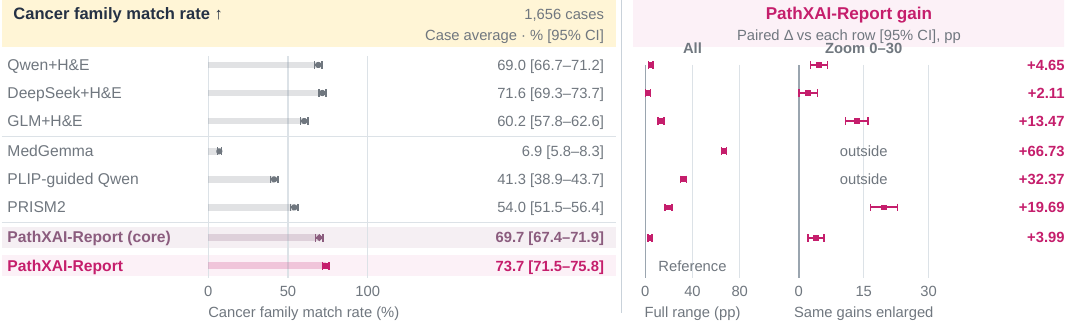}}
    \par\vspace{2mm}
    \manuscriptpanelblock{5mm}{180.0mm}{\manuscriptpanelmark{pan:report-public-grade-results}}{Balanced grade accuracy and prediction distributions}{\includegraphics[width=\linewidth]{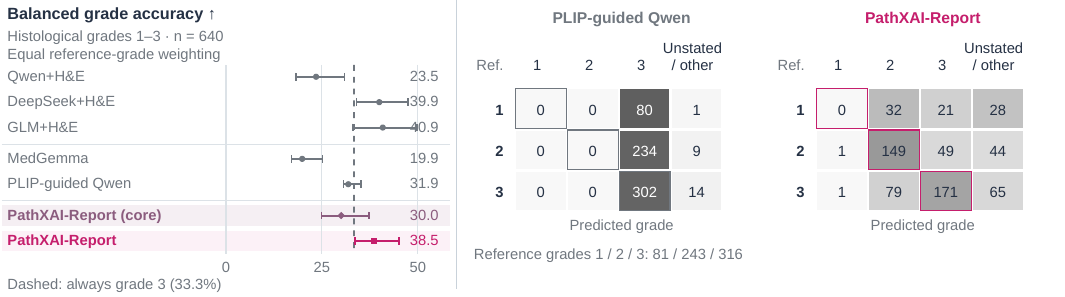}}
    \par\vspace{2mm}
    \manuscriptpanelblock{5mm}{180.0mm}{\manuscriptpanelmark{pan:report-public-finding-results}}{Finding assertion, coverage and contradiction outcomes}{\includegraphics[width=\linewidth]{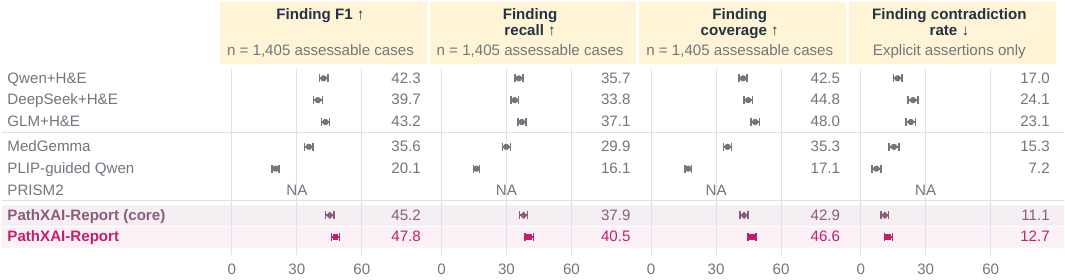}}
    \par\vspace{2mm}
    \manuscriptpanelblock{5mm}{180.0mm}{\manuscriptpanelmark{pan:report-public-examples}}{Selected examples of diagnostic specificity and grade}{\includegraphics[width=\linewidth]{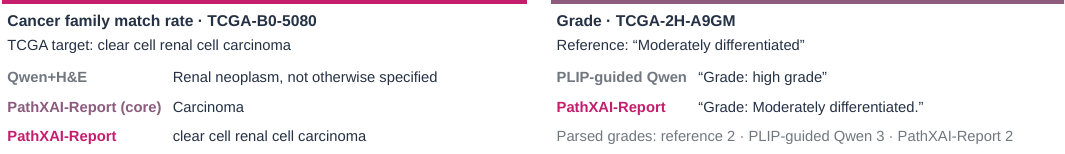}}
    \end{minipage}%
  }{%
\textbf{\panelletterref{pan:report-public-systems} Systems and output scope.} PRISM2 supplies diagnosis only.
\textbf{\panelletterref{pan:report-public-compatibility-gains} Cancer family match rate.} Gains subtract each comparator from PathXAI-Report.
\textbf{\panelletterref{pan:report-public-grade-results} Histological grade.} Balanced grade accuracy weights reference grades equally; missing or incompatible predictions are incorrect. Matrices show raw counts.
\textbf{\panelletterref{pan:report-public-finding-results} Findings and coverage.} Finding contradiction rate requires comparable assertions and has method-specific denominators; interpret it with finding coverage.
\textbf{\panelletterref{pan:report-public-examples} Diagnostic specificity and grade examples.}
Full estimates are in \crefrange{tab:source_fig8:part-1}{tab:source_fig8:part-7}.
(Notation: arrows, preferred directions; dashed line, always predicting grade 3. Whiskers show \textcolor{black}{pointwise 95\% confidence intervals (\cref{supp:report-statistics})}.)
}

\begin{supplementarytables}
\input{supplementary/generated_report_v4_public}
\end{supplementarytables}

{\color{black}
\paragraph{Overall grade accuracy and answering}
\textcolor{black}{In the public evaluation cohort, overall histological-grade accuracy was 222/472 (47.0\%)
for PathXAI-Report and 146/472 (30.9\%) for Qwen+H\&E, counting unanswered cases as incorrect.
Balanced grade accuracy was 37.0\% and 24.1\%, respectively.
Compatible-grade coverage was 375/472 (79.4\%) and 222/472 (47.0\%);
conditional accuracy among answered cases was 222/375 (59.2\%) and 146/222 (65.8\%).}
The conditional comparison concerns different answering subsets and is interpreted alongside
the overall measures (\cref{fig:report-grade-answering,tab:report-grade-answering}).
\manuscriptnativefigure{\textcolor{black}{\textbf{Histological-grade accuracy and answering.}}}{fig:report-grade-answering}{%
  \begin{minipage}{180mm}\color{black}
  \manuscriptpanelblock{5mm}{180mm}{\manuscriptpanelmark{pan:grade-answering-evaluation}}{Public evaluation cohort}{\includegraphics[width=\linewidth]{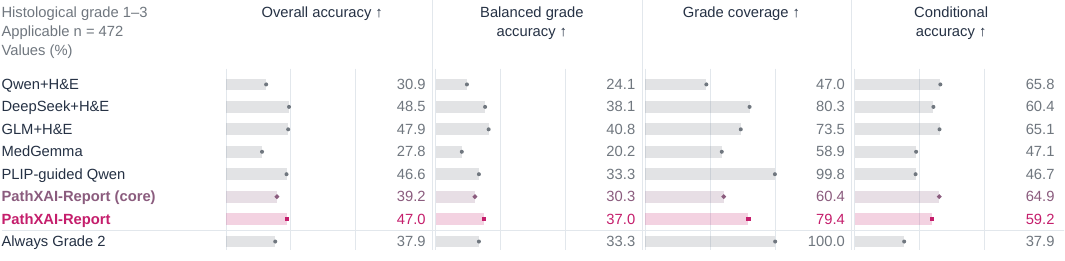}}
  \par\vspace{2mm}
  \manuscriptpanelblock{5mm}{180mm}{\manuscriptpanelmark{pan:grade-answering-development}}{PathXAI-Report pipeline development subset}{\includegraphics[width=\linewidth]{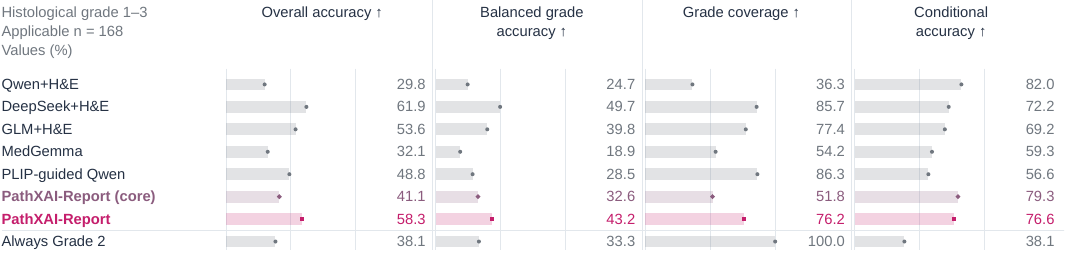}}
  \par\vspace{2mm}
  \manuscriptpanelblock{5mm}{180mm}{\manuscriptpanelmark{pan:grade-answering-p1}}{Private cohort, reference set P1}{\includegraphics[width=\linewidth]{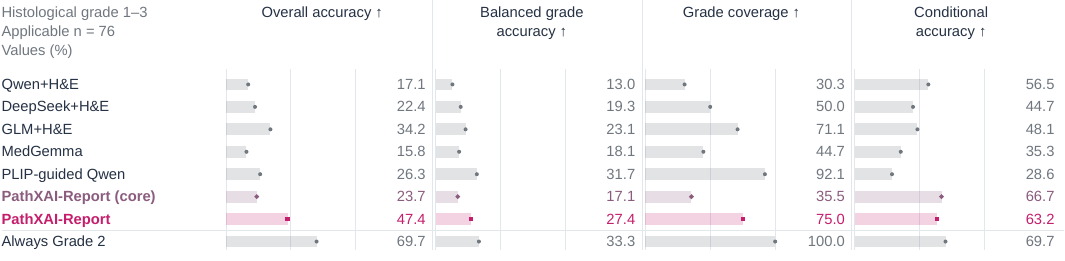}}
  \par\vspace{2mm}
  \manuscriptpanelblock{5mm}{180mm}{\manuscriptpanelmark{pan:grade-answering-p2}}{Private cohort, reference set P2}{\includegraphics[width=\linewidth]{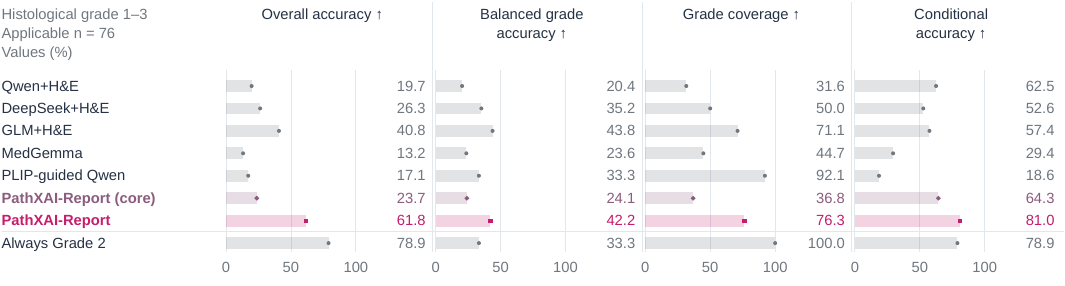}}
  \end{minipage}%
}{\textcolor{black}{Overall accuracy counts unanswered or incompatible grades as incorrect; balanced grade accuracy additionally weights reference grades equally. Coverage is the fraction receiving a compatible grade. Conditional accuracy uses each method's answered cases. All panels use the same percentage scale. Values are descriptive estimates; counts are in \cref{tab:report-grade-answering}. The Grade 2 control always predicts Grade 2. PRISM2 is omitted because it provides diagnosis-only outputs.}}

\begin{supplementarytables}
\begin{longtable}{@{}>{\raggedright\arraybackslash}p{.24\linewidth}>{\raggedright\arraybackslash}p{.18\linewidth}>{\raggedright\arraybackslash}p{.18\linewidth}>{\raggedright\arraybackslash}p{.18\linewidth}>{\raggedright\arraybackslash}p{.12\linewidth}@{}}
\caption{Histological-grade accuracy and answering. Overall accuracy uses all applicable cases; unanswered or incompatible grades count as incorrect. Coverage is a definite answer in the reference grading system; conditional accuracy uses answered cases only. Counts and percentages are shown. Balanced grade accuracy also includes unanswered cases as incorrect and weights reference grades equally. These are descriptive summaries of existing predictions. Conditional accuracy uses method-specific answering subsets. PRISM2 has diagnosis-only outputs. Other grading systems are supplied in Source Data.}\label{tab:report-grade-answering}\\
\toprule Method & Overall accuracy\newline correct/N (\%) & Coverage\newline n/N (\%) & Conditional accuracy\newline correct/n (\%) & Balanced accuracy (\%)\\\midrule
\endfirsthead
\toprule Method & Overall accuracy\newline correct/N (\%) & Coverage\newline n/N (\%) & Conditional accuracy\newline correct/n (\%) & Balanced accuracy (\%)\\\midrule
\endhead
\bottomrule\endlastfoot
\multicolumn{5}{@{}l}{\textbf{\textcolor{black}{PathXAI-Report pipeline development subset} (applicable n=\textcolor{black}{168})}}\\*[2pt]
Qwen+H\&E & \textcolor{black}{50/168 (29.8)} & \textcolor{black}{61/168 (36.3)} & \textcolor{black}{50/61 (82.0)} & \textcolor{black}{24.7}\\*[2pt]
DeepSeek+H\&E & \textcolor{black}{104/168 (61.9)} & \textcolor{black}{144/168 (85.7)} & \textcolor{black}{104/144 (72.2)} & \textcolor{black}{49.7}\\*[2pt]
GLM+H\&E & \textcolor{black}{90/168 (53.6)} & \textcolor{black}{130/168 (77.4)} & \textcolor{black}{90/130 (69.2)} & \textcolor{black}{39.8}\\*[2pt]
MedGemma & \textcolor{black}{54/168 (32.1)} & \textcolor{black}{91/168 (54.2)} & \textcolor{black}{54/91 (59.3)} & \textcolor{black}{18.9}\\*[2pt]
PLIP-guided Qwen & \textcolor{black}{82/168 (48.8)} & \textcolor{black}{145/168 (86.3)} & \textcolor{black}{82/145 (56.6)} & \textcolor{black}{28.5}\\*[2pt]
PathXAI-Report (core) & \textcolor{black}{69/168 (41.1)} & \textcolor{black}{87/168 (51.8)} & \textcolor{black}{69/87 (79.3)} & \textcolor{black}{32.6}\\*[2pt]
PathXAI-Report & \textcolor{black}{98/168 (58.3)} & \textcolor{black}{128/168 (76.2)} & \textcolor{black}{98/128 (76.6)} & \textcolor{black}{43.2}\\
\end{longtable}
\end{supplementarytables}

\paragraph{Ordinal agreement and grading-system availability}
\textcolor{black}{Among compatible histological-grade answers, linear weighted $\kappa$ was 0.266 for
PathXAI-Report and 0.387 for Qwen+H\&E in the public evaluation cohort.
The corresponding values were 0.114 and 0.173 against private reference set P1,
and 0.481 and 0.299 against P2 (\cref{tab:report-grade-agreement}).
The zero Fuhrman and nuclear-grade accuracy in the combined public cohort reflected
the absence of compatible answers in the extracted predictions. Of 118 Fuhrman cases,
PathXAI-Report had no definite grade in 110 and used a different grading system in eight.
The corresponding counts were four and eight among 12 three-level nuclear-grade cases;
all 13 four-level nuclear-grade cases lacked a definite prediction.
These strata therefore indicate limited coverage of the named grading systems;
conditional accuracy is undefined when no compatible prediction is available.
For systems represented by only one or two cases, the wide confidence intervals
reflect sparse observations, and the estimates provide little information about
performance across grades.}
In the private comparison, neither reference set provided an assessable grade for the
62 prostate cases, so these cases did not contribute to grade accuracy.
Source Data provide all grading-system and cancer/organ strata, reference availability
and prediction-system counts.
\begin{supplementarytables}
\begin{longtable}{@{}p{.30\linewidth}*{2}{>{\centering\arraybackslash}p{.16\linewidth}}>{\centering\arraybackslash}p{.30\linewidth}@{}}
\caption{\textcolor{black}{Histological-grade agreement among compatible answers. Linear weighted $\kappa$ uses definite grades 1--3 in the reference grading system. Answered/eligible counts show each method\textquotesingle s coverage; overall accuracy counts unanswered and incompatible grades as incorrect. Brackets show unadjusted 95\% percentile intervals from 4,000 bootstrap draws of answered case pairs. Answering subsets differ between methods. \textcolor{black}{PathXAI-Report pipeline development subset and combined-cohort estimates}, other grading systems and cancer/organ strata are supplied in Source Data. PRISM2 is diagnosis-only.}}\label{tab:report-grade-agreement}\\
\toprule \textcolor{black}{Method} & \textcolor{black}{Answered / eligible $n$} & \textcolor{black}{Overall accuracy (\%)} & \textcolor{black}{Weighted $\kappa$ [95\% CI]} \\ \midrule
\endfirsthead
\multicolumn{4}{@{}l}{\manuscripttablecontinued{}}\\[3pt]
\toprule \textcolor{black}{Method} & \textcolor{black}{Answered / eligible $n$} & \textcolor{black}{Overall accuracy (\%)} & \textcolor{black}{Weighted $\kappa$ [95\% CI]} \\ \midrule
\endhead
\midrule\multicolumn{4}{r@{}}{Continued on next page}\\\endfoot
\bottomrule\endlastfoot
\multicolumn{4}{@{}l}{\textcolor{black}{\textbf{Public evaluation cohort}}}\\*[3pt]
\textcolor{black}{Qwen+H\&E} & \textcolor{black}{\textcolor{black}{222 / 472}} & \textcolor{black}{\textcolor{black}{30.9}} & \textcolor{black}{\textcolor{black}{0.387 [0.288, 0.486]}}\\[2pt]
\textcolor{black}{DeepSeek+H\&E} & \textcolor{black}{\textcolor{black}{379 / 472}} & \textcolor{black}{\textcolor{black}{48.5}} & \textcolor{black}{\textcolor{black}{0.298 [0.220, 0.376]}}\\[2pt]
\textcolor{black}{GLM+H\&E} & \textcolor{black}{\textcolor{black}{347 / 472}} & \textcolor{black}{\textcolor{black}{47.9}} & \textcolor{black}{\textcolor{black}{0.415 [0.328, 0.500]}}\\[2pt]
\textcolor{black}{MedGemma} & \textcolor{black}{\textcolor{black}{278 / 472}} & \textcolor{black}{\textcolor{black}{27.8}} & \textcolor{black}{\textcolor{black}{-0.021 [-0.062, 0.025]}}\\[2pt]
\textcolor{black}{PLIP-guided Qwen} & \textcolor{black}{\textcolor{black}{471 / 472}} & \textcolor{black}{\textcolor{black}{46.6}} & \textcolor{black}{0.000 [0.000, 0.000]}\\[2pt]
\textcolor{black}{PathXAI-Report (core)} & \textcolor{black}{\textcolor{black}{285 / 472}} & \textcolor{black}{\textcolor{black}{39.2}} & \textcolor{black}{\textcolor{black}{0.360 [0.277, 0.446]}}\\[2pt]
\textcolor{black}{PathXAI-Report} & \textcolor{black}{\textcolor{black}{375 / 472}} & \textcolor{black}{\textcolor{black}{47.0}} & \textcolor{black}{\textcolor{black}{0.266 [0.188, 0.342]}}\\
\end{longtable}
\end{supplementarytables}
}

\supplementaryresultsubsection{Public-case membership}{supp:results-public-site-sensitivity}
\Cref{tab:site-prior-sensitivity} records the exclusions applied to the public cohort and
its overlapping evaluation subsets. Cohort construction and screening are described in
\cref{supp:report-cohorts}.
\begin{supplementarytables}
\begin{table}[htbp]
\manuscripttablefont\centering\setlength{\tabcolsep}{2pt}\renewcommand{\arraystretch}{1.08}
\caption{Public-case selection and evaluation subsets. \textcolor{black}{Checks excluded 51 cases whose supplied site descriptions contained disease names and one case with a corrupt source WSI. These exclusions apply to every public report analysis before aggregation and resampling. \textcolor{black}{The input-policy and writing comparisons use all 1,200 public evaluation cases.}} \textcolor{black}{Evaluation subsets overlap.}}\label{tab:site-prior-sensitivity}
\begin{tabular}{@{}p{.43\linewidth}rrr@{}}
\toprule Evaluation set & Before exclusion & Excluded & Evaluated \\ \midrule
Public benchmark & 1,708 & \textcolor{black}{52} & \textcolor{black}{1,656} \\
\textcolor{black}{\textcolor{black}{PathXAI-Report pipeline development subset}} & 508 & \textcolor{black}{52} & \textcolor{black}{456} \\
\textcolor{black}{\textcolor{black}{Public evaluation cohort}} & 1,200 & 0 & 1,200 \\
Four reporting policies & \textcolor{black}{1,200} & \textcolor{black}{0} & \textcolor{black}{1,200} \\
Pathologist ranking and correction assessment & 145 & 13 & 132 \\
HistGen/HistoGPT four-method comparison & 324 & 12 & 312 \\
HistGen three-method comparison & 327 & 13 & 314 \\
Output constraints comparison & 46 & 4 & 42 \\
Region acquisition comparison & 471 & 39 & 432 \\
Writing-evidence comparison & \textcolor{black}{1,200} & \textcolor{black}{0} & \textcolor{black}{1,200} \\
\bottomrule
\end{tabular}
\end{table}
\end{supplementarytables}

{\color{black}
\supplementaryresultsubsection{Report performance after excluding segmentation training cases}{supp:results-report-training-overlap}
TCGA identifiers linked 28 public report cases to segmentation training pools, including
three \textcolor{black}{cases used for PathXAI-Report pipeline development} and 25 cases from the public evaluation cohort.
Excluding the latter left 1,175 evaluation cases. PathXAI-Report retained higher cancer
family match rate and exact diagnosis match rate than Qwen+H\&E, with paired differences
of 4.60~pp and 3.54~pp, respectively. The corresponding differences versus DeepSeek+H\&E
were 0.68~pp and 2.19~pp, with intervals spanning zero
(\cref{tab:report-training-overlap}). Finding F1 and finding recall remained higher,
and finding contradiction rate lower, than both comparators.
These exclusions address identifiable training-pool overlap; membership of a training
pool does not establish that a sample was drawn during training.
\begin{supplementarytables}
\begin{longtable}{@{}p{.30\linewidth}*{2}{>{\centering\arraybackslash}p{.16\linewidth}}>{\centering\arraybackslash}p{.30\linewidth}@{}}
\caption{\textcolor{black}{Report comparisons after excluding cases shared with segmentation training pools. The primary public evaluation includes 1,200 cases; the sensitivity analysis excludes 25 cases with matching TCGA patient identifiers, leaving 1,175. Differences are PathXAI-Report minus the comparator in percentage points. The two $n$ values are the applicable paired counts before and after exclusion. Brackets give unadjusted 95\% intervals from 4,000 paired-case bootstrap draws after exclusion. Positive differences favour PathXAI-Report except for finding contradiction rate, where negative differences favour it. PRISM2 has no finding-level measures.}}\label{tab:report-training-overlap}\\
\toprule \textcolor{black}{Comparator} & \textcolor{black}{Paired $n$ before / after} & \textcolor{black}{Primary difference (pp)} & \textcolor{black}{After exclusion (pp) [95\% CI]} \\ \midrule
\endfirsthead
\multicolumn{4}{@{}l}{\manuscripttablecontinued{}}\\[3pt]
\toprule \textcolor{black}{Comparator} & \textcolor{black}{Paired $n$ before / after} & \textcolor{black}{Primary difference (pp)} & \textcolor{black}{After exclusion (pp) [95\% CI]} \\ \midrule
\endhead
\midrule\multicolumn{4}{r@{}}{Continued on next page}\\\endfoot
\bottomrule\endlastfoot
\multicolumn{4}{@{}l}{\textcolor{black}{\textbf{Cancer family match rate}}}\\*[3pt]
\textcolor{black}{Qwen+H\&E} & \textcolor{black}{1200 / 1175} & \textcolor{black}{+4.42} & \textcolor{black}{+4.60 [+2.21, +7.06]}\\[2pt]
\textcolor{black}{DeepSeek+H\&E} & \textcolor{black}{1200 / 1175} & \textcolor{black}{+0.58} & \textcolor{black}{+0.68 [-1.96, +3.32]}\\[2pt]
\textcolor{black}{GLM+H\&E} & \textcolor{black}{1200 / 1175} & \textcolor{black}{+10.67} & \textcolor{black}{+10.30 [+7.06, +13.53]}\\[2pt]
\textcolor{black}{MedGemma} & \textcolor{black}{1200 / 1175} & \textcolor{black}{+65.83} & \textcolor{black}{+65.70 [+62.98, +68.43]}\\[2pt]
\textcolor{black}{PLIP-guided Qwen} & \textcolor{black}{1200 / 1175} & \textcolor{black}{+26.83} & \textcolor{black}{+27.32 [+24.60, +30.04]}\\[2pt]
\textcolor{black}{PRISM2} & \textcolor{black}{1200 / 1175} & \textcolor{black}{+19.25} & \textcolor{black}{+19.23 [+15.40, +23.06]}\\[2pt]
\textcolor{black}{PathXAI-Report (core)} & \textcolor{black}{1200 / 1175} & \textcolor{black}{+4.33} & \textcolor{black}{+4.17 [+2.04, +6.30]}\\
\end{longtable}
\end{supplementarytables}
}

\supplementaryresultsubsection{Finding assertion, coverage and contradiction}{supp:results-public-findings}
Finding-level coverage and contradiction are reported together in
\cref{tab:contradiction_coverage}, including each method's number of reports with no
comparable assertion.
Comparing PathXAI-Report with PathXAI-Report (core), the paired difference in finding contradiction rate was +\FullVsCorePublicContradiction{}~pp
(95\% CI, \FullVsCorePublicContradictionCI; n=\FullVsCorePublicContradictionN{} cases
with assessable assertions in both methods).
Across the broader set of applicable diagnostic, grade,
invasion and necrosis fields, PathXAI-Report supplied definite answers in \PublicCoverageFull\%
of fields, irrespective of correctness (\cref{tab:source_fig8:part-3}).
\begingroup
\manuscripttablefont
\renewcommand{\arraystretch}{1.08}
\setlength{\LTcapwidth}{\linewidth}
\setlength{\LTpre}{7pt}\setlength{\LTpost}{10pt}
\setlength{\tabcolsep}{3pt}
\input{supplementary/generated_report_v4_findings.tex}
\endgroup

{\color{black}
\paragraph{Fixed-reference and shared-finding comparisons}
In the public evaluation cohort, contradictions per documented finding ($E/D$) averaged 5.5\% for PathXAI-Report and 6.3\% for Qwen+H\&E (\cref{tab:finding-fixed-denominator}). The paired difference was -0.8~pp, with an interval spanning zero (\cref{tab:finding-matched-contrasts}). On identical finding types asserted by both methods, PathXAI-Report had fewer contradictions, with a paired difference of -6.0~pp across 656 cases. Against private reference sets P1/P2, $E/D$ was 17.7\% / 14.2\%, compared with 11.0\% / 8.1\% for Qwen+H\&E; the shared-finding comparisons also favoured Qwen+H\&E. Thus the direction of the private contradiction comparison persists when both methods are scored on the same asserted findings.

Public references predominantly recorded positive findings (\cref{tab:finding-reference-polarity}). Always asserting presence yielded a finding F1 of 86.3\%, with 100\% coverage and an $E/D$ of 13.7\%. This control shows that finding F1 can reward indiscriminate assertion when undocumented types remain unscored. Finding agreement therefore needs to be interpreted with contradiction and coverage; it does not measure the truth of assertions absent from the reference.
\begin{supplementarytables}
\input{supplementary/generated_finding_fixed_denominator}
\input{supplementary/generated_finding_matched_contrasts}
\input{supplementary/generated_finding_reference_polarity}
\end{supplementarytables}
}

\supplementaryresultsubsection{Comparisons with HistGen and HistoGPT}{supp:results-public-native}
\textcolor{black}{\Cref{fig:report-native-public} presents both the \EvaluationNativeN{}-case shared evaluation comparison and the 312-case combined-cohort comparison. On the shared evaluation cases, PathXAI-Report had higher cancer family match rate, finding F1 and finding recall than HistGen and HistoGPT. Finding F1 reached \EvaluationNativeFullFindingFOneProse\%, compared with \EvaluationNativeHistGenFindingFOneProse\% for HistGen and \EvaluationNativeHistoGPTFindingFOneProse\% for HistoGPT. HistoGPT's lower finding contradiction rate accompanied lower finding coverage (\cref{tab:report-primary-native}).}
\textcolor{black}{For the combined cohort,}
\crefrange{tab:report-native-public:part-1}{tab:report-native-public:part-13} give all estimates and the separate 314-case HistGen
sensitivity analysis. HistoGPT's non-skin evaluation is outside its original dermatopathology domain.
\manuscriptnativefigure{\textbf{Dedicated report generators on shared public cases.}}{fig:report-native-public}{%
  \begin{minipage}{180mm}
  \manuscriptpanelblock{5mm}{180mm}{\manuscriptpanelmark{pan:report-native-public-eligibility}}{Eligibility and completed reports}{\includegraphics[width=\linewidth]{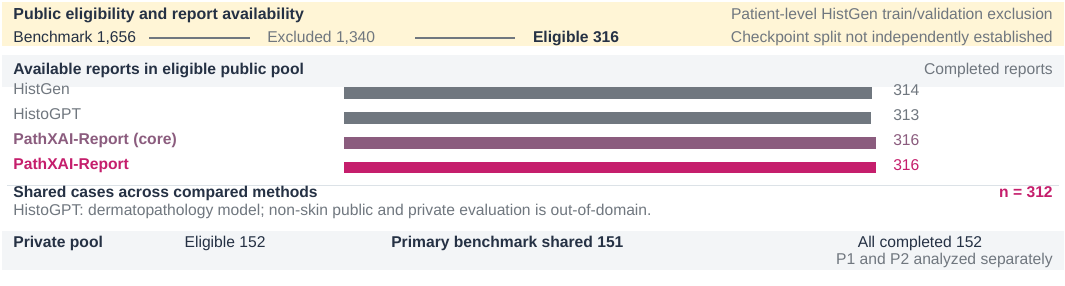}}
  \par\vspace{2mm}
  \manuscriptpanelblock{5mm}{180mm}{\manuscriptpanelmark{pan:report-native-public-evaluation}}{Public evaluation cohort}{\includegraphics[width=\linewidth]{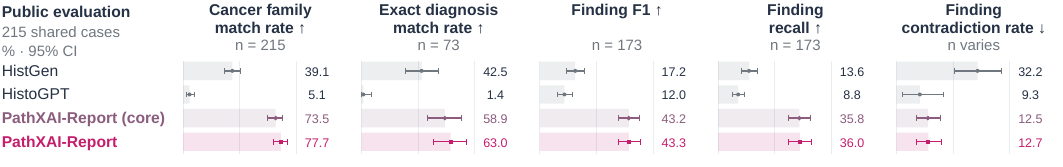}}
  \par\vspace{2mm}
  \manuscriptpanelblock{5mm}{180mm}{\manuscriptpanelmark{pan:report-native-public-performance}}{Combined public cohort}{\includegraphics[width=\linewidth]{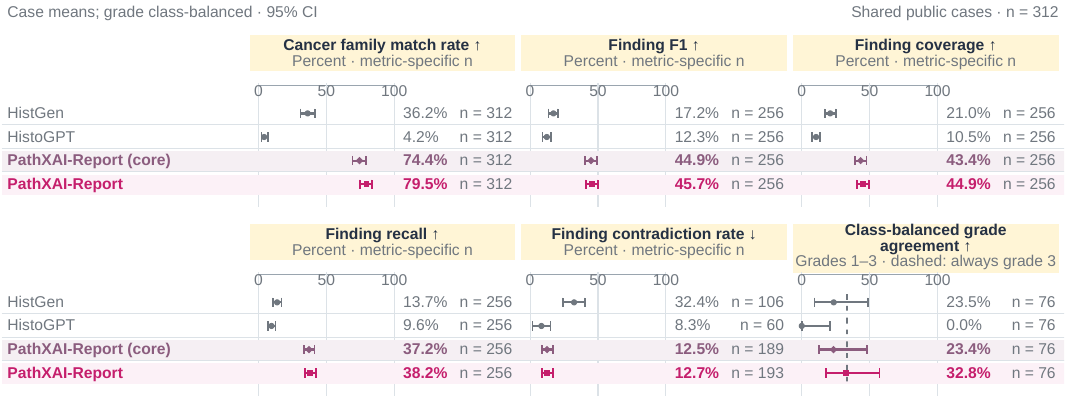}}
  \end{minipage}%
}{%
\textbf{\panelletterref{pan:report-native-public-eligibility} Eligibility.} Patients in HistGen's released training or validation lists are excluded; the checkpoint's exact training split is unverified. Evaluation uses cases with reports from every compared method. Non-skin cases fall outside HistoGPT's dermatopathology development domain.
\textcolor{black}{\textbf{\panelletterref{pan:report-native-public-evaluation} Public evaluation cohort.} The comparison uses \EvaluationNativeN{} cases shared by all four methods from the 1,200-case evaluation cohort; full estimates are in \cref{tab:report-primary-native}.}
\textbf{\panelletterref{pan:report-native-public-performance} \textcolor{black}{Combined public cohort.}} \textcolor{black}{This 312-case comparison also includes \textcolor{black}{cases used for PathXAI-Report pipeline development}.} Ordinary metrics weight applicable cases equally; balanced grade accuracy weights reference grades equally. Finding contradiction rate is interpreted with finding coverage and recall. Full estimates and the HistGen-only sensitivity analysis are in \crefrange{tab:report-native-public:part-1}{tab:report-native-public:part-13}.
(Notation: arrows, preferred directions; dashed line, always predicting grade 3. Whiskers show \textcolor{black}{pointwise 95\% confidence intervals (\cref{supp:report-statistics})}.)
}

\begin{supplementarynativetables}
\input{supplementary/generated_report_native_public}
\end{supplementarynativetables}

\supplementaryresultsection{Automated report evaluation on the private cohort}{supp:private-report-results}
\supplementaryresultsubsection{Evaluation by reference set}{supp:results-private-full}
The primary comparison uses the 151 cases with reports from all methods in the 152-case private
cohort. Reference sets P1 and P2 each contain one report per case from one of the two pathologists.
\Cref{fig:report-private} shows diagnostic and finding performance and selected cases.
\Cref{fig:report-private-metrics} compares finding coverage, report field completeness and histological-grade agreement across all ten methods;
\crefrange{tab:source_fig9:part-1}{tab:source_fig9:part-11} give the complete estimates against each reference set and a sensitivity
analysis using each method's available reports across all 152 cases.
\textcolor{black}{Finding comparisons with a fixed reference denominator and with identical asserted finding types
are reported in \cref{tab:finding-fixed-denominator,tab:finding-matched-contrasts}.}

\textcolor{black}{The reference sets agreed on specific diagnosis in 122/127 comparable cases ($\kappa=0.94$), necrosis in 126/135 ($\kappa=0.58$), and grade in 45/75 (linear weighted $\kappa=0.04$; \cref{tab:private-reference-agreement}). Of the 30 grade disagreements, 27 were between Grade 2 and Grade 3. Comparable grades came from 74 breast cases and one colorectal case. Model performance is reported separately against P1 and P2 to retain these reference differences.}

\begin{supplementarytables}
\input{supplementary/generated_private_reference_agreement}
\end{supplementarytables}

Cancer family match rate uses \PrivateOneCancerFullN{} applicable cases against P1
(74 invasive-breast, 46 prostate and one colorectal cancer reference) and
\PrivateTwoCancerFullN{} against P2 (76, 46 and one, respectively).
PathXAI-Report achieved \PrivateOneCancerFull\% (95\% CI, \PrivateOneCancerFullCI) and
\PrivateTwoCancerFull\% (\PrivateTwoCancerFullCI), compared with \PrivateOneCancerQwen\%
and \PrivateTwoCancerQwen\% for Qwen+H\&E (\cref{pan:public-summary-private-performance}).
The accompanying Source Data file \texttt{private\_reference\_cancer\_compatibility.csv}
provides all ten methods' estimates, intervals and denominators.

\textcolor{black}{For exact diagnosis matching, PathXAI-Report had fewer definite
non-specific diagnoses than Qwen+H\&E against P1 (20 versus 36) and P2 (16 versus 34).
Different specific diagnoses occurred equally often against P1 (22 each) and slightly
more often for PathXAI-Report against P2 (26 versus 24). The corresponding match counts
were 91 versus 73 among 135 applicable P1 references and 92 versus 74 among 136
applicable P2 references (Source Data, \texttt{diagnostic\_outcomes}).}

\supplementaryresultsubsection{Contrasting diagnostic and grade assignments}{supp:results-private-cases}
{\color{black}
\paragraph{Malignant predictions in non-malignant reference cases}
The existing extraction classified all 151 PathXAI-Report outputs as malignant,
including all 13 reference-benign cases in each reference set; Qwen+H\&E had malignant
predictions in 11 of these 13 cases.
\Cref{fig:report-benign-calls,tab:report-benign-calls} also report borderline reference diagnoses separately.
Reference-benign cases were confined to prostate and colorectal samples, and all
reference-borderline cases were colorectal. \Cref{tab:private-malignancy-confusion}
reports all extracted categories, with missing, uncertain and conflicting predictions
kept separate; organ-stratified counts are supplied in Source Data.
These reference-defined comparisons identify overcalls for review; they are not an
independently adjudicated estimate of clinical specificity.
\manuscriptnativefigure{\textcolor{black}{\textbf{Malignant predictions in non-malignant reference cases.}}}{fig:report-benign-calls}{%
  \begin{minipage}{180mm}\color{black}
  \includegraphics[width=\linewidth]{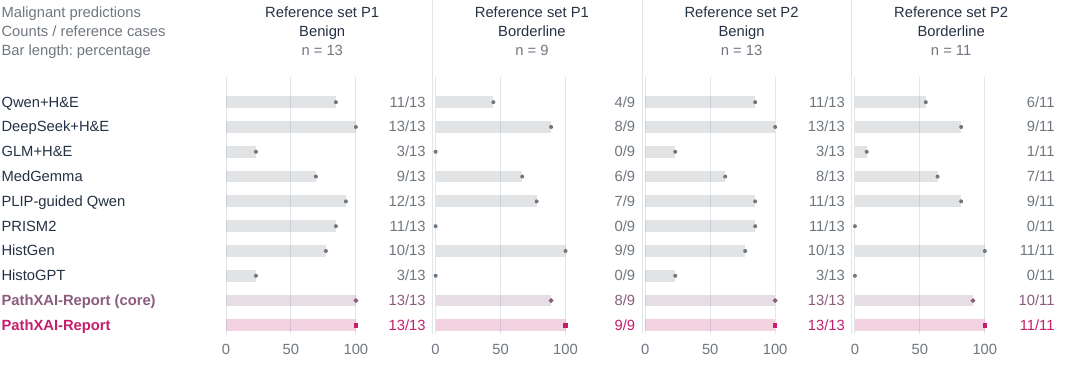}
  \end{minipage}%
}{\textcolor{black}{Bars show the percentage classified as malignant for all ten methods; labels give malignant predictions/reference cases. Benign and borderline categories are shown separately for P1 and P2, the two reference sets for the same private cohort. All reference-defined cases remain in the denominator, including missing, uncertain or conflicting predictions. These descriptive counts use the existing automated extraction and have not undergone independent adjudication. Detailed counts are in \cref{tab:report-benign-calls,tab:private-malignancy-confusion}.}}

\begin{supplementarytables}
\begin{longtable}{@{}p{.34\linewidth}rrrr@{}}
\caption{Definite malignant predictions in private cases classified as benign or borderline in each reference set. Each cell gives malignant predictions/reference cases. The denominator includes all cases in the stated reference category; missing or uncertain predictions remain in the denominator and are separately recorded in Source Data. Categories use the existing automated extraction rules and have not undergone independent adjudication. P1 and P2 concern the same 151-case cohort and must not be added as independent samples.}\label{tab:report-benign-calls}\\
\toprule Method & P1 benign & P1 borderline & P2 benign & P2 borderline\\\midrule
\endfirsthead
\toprule Method & P1 benign & P1 borderline & P2 benign & P2 borderline\\\midrule
\endhead
\bottomrule\endlastfoot
Qwen+H\&E & 11/13 & 4/9 & 11/13 & 6/11\\[2pt]
DeepSeek+H\&E & 13/13 & 8/9 & 13/13 & 9/11\\[2pt]
GLM+H\&E & 3/13 & 0/9 & 3/13 & 1/11\\[2pt]
MedGemma & 9/13 & 6/9 & 8/13 & 7/11\\[2pt]
PLIP-guided Qwen & 12/13 & 7/9 & 11/13 & 9/11\\[2pt]
PRISM2 & 11/13 & 0/9 & 11/13 & 0/11\\[2pt]
HistGen & 10/13 & 9/9 & 10/13 & 11/11\\[2pt]
HistoGPT & 3/13 & 0/9 & 3/13 & 0/11\\[2pt]
PathXAI-Report (core) & 13/13 & 8/9 & 13/13 & 10/11\\[2pt]
PathXAI-Report & 13/13 & 9/9 & 13/13 & 11/11\\[2pt]
\end{longtable}
\input{supplementary/generated_private_malignancy_confusion}
\end{supplementarytables}
}

Two cases illustrate subtype reporting and a diagnostic overcall
(\cref{pan:report-private-examples}). In C035, Qwen+H\&E and the core configuration also mention
ductal carcinoma of no special type in summary prose. The core gives Grade 2, like PathXAI-Report
and both pathologists' reports, whereas Qwen+H\&E gives Grade 3. The references disagree on necrosis;
PLIP-guided Qwen has no parsed final diagnosis or grade.

In the colorectal case C006, all displayed
systems score zero for diagnosis and subtype against both pathologists' reports. PathXAI-Report labels
colorectal adenocarcinoma, whereas the references describe a sessile serrated lesion with
dysplasia or low-grade intraepithelial neoplasia. Both support a precancerous classification
but differ on whether malignancy can be assessed.

\manuscriptnativefigure{\textbf{Automated report evaluation using two pathologist reference sets.}}{fig:report-private}{%
    \begin{minipage}{180mm}
\manuscriptpanelblock{4.2mm}{180.0mm}{\manuscriptpanelmark{pan:report-private-performance}}{Diagnostic and finding agreement by reference set}{\includegraphics[width=\linewidth]{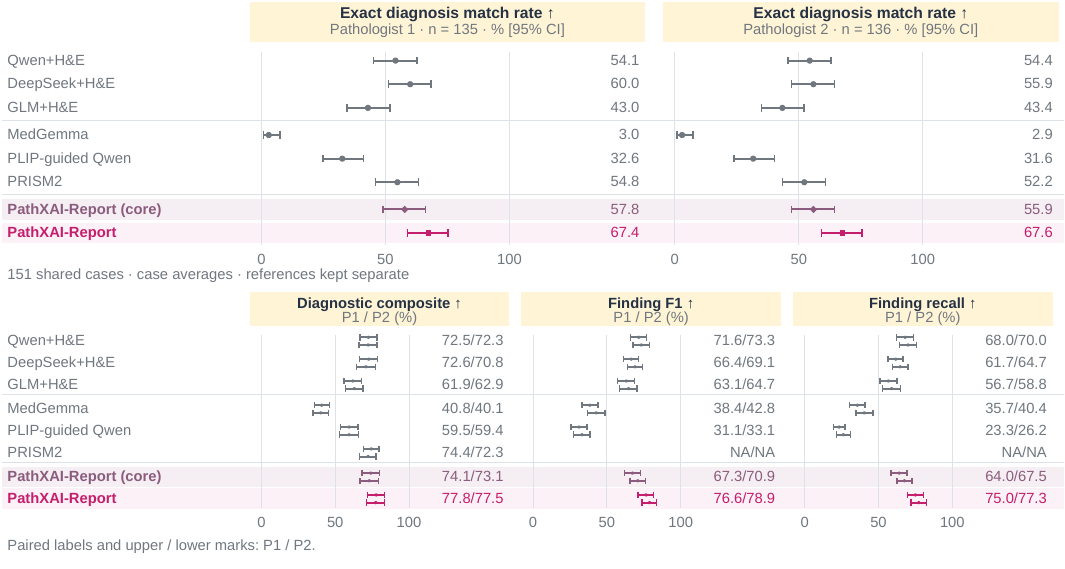}}
    \par\vspace{1.5mm}
    \manuscriptpanelblock{4.2mm}{180.0mm}{\manuscriptpanelmark{pan:report-private-examples}}{Private-cohort report examples}{\includegraphics[width=\linewidth]{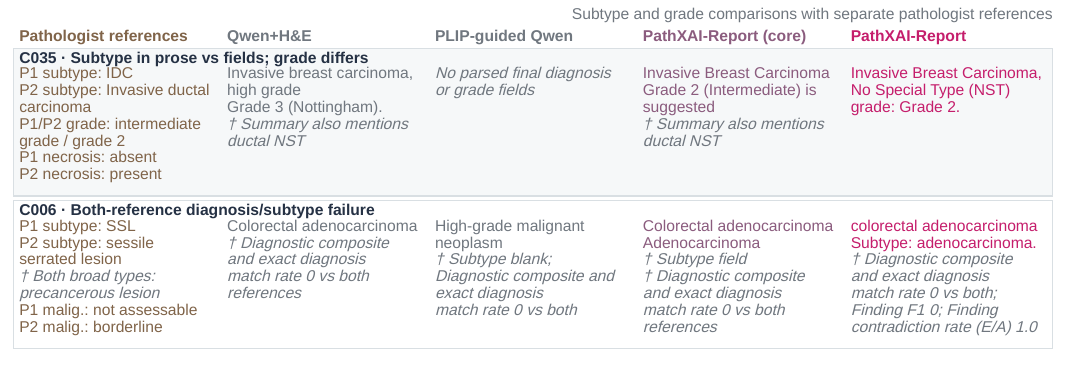}}
    \end{minipage}%
  }{%
\textbf{\panelletterref{pan:report-private-performance} Automated report evaluation.} All methods use shared cases, evaluated separately against P1 and P2. Exact diagnosis match rate includes assessable malignant and benign diagnoses. Cases receive equal weight; PRISM2 supplies diagnosis only. Available-case and grade estimates are in \crefrange{tab:source_fig9:part-1}{tab:source_fig9:part-11}.
\textbf{\panelletterref{pan:report-private-examples} Report examples.} Diagnostic specificity, grade differences and a diagnostic overcall relative to both reference sets.
(Notation: P1/P2, pathologists' reference sets; $\uparrow$, higher preferred. Whiskers show \textcolor{black}{pointwise 95\% confidence intervals (\cref{supp:report-statistics})}.)
}

\FloatBarrier
\manuscriptnativefigure{\textbf{Private-cohort report coverage and grade accuracy.}}{fig:report-private-metrics}{%
  \begin{minipage}{180mm}
  \includegraphics[width=\linewidth]{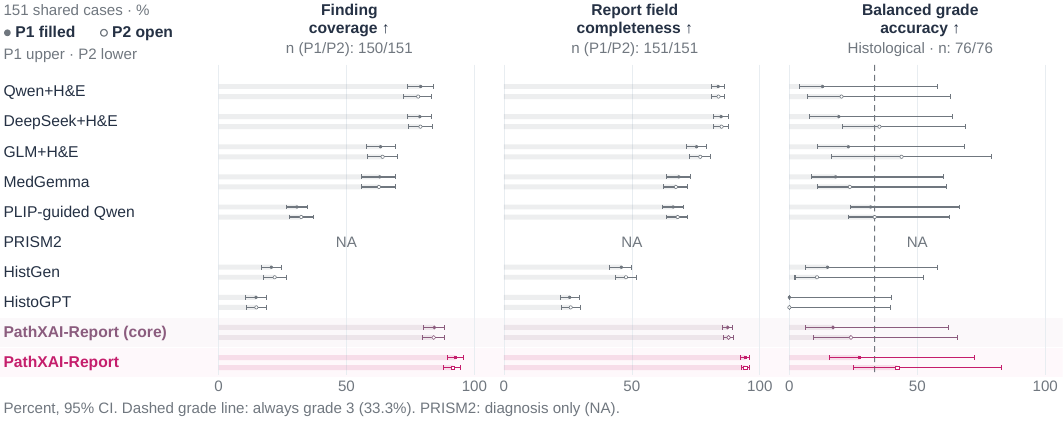}
  \end{minipage}%
}{%
Methods use shared cases and separate pathologist reference sets. Finding coverage concerns documented binary findings; report field completeness concerns applicable structured fields. Balanced grade accuracy weights histological grades equally, counting missing or incompatible predictions as incorrect. These measures are inapplicable to PRISM2's diagnosis-only outputs. Non-skin cases fall outside HistoGPT's dermatopathology domain. Definitions and full estimates are in \cref{supp:report-grade-findings} and \crefrange{tab:source_fig9:part-1}{tab:source_fig9:part-11}.
(Notation: filled upper marks, P1; outlined lower marks, P2; NA, not applicable; dashed line, always predicting grade 3. Bars extend from zero to the estimate; whiskers show \textcolor{black}{pointwise 95\% confidence intervals}.)
}

\begin{supplementarytables}
\input{supplementary/generated_report_v4_private}
\end{supplementarytables}

\supplementaryresultsubsection{Comparisons with HistGen and HistoGPT}{supp:results-private-native}
The HistGen/HistoGPT comparison uses the same 151 cases and separate reference sets
(\textcolor{black}{\cref{fig:report-native-private}}); full estimates and the 152-case sensitivity analysis
appear in \crefrange{tab:report-native-private:part-1}{tab:report-native-private:part-8}.
\textcolor{black}{PathXAI-Report achieved higher cancer family match rate, exact diagnosis match rate, finding F1 and finding recall than both report generators against each reference set.}
\manuscriptnativefigure{\textbf{Automated report evaluation against dedicated report generators on shared private cases.}}{fig:report-native-private}{%
  \begin{minipage}{180mm}
  \manuscriptpanelblock{5mm}{180mm}{\manuscriptpanelmark{pan:report-native-private-performance}}{Comparison against separate pathologist references}{\includegraphics[width=\linewidth]{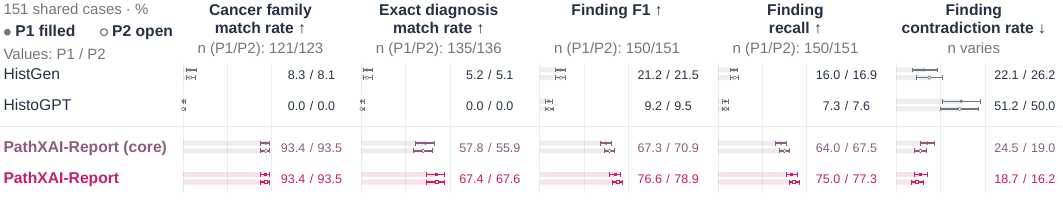}}
  \end{minipage}%
}{%
\textcolor{black}{\textbf{\panelletterref{pan:report-native-private-performance} Diagnostic and morphological comparison.} All methods use the same 151 shared cases, evaluated separately against reference sets P1 and P2. Cancer family match rate uses definite cancer-family diagnoses; exact diagnosis match rate also includes applicable benign diagnoses. Finding contradiction rate is interpreted with finding coverage and recall. Non-skin cases are outside HistoGPT's dermatopathology development domain. Finding coverage, grading results and the 152-case sensitivity analysis are in \crefrange{tab:report-native-private:part-1}{tab:report-native-private:part-8}; cancer family match estimates are in Source Data, \texttt{private\_reference\_cancer\_compatibility.csv}.
(Notation: P1/P2, pathologists' reference sets; filled/open marks, P1/P2. Bars show point estimates; whiskers show pointwise 95\% confidence intervals. Axis ranges vary.)}
}

\begin{supplementarynativetables}
\input{supplementary/generated_report_native_private}
\end{supplementarynativetables}

\supplementaryresultsection{Pathologist ranking and correction assessment and report-content checks}{supp:pathologist-results}
\supplementaryresultsubsection{Report preference, corrections and field coverage}{supp:results-pathologist-assessment}
\Cref{fig:report-physician,tab:source_fig10} report preference, correction categories,
original concern flags and report-content checks for all 264 ratings per method.

The two readers differed in both correction requirements and relative report preference.
Reader 1 rated 97 of 132 PathXAI-Report reports as needing no or only minor correction
(73.5\%), compared with 74 for the core method (56.1\%) and 70 for Qwen+H\&E (53.0\%).
Reader 2 gave this rating to 123 PathXAI-Report reports (93.2\%), 122 core reports
(92.4\%) and 123 Qwen+H\&E reports (93.2\%).
Reader 1 assigned lower mean ranks to the core method and Qwen+H\&E (1.80 and 1.80)
than to PathXAI-Report (1.92), whereas Reader 2 assigned the lowest mean rank to
PathXAI-Report (1.36, versus 1.40 for core and 1.46 for Qwen+H\&E).

\textcolor{black}{Correction-category agreement was limited, with linear weighted $\kappa$ ranging from 0.017 to 0.077 across methods (\cref{tab:reader-agreement}). Reader 1 assigned no correction to 0--1 reports per method, compared with 103--119 for Reader 2. For the five-method preference rankings, mean Kendall $\tau_b$ was 0.52 (95\% CI, 0.46--0.57) across 123 cases with estimable rankings; 9 cases had all methods tied by at least one reader.}

\textcolor{black}{On the same 132 reader-assessed cases, MedGemma received no/minor-correction ratings in 232/264 assessments (87.9\%), whereas its cancer family match rate was 14/132 (10.6\%) and exact diagnosis match rate was 4/99 (4.0\%) (\cref{tab:reader-diagnostic-alignment}). Among its no/minor ratings with applicable targets, 27/232 were linked to a matching cancer family and 7/168 to a matching exact diagnosis. The original Diagnosis fields contained only ``carcinoma'' in 57 reports and ``neoplasm'' in 32; these broad labels did not meet the cancer-family matching rules. Thus, a low correction requirement did not ensure diagnostic agreement.}

PathXAI-Report more often combined a no/minor-correction rating
with complete field coverage than PathXAI-Report (core) or Qwen+H\&E. Because reports can require
few corrections while leaving fields unanswered, this analysis combined the rating with
checks for output failure and definite answers to applicable diagnostic, grade, invasion
and necrosis fields (\cref{supp:rating-protocol,supp:rating-checks,tab:source_fig10}).
Reports met all three criteria in
\ReaderCompleteFullSuccess\% of PathXAI-Report assessments
(\ReaderCompleteFullSuccessCount/\ReaderCompleteFullN{} of all assessments; 95\% CI,
\ReaderCompleteFullSuccessCI), versus \ReaderCompleteCoreSuccess\% for PathXAI-Report (core),
\ReaderCompleteQwenSuccess\% for Qwen+H\&E and \ReaderCompleteMedGemmaSuccess\% for MedGemma
(\ReaderCompleteMedGemmaSuccessCount/\ReaderCompleteMedGemmaN;
\ReaderCompleteMedGemmaSuccessCI). The complementary failure rates were
\ReaderCompleteFullFailure\% for PathXAI-Report and \ReaderCompleteMedGemmaFailure\% for MedGemma,
counting overlapping failures once. The field check measured whether fields were answered,
not whether answers were correct; clinically appropriate abstentions also counted as
incomplete. These automated checks used the original ratings without additional
pathologist review.

\supplementaryresultsubsection{Observed output-screening findings}{supp:results-pathologist-screening}
Among the reviewed texts, none was empty and no recognized diagnosis section was empty.
The output check identified one MedGemma report with ``not specified'' and no Qwen+H\&E
report explicitly stating that no diagnosis was generated. Ten PLIP-guided Qwen reports
lacked a recognized diagnosis heading and were flagged for format review. This alone did not
count as a missing diagnosis; the screen did not assess whether their provisional narrative
impressions provided an adequate final diagnosis
(\cref{supp:rating-checks}).

\manuscriptnativefigure{\textbf{Pathologist ranking and correction assessment of generated reports.}}{fig:report-physician}{%
    \begin{minipage}{180mm}
    \manuscriptpanelblock{5mm}{180.0mm}{\manuscriptpanelmark{pan:report-physician-review-design}}{Ranking and correction assessment with anonymous method labels}{\includegraphics[width=\linewidth]{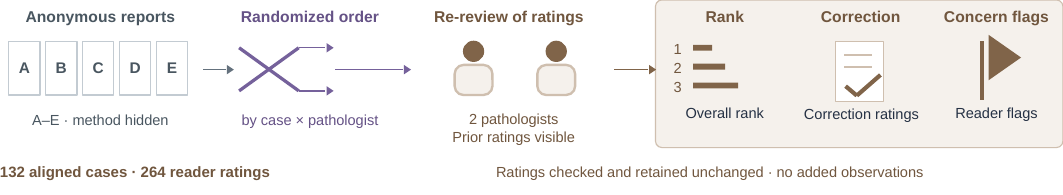}}
    \par\vspace{2mm}
    \manuscriptpanelblock{5mm}{180.0mm}{\manuscriptpanelmark{pan:report-physician-correction-ratings}}{Correction ratings with output and field-coverage checks}{\includegraphics[width=\linewidth]{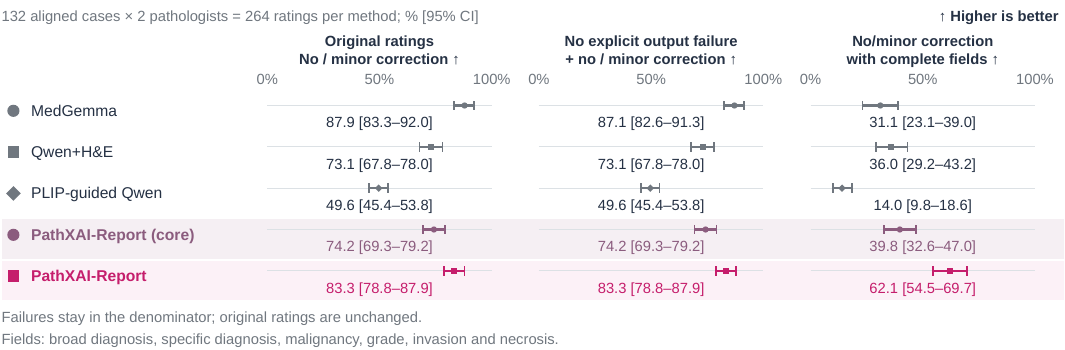}}
    \end{minipage}%
  }{%
\textbf{\panelletterref{pan:report-physician-review-design} Assessment design.} \textcolor{black}{Reports were presented under anonymous method labels in randomized order; prior ratings remained visible and were retained during re-review.}
\textbf{\panelletterref{pan:report-physician-correction-ratings} Correction ratings and content checks.} Cumulative checks assess output failures and definite answers in applicable reference fields. Field coverage measures answer presence, not correctness. Checks retain all original ratings and weights without additional pathologist ratings. Procedures are in \cref{supp:rating-protocol,supp:rating-checks}; rating categories and concern flags are in \cref{tab:source_fig10}.
(Notation: arrows, preferred directions. Whiskers show unadjusted 95\% case-cluster bootstrap intervals, retaining both readers and all methods.)
}

\begin{supplementarytables}
\input{supplementary/generated_human_ratings}
\input{supplementary/generated_reader_agreement}
\input{supplementary/generated_reader_diagnostic_alignment}
\end{supplementarytables}

\supplementaryresultsection{Tissue evidence and its use in reporting}{supp:report-policy-results}
\supplementaryresultsubsection{Direct-input policy comparison}{supp:results-policies-input}
The four-policy comparison uses \textcolor{black}{the same 1,200 public evaluation cases as \cref{fig:public-summary}}
(\cref{fig:report-input}); \crefrange{tab:source_fig7:part-1}{tab:source_fig7:part-2} give all estimates and applicable counts.
All four receive known-site context.

\textcolor{black}{H\&E with tissue statistics yielded small increases in cancer family match rate and
exact diagnosis match rate over H\&E alone, with both intervals including zero.
Among 604 applicable cases, tissue statistics corrected 38 previously mismatched diagnoses
and changed 25 correct predictions to incorrect ones; overlays gave 39 and 56 such changes,
respectively, alongside a lower finding-F1 point estimate than H\&E alone.}
The diagnostic composite (cohort target)
combines broad-diagnosis agreement, cancer family match rate and malignancy agreement
(\cref{supp:report-scoring}). \textcolor{black}{Tissue statistics changed this score little, whereas overlays lowered it.}
\textcolor{black}{In the \textcolor{black}{PathXAI-Report pipeline development subset} sensitivity analysis, including final-summary diagnoses reduced the tissue-statistics gain in exact diagnosis match rate to}
\ScopeStatisticsVsHE{}~pp (95\% CI, \ScopeStatisticsVsHECI;
\cref{tab:report-scope-sensitivity:part-2}).

PathXAI-Report had higher finding agreement and report field completeness than direct tissue-statistics or overlay input. \textcolor{black}{Its diagnostic-composite intervals excluded zero for both comparisons.} The
diagnostic-composite difference was
\FullVsStatisticsDiagnostic{}~pp over statistics
(95\% CI, \FullVsStatisticsDiagnosticCI) and \FullVsOverlaysDiagnostic{}~pp over overlays
(\FullVsOverlaysDiagnosticCI). \textcolor{black}{Cancer family match rate also increased over statistics}
(\FullVsStatisticsCancer{}~pp; \FullVsStatisticsCancerCI).
Report field completeness reached \PolicyCoverageFull\%, compared with
\PolicyCoverageStatistics\% for statistics and \PolicyCoverageOverlays\% for overlays.
\manuscriptnativefigure{\textbf{Tissue-input and reporting-workflow comparisons.}}{fig:report-input}{%
    \begin{minipage}{180mm}
    \manuscriptpanelblock{5mm}{180.0mm}{\manuscriptpanelmark{pan:report-input-policies}}{Four reporting policies on the public evaluation cohort}{\includegraphics[width=\linewidth]{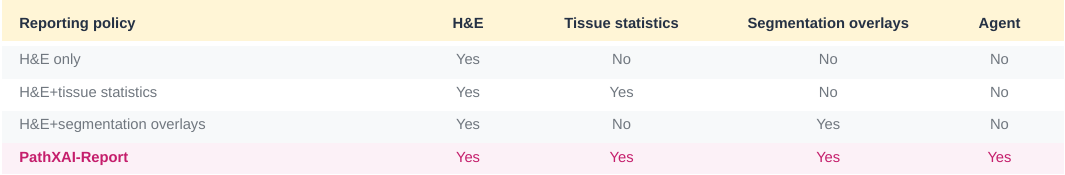}}
    \par\vspace{2mm}
    \manuscriptpanelblock{5mm}{180.0mm}{\manuscriptpanelmark{pan:report-input-shared-case-performance}}{Input-policy comparison}{\includegraphics[width=\linewidth]{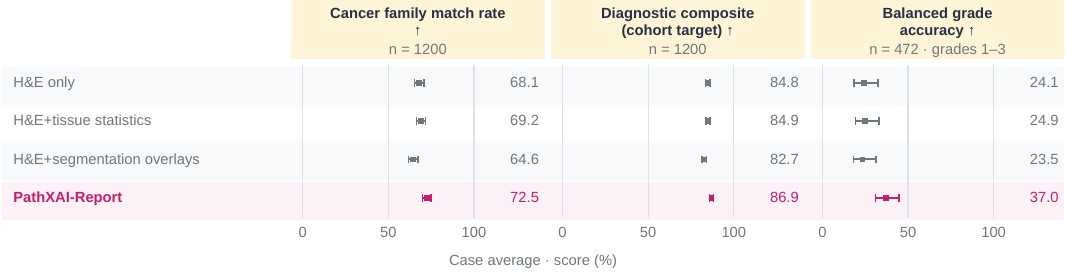}}
    \par\vspace{2mm}
    \manuscriptpanelblock{5mm}{180.0mm}{\manuscriptpanelmark{pan:report-input-tissue-input-gains}}{Layer 1 | Direct tissue-evidence policies versus H\&E}{\includegraphics[width=\linewidth]{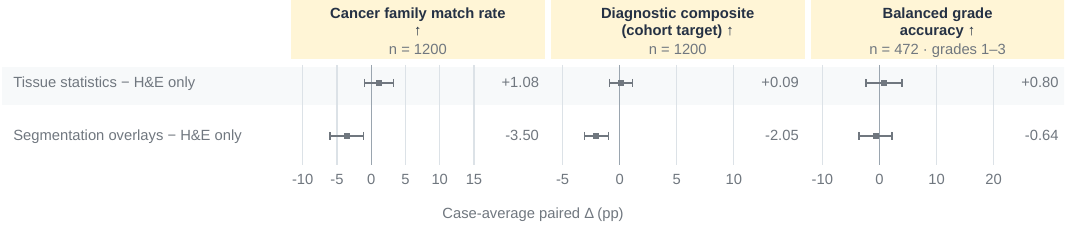}}
    \par\vspace{2mm}
    \manuscriptpanelblock{5mm}{180.0mm}{\manuscriptpanelmark{pan:report-input-workflow-gains}}{Layer 2 | Complete reporting strategy versus direct inputs}{\includegraphics[width=\linewidth]{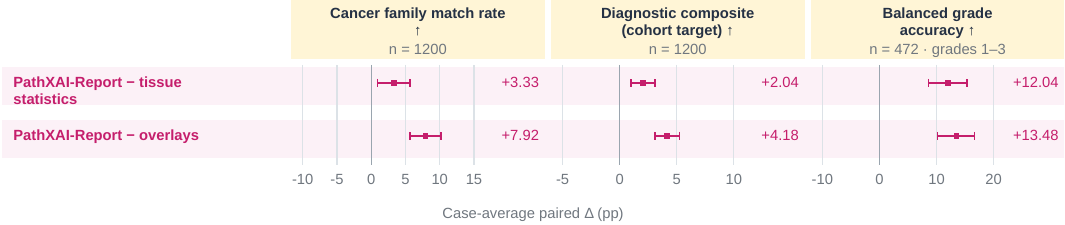}}
    \par\vspace{2mm}
    \manuscriptpanelblock{5mm}{180.0mm}{\manuscriptpanelmark{pan:report-input-report-content}}{Report content, finding coverage and contradictions}{\includegraphics[width=\linewidth]{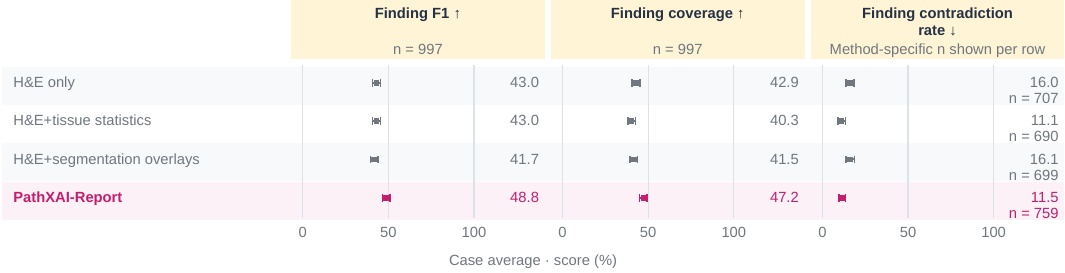}}
    \end{minipage}%
  }{%
\textbf{\panelletterref{pan:report-input-policies} Reporting policies.} \textcolor{black}{Comparisons use the same 1,200 public evaluation cases as the main analysis.}
\textbf{\panelletterref{pan:report-input-shared-case-performance} Diagnostic and grading performance.} Cancer-family targets are TCGA labels; balanced grade accuracy uses three-level histological grades.
\textbf{\panelletterref{pan:report-input-tissue-input-gains},\panelletterref{pan:report-input-workflow-gains} Paired gains.} Differences subtract the reference policy from the compared policy.
\textbf{\panelletterref{pan:report-input-report-content} Report content.} Finding contradiction rate is interpreted with finding coverage. Inapplicable reference fields are excluded; protocols are in \cref{supp:report-input-comparisons}.
(Notation: H\&E, haematoxylin and eosin; ROI, region of interest; arrows, preferred directions; pp, percentage points. \textcolor{black}{Whiskers show pointwise 95\% confidence intervals (\cref{supp:report-statistics}). Paired contrasts use case-bootstrap intervals, stratified by reference grade for grading.})
}

\begin{supplementarytables}
\input{supplementary/generated_report_v4_policy}
\end{supplementarytables}

\textcolor{black}{\textbf{Sensitivity to tissue-statistics availability.}
Restricting the \textcolor{black}{\textcolor{black}{PathXAI-Report pipeline development subset}} four-policy comparison to the 443 cases with an available tissue-statistics
context flag changed each paired difference by less than 0.6~pp
(\cref{tab:tissue-context-sensitivity}). For tissue statistics versus H\&E alone,
the cancer family match rate difference was 4.06~pp and the exact diagnosis match rate
difference was 4.71~pp, compared with 4.17 and 4.55~pp in the full \textcolor{black}{PathXAI-Report pipeline development subset}.
Both restricted diagnostic intervals excluded zero. Nonzero paired differences retained
their directions; the previously zero exact-diagnosis difference between PathXAI-Report
and direct tissue statistics became 0.59~pp, with an interval including zero.
Source Data includes cancer-type counts and diagnostic transitions for both case sets.}
\clearpage
\begin{supplementarytables}
\input{supplementary/generated_tissue_context_sensitivity}
\end{supplementarytables}

\supplementaryresultsubsection{Grade-specific outcomes of the reporting policies}{supp:results-policies-grade}
The grade-agreement gain of PathXAI-Report over direct tissue statistics was largest in
reference histological Grade 2 cases. \Cref{fig:report-grade-policy} breaks down correct,
incorrect and missing grade assignments in the same \textcolor{black}{472} applicable cases. In Grade 2,
correct assignments rose from
\textcolor{black}{46/179 to 96/179, including 47 cases changing from a missing or incompatible prediction to a correct
Grade 2. Grade 1 remained poorly recovered (1/73 with statistics and 0/73 with PathXAI-Report).
Among reference Grade 3 cases, correct assignments increased from 105/220 to 126/220,
although incorrect Grade 2 assignments increased from 20/220 to 55/220.}
\manuscriptnativefigure{\textbf{Grade predictions under four reporting policies.}}{fig:report-grade-policy}{%
  \begin{minipage}{180mm}
  \manuscriptpanelblock{5mm}{88mm}{\manuscriptpanelmark{pan:report-grade-policy-he}}{H\&E only}{\includegraphics[width=\linewidth]{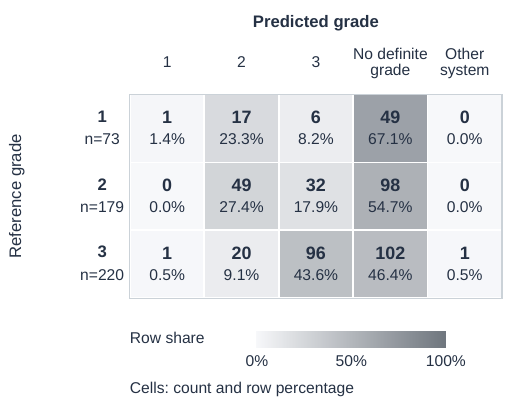}}\hfill
  \manuscriptpanelblock{5mm}{88mm}{\manuscriptpanelmark{pan:report-grade-policy-statistics}}{H\&E+tissue statistics}{\includegraphics[width=\linewidth]{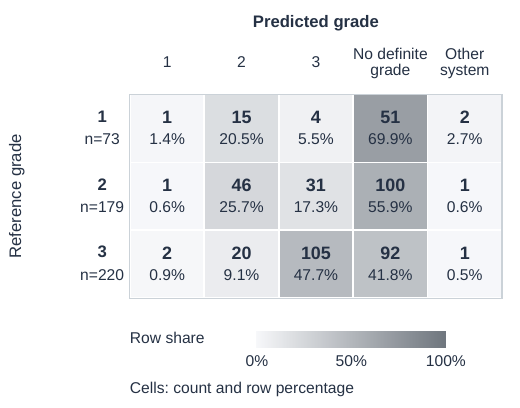}}
  \par\vspace{3mm}
  \manuscriptpanelblock{5mm}{88mm}{\manuscriptpanelmark{pan:report-grade-policy-overlays}}{H\&E+segmentation overlays}{\includegraphics[width=\linewidth]{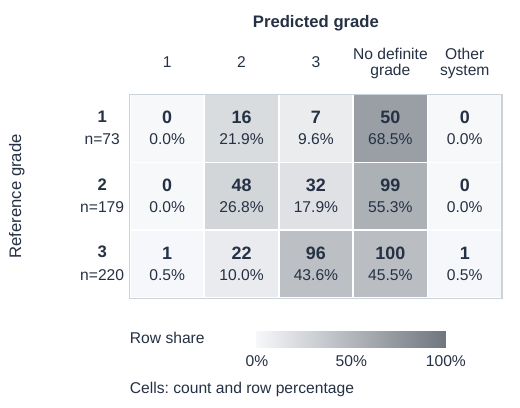}}\hfill
  \manuscriptpanelblock{5mm}{88mm}{\manuscriptpanelmark{pan:report-grade-policy-workflow}}{PathXAI-Report}{\includegraphics[width=\linewidth]{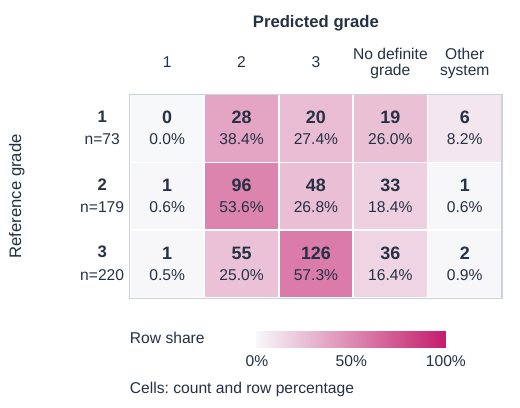}}
  \end{minipage}%
}{%
\textbf{\panelletterref{pan:report-grade-policy-he}--\panelletterref{pan:report-grade-policy-workflow}} compare H\&E alone, tissue-statistics input, segmentation-overlay input and PathXAI-Report on the same cases with assessable three-level histological grades (\cref{supp:report-input-comparisons,supp:report-grade-findings}).
Rows denote reference grades and columns denote predictions; cells give counts and row percentages on a common shading scale. ``No definite grade'' includes missing, uncertain and conflicting outputs. ``Other system'' denotes incompatible grading systems, whose values do not count as correct histological grades.
(Notation: H\&E, haematoxylin and eosin.)
}

\supplementaryresultsubsection{Report-workflow comparisons}{supp:results-policies-ablations}
The comparisons examine evidence organization, acquisition and writing. \textcolor{black}{The earlier organization
and acquisition comparisons use shared \textcolor{black}{cases used for PathXAI-Report pipeline development}; writing comparisons use the 1,200 public evaluation cases.} \Cref{tab:source_supp3} gives all estimates and applicable counts.
The 42-case evidence-organization contrasts were imprecise across all plotted outcomes
(\cref{pan:report-ablation-evidence-organization}).

Conflict-triggered expansion achieved
higher balanced grade accuracy than guarded dual-path reasoning by
\ExpansionVsGuardGrade{}~pp (95\% CI, \ExpansionVsGuardGradeCI; \ExpansionVsGuardGradeN{} applicable cases). Zoom-enabled reinspection
exceeded conflict-triggered expansion by \ZoomVsExpansionGrade{}~pp, with an interval spanning zero
(\ZoomVsExpansionGradeCI; the same \ZoomVsExpansionGradeN{} cases; \cref{pan:report-ablation-acquisition}).

\Needspace{10\baselineskip}
Combining knowledge\textcolor{black}{, diagnostic safeguards and} diagnosis-constrained writing improved grade agreement
and finding F1 over writer-only grade images. The grade-agreement difference was \KnowledgeVsGradeGrade{}~pp
(95\% CI, \KnowledgeVsGradeGradeCI; \KnowledgeVsGradeGradeN{} applicable cases).
The finding-F1 difference for this combination was \KnowledgeVsGradeFinding{}~pp (95\% CI, \KnowledgeVsGradeFindingCI;
\KnowledgeVsGradeFindingN{} applicable cases); \textcolor{black}{the diagnostic-composite difference was
\KnowledgeVsGradeDiagnostic{}~pp (95\% CI, \KnowledgeVsGradeDiagnosticCI)} (\cref{pan:report-ablation-writing-constraints}).
In the same \textcolor{black}{1,200 public evaluation} cases, supplying grade-focused images only to the
writer changed balanced grade accuracy by \GradeEvidenceVsCoreGrade{}~pp relative
to PathXAI-Report (core) (95\% CI, \GradeEvidenceVsCoreGradeCI; \GradeEvidenceVsCoreGradeN{} applicable cases), \textcolor{black}{with an interval including zero}.
\manuscriptnativefigure{\textbf{Evidence organization, acquisition and writing-policy comparisons.}}{fig:report-ablation}{%
    \begin{minipage}{180mm}
    \manuscriptpanelblock{5mm}{180.0mm}{\manuscriptpanelmark{pan:report-ablation-policy-comparisons}}{Agent-policy comparisons}{\includegraphics[width=\linewidth]{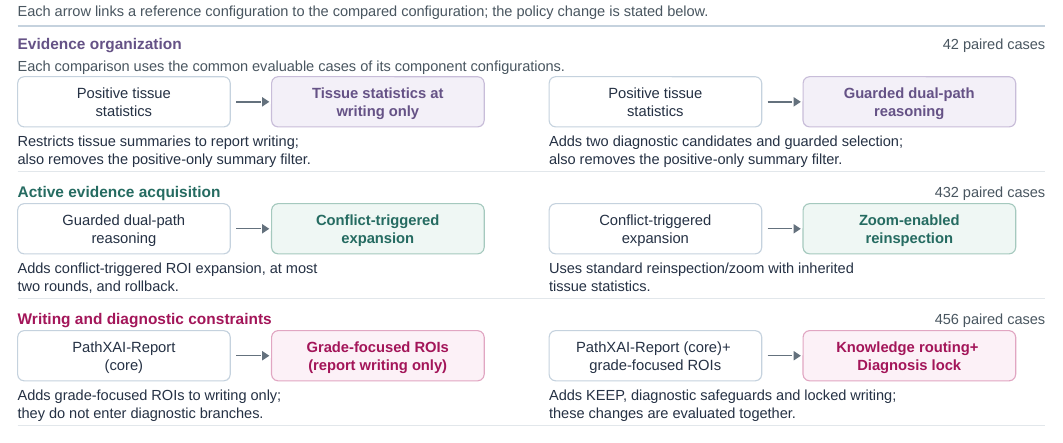}}
    \par\vspace{1.5mm}
    \manuscriptpanelblock{5mm}{180.0mm}{\manuscriptpanelmark{pan:report-ablation-evidence-organization}}{Evidence organization}{\includegraphics[width=\linewidth]{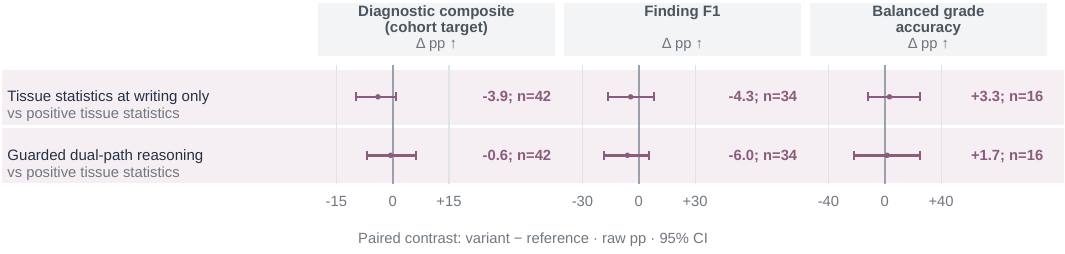}}
    \par\vspace{1.5mm}
    \manuscriptpanelblock{5mm}{180.0mm}{\manuscriptpanelmark{pan:report-ablation-acquisition}}{Active evidence acquisition}{\includegraphics[width=\linewidth]{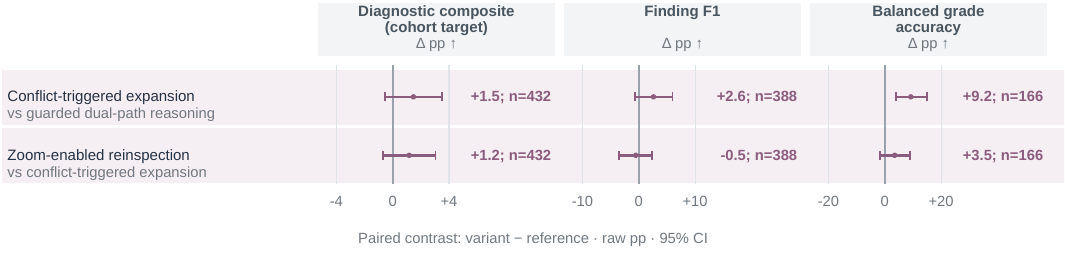}}
    \par\vspace{1.5mm}
    \manuscriptpanelblock{5mm}{180.0mm}{\manuscriptpanelmark{pan:report-ablation-writing-constraints}}{Writing and diagnostic constraints}{\includegraphics[width=\linewidth]{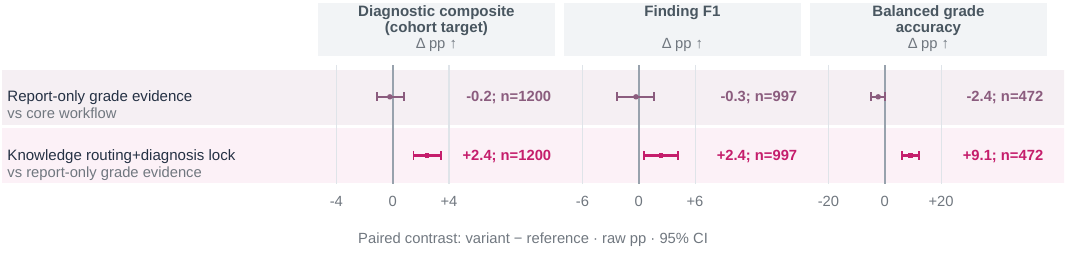}}
    \end{minipage}%
  }{%
\textbf{\panelletterref{pan:report-ablation-policy-comparisons} Policy contrasts.}
\textbf{\panelletterref{pan:report-ablation-evidence-organization}--\panelletterref{pan:report-ablation-writing-constraints} Effects of evidence organization, acquisition and writing constraints.} \textcolor{black}{Evidence-organization and acquisition comparisons use 42 and 432 paired cases, respectively, from the 456-case PathXAI-Report pipeline development subset. Writing-policy comparisons use the same 1,200 public evaluation cases as the input-policy comparison.} \textcolor{black}{The final contrast changes knowledge routing, candidate selection and diagnosis constraints together.}
Balanced grade accuracy uses three-level histological grading, with two represented grades in \panelletterref{pan:report-ablation-evidence-organization}. Policies are defined in \cref{supp:report-policy-comparisons}; full estimates are in Source Data.
(Notation: $n$, metric-applicable cases; arrows in \panelletterref{pan:report-ablation-policy-comparisons}, policy changes; pp, percentage points. Points show paired differences; whiskers show unadjusted 95\% case-bootstrap intervals, stratified by reference grade for grading. Higher values are preferred.)
}

\begin{supplementarytables}
\input{supplementary/generated_report_v4_ablation}
\end{supplementarytables}

\textcolor{black}{\textbf{Sensitivity to diagnostic-component weights.}
With equal component weights, PathXAI-Report retained higher diagnostic-composite
point estimates than Qwen+H\&E and PathXAI-Report (core) in the public evaluation
cohort and using both private reference sets (\cref{tab:diagnostic-weight-sensitivity}).
No paired comparison reversed direction. In the \textcolor{black}{PathXAI-Report pipeline development comparisons},
the cohort-target difference between guarded dual-path reasoning and positive
tissue statistics shifted from $-0.56$~pp to zero; both intervals included zero.
The direct-input, acquisition and writing comparisons retained their directions.
Individual diagnostic measures remain the primary outcomes.}
\begin{supplementarytables}
\input{supplementary/generated_diagnostic_weight_sensitivity}
\end{supplementarytables}

\supplementaryresultsection{Sensitivity to diagnostic extraction scope}{supp:report-scope-results}\label{supp:results-report-scope}
\textcolor{black}{Final summaries combine diagnostic, grading and morphological statements, so disease
terms are not always clearly attributable to a final diagnosis. The alternative extraction
applies the same diagnostic assertion rules to these additional sections.}
Including final-summary diagnoses tests whether diagnostic agreement depends on the report
sections used for extraction (\cref{tab:report-scope-sensitivity:part-2}). The added text can
refine a generic diagnosis or introduce a conflicting assertion, so agreement can increase
or decrease. PRISM2 and PLIP-guided Qwen have no recognized summary sections in these
comparisons and are unchanged. The table reports summary availability alongside score
changes; these results assess sensitivity to extraction scope, not the clinical accuracy
of either extraction rule.

\textcolor{black}{With summary-inclusive extraction, five PathXAI-Report exact-diagnosis matches
became mismatches in each public group, with no changes in the opposite direction
(\CuratedEntityFullN{} applicable \textcolor{black}{cases used for PathXAI-Report pipeline development} and \EvaluationEntityFullN{} applicable
evaluation cases). Its paired difference from Qwen+H\&E was \ScopeCuratedVsQwen{}~pp
in the \textcolor{black}{PathXAI-Report pipeline development subset} (95\% CI, \ScopeCuratedVsQwenCI) and \ScopeOCRVsQwen{}~pp in
the evaluation cohort (\ScopeOCRVsQwenCI); both intervals included zero.
The corresponding private differences were \ScopePrivateOneVsQwen{}~pp for P1
(\ScopePrivateOneVsQwenCI) and \ScopePrivateTwoVsQwen{}~pp for P2
(\ScopePrivateTwoVsQwenCI). Private differences from core were smaller and their
intervals included zero. In the \textcolor{black}{\textcolor{black}{PathXAI-Report pipeline development subset}} direct-input comparison, the tissue-statistics difference
from H\&E alone was \ScopeStatisticsVsHE{}~pp (\ScopeStatisticsVsHECI).}

\textcolor{black}{For cancer family match rate in the combined public cohort, the
summary-inclusive differences were \ScopeCancerVsQwen{}~pp versus Qwen+H\&E
(95\% CI, \ScopeCancerVsQwenCI) and \ScopeCancerVsDeepSeek{}~pp versus DeepSeek+H\&E
(\ScopeCancerVsDeepSeekCI). In the public evaluation cohort alone, PathXAI-Report
retained a higher rate than Qwen+H\&E (\ScopeCancerOCRVsQwen{}~pp;
\ScopeCancerOCRVsQwenCI), but had a lower point estimate than DeepSeek+H\&E;
the latter paired interval included zero. The tissue-statistics versus H\&E-only
difference in the \textcolor{black}{PathXAI-Report pipeline development subset} was \ScopeCancerStatisticsVsHE{}~pp
(\ScopeCancerStatisticsVsHECI; \cref{tab:cohort-scope-sensitivity:part-2}).}

\begin{supplementarytables}
\input{supplementary/generated_report_scope_sensitivity}
\input{supplementary/generated_cohort_scope_sensitivity}
\end{supplementarytables}

{\color{black}
\supplementaryresultsection{Multiple-comparison sensitivity}{supp:results-report-multiplicity}
After Holm adjustment across 126 comparisons, tests of PathXAI-Report versus Qwen+H\&E
in the public evaluation cohort gave $P<0.05$ for cancer family match rate, finding F1,
finding recall and finding contradiction rate
(\cref{tab:report-multiplicity}). The exact diagnosis match rate differences from Qwen+H\&E,
DeepSeek+H\&E and PathXAI-Report (core) did not meet this threshold. None of the five
comparisons with Qwen+H\&E in either private reference set met the adjusted threshold.
For the dedicated report generators, the public cancer family match rate, finding F1 and
finding recall comparisons gave adjusted $P<0.05$ for both comparators; the exact diagnosis
match rate comparison did so for HistoGPT but not HistGen. PathXAI-Report also had a higher
finding contradiction rate than PLIP-guided Qwen in the public evaluation cohort
(adjusted $P=0.035$). The table includes all comparisons; the adjustment leaves their
point estimates unchanged.
}
\begin{supplementarytables}
\input{supplementary/generated_report_multiplicity}
\end{supplementarytables}

{\color{black}
\supplementaryresultsection{Report availability and sensitivity to unavailable outputs}{supp:results-report-availability}
All eight methods in the main public evaluation had saved outputs for all 1,200 cases.
Among the 217 public cases eligible for the dedicated-generator comparison, HistGen and
HistoGPT each had 216 reports, with 215 cases shared by all four methods. In the private
cohort, GLM+H\&E had reports for 151 of 152 cases; the other nine methods had all 152
(\cref{tab:report-availability}).

Including all eligible cases and assigning zero to unavailable outputs changed each paired
difference by less than 1~pp. Across the four measures, no nonzero difference between
PathXAI-Report and a comparator changed direction
(\cref{tab:report-unavailable-sensitivity}).
}
\begin{supplementarytables}
\input{supplementary/generated_report_availability}
\end{supplementarytables}

{\color{black}
\supplementaryresultsection{Cancer-type- and organ-stratified comparisons}{supp:results-report-stratification}
Stratified resampling retained the original paired estimates and conditioned their
intervals on the observed cancer-type or organ composition
(\cref{tab:report-stratification}). In the public evaluation cohort, all five
pointwise intervals for differences from Qwen+H\&E excluded zero in favour of
PathXAI-Report.

In the private cohort, exact diagnosis match rate and finding recall differences
from Qwen+H\&E had intervals above zero for both reference sets, while cancer
family match rate and finding F1 intervals included zero. The higher finding
contradiction rate had an interval above zero for P2 and included zero for P1.
These are pointwise intervals; the separate multiplicity analysis is reported in
\cref{supp:results-report-multiplicity}. Source Data gives descriptive results
within each of the 22 public cancer types and three private organs, with
metric-specific counts. Three public cancer types had only one case each.
}
\clearpage
\begin{supplementarytables}
\input{supplementary/generated_report_stratification}
\end{supplementarytables}

\FloatBarrier
\supplementaryresultsection{Evidence-linked public cases and failure analysis}{supp:public-case-results}
Two public cases illustrate agreement and disagreement with TCGA cohort targets.

\manuscriptnativefigure{\textbf{A public report with a target-concordant diagnosis.}}{fig:report-case-success}{%
    \begin{minipage}{180mm}
    \manuscriptpanelblock{5mm}{180mm}{\manuscriptpanelmark{pan:report-case-success-spatial-evidence}}{Evidence acquisition and reporting timeline}{\includegraphics[width=\linewidth]{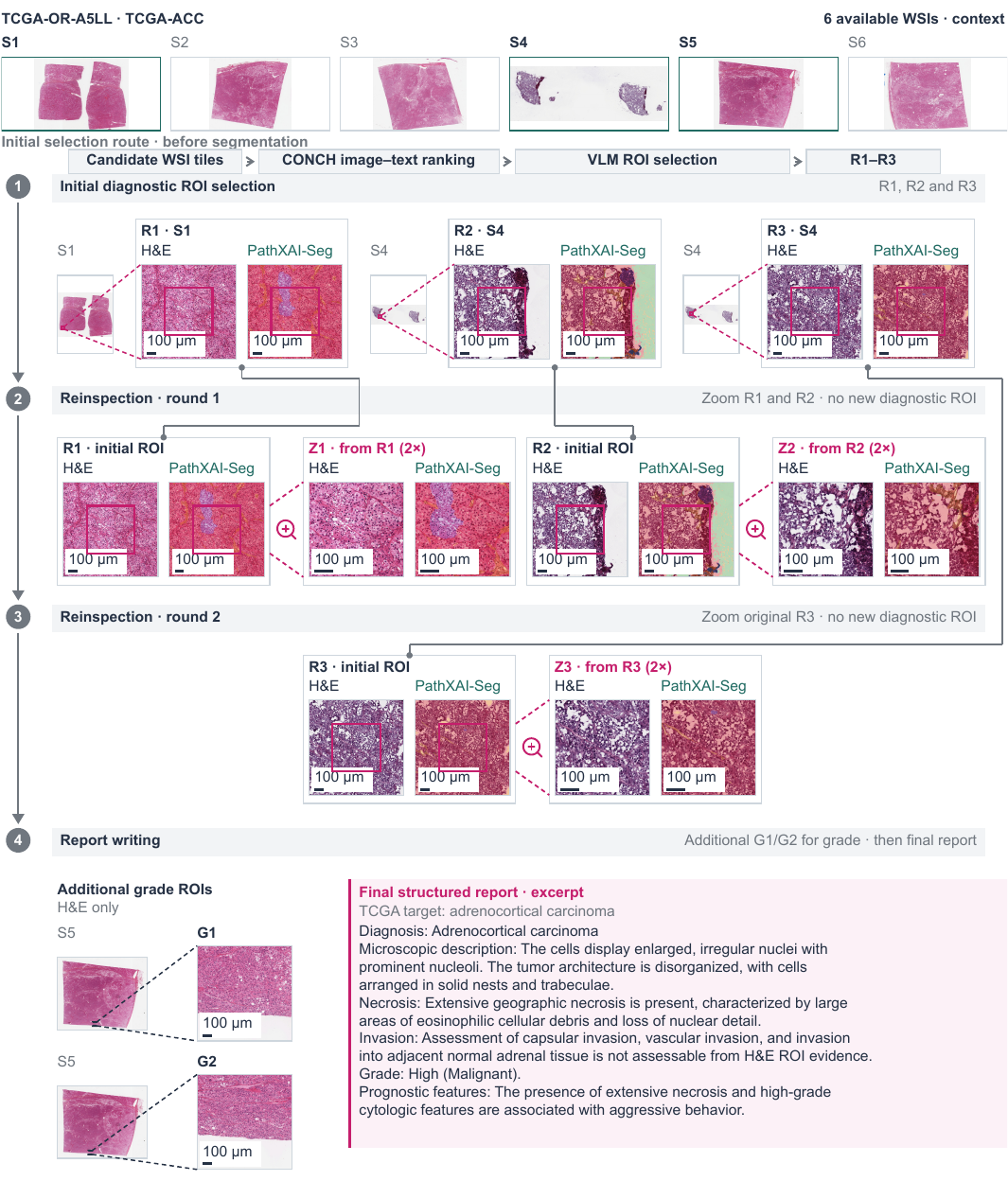}}
    \par\vspace{1mm}
    \includegraphics[width=180mm]{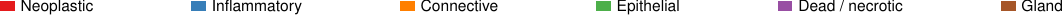}
    \end{minipage}%
  }{%
\textbf{\panelletterref{pan:report-case-success-spatial-evidence} Evidence acquisition and reporting.} \textcolor{black}{CONCH ranking precedes VLM region selection.} Reinspection uses central-crop 2$\times$ digital zooms of the original diagnostic ROIs. G1/G2 are unmasked writer-only grade images; high grade was already assigned before writing. Abridged report excerpts match the TCGA cancer-family target.
\textcolor{black}{Report excerpts and verification records correspond to the \href{https://github.com/zhaorui-tan/PathXAI/blob/0bb72386afb2c21bde5c912bf72a663bd6f48175/examples/SO4/report.html}{released Full example}}\textcolor{black}{; \cref{fig:case-summary} shows the archived evaluation run of the same case, which used the same ROIs and recorded the same necrosis tissue check.}
Retrieved FNA text in \cref{fig:case-summary} supplies external disease context. \textcolor{black}{Tissue checks assess tumour presence and stromal evidence; architectural and cytological descriptions require image interpretation.}
(Notation: WSI, whole-slide image; ROI, region of interest; VLM, vision-language model; H\&E, haematoxylin and eosin; FNA, fine-needle aspiration. Display conventions and scale bars follow \cref{fig:report-case-failure}.)
}

\manuscriptnativefigure{\textbf{A public report with persistent target discordance.}}{fig:report-case-failure}{%
    \begin{minipage}{180mm}
    \manuscriptpanelblock{5mm}{180mm}{\manuscriptpanelmark{pan:report-case-failure-spatial-evidence}}{Evidence acquisition and reporting timeline}{\includegraphics[width=\linewidth]{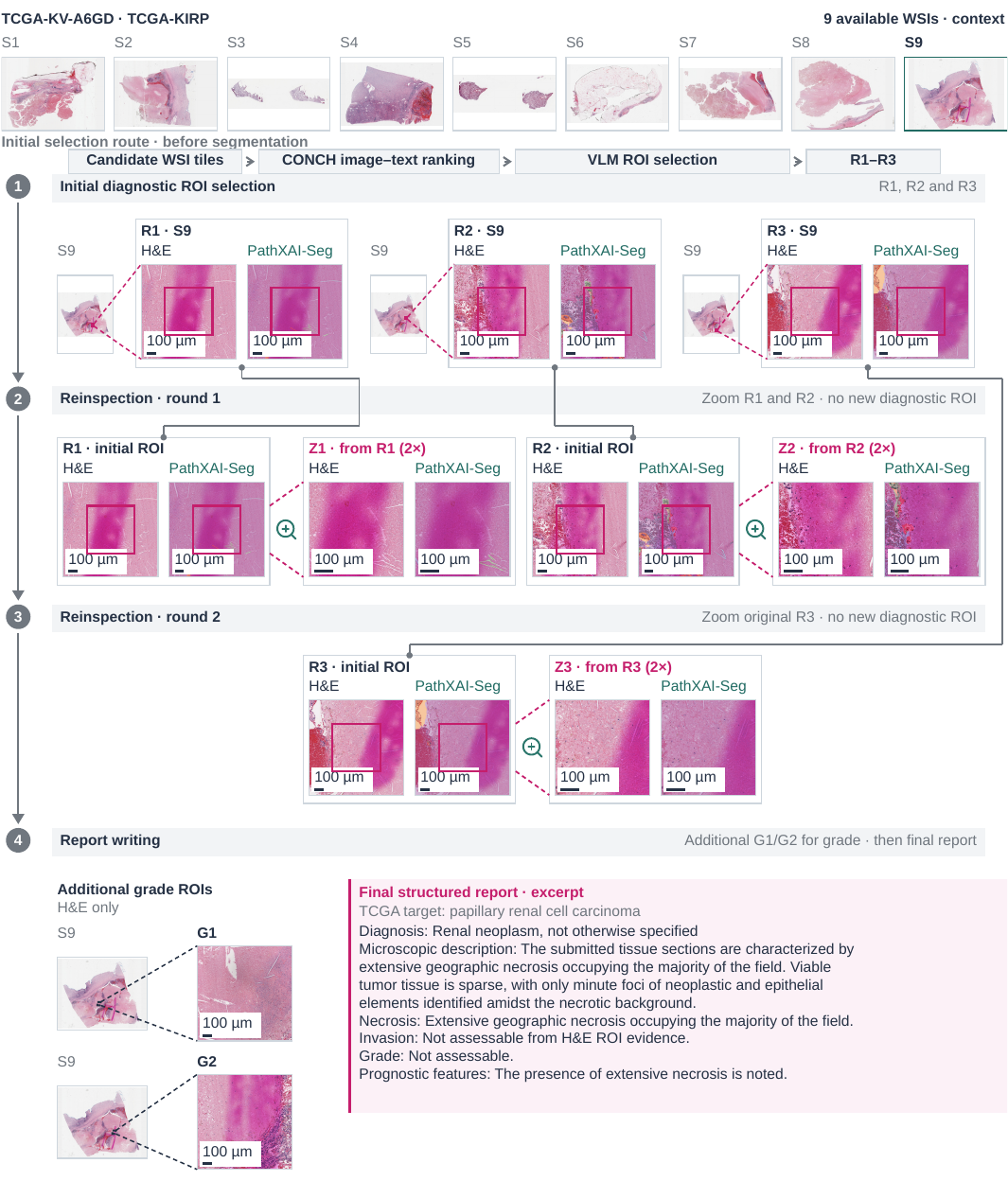}}
    \par\vspace{1mm}
    \includegraphics[width=180mm]{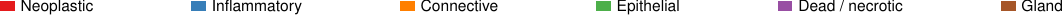}
    \end{minipage}%
  }{%
\textbf{\panelletterref{pan:report-case-failure-spatial-evidence} Evidence acquisition and reporting.} CONCH ranking precedes VLM region selection. Reinspection uses central-crop 2$\times$ digital zooms; G1/G2 supply unmasked H\&E for writing only. The final diagnosis remains broader than the papillary renal-cell-carcinoma TCGA target.
(Notation: WSI, whole-slide image; ROI, region of interest; VLM, vision-language model; H\&E, haematoxylin and eosin. Grey solid links denote repeated images; dashed guides link slides to ROIs. Crop boxes and magnifiers mark zooms; numbered boxes identify slides or ROIs. Predicted overlays use the shared tissue palette. Scale bars, 100~$\mu$m, calibrated from source WSIs.)
}

\FloatBarrier
\supplementaryresultsection{Segmentation transfer and training-objective comparisons}{supp:segmentation-results}
\supplementaryresultsubsection{Validation across datasets and organs}{supp:results-seg-validation}
\Crefrange{tab:source_seg:part-1}{tab:source_seg:part-4} give the estimates and confidence intervals underlying
\cref{pan:seg-summary-dataset-dice,pan:seg-summary-shared-external-comparison,pan:seg-summary-fine-tuning-gains,pan:seg-summary-organ-dice}.
Per-dataset Dice and IoU, sample counts and source-cluster counts are in Source Data.
Relative to same-backbone fine-tuning, PathXAI-Seg (SAM2) has higher internal lung Dice and lower internal colorectal
Dice.

\textcolor{black}{Class-specific results show different changes across tissues
(\cref{tab:seg-main-classes}). On RINGS gland, PathXAI-Seg (MedSAM3) improved both
gland and background mean Dice relative to same-backbone fine-tuning. The SAM2 variant
increased background Dice from 52.27\% to 69.60\% but reduced gland Dice from
56.72\% to 37.05\%. Class support also varied: epithelial Dice was evaluated on only
two WSSS4LUAD images.}

Coupling the bootstrap draws for the shared RINGS sources retained the MedSAM3-versus-fine-tuning
external gain of 0.83~pp; its 95\% interval was 0.41--1.35, compared with
0.44--1.32 under task-stratified resampling (Source Data). The sampling procedures are described
in \cref{supp:seg-validation-protocol}.

MedSAM3 fine-tuning and PathXAI-Seg (MedSAM3) had similar mean Dice on the complete PAIP2019
test set (\cref{pan:seg-summary-dataset-dice}).

\textcolor{black}{In the shared external comparison, both PathXAI-Seg variants had higher equal-dataset
foreground Dice than PathSegmentor and VISTA-PATH, while VISTA-PATH had the highest
background Dice on each dataset (\cref{tab:seg-shared-class}). Under the reference-class-presence
condition, all four methods scored nearly 100\% Dice on the 7,584 images containing only one
class. On the 2,432 images containing both foreground and background, equal-dataset mean
Dice was 55.70\% for PathXAI-Seg (SAM2), 54.21\% for PathXAI-Seg (MedSAM3), 51.01\% for
VISTA-PATH and 31.59\% for PathSegmentor. These mixed-class estimates describe segmentation
where a spatial boundary is present.}
\begin{supplementarytables}
\input{supplementary/generated_seg_validation}
\end{supplementarytables}
\input{supplementary/generated_main_seg_classes}
\input{supplementary/generated_seg_shared_class}

\Needspace{0.7\textheight}
\supplementaryresultsubsection{Training stages, objective configurations and refinement branches}{supp:results-seg-training}
\Cref{pan:seg-summary-component-deletions} shows the objective-weight comparisons; \cref{tab:source_supp2} also reports training checkpoints and refinement branches.
The alternative refinement branches were evaluated only internally, with Dice ranging from
61.03\% to 65.83\%; these runs are distinct from the final focal-refinement checkpoint.
Comparison designs and checkpoint-selection conditions are described in
\cref{supp:seg-training-comparisons}.
\begin{supplementarytables}
\input{supplementary/generated_seg_ablation}
\end{supplementarytables}

\FloatBarrier

\FloatBarrier
\input{supplementary/visual_conventions}

\end{document}